\pdfoutput=1
\documentclass[10pt,twocolumn,letterpaper]{article}

\usepackage{iccv}
\usepackage{times}
\usepackage{epsfig}
\usepackage{graphicx}
\usepackage{amsmath}
\usepackage{amssymb}

\usepackage[pagebackref=true,breaklinks=true,letterpaper=true,colorlinks,bookmarks=false]{hyperref}

\iccvfinalcopy 

\def\iccvPaperID{9450} 

\ificcvfinal\fi
\usepackage{scalerel}
\usepackage{kotex}
\usepackage{multirow}
\usepackage{graphicx}
\usepackage{textgreek}
\usepackage{amssymb}
\usepackage{amsmath}
\usepackage{array}
\usepackage[dvipsnames]{xcolor}
\usepackage{soul}
\usepackage{booktabs}
\usepackage{adjustbox}
\usepackage{makecell}
\usepackage{stmaryrd}
\usepackage{multibib}
\newcites{SM}{References not in the main paper}

\newcolumntype{N}{>{\centering\arraybackslash}m{.5in}}
\newcolumntype{G}{>{\centering\arraybackslash}m{2in}}

\begin{document}

\newcommand{\Eq}[1]  {Eq.\ (\ref{eq:#1})}
\newcommand{\Eqs}[1] {Eqs.\ (\ref{eq:#1})}
\newcommand{\Fig}[1] {Fig.\ \ref{fig:#1}}
\newcommand{\Figs}[1]{Figs.\ \ref{fig:#1}}
\newcommand{\Tbl}[1]  {Table \ref{tbl:#1}}
\newcommand{\Tbls}[1] {Tables \ \ref{tbl:#1}}
\newcommand{\Sec}[1] {Sec.\ \ref{sec:#1}}
\newcommand{\SSec}[1] {Sec.\ \ref{ssec:#1}}
\newcommand{\SSecs}[1] {Sec.\ \ref{ssec:#1}}
\newcommand{\Secs}[1] {Secs.\ \ref{sec:#1}}
\newcommand{\Etal}   {{\textit{et al.}}}
\newcommand{\IE}   {{\textit{i.e.}}}

\newcommand{\setone}[1] {\left\{ #1 \right\}} 
\newcommand{\settwo}[2] {\left\{ #1 \mid #2 \right\}} 

\newcommand{\todo}[1]{{\textcolor{red}{TODO: #1}}}
\newcommand{\son}[1]{{\textcolor{magenta}{hyeongseok: #1}}}
\newcommand{\jy}[1]{{\textbf{\textcolor{MidnightBlue}{[JY] }}\textcolor{MidnightBlue}{#1}}}
\newcommand{\sean}[1]{{\textcolor{green}{sean: #1}}}
\newcommand{\sunghyun}[1]{{\textcolor[rgb]{0.6,0.0,0.6}{sunghyun: #1}}}
\newcommand{\change}[1]{{\color{black}#1}}
\newcommand{\changed}[1]{{\color{MidnightBlue}#1}}
\newcommand{\bb}[1]{\textbf{\textit{#1}}}

\long\def\myskip#1{}

\renewcommand{\topfraction}{0.95}
\setcounter{bottomnumber}{1}
\renewcommand{\bottomfraction}{0.95}
\setcounter{totalnumber}{3}
\renewcommand{\textfraction}{0.05}
\renewcommand{\floatpagefraction}{0.95}
\setcounter{dbltopnumber}{2}
\renewcommand{\dbltopfraction}{0.95}
\renewcommand{\dblfloatpagefraction}{0.95}




\title{Single Image Defocus Deblurring\\
Using Kernel-Sharing Parallel Atrous Convolutions}

%

\author{
    \vspace*{-10pt}\\
    Hyeongseok Son
    \qquad
    Junyong Lee
    \qquad
    Sunghyun Cho
    \qquad
    Seungyong Lee\\
    \vspace*{-10pt}\\
    POSTECH\\
    {\tt\small
    \{sonhs, junyonglee, s.cho, leesy\}@postech.ac.kr}
}


\maketitle
\ificcvfinal\thispagestyle{empty}\fi

\begin{abstract}

This paper proposes a novel deep learning approach for single image defocus deblurring based on inverse kernels.
In a defocused image, the blur shapes are similar among pixels although the blur sizes can spatially vary.
To utilize the property with inverse kernels, we exploit the observation that when only the size of a defocus blur changes while keeping the shape, the shape of the corresponding inverse kernel remains the same and only the scale changes.
Based on the observation, we propose a kernel-sharing parallel atrous convolutional (KPAC) block specifically designed by incorporating the property of inverse kernels for single image defocus deblurring.
To effectively simulate the invariant shapes of inverse kernels with different scales,
KPAC shares the same convolutional weights among multiple atrous convolution layers.
To efficiently simulate the varying scales of inverse kernels, 
KPAC consists of only a few atrous convolution layers with different dilations and learns per-pixel scale attentions to aggregate the outputs of the layers.
KPAC also utilizes the shape attention to combine the outputs of multiple convolution filters in each atrous convolution layer, to deal with defocus blur with a slightly varying shape.
We demonstrate that our approach achieves state-of-the-art performance with a much smaller number of parameters than previous methods.

\end{abstract}
\section{Introduction}
\label{sec:intro}

Defocus blur of an image occurs when the light ray from a point in the scene forms a circle of confusion (COC) on the camera sensor.
The aperture shape and lens design of the camera determine the blur shape,
and the blur size varies upon the depth of a scene point and intrinsic camera parameters.
In a defocused image, the spatial variance of the blur size is large, while that of the blur shape is relatively small.
Single image defocus deblurring remains a challenging problem as it is hard to accurately estimate and remove defocus blur spatially varying in both size and shape.

The conventional two-step approach \cite{Bando2007refocus,Shi:2015:JNB,Andres:2016:RTF,Park:2017:unified,Cho:2017:Convergence,Karaali:2018:DMEAdaptive,Lee:2019:DMENet} reduces the complexity of defocus deblurring by assuming an isotropic kernel for the blur shape,
such as disc \cite{Bando2007refocus,Andres:2016:RTF} or Gaussian \cite{Shi:2015:JNB, Park:2017:unified, Karaali:2018:DMEAdaptive, Lee:2019:DMENet}.
Based on the assumption, the approach first estimates a defocus map containing the per-pixel blur size of a defocused image,
then uses the defocus map to perform non-blind deconvolution \cite{Fish:95:BD, Levin:2007:Coded, Krishnan:2008:deconvolution} on the image.
However, real-world defocused images may have more complex kernel shapes than disc or Gaussian, and the discrepancy often hinders accurate defocus map estimation and successful defocus deblurring.

Recently, Abuolaim and Brown~\cite{Abuolaim:2020:DPDNet} proposed DPDNet, the first end-to-end 
defocus deblurring network, that learns to directly deblur a defocused image without relying on a restrictive blur model.
They also presented a defocus deblurring dataset that includes stereo images attainable from a dual-pixel sensor camera.
Thanks to the end-to-end learning and the strong supervision provided by the dual-pixel dataset, DPDNet outperforms two-step approaches on deblurring of real-world defocused images.
Still, the deblurred results tend to include ringing artifacts or remaining blur, as the conventional encoder-decoder architecture of DPDNet confines its capability in handling spatially variant blur \cite{Zhou:2019:STFAN}.

In this paper, we propose a novel deep learning approach for single image defocus deblurring based on inverse kernels. 
It was shown that deconvolution of an image with a given blur kernel can be performed by convolving the image with an inverse kernel~\cite{Xu:2014:DECONV}, where the inverse kernel could be computed from the given blur kernel using Fourier transform.
Xu \Etal~\cite{NIPS2014inverse} trained a deep network to learn uniform deconvolution by introducing the property of pseudo inverse kernel into the network. 
Similarly, we train our network to learn the deconvolution operation by capitalizing the specific characteristics of inverse kernels needed for defocus deblurring. 
However, due to the spatially varying nature of defocus blur, the inverse kernel required for defocus deblurring also changes per pixel. 
Training a deep network to learn deconvolution operations of varying inverse kernels would be challenging even with the guidance of defocused and sharp image pairs.

To reduce the complexity, we use the property of defocus blur that the blur shape is similar in a defocused image although the blur size can drastically change. However, instead of assuming any specific blur shape as in the two-step approach,
we exploit our observation on inverse kernels;
When only the size of a blur changes while keeping the shape,
the shape of the corresponding inverse kernel remains the same and the size changes in the same way as the blur (\SSec{key_idea}).
Then, we may constrain our network to simulate inverse kernels with a single shape but with different sizes to cover spatially varying defocus blur.
However, it is hard to directly simulate inverse kernels with all possible sizes in practice.
Instead, we design our network to carry a few convolutional layers to cover inverse kernels with a discrete set of sizes, and aggregate the outputs of the layers for handling blur with arbitrary size.
As a result, compared to the conventional two-step approach and the recent deep learning approach~\cite{Abuolaim:2020:DPDNet},
our method can perform defocus deblurring more accurately
by exploiting the properties of defocus blur in the form of inverse kernel.

To implement the network design, we propose a novel {\em kernel-sharing parallel atrous convolutional (KPAC) block.} 
The KPAC block consists of multiple atrous convolution layers~\cite{Holschneider1990Atrous,Chen2018Atrous} with different dilation rates and additional layers for the scale and shape attentions.
The multiple atrous convolution layers share the same convolution kernels representing the invariant shape of an inverse kernel,
and different dilation rates of the layers correspond to inverse kernels with a discrete set of sizes.
To simulate deconvolution using inverse kernels of other sizes, the KPAC block equips a spatial attention mechanism~\cite{Kelvin2015Spatial}, which we call {\em the scale attention}, to aggregate the outputs of the atrous convolution layers.
By combining the per-pixel scale attention with multiple atrous convolution layers,
the KPAC block
can handle the spatially varying size of defocus blur.
In addition,
the shape of defocus blur may slightly change in a defocused image due to the non-linearity of a camera image pipeline.
To handle the variance, we include a channel attention mechanism~\cite{zhang2018rcan}, which we call {\em the shape attention}, in a KPAC block
to support the slight shape change of the inverse kernel.

An important benefit of our KPAC block is its small number of parameters, enabled by kernel sharing of the multiple atrous convolution layers.
As a result, our defocus deblurring network is lighter-weighted than the previous work~\cite{Abuolaim:2020:DPDNet}, showing the better performance (\SSec{comparison}).

To summarize, our contributions include:
\begin{itemize}
    \vspace{-5pt}
    \item Novel deep learning approach for single image defocus deblurring based on inverse kernels,
    \vspace{-5pt}
    \item Novel Kernel-Sharing Parallel Atrous Convolutional (KPAC) block designed upon the properties of spatially varying inverse kernels for defocus deblurring,
    \vspace{-10pt}
    \item Light-weight single image defocus deblurring network, which shows state-of-the-art performance.
\end{itemize}

\section{Related Work}
\subsection{Defocus deblurring}
Conventional approaches~\cite{Bando2007refocus,Shi:2015:JNB,Andres:2016:RTF,Park:2017:unified,Cho:2017:Convergence,Karaali:2018:DMEAdaptive,Lee:2019:DMENet} perform defocus deblurring in two steps of defocus map estimation and non-blind deconvolution.
As they use existing non-blind deconvolution methods~\cite{Levin:2007:Coded, Krishnan:2008:deconvolution} for deblurring, they focus on improving the accuracy of defocus map estimation
based on parametric blur models such as disc and Gaussian blur.
Various methods were proposed for defocus map estimation using hand-crafted features such as edge gradients~\cite{ Karaali:2018:DMEAdaptive}, sparse coded features~\cite{Shi:2015:JNB}, machine learning features \cite{Andres:2016:RTF}, combination of hand-crafted and deep learning features \cite{Park:2017:unified}, and an end-to-end deep learning model \cite{Lee:2019:DMENet}.
These two-step approaches often fail to produce faithful deblurring results due to the restricted blur model as well as the errors of defocus map estimation.

Recently, Abuolaim and Brown~\cite{Abuolaim:2020:DPDNet} proposed the first end-to-end model for deep learning-based defocus deblurring and a dataset for supervised training.
The model outperforms conventional two-step approaches,
and shows that a dual-pixel image input significantly improves defocus deblurring performance.
However, their network structure does not explicitly consider the spatially varying nature of defocus blur with large variance in size but small variance in shape, and the performance has rooms for improvements.


\subsection{Inverse kernel}
Convolution of an inverse kernel can be used to perform deblurring on an image~\cite{Wiener,Yuan2008progressive,Xu:2014:DECONV}.
However, a na\"ive inverse kernel that could be obtained from Wiener deconvolution~\cite{Wiener} usually introduces unwanted artifacts such as ringing and amplified noise.
To suppress artifacts, previous works took a progressive approach using an image pyramid~\cite{Yuan2008progressive}, regularization using sparse priors~\cite{Levin:2007:Coded,Xu:2014:DECONV}, a neural network for postprocessing~\cite{Son2017}, and feature space processing~\cite{Dong2020Wiener}.
Xu \Etal~\cite{NIPS2014inverse} and Ren \Etal~\cite{Ren2018deconv} directly used a CNN to perform non-blind deconvoluton by adapting a separable property of a large inverse kernel, and showed that their approaches are 
effective for suppressing artifacts.
Distinct from \cite{NIPS2014inverse,Ren2018deconv} that simulate a single inverse kernel for uniform deconvolution, our approach simulates spatially varying inverse kernels needed for defocus deblurring.

\section{Key Idea}
In this section, we first present our observation on inverse kernels for deconvolution, and propose inverse kernel-based deconvolution to deal with spatially varying defocus blur (\SSec{key_idea}).
We then present experiments to verify the observation and the proposition (\SSec{verification}). 

\begin{figure}[t]
\centering
\scriptsize
\setlength\tabcolsep{0.9 pt}
  \begin{tabular}{cccccc}
    \multicolumn{2}{c}{\includegraphics[width=0.33\linewidth]{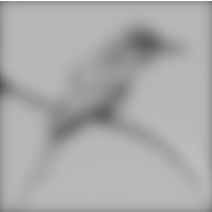}} & 
    \multicolumn{2}{c}{\includegraphics[width=0.33\linewidth]{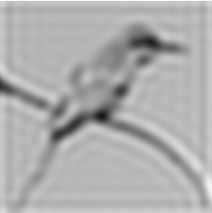}} & 
    \multicolumn{2}{c}{\includegraphics[width=0.33\linewidth]{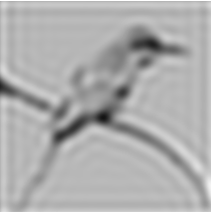}} \\

    \multicolumn{2}{c}{(a) blurred by \tiny{\(\frac{1}{5^2}k_{\uparrow{5}}\)}}  & \multicolumn{2}{c}{(b) deblurred by \tiny{\((\frac{1}{5^{2}}k_{\uparrow{5}})^{\dagger}\)}}
     & \multicolumn{2}{c}{(c) deblurred by \tiny{\(\frac{1}{5^{2}}({k^{\dagger}}_{\!\!\uparrow{5}})\)}} \\

  \end{tabular}
  \vspace{0.01cm}
  \caption{Deblurring examples using the two kernels in \Eq{sharing1}. Wiener deconvolution is used for computing the inverse kernels. The results in (b) and (c) are almost equivalent. Their PSNRs to the original sharp image are 21.64dB and 21.59dB, respectively, and the PSNR between them is 51.09dB. 
}
\label{fig:kernel_example}
\vspace{-10pt}
\end{figure}

\subsection{Inverse kernel-based deconvolution for\\spatially varying defocus blur}
\label{ssec:key_idea}

The inverse kernel-based deconvolution approach is closely related to convolution neural networks (CNN) by nature, as they are based on convolutional operations in the spatial domain.
We aim to design a network architecture that takes the benefits of both CNN and inverse kernels for defocus deblurring.
\change{To this end, we first introduce the concept and general derivation of pseudo inverse kernel, as in previous works \cite{NIPS2014inverse,Ren2018deconv}.

We consider a simple blur model defined with convolution operation \(*\) as}
\vspace{-4pt}
\begin{equation}
\vspace{-4pt}
    y = k*x,
\label{eq:blur_model}
\end{equation}
where \(k\) is a blur kernel, and \(y\) and \(x\) are a blurry image and a latent sharp image, respectively.
The spatial convolution can be transformed to an element-wise multiplication in the frequency domain as
\vspace{-4pt}
\begin{equation}
\vspace{-4pt}
    F(y) = F(k)\cdot F(x),
\label{eq:fourier_blur_model}
\end{equation}
where \(F(\cdot)\) denotes the discrete Fourier transform. Then, the latent sharp image can be derived with convolution operation as 
\vspace{-4pt}
\begin{equation}
\vspace{-4pt}
    x = F^{-1}(1/F(k))*y =k^{\dagger}*y,
\label{eq:spatial_deconv}
\end{equation}
where \(F^{-1}(\cdot)\) denotes the inverse discrete Fourier transform, and $k^{\dagger}$ is the spatial pseudo inverse kernel.

We observe that the shape of the corresponding inverse kernel \(k^{\dagger}\) remains the same when the spatial support size of a blur kernel \(k\) changes, \IE, 
\vspace{-4pt}
\begin{equation}
\vspace{-4pt}
(\frac{1}{s^{2}}k_{\uparrow{s}})^{\dagger}=\frac{1}{s^{2}}({k^{\dagger}}_{\!\uparrow{s}}),
\label{eq:sharing1}
\end{equation}
\change{where \(\uparrow\!\!{s}\) denotes general upsampling operation with a scale factor \(s\) in the spatial domain (refer to the supplementary material for more details on kernel upsampling).}
In \Eq{sharing1}, \(\frac{1}{s^{2}} k_{\uparrow{s}}\) is the kernel \(k'\) with the upsampled resolution, where \(\frac{1}{s^{2}}\) is used for normalizing the kernel weights.
Then, the inverse kernel of \(k'\) is the upsampled version of \(k^{\dagger}\) with the same scale factor \(s\).

Based on the observation, to handle defocus blur with varying sizes but with the same shape, we may use inverse kernels sharing a single shape but with different sizes.
However, as we aim to utilize a CNN for defocus deblurring, it is not practical to implement a network that carries the inverse kernels of all possible sizes.
To reduce the complexity,
we may approximate deconvolution of an image by combining the results of applying inverse kernels with a discrete set of sizes to the image.
Similarly to a classical approach~\cite{Bando2007refocus} that handles non-uniform blur using a linear combination of differently deblurred images,
spatially varying deconvolution for defocus blur can be approximated as
\vspace{-5pt}
\begin{equation}
\vspace{-5pt}
    x \approx
    \sum_{i}\{\alpha_{i}\cdot (\frac{1}{{s_i}^{2}}{k^{\dagger}}_{\!\uparrow{s_{i}}}*y)\},
    \label{eq:approx1}
\end{equation}
where \(s_{i}\in\{1,\cdots,n\}\) is an upsampling factor,
and \(\alpha_{i}\) is the per-pixel weight map for the result image obtained using the inverse kernel with an upsampling factor \(s_{i}\).

\begin{figure}[t]
\centering
\scriptsize
\setlength\tabcolsep{1 pt}
  \begin{tabular}{cccccccccc}
    \multicolumn{2}{c}{\includegraphics[width=0.18\linewidth]{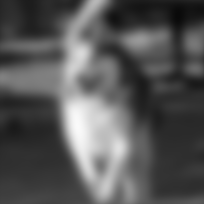}} & 
    \multicolumn{2}{c}{\includegraphics[width=0.18\linewidth]{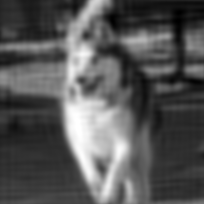}} & 
    \multicolumn{2}{c}{\includegraphics[width=0.18\linewidth]{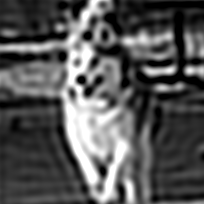}} & 
    \multicolumn{2}{c}{\includegraphics[width=0.18\linewidth]{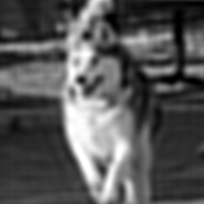}} & 
    \multicolumn{2}{c}{\includegraphics[width=0.18\linewidth]{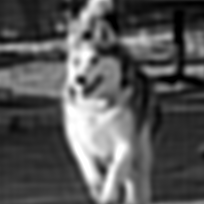}} \\

    \multicolumn{2}{c}{(a) blurred by }  & \multicolumn{2}{c}{(b) deblurred by}
     & \multicolumn{2}{c}{(c) deblurred by} & \multicolumn{2}{c}{(d) blended using} & \multicolumn{2}{c}{(e) deblurred by} \\
    \multicolumn{2}{c}{\tiny{\(\frac{1}{3.5^2}k_{\uparrow{3.5}}\)}} & \multicolumn{2}{c}{\tiny{\(\frac{1}{3.0^{2}}({k^{\dagger}}_{\!\!\uparrow{3.0}})\)}} & \multicolumn{2}{c}{\tiny{\(\frac{1}{4.0^{2}}({k^{\dagger}}_{\!\!\uparrow{4.0}})\)}} & \multicolumn{2}{c}{0.5*(b)+0.5*(c)} & \multicolumn{2}{c}{\tiny{\(\frac{1}{3.5^{2}}({k^{\dagger}}_{\!\!\uparrow{3.5}})\)}} \\
  \end{tabular}
  \vspace{0.01cm}
  \caption{Linear combination of deblurring results from different inverse kernels. 
  }
  \vspace{-12pt}
\label{fig:deblur_example}
\end{figure}

\vspace{-3pt}
\subsection{Experimental validation of the key idea}
\label{ssec:verification}
In this section, we experimentally validate the observed property of an inverse kernel for defocus deblurring (\Eq{sharing1}) and the approach for spatially varying deconvolution based on the property (\Eq{approx1}).
In the validation, we use Wiener deconvolution~\cite{Wiener} to compute a spatial inverse kernel and Lanczos upsampling~\cite{Duchon1979Lanczos} to scale inverse kernels.
We use Wiener inverse kernel as it has a finite spatial support analogous to the finite receptive field of a CNN, due to the involvement of signal-to-noise ratio (SNR) as regularization~\cite{NIPS2014inverse}.

Regarding \Eq{sharing1}, \Fig{kernel_example} shows an example that two kernels on both sides of \Eq{sharing1} produce equivalent deblurring results.
We also include the proof of \Eq{sharing1} in the supplementary material.
Regarding \Eq{approx1},
\Fig{deblur_example} shows that
a linear combination of resulting images
from the inverse kernels with a discrete set of sizes
can approximate deconvolution with an inverse kernel of a different size.
\Figs{deblur_example}b and \ref{fig:deblur_example}c are the deblurred results obtained by convolving inverse kernels of different sizes.
While neither kernel fits the actual blur size,
we can still obtain a visually decent result (\Fig{deblur_example}d) with an approximation accuracy\footnote{The accuracy is computed by 
$1\!-\!\mathit{MAE}(1, \hat{x}/{x_s})$,
where $\mathit{MAE}$ is the mean absolute error, $/$ is pixel-wise division, $x_s$ is the deconvolution result using an inverse kernel of a target scale (\textit{e.g.}, \Fig{deblur_example}e), and $\hat{x}$ is the approximated deconvolution result computed using \Eq{approx1} (\textit{e.g.}, \Fig{deblur_example}d).}
of $87.4\%$, by simply applying \Eq{approx1} with $\alpha_i\!=\!0.5$.
When we optimize $\alpha_i$ using the method of non-negative least squares, the accuracy further increases up to $92.5\%$.
These experiments confirm the validity of expanding \Eqs{sharing1} and (\ref{eq:approx1}) to a CNN architecture, where the remaining errors could be compensated through deep learning process (\Sec{network_design}).
Refer to the supplementary material for more examples.

\begin{figure}[t]
\begin{center}
\includegraphics [width=1.0\linewidth] {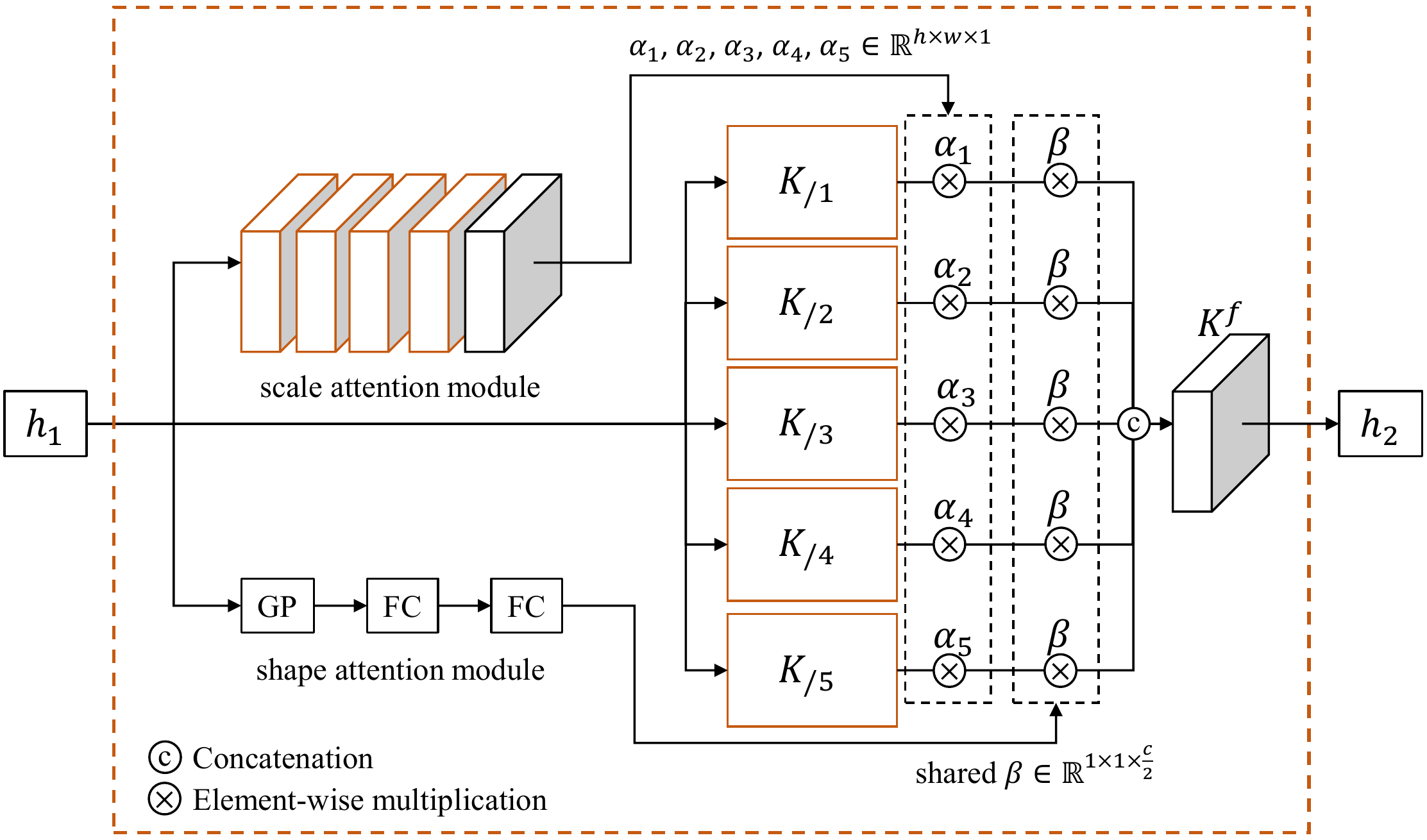}
\end{center}
\vspace{-12pt}
  \caption{Architecture of Kernel-sharing Parallel Atrous Convolutional (KPAC) block. 
  GP and FC denote global average pooling and a fully connected layer, respectively.
}
\label{fig:KPAC}
\vspace{-12pt}
\end{figure}

\section{Network Design}
\label{sec:network_design}

Based on the key idea, we design a network reflecting the property of inverse kernels for defocus deblurring.
Instead of directly adopting the linear combination in \Eq{approx1},
we extend the concept of combination
with a convolution layer that aggregates the results of multiple inverse kernels,
to fully exploit the non-linear nature of deep learning.

Let \(h_1\!\in\! \mathbb{R}^{h\times w\times c}\) and \(h_2\!\in\! \mathbb{R}^{h\times w\times c}\) be the input and output feature maps that correspond to images \(y\) and \(x\) in \Eq{spatial_deconv}, respectively.
We modify \Eq{approx1} to represent defocus deblurring 
using multiple convolution layers as 
\vspace{-4pt}
\begin{equation}
\vspace{-4pt}
    h_2  = K^{f}*\bigparallel_{i=1}^{n}\{\alpha_i\cdot (K_{s_i}*h_1)\},
    \label{eq:approx1_net}
\end{equation}
where \(K_{s_i}\!\in\! \mathbb{R}^{s_ik\times s_ik\times c \times \frac{c}{2}}\) are the convolutional kernels representing an inverse kernel with an upsampling factor \(s_i\).
\(K^f\!\in\! \mathbb{R}^{3\times 3\times \frac{nc}{2} \times c}\) is for non-linear aggregation of multiple scales of the outputs of \{\(K_{s_i}\)\}.
\(\alpha_i\!\in\! \mathbb{R}^{h\times w\times 1}\) is a per-pixel weight map for scale \(s_i\), and \(\bigparallel\) denotes concatenation operation.
Based on this modification, we propose a kernel-sharing parallel atrous convolutional (KPAC) block.

\subsection{Kernel-sharing Parallel Atrous Convolutional (KPAC) block }

Our KPAC block consists of multiple kernel-sharing atrous convolution layers, and modules for the scale and shape attentions (\Fig{KPAC}).

\vspace{-12pt}
\paragraph{Multiple kernel-sharing atrous convolutions}
In \Eq{approx1_net}, convolutional kernels \{\(K_{s_i}\)\} should represent inverse kernels of the same shape but with different sizes as observed in \Eq{sharing1}.
We may share the weights of \{\(K_{s_i}\)\} to enforce the constraint of the same inverse kernel shape.
However, in practice, the weight sharing is not straightforward because of the different sizes of \{\(K_{s_i}\)\}.

We resolve this problem with a simple but effective solution 
based on 
another important observation; For a blurred region, which is spatially smooth, filtering operations on sparsely sampled pixels (with a dilated kernel) and on densely sampled pixels (with a rescaled kernel) produce similar results.
Thus, the upsampling operation for \(\frac{1}{s^{2}}({k^{\dagger}}_{\!\uparrow{s}})\) in \Eq{sharing1} 
can be replaced
by a dilation operation, yielding an inverse kernel applied to sparsely sampled pixels.
The dilation operation does not change the number of filter weights 
but only scales the spatial support of the kernel without resampling, 
enabling direct weight sharing for the convolutional kernels \{\(K_{s_i}\)\}.

\begin{figure}[t]
\centering
\scriptsize
\setlength\tabcolsep{1 pt}
  \begin{tabular}{cccccc}
    \multicolumn{2}{c}{\includegraphics[width=0.325\linewidth]{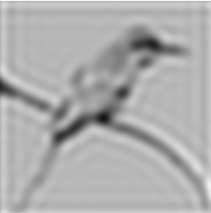}} &
    \multicolumn{2}{c}{\includegraphics[width=0.325\linewidth]{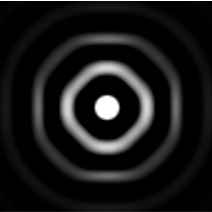}} & 
    \multicolumn{2}{c}{\includegraphics[width=0.325\linewidth]{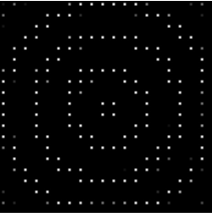}} \\
    
    \multicolumn{2}{c}{(a) deblurred by \({k^{\dagger}}_{\!\!\uparrow{\!/5}}\)}  & \multicolumn{2}{c}{(b) \(\frac{1}{5^{2}}({k^{\dagger}}_{\!\!\uparrow{5}})\) }
     & \multicolumn{2}{c}{(c) \({k^{\dagger}}_{\!\!\uparrow{\!/5}}\)} \\
  \end{tabular}
  \vspace{0.01cm}
  \caption{All settings are the same as in \Fig{kernel_example}. (a) deblurring result using a dilated inverse kernel. (a) is almost same as (b) and (c) in \Fig{kernel_example}, and the PSNR between (a) and \Fig{kernel_example}(c) is 53.74dB. (b) \& (c) shapes of the scaled and dilated inverse kernels.}
\label{fig:kernel_example2}
\vspace{-10pt}
\end{figure}

\Fig{kernel_example2} shows an example that \(\frac{1}{s^{2}}({k^{\dagger}}_{\!\uparrow{s}})\) and \({k^{\dagger}}_{\!\uparrow{/s}}\) produce almost equivalent deblurring results, where \({\uparrow\!\!{/s}}\) denotes the dilation operation.
We also experimentally verified that a modified version of \Eq{approx1} using \({k^{\dagger}}_{\!\uparrow{/s}}\) produces results almost equivalent to \Fig{deblur_example}. Refer to the supplementary material for the experiment on dilated inverse kernels.

Based on the observation,
our KPAC block includes multiple atrous convolution layers with different dilation rates, placed in parallel (\Fig{KPAC}).
The atrous convolution kernels \(K_{/{s_i}}\) are composed of the same number of kernel weights regardless of scales and share the same kernel weights, satisfying the constraint of shared inverse kernel shape.
That is,
we substitute the standard convolution in \Eq{approx1_net} by an atrous convolution layer, obtaining 
\vspace{-5pt}
\begin{equation}
\vspace{-5pt}
    h_2 = K^{f}*\bigparallel_{i=1}^{n}\{\alpha_i\cdot (K_{/{s_i}}*h_1)\},
    \label{eq:approx3_net}
\end{equation}
where \(K_{/{s_i}}\!\in\! \mathbb{R}^{k\times k\times c \times \frac{c}{2}}\) denotes the kernel with a dilation rate \(s_i\) in the kernel-sharing atrous convolution layer. 



\begin{figure*}[t]
\begin{center}
\includegraphics [width=0.93\linewidth] {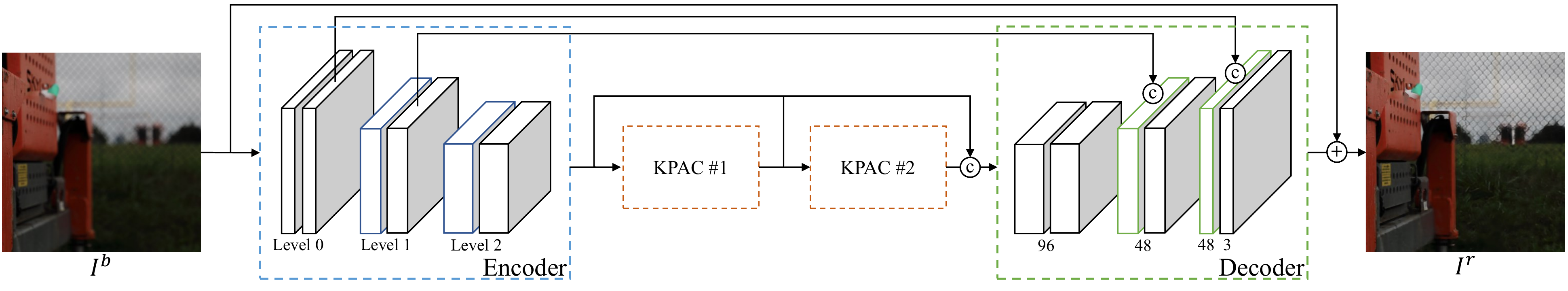}
\end{center}
  \vspace{-14pt}
  \caption{Overall network architecture. Refer to the supplementary material for the detailed architecture of our network.
}
\label{fig:overall}
\vspace{-12pt}
\end{figure*}

\vspace{-12pt}
\paragraph{Scale attention}
Our KPAC block represents an inverse kernel with an arbitrary scale by aggregating resulting feature maps from multiple scales of atrous convolution layers.
To dynamically determine pixel-wise weights for aggregation, we use a scale attention $\alpha_i\!\in\! \mathbb{R}^{h\times w\times 1}$ based on spatial attention mechanism~\cite{Kelvin2015Spatial}.
As the scale of defocus deblurring is determined by a combination of atrous convolution layers,
a different scale attention map should be applied to the resulting feature map of each atrous convolution layer.

\vspace{-15pt}
\paragraph{Shape attention}

Although we have assumed that inverse kernels share the same shape,
the shape of defocus blur and then the corresponding shape of the inverse kernel
can change in a defocused image due to the non-linearity in a camera imaging pipeline.
As an atrous convolution layer consists of multiple convolution filters, different combinations of the filters can represent different inverse kernels.
To support shape variations of inverse kernels,
we use a shape attention module based on channel attention mechanism~\cite{zhang2018rcan} that determines combination weights for filters on the resulting feature maps. 
As the atrous convolution layers in our KPAC block should share the inverse kernel shape, they share the channel-wise weight vector $\beta\!\in\! \mathbb{R}^{1\times 1\times \frac{c}{2}}$ from the shape attention module.


\begin{figure*}[t]
\begin{center}
\includegraphics [trim=0 0px 0 12px, clip, width=1.0\linewidth] {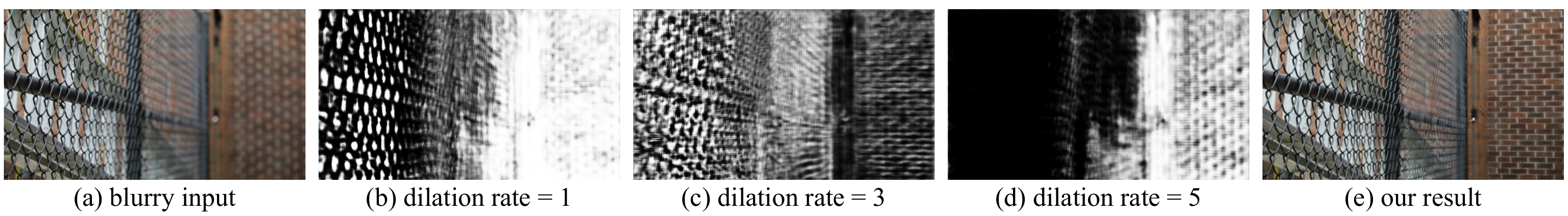}
\end{center}
\vspace{-14pt}
  \caption{Visualization of scale attention maps for different atrous convolution layers in the first block of our 2-block network. 
}
\label{fig:attention_visualization}
\vspace{-7pt}
\end{figure*}

\subsection{Defocus deblurring network}
\label{ssec:deblur_network}

For effective multi-scale processing in the feature space, we adopt an encoder-decoder structure~\cite{Ronneberger:2015:Unet} for our defocus deblurring network (\Fig{overall}).
The network consists of three parts: encoder, KPAC blocks, and decoder.
Except for the final convolution, all convolution layers include LeakyReLU~\cite{Maas:2013:Leaky} for the non-linear activation layer.

As our KPAC block is designed upon inverse kernels defined in the linear space, one question naturally follows whether it is proper for the KPAC block to run in the non-linear feature space.
We found that the KPAC block still works with non-linear features as CNNs are locally linear~\cite{Montufar2014linear,Lee2019locally}.
It has also been shown in a recent work~\cite{Dong2020Wiener} that Wiener deconvolution, which is a kind of inverse filters, can be successfully extended to the feature space
due to the piecewise linearity of the feature space of CNN.

The proposed KPAC block may not operate exactly the same as conventional inverse kernel-based approaches for deblurring, 
as it does not explicitly employ inverse kernels.
However, the architecture is still constrained with atrous convolutional layers with shared kernels and non-linear aggregation of resulting features, which are designed upon the property of inverse kernels.
As a result, the KPAC block would learn restoration kernels that are
more robust and effective for defocus deblurring.
In addition, while a single KPAC block is designed to model the entire process of deblurring, we can stack multiple KPAC blocks to exploit the iterative nature for removing residual blurs.

\vspace{-15pt}
\paragraph{Training}
For training the defocus deblurring network, we use the mean absolute error (MAE) between a network output and the corresponding ground-truth sharp image as a loss function.
We also employ the perceptual loss~\cite{Johnson2016Perceptual} for restoring more realistic textures.
For the perceptual loss, we use the feature map extracted at the `conv4\_4' layer in the pre-trained VGG-19 network~\cite{Simonyan15}.
When the perceptual loss is used, it is combined with the MAE loss, where the balancing factor is $7\times 10^{-4}$ for the perceptual loss.

\section{Experiments}
\label{ssec:experiments}

\begin{figure*}[tp]
\centering
\small
\setlength\tabcolsep{1 pt}
  \begin{tabular}{cccccccccccc}

    \multicolumn{2}{c}{\includegraphics[trim=0 0px 0 150px, clip, width=0.162\linewidth]{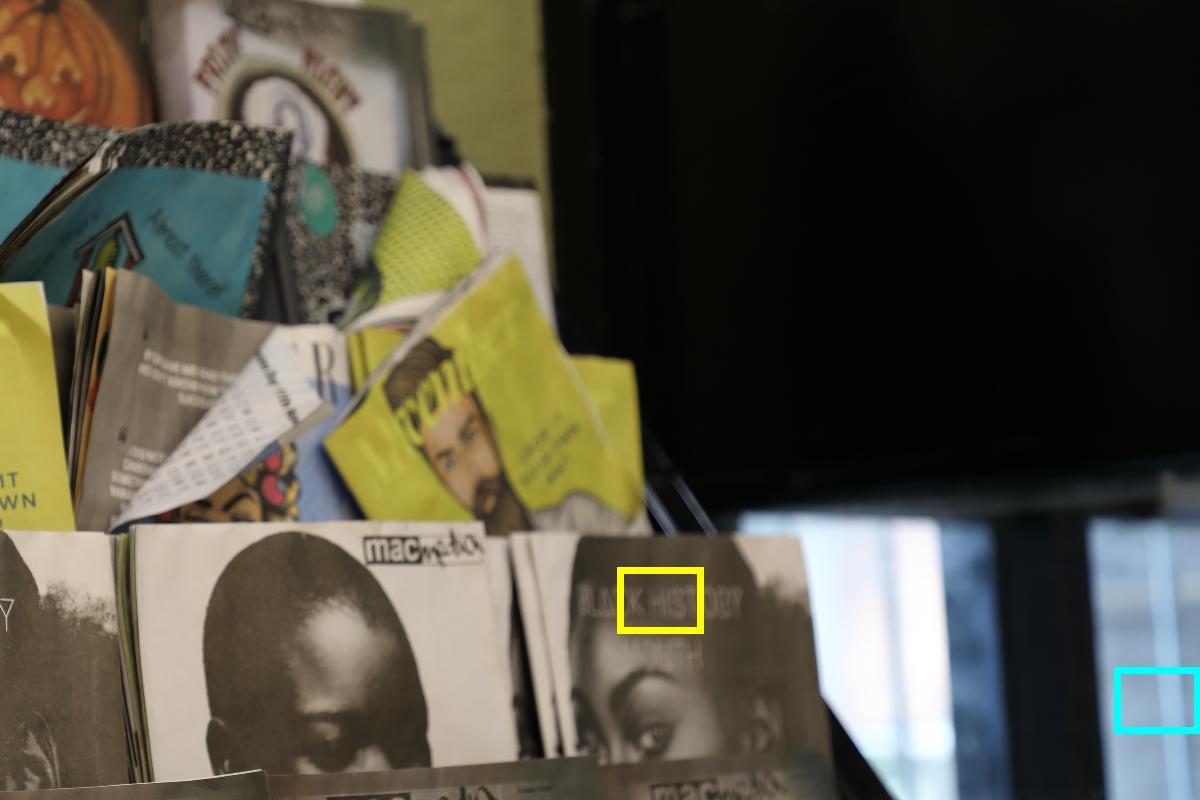}} &
    \multicolumn{2}{c}{\includegraphics[trim=0 0px 0 150px, clip, width=0.162\linewidth]{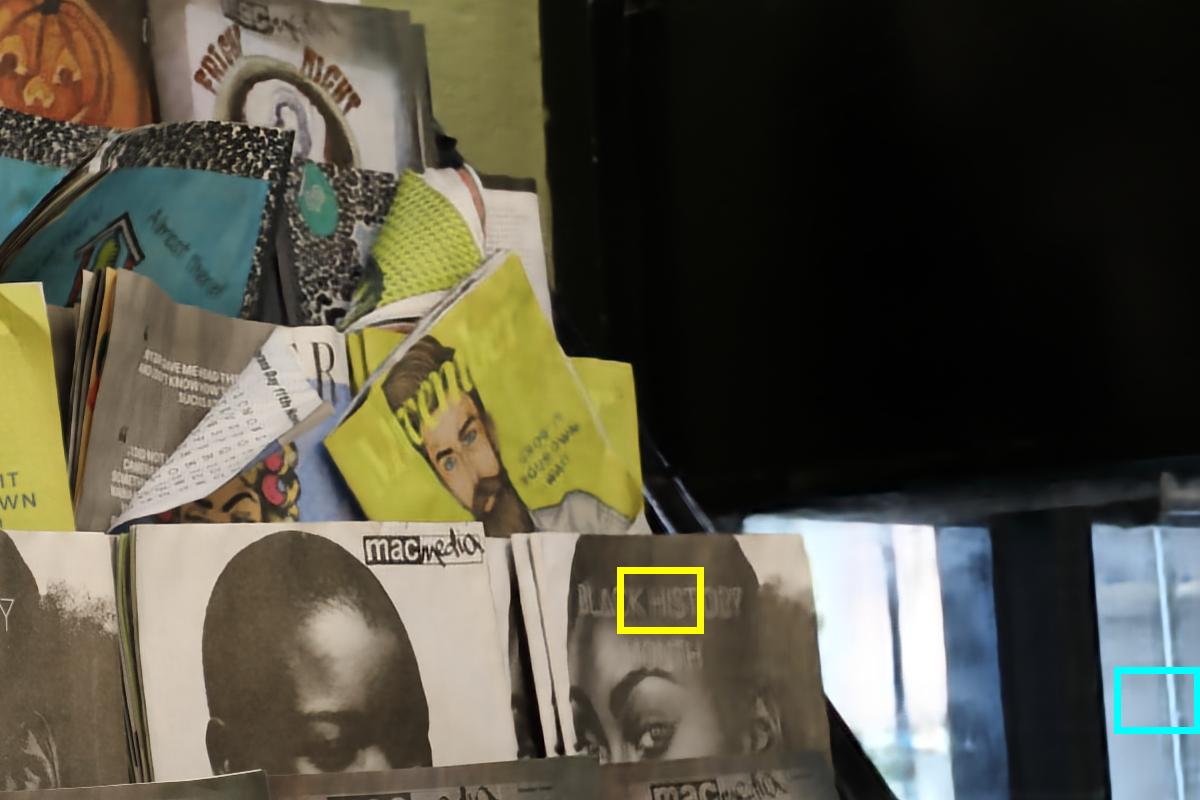}} &
    \multicolumn{2}{c}{\includegraphics[trim=0 0px 0 150px, clip, width=0.162\linewidth]{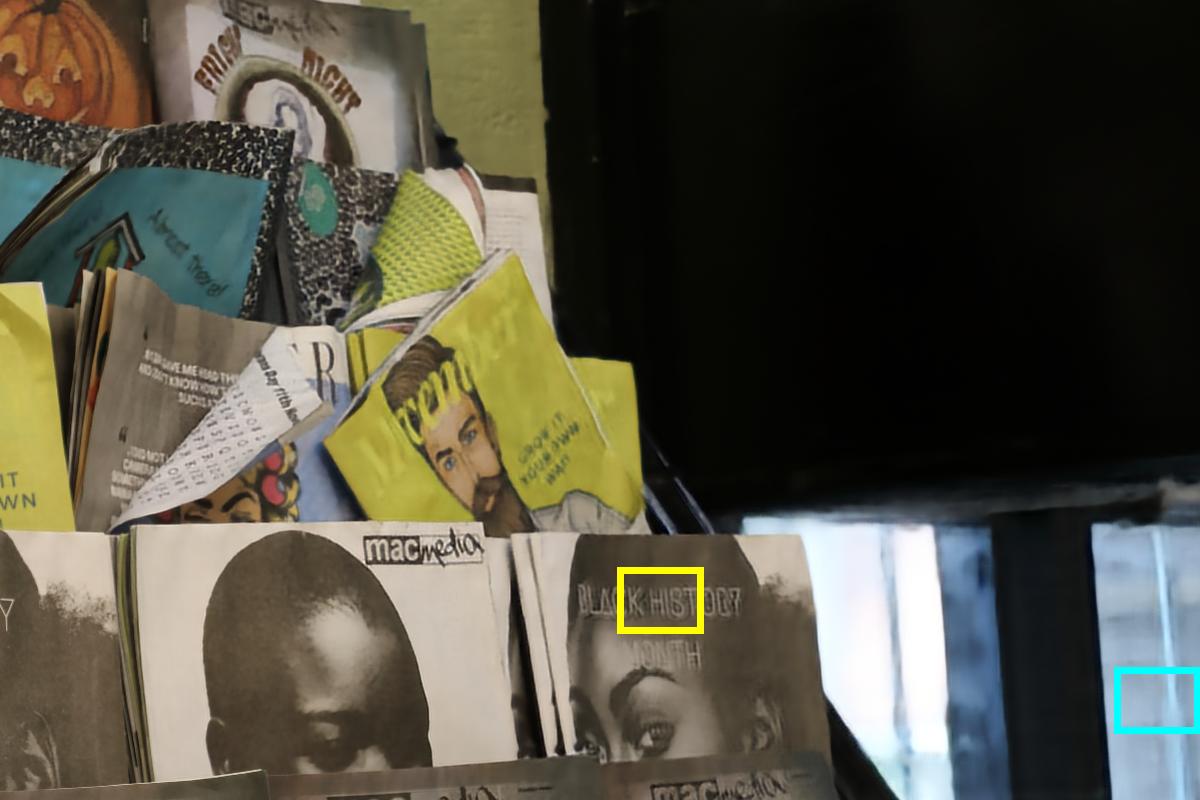}} &
    \multicolumn{2}{c}{\includegraphics[trim=0 0px 0 150px, clip, width=0.162\linewidth]{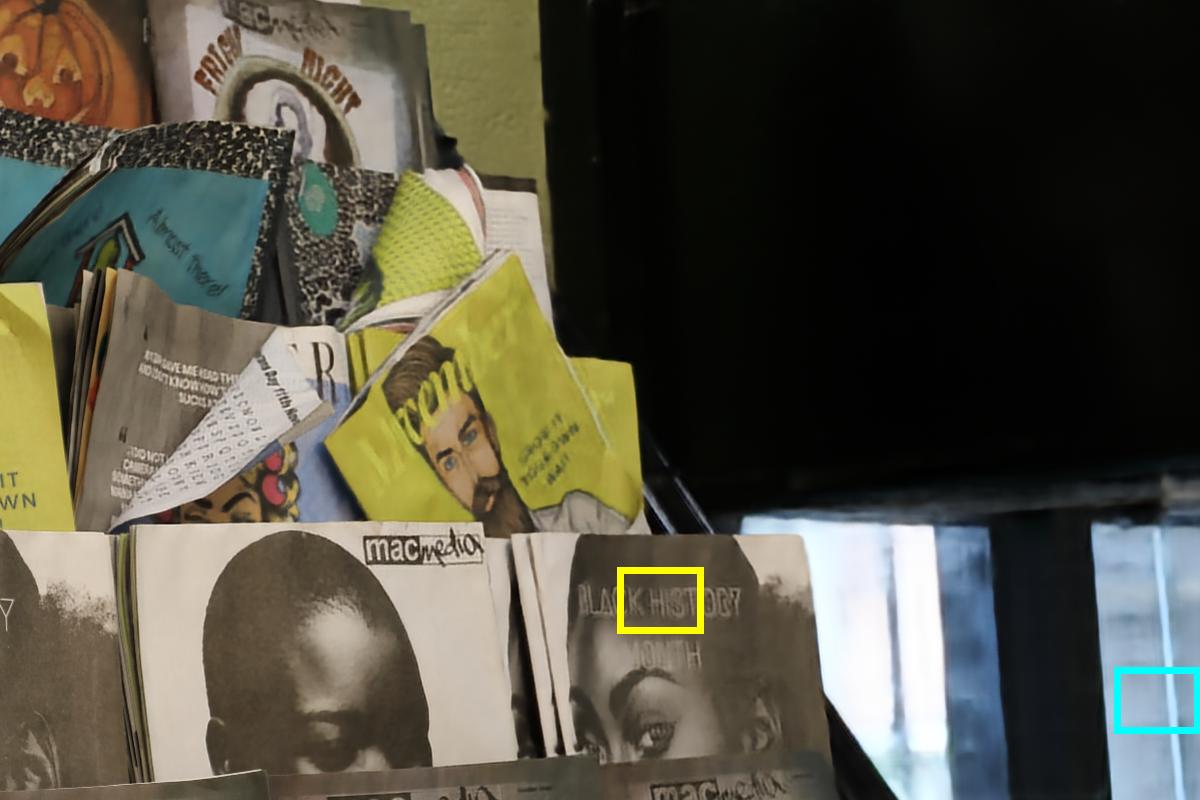}} &
    \multicolumn{2}{c}{\includegraphics[trim=0 0px 0 150px, clip, width=0.162\linewidth]{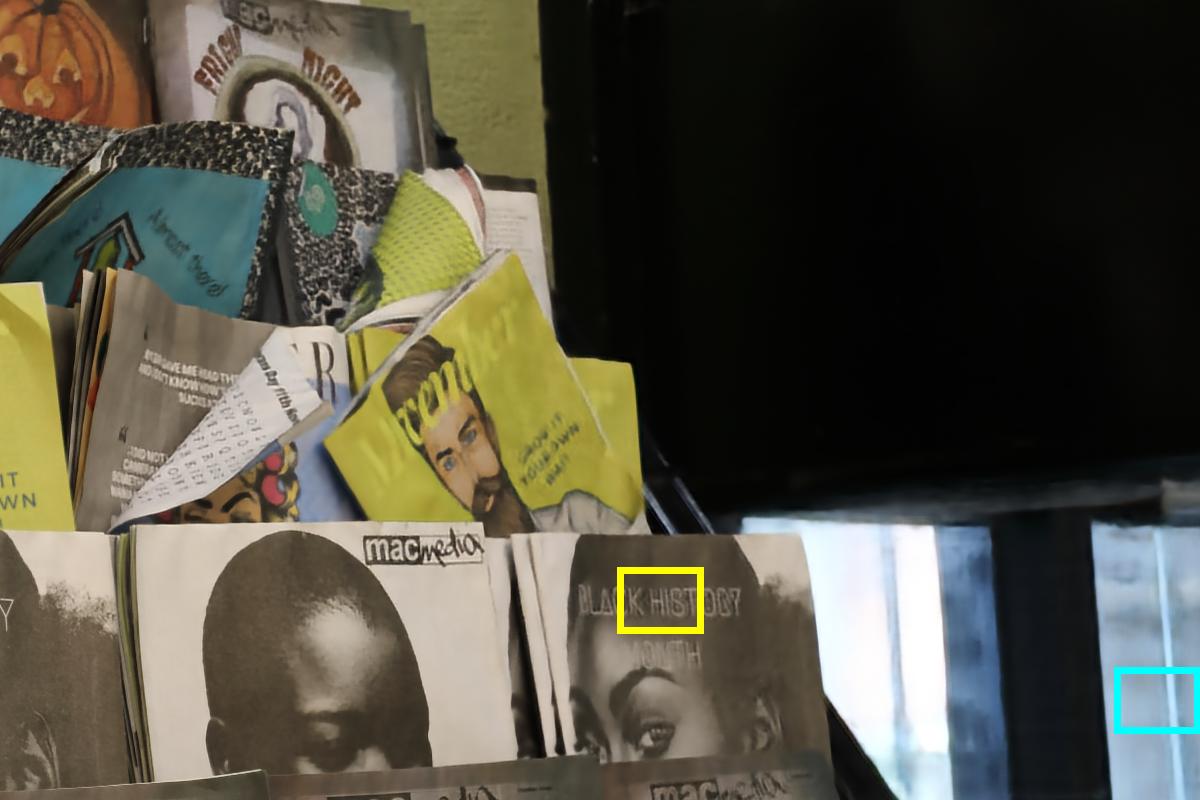}} &
    \multicolumn{2}{c}{\includegraphics[trim=0 0px 0 150px, clip, width=0.162\linewidth]{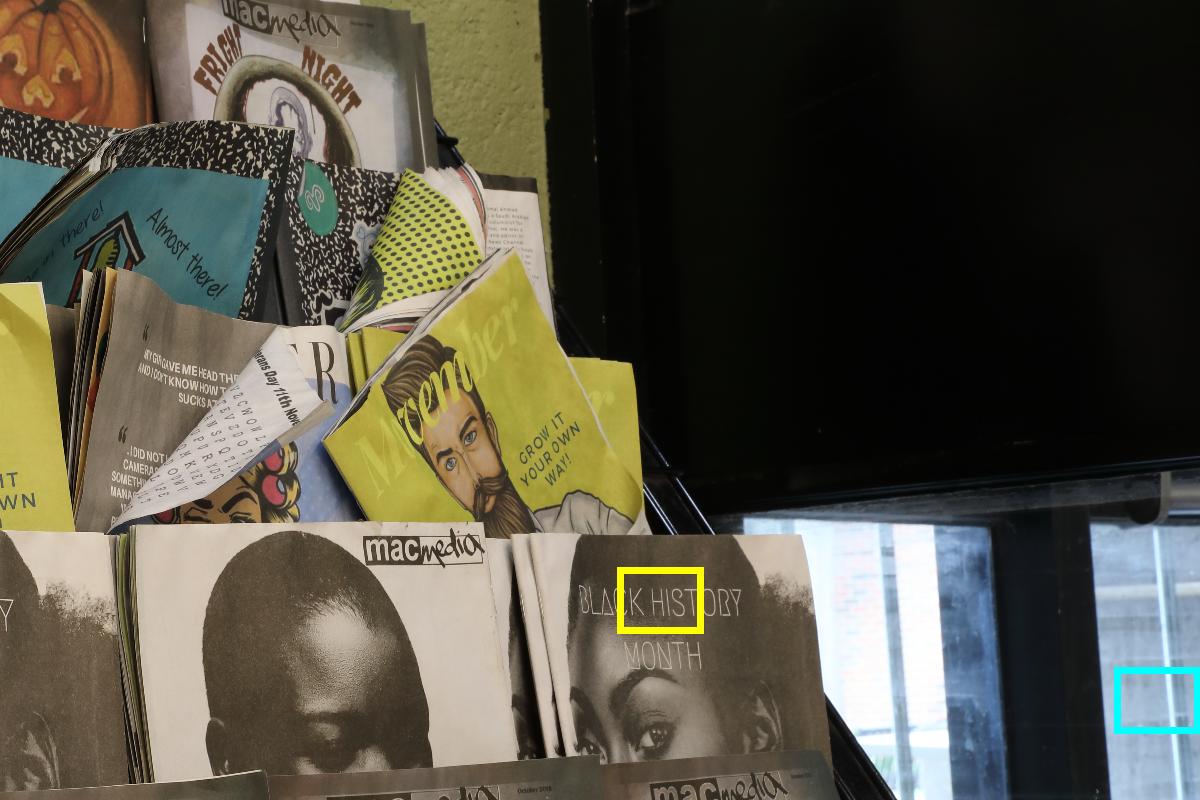}} \\ [-0.02in]

    \multicolumn{1}{c}{\includegraphics[width=0.08\linewidth]{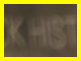}} &
    \multicolumn{1}{c}{\includegraphics[width=0.08\linewidth]{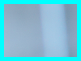}} &
    \multicolumn{1}{c}{\includegraphics[width=0.08\linewidth]{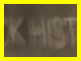}} &
    \multicolumn{1}{c}{\includegraphics[width=0.08\linewidth]{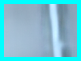}} &
    \multicolumn{1}{c}{\includegraphics[width=0.08\linewidth]{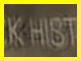}} &
    \multicolumn{1}{c}{\includegraphics[width=0.08\linewidth]{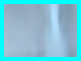}} &
    \multicolumn{1}{c}{\includegraphics[width=0.08\linewidth]{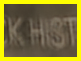}} &
    \multicolumn{1}{c}{\includegraphics[width=0.08\linewidth]{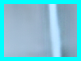}} &
    \multicolumn{1}{c}{\includegraphics[width=0.08\linewidth]{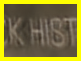}} &
    \multicolumn{1}{c}{\includegraphics[width=0.08\linewidth]{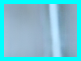}} &
    \multicolumn{1}{c}{\includegraphics[width=0.08\linewidth]{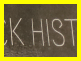}} &
    \multicolumn{1}{c}{\includegraphics[width=0.08\linewidth]{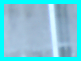}} \\

    \multicolumn{2}{c}{(a) Input}  & \multicolumn{2}{c}{(b) baseline }
     & \multicolumn{2}{c}{(c) no attention } & \multicolumn{2}{c}{(d) w/ scale } & \multicolumn{2}{c}{(e) w/ scale \& shape} & \multicolumn{2}{c}{(f) GT }\\

  \end{tabular}
  \vspace{-0.06cm}
  \caption{Qualitative examples of the ablation study. The baseline denotes a network model using conventional residual convolution blocks without our KPAC blocks. `scale' and `shape' denote scale attention and shape attention, respectively.}
\label{fig:ablation_example}
\vspace{-12pt}
\end{figure*}

We implemented and evaluated our models using Tensorflow 1.10.0 with NVIDIA Titan Xp GPU.
Our final model has two KPAC blocks with a kernel size~$k=5$ and the number of atrous convolution layers~$n=5$, as we empirically found it to work well in most cases.
We use negative slope coefficient \(\lambda=0.2\) for LeakyReLU layers.
We use the Adam optimizer~\cite{paszke2017automatic} with \(\beta_{1}=0.9\) and \(\beta_{2}=0.99\) to train our models.
We train our models for 200k iterations with the fixed learning rate of \(1\times10^{-4}\).
We tested a model trained for more iterations with learning rate decay, but its improvement was marginal in PSNR.
For evaluation in \SSec{comparison}, we train our models with the perceptual loss~\cite{Johnson2016Perceptual}.
For those models, we initialize them with pre-trained models trained with the MAE loss for 200k iterations.
Then, we fine-tune the networks with both MAE and perceptual losses for additional 100k iterations with the fixed learning rate of $5\times10^{-5}$.
We use the batch size of 4.
Each image in a batch is randomly cropped to $512 \times 512$.

\vspace{-15pt}
\paragraph{Dataset}
We use the DPDD dataset~\cite{Abuolaim:2020:DPDNet} for evaluation of our models.
The dataset provides 500 image pairs of a real-world defocused image and the corresponding all-in-focus ground-truth image captured by a Canon EOS 5D Mark IV.
The dataset consists of training, validation, and testing sets of 350, 74, and 76 pairs of images, respectively.
In our experiments, we train and evaluate our models using the training and testing sets, respectively.
While the dataset also provides dual-pixel data, we do not use them in our experiments.
The dataset provides 16-bit images in the PNG format. We convert them to 8-bit images for our experiments.


\begin{table}[t]
\centering
\scalebox{0.9}{
\begin{tabular}{ c  c c }
\toprule
 & PSNR (dB) & Parameters (M)\\
 \cmidrule(r){1-1} \cmidrule(l){2-3}
 w/o weight sharing & 24.78 & 2.50 \\
 w/ weight sharing &  25.21 & 1.58 \\
\bottomrule

\end{tabular}
}
\vspace{0.05cm}
\caption{Effect of weight sharing. The weight sharing improves the deblurring quality while reducing the number of parameters.}
\label{tbl:weight_sharing}
\vspace{-12pt}
\end{table}

\subsection{Analysis}

\paragraph{Effect of weight sharing}
Our KPAC blocks learn spatially varying inverse kernels whose shapes remain the same, but their sizes vary.
For effective learning of such inverse kernels, our network shares the convolution weights across multiple atrous convolution layers.
In this experiment, we verify the effect of the weight sharing between atrous convolution layers by comparing the performance of models with and without the weight sharing.
Both models have two KPAC blocks with $5\times 5$ kernels.
\Tbl{weight_sharing} shows the deblurring quality and the number of parameters of each model.
As shown in the table, our model with the weight sharing not only reduces the number of learning parameters, but also improves the deblurring quality, as its weight sharing structure properly constrains and guides the learning process.


\vspace{-14pt}
\paragraph{Scale attention}
The atrous convolution layers in our KPAC block simulate inverse kernels of different sizes to effectively handle the spatially varying nature of defocus blur.
To analyze how they are activated for defocus blur with different sizes, we visualize the scale attention maps of different atrous convolution layers (\Fig{attention_visualization}).
The roles of different attention maps may not be strictly distinguished because of the nature of the learning process that implicitly learns the use of different layers.
Nevertheless, we can observe a clear tendency that the attention maps of different dilation rates are activated for different blur sizes.
For example, the attention map of the dilation rate 1 is activated for pixels with blur of almost any size.
On the other hand, the attention map of the dilation rate 5 is activated only for pixels with large blur.
This shows that our scheme properly works for handling spatially varying size of defocus blur.
\begin{table}[t]
\centering
\small
\scalebox{0.9}{
\begin{tabular}{ c c c c c c }
\toprule
\multicolumn{3}{c}{KPAC components} & \multirow{2}{*}{Baseline} & \multirow{2}{*}{PSNR (dB)} & \multirow{2}{*}{Parameters (M)} \\
\cmidrule(r){1-3} 
ACs & Scale & Shape & &  & \\
\toprule
\multicolumn{3}{c}{} & \(\checkmark\) & 24.59 & 2.26\\ 
\cmidrule(lr){1-6} 
\(\checkmark\) & & & & 24.74 & 1.33 \\
\(\checkmark\) & & \(\checkmark\) & & 24.98 & 1.33 \\
\(\checkmark\) & \(\checkmark\) & & & 25.03 & 1.58 \\
\(\checkmark\) & \(\checkmark\) & \(\checkmark\) & & 25.21 & 1.58\\

\bottomrule

\end{tabular}
}
\vspace{0.05cm}
\caption{Ablation study. ACs: atrous convolution layers. Scale: scale attention. Shape: shape attention.}
\label{tbl:ablation}
\vspace{-12pt}
\end{table}



\vspace{-14pt}
\paragraph{Ablation study}
To quantitatively analyze the effect of each component in our KPAC block, we conduct an ablation study (\Tbl{ablation}).
We first prepare a baseline model, which uses na\"ive convolution blocks instead of our KPAC blocks.
For the baseline model, we use a conventional residual block that consists of two convolution layers with the filter size of \(3\times 3\).
For a fair comparison, the baseline model includes multiple convolution blocks so that its model size is similar to our model without the weight sharing.
We also prepare four variants of the baseline model using two KPAC blocks with the kernel size of \(5\times5\), then measure the deblurring performances of the models.
\Tbl{ablation} summarizes the ablation study result.
As shown in the table, every component of our proposed approach increases the deblurring quality significantly.
\Fig{ablation_example} presents a qualitative comparison, which shows that both scale attention and shape attention help our network better handle spatially varying blur and restore fine structures while the models with no attentions suffer from spatially varying blurs.



\begin{figure}[t]
\begin{center}
\includegraphics[width=1.0\linewidth] {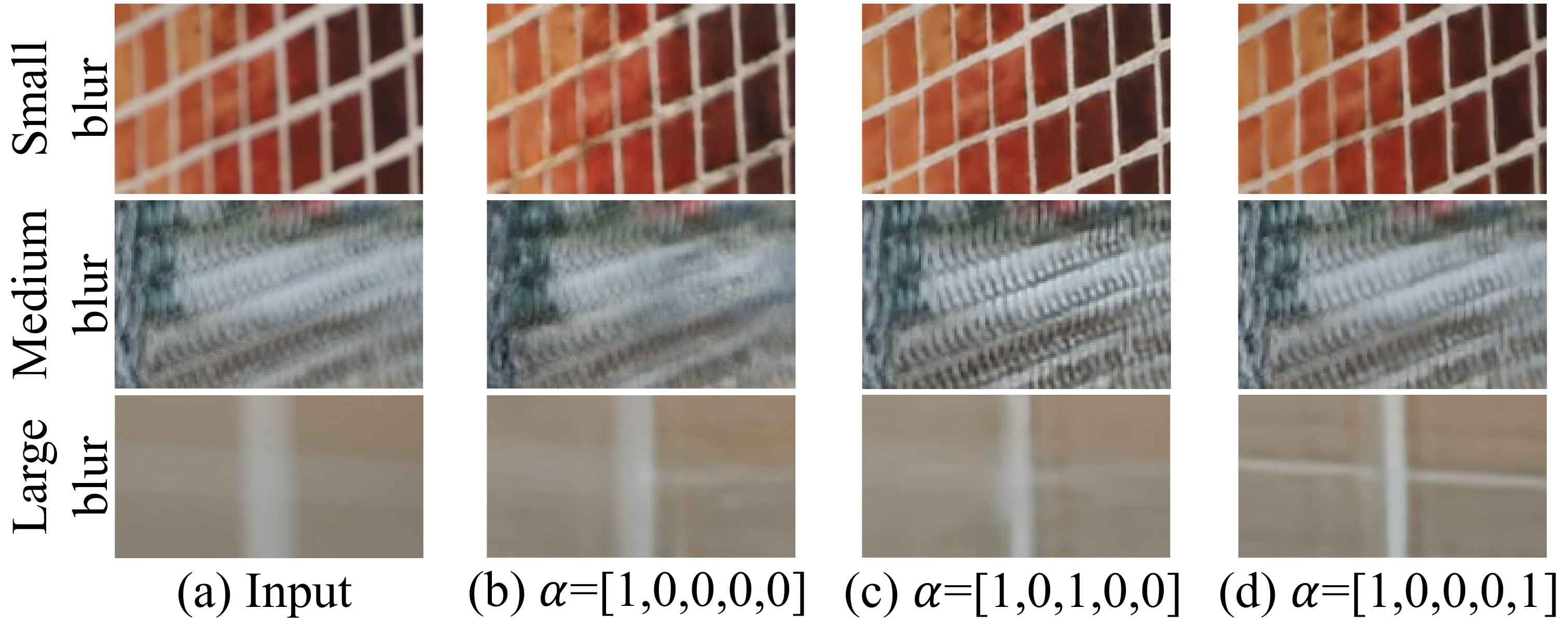}
\end{center}
\vspace{-12pt}
  \caption{Deblurring results with modulated scale attention maps $[\alpha_1, \alpha_2, \alpha_3, \alpha_4, \alpha_5]$.}
\label{fig:alpha_ctrl}
\vspace{-7pt}
\end{figure}

\vspace{-12pt}
\paragraph{Atrous convolutions with different dilation rates}
We analyze the effect of dilation rates of atrous convolution layers in handling defocus blur with varying scales.
In the test, we manually modulated the attention weight maps [\(\alpha_{1\sim5}\)] to make our pretrained network use only the feature maps produced by atrous convolution layers 
of specified scales. 
With the similar tendency to \Fig{attention_visualization}, we observed that atrous convolutions with dilation rates of 1, 3 and 5 help remove blur of any, medium, and large sizes, respectively (\Fig{alpha_ctrl}).

\begin{table}[t]
\centering
\small
\begin{tabular}{ c c c c c c}
\toprule
 &Blurry&\multicolumn{4}{c}{Number of KPAC blocks}\\
   & input & 1 & 2 & 3 & 4\\
 \cmidrule(r){1-1} \cmidrule(lr){2-2} \cmidrule(l){3-6}
 PSNR (dB) & 23.92  & 24.82 &  25.21 & 25.25 & 25.14\\
 SSIM (dB) & 0.812 & 0.836 & 0.842 & 0.842 & 0.841\\
 Params (M) & - &1.05 & 1.58 & 2.11 & 2.64 \\
\bottomrule

\end{tabular}
\caption{Comparison on the effect of different numbers of KPAC blocks with $5\times 5$ kernels.}
\label{tbl:num_block}
\vspace{-12pt}
\end{table}

\vspace{-12pt}
\paragraph{Number of KPAC blocks}
Our KPAC block can be stacked together so that the network can iteratively remove defocus blur to achieve higher-quality deblurring results.
We investigate the performance of different numbers of KPAC blocks.
\Tbl{num_block} shows that even a single KPAC block can effectively remove defocus blur and increase the PNSR by 0.90 dB.
As we adopt more KPAC blocks, the PSNR increases although the improvement becomes smaller.
After three KPAC blocks, the PSNR starts to decrease possibly due to the increased training complexity.
Based on this experiment, we design our final model to have two KPAC blocks as two blocks provide relatively high deblurring quality with small model size.

\subsection{Evaluation}
\label{ssec:comparison}
\paragraph{Deblurring quality comparison with state-of-the-art}
We compare our method with state-of-the-art defocus deblurring methods, including both conventional two-step approaches \cite{Shi:2015:JNB,Karaali:2018:DMEAdaptive,Lee:2019:DMENet} and the recent end-to-end deep learning-based approach~\cite{Abuolaim:2020:DPDNet}. 
For all the methods, we produced result images using the source code provided by the authors.
For JNB~\cite{Shi:2015:JNB}, EBDB~\cite{Karaali:2018:DMEAdaptive} and DMENet~\cite{Lee:2019:DMENet}, we used the non-blind deconvolution method~\cite{Krishnan:2008:deconvolution}
for generating deblurred images using estimated defocus maps.
For DPDNet~\cite{Abuolaim:2020:DPDNet}, we used the source code and pre-trained models provided by the author.
DPDNet provides two versions of models, each of which takes a single input image and dual-pixel data, which is a pair of sub-aperture images, respectively.
We include both of them in our comparison.
For evaluation, we measure PSNR and SSIM~\cite{wang2004ssim}.
We also measure LPIPS~\cite{zhang2018perceptual} for evaluating the perceptual quality as done in \cite{Abuolaim:2020:DPDNet}.

We include two variants of our model, each of which has a different number of encoding levels, or a different number of downsampling layers in the encoder.
By increasing the encoding levels, we can more easily handle large blur with small filters and with a small amount of computations.
On the other hand, with fewer encoding levels, it is easier to restore fine-scale details.
To inspect the difference between models with different numbers of encoding levels,
we include two variants of our model, which have two and three levels, respectively.
Both models are trained with both MAE and perceptual loss functions.

\begin{figure}[tp]
\centering
\small
\setlength\tabcolsep{1 pt}
  \begin{tabular}{cccccccc}
    \multicolumn{2}{c}{\includegraphics[width=0.243\linewidth]{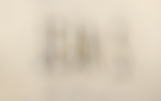}} & 
    \multicolumn{2}{c}{\includegraphics[width=0.243\linewidth]{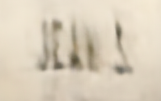}} & 
    \multicolumn{2}{c}{\includegraphics[width=0.243\linewidth]{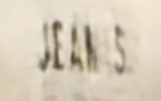}} &
    \multicolumn{2}{c}{\includegraphics[width=0.243\linewidth]{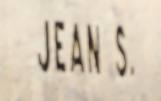}} \\

    \multicolumn{2}{c}{(a) Input}  & \multicolumn{2}{c}{(b) 2-level}
     & \multicolumn{2}{c}{(c) 3-level} & \multicolumn{2}{c}{(d) GT} \\

  \end{tabular}
  \vspace{-0.11cm}
  \caption{Visual comparison between our 2- and 3-level models.}
\label{fig:encoding_level}
\vspace{-7pt}
\end{figure}

\begin{figure*}[tp]
\centering
\small
\setlength\tabcolsep{1 pt}
  \begin{tabular}{cccccccccccccc}
    \multicolumn{3}{c}{\includegraphics[trim=0 60px 0 0px, clip, width=0.244\linewidth]{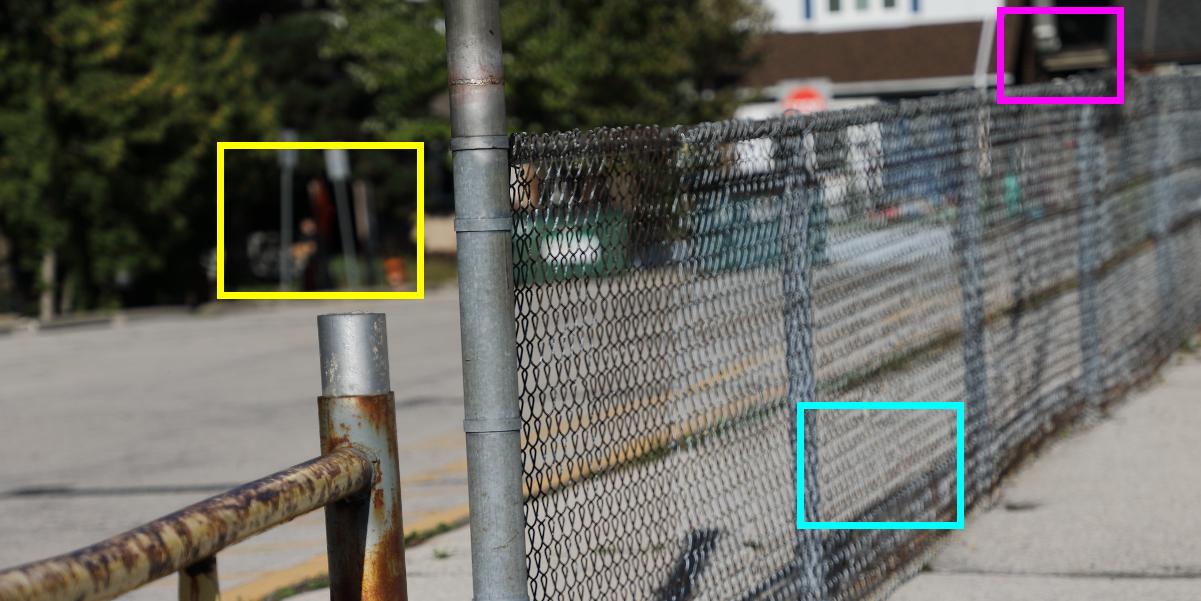}} &
    \multicolumn{3}{c}{\includegraphics[trim=0 60px 0 0px, clip, width=0.244\linewidth]{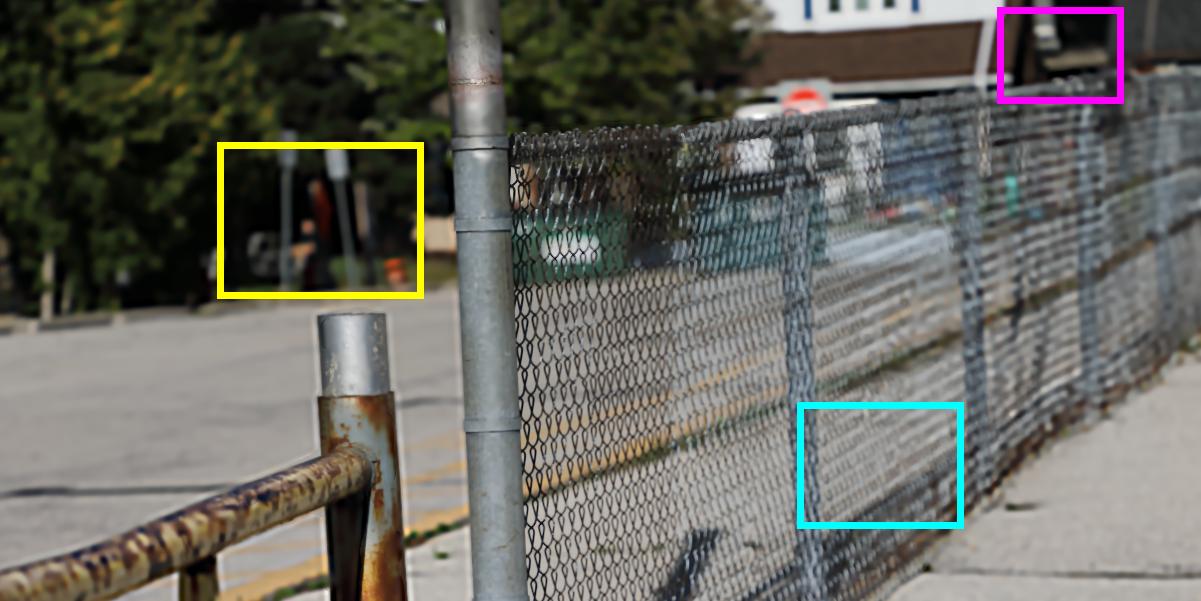}} &
    \multicolumn{3}{c}{\includegraphics[trim=0 60px 0 0px, clip, width=0.244\linewidth]{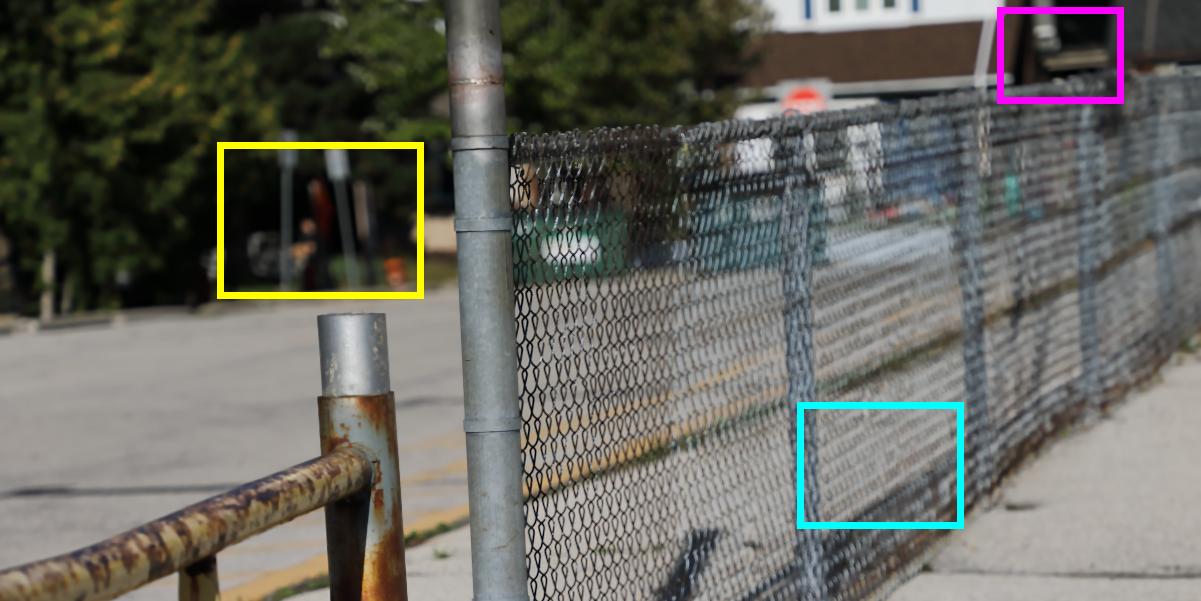}} &
    \multicolumn{3}{c}{\includegraphics[trim=0 60px 0 0px, clip, width=0.244\linewidth]{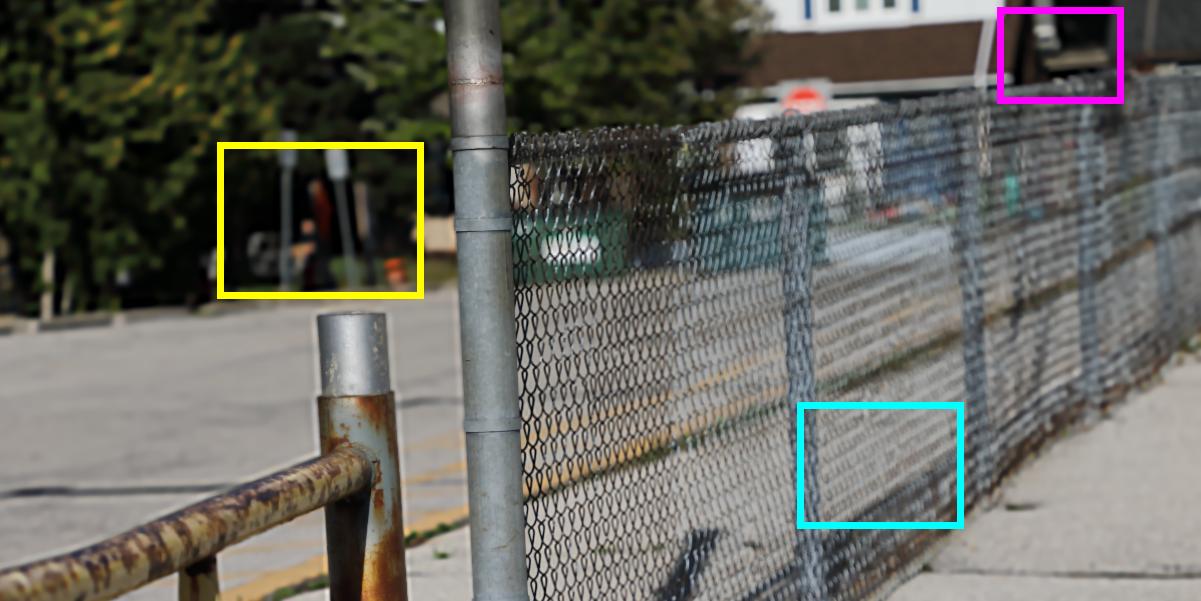}} \\ [-0.02in]
    \multicolumn{1}{c}{\includegraphics[width=0.08\linewidth]{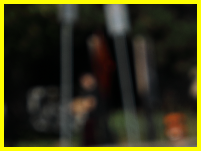}} &
    \multicolumn{1}{c}{\includegraphics[width=0.08\linewidth]{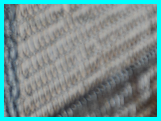}} &
    \multicolumn{1}{c}{\includegraphics[width=0.08\linewidth]{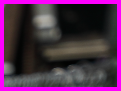}} &
    \multicolumn{1}{c}{\includegraphics[width=0.08\linewidth]{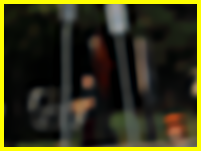}} &
    \multicolumn{1}{c}{\includegraphics[width=0.08\linewidth]{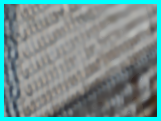}} &
    \multicolumn{1}{c}{\includegraphics[width=0.08\linewidth]{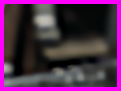}} &
    \multicolumn{1}{c}{\includegraphics[width=0.08\linewidth]{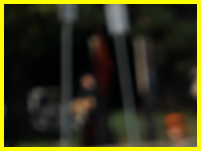}} &
    \multicolumn{1}{c}{\includegraphics[width=0.08\linewidth]{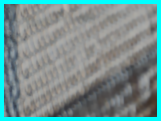}} &
    \multicolumn{1}{c}{\includegraphics[width=0.08\linewidth]{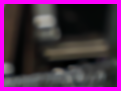}} &
    \multicolumn{1}{c}{\includegraphics[width=0.08\linewidth]{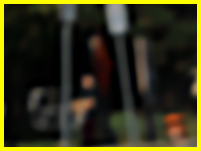}} &
    \multicolumn{1}{c}{\includegraphics[width=0.08\linewidth]{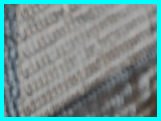}} &
    \multicolumn{1}{c}{\includegraphics[width=0.08\linewidth]{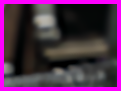}} \\ 
    \multicolumn{3}{c}{(a) Input}  & \multicolumn{3}{c}{(b) JNB~\cite{Shi:2015:JNB}}
     & \multicolumn{3}{c}{(c) EBDB~\cite{Karaali:2018:DMEAdaptive} } & \multicolumn{3}{c}{(d) DMENet~\cite{Lee:2019:DMENet} } \\ [0.03in]

    \multicolumn{3}{c}{\includegraphics[trim=0 60px 0 0px, clip, width=0.244\linewidth]{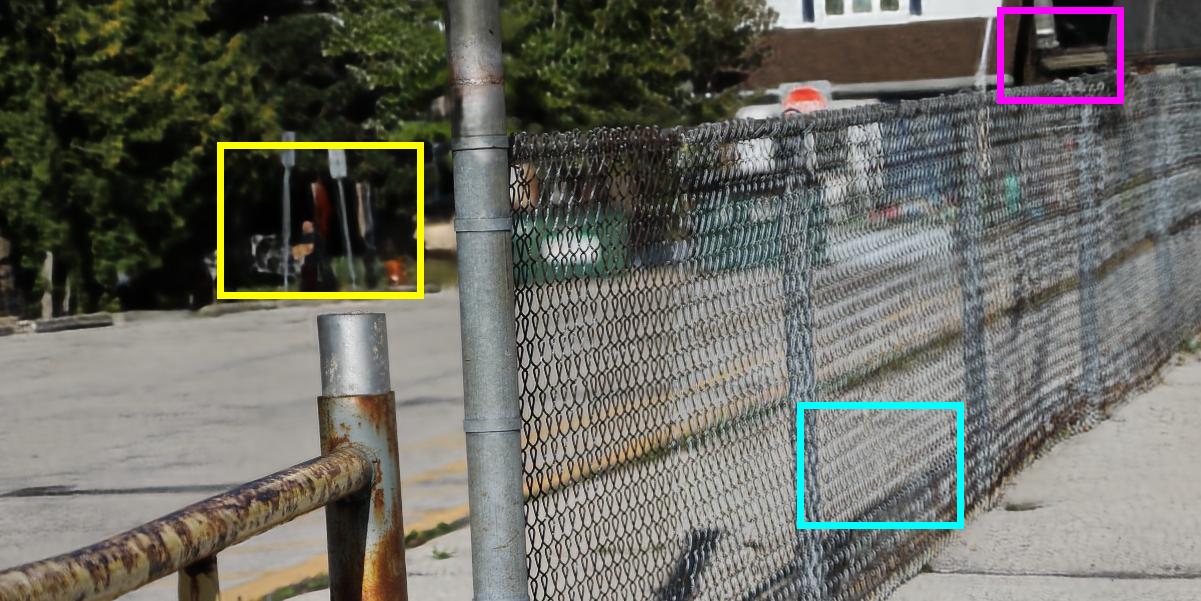}} &
    \multicolumn{3}{c}{\includegraphics[trim=0 60px 0 0px, clip, width=0.244\linewidth]{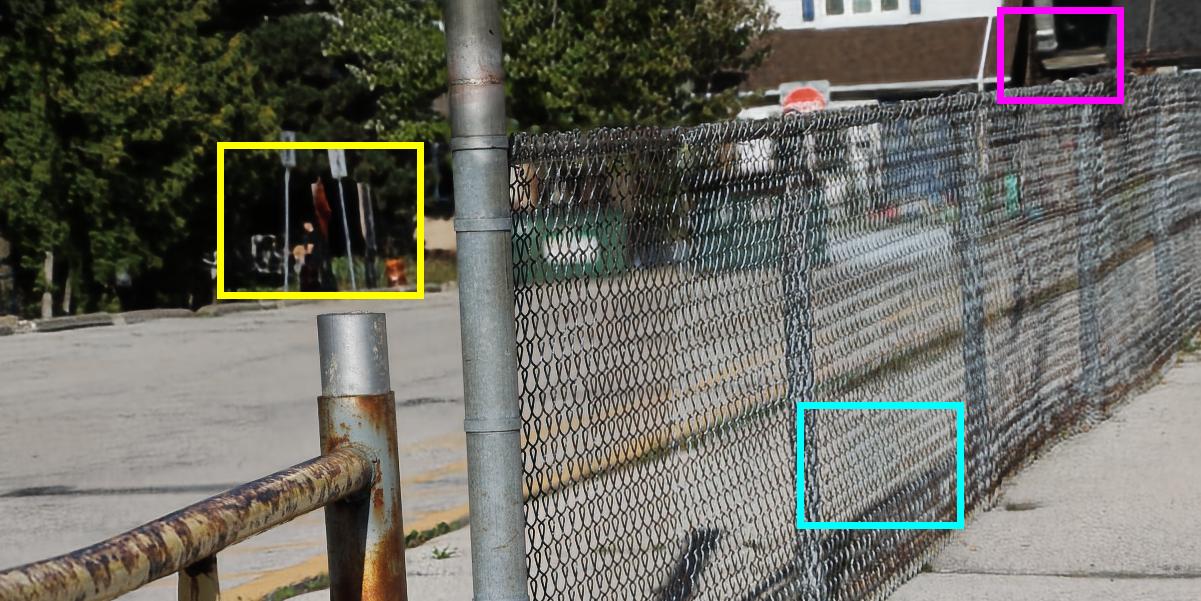}} &
    \multicolumn{3}{c}{\includegraphics[trim=0 60px 0 0px, clip, width=0.244\linewidth]{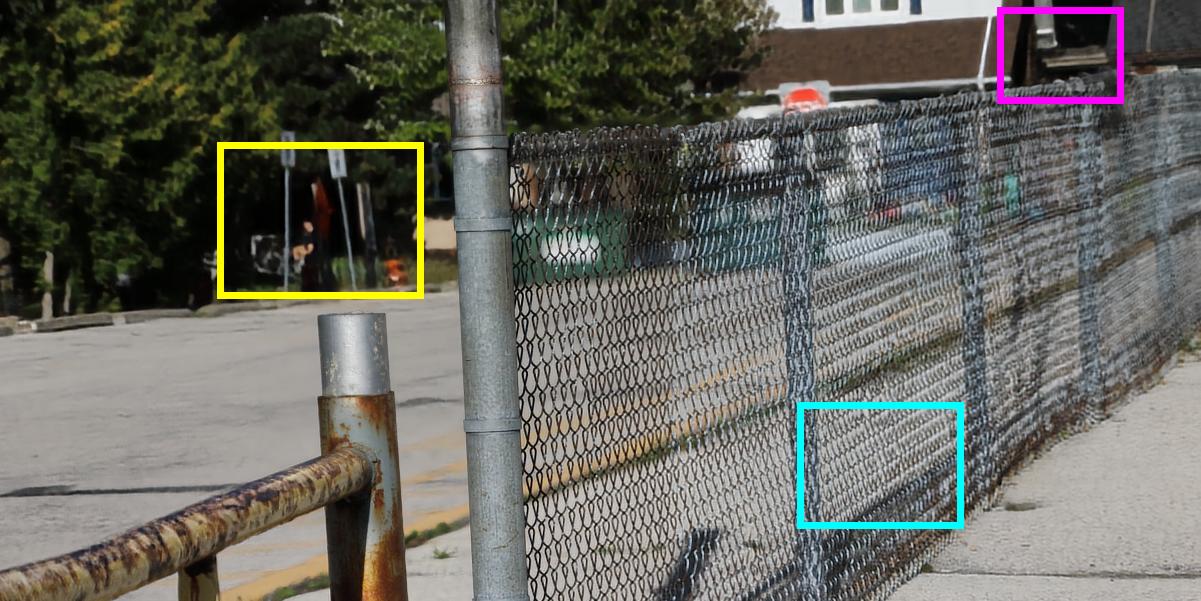}} &
    \multicolumn{3}{c}{\includegraphics[trim=0 60px 0 0px, clip, width=0.244\linewidth]{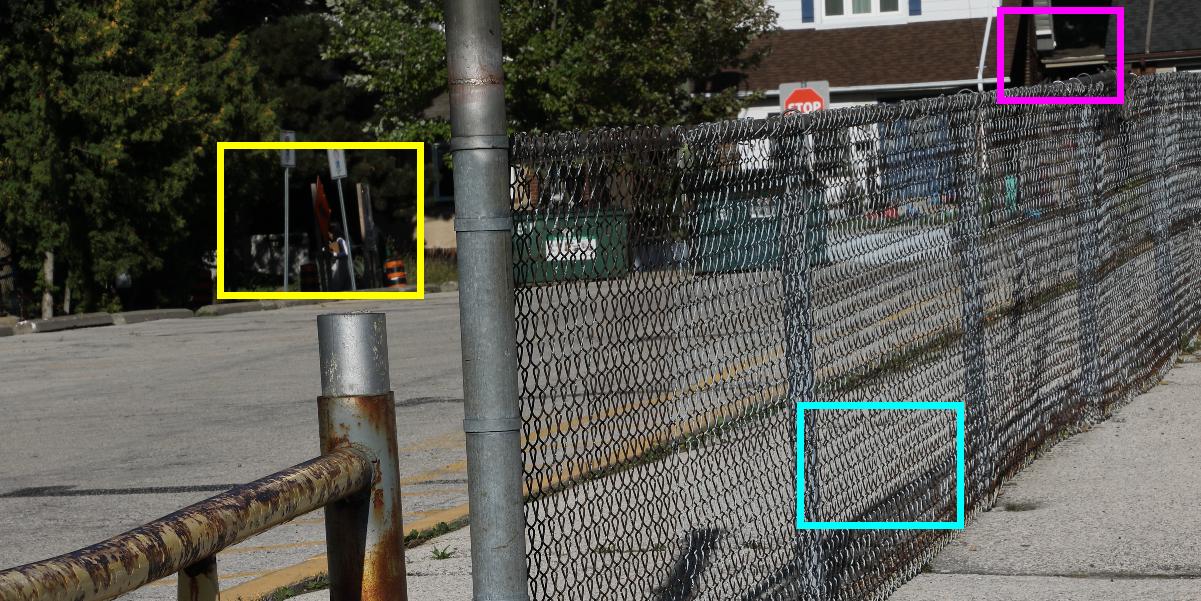}} \\ [-0.02in]
    \multicolumn{1}{c}{\includegraphics[width=0.08\linewidth]{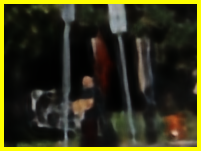}} &
    \multicolumn{1}{c}{\includegraphics[width=0.08\linewidth]{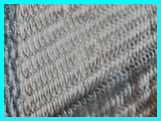}} &
    \multicolumn{1}{c}{\includegraphics[width=0.08\linewidth]{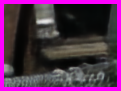}} &
    \multicolumn{1}{c}{\includegraphics[width=0.08\linewidth]{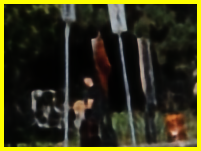}} &
    \multicolumn{1}{c}{\includegraphics[width=0.08\linewidth]{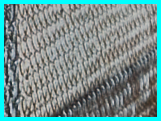}} &
    \multicolumn{1}{c}{\includegraphics[width=0.08\linewidth]{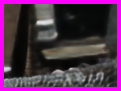}} &
    \multicolumn{1}{c}{\includegraphics[width=0.08\linewidth]{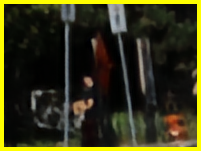}} &
    \multicolumn{1}{c}{\includegraphics[width=0.08\linewidth]{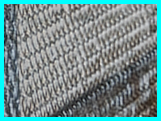}} &
    \multicolumn{1}{c}{\includegraphics[width=0.08\linewidth]{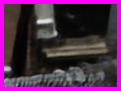}} &
    \multicolumn{1}{c}{\includegraphics[width=0.08\linewidth]{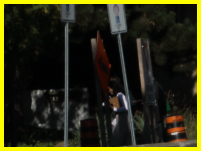}} &
    \multicolumn{1}{c}{\includegraphics[width=0.08\linewidth]{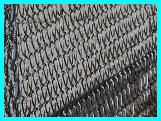}} &
    \multicolumn{1}{c}{\includegraphics[width=0.08\linewidth]{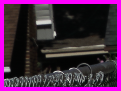}} \\

    \multicolumn{3}{c}{(e) DPDNet (single)~\cite{Abuolaim:2020:DPDNet}}  & \multicolumn{3}{c}{(f) DPDNet (dual)~\cite{Abuolaim:2020:DPDNet}}
     & \multicolumn{3}{c}{(g) Ours } & \multicolumn{3}{c}{(h) GT }  \\

  \end{tabular}
  \vspace{-0.08cm}
  \caption{Qualitative comparison on the test set of the DPDD dataset \cite{Abuolaim:2020:DPDNet}. Note that DPDNet (dual) uses dual-pixel stereo images.}
\label{fig:comparison}
\vspace{-12pt}
\end{figure*}
\begin{table}[t]
\centering
\scalebox{0.78}{
\begin{tabular}{ c  c c c c }
\toprule
Model & PSNR\(\uparrow\) & SSIM\(\uparrow\) & LPIPS\(\downarrow\) & Parameters (M)\\
\cmidrule(r){1-1} \cmidrule(l){2-5}
JNB~\cite{Shi:2015:JNB}  & 23.70 & 0.799 & 0.442 & -\\
EBDB~\cite{Karaali:2018:DMEAdaptive} & 23.96& 0.819 & 0.402 & -\\
DMENet~\cite{Lee:2019:DMENet} & 23.92 & 0.808 & 0.410 & 26.94 \\
\cmidrule(r){1-1} \cmidrule(l){2-5}
DPDNet (single)~\cite{Abuolaim:2020:DPDNet} & 24.42 & 0.827 & 0.277 & 32.25\\
DPDNet (dual)~\cite{Abuolaim:2020:DPDNet}& 25.12 & \textbf{0.850} & \textbf{0.223}  & 32.25\\
Ours (2-level)& \textbf{25.24} & 0.845 & 0.229 & 1.58 \\
Ours (3-level)& \textbf{25.24} & 0.842 & 0.225 & 2.06 \\
\bottomrule
\end{tabular}
}
\caption{Quantitative comparison. The numbers of parameters of JNB and EBDB are not available as they are not deep learning-based methods. Note that DPDNet (dual) uses dual-pixel images.}
\label{tbl:comparison}
\vspace{-7pt}
\end{table}
\begin{table}[t]
\centering
\scalebox{0.77}{
\begin{tabular}{ c  c c c }
\toprule
Model & FLOPs (B) & running time (s) & Parameters (M)\\
\cmidrule(r){1-1} \cmidrule(l){2-4}
DPDNet (single)~\cite{Abuolaim:2020:DPDNet} & 1980 & 0.17 & 32.25\\
DPDNet (dual)~\cite{Abuolaim:2020:DPDNet} & 1983 & 0.17 & 32.25\\
Ours (2-level) & 358 & 0.09 & 1.58 \\
Ours (3-level) & 197 & 0.07 & 2.06 \\
\bottomrule
\end{tabular}
}
\caption{Computational cost comparison. The average FLOPs and running times are measured on images of size \(1280\times 720\).}
\label{tbl:computation}
\vspace{-12pt}
\end{table}

\Tbl{comparison} reports the quantitative comparison.
As shown in the table, the classical two-step approaches \cite{Shi:2015:JNB,Karaali:2018:DMEAdaptive,Lee:2019:DMENet} perform worse than the recent deep-learning based approach \cite{Abuolaim:2020:DPDNet}.
While the DPDNet model with a single input image performs better than the classical approaches,
our models outperform both the classical approaches and the DPDNet model with a single input image by a large margin.
Moreover, our models outperform the dual-pixel-based DPDNet model even without the strong cue to defocus blur provided by dual-pixel data and with a much small number of parameters.
This result clearly proves the effectiveness of our approach.
In the supplementary material, we report the performance of the dual-pixel-based variant of our model, which outperforms the dual-pixel-based DPDNet model.
\Tbl{comparison} also shows that our 2- and 3-level models perform similarly.
However, we found that the 3-level model tends to better handle extremely large blur as shown in \Fig{encoding_level} due to its larger receptive fields.

\Fig{comparison} shows a qualitative comparison.
Our result is produced by the 3-level model.
As the figure shows, our method produces sharper results with more details.
Even compared to the result of the dual-pixel-based DPDNet~\cite{Abuolaim:2020:DPDNet}, our result has comparably clear details. 

\vspace{-12pt}
\paragraph{Computational cost}
We compare the computational cost of our models and DPDNet~\cite{Abuolaim:2020:DPDNet}.
Classical two-step approaches rely on computationally heavy non-blind deconvolution algorithms, so we do not include them in this comparison.
For the comparison, we measure FLOPs and the average running time per image of size $1280\times 720$.
\Tbl{computation} shows that our 3-level model requires small computational cost in FLOPs, which is 10 times smaller than \cite{Abuolaim:2020:DPDNet}.
The table also shows that our 3-level model is slightly faster than the 2-level model as features are more downsampled, even though it has more parameters.

\begin{figure}[tp]
\centering
\setlength\tabcolsep{1 pt}
  \begin{tabular}{cccccc}
    

    \multicolumn{2}{c}{\includegraphics[trim=0 20px 0 20px, clip, width=0.331\linewidth]{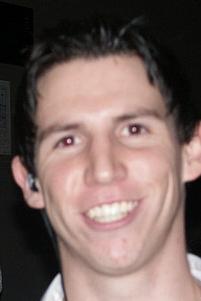}} & 
    \multicolumn{2}{c}{\includegraphics[trim=0 20px 0 20px, clip, width=0.331\linewidth]{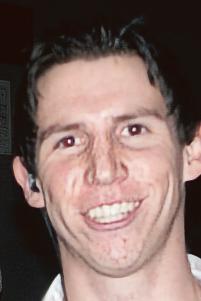}} & 
    \multicolumn{2}{c}{\includegraphics[trim=0 20px 0 20px, clip, width=0.331\linewidth]{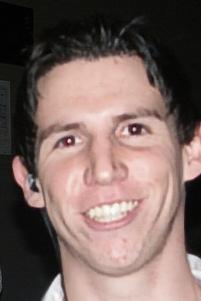}} \\

    \multicolumn{2}{c}{(a) Input}  & \multicolumn{2}{c}{(b) DPDNet~\cite{Abuolaim:2020:DPDNet}  }
     & \multicolumn{2}{c}{(c) Ours } \\

  \end{tabular}
  \vspace{-0.07cm}
  \caption{Defocus deblurring results on the defocused images in the CUHK blur detection dataset~\cite{Shi:2014:CUHK}.}
\label{fig:CUHK}
\vspace{-12pt}
\end{figure}

\vspace{-12pt}
\paragraph{Generalization to other images}
Our models are trained using the DPDD dataset \cite{Abuolaim:2020:DPDNet}, which was generated using one camera. Thus, lastly, we inspect how well our model generalizes to images from other cameras.
To this end, we use the CUHK blur detection dataset \cite{Shi:2014:CUHK}, which provides 704 defocused images without ground-truth all-in-focus images.
The defocused images in the dataset are collected from various sources on the internet.
As there are no ground-truth images, we qualitatively inspect the generalization ability.
\Fig{CUHK} shows that our method can successfully restore fine details with less visual artifacts compared to the single image-based model of DPDNet~\cite{Abuolaim:2020:DPDNet}. 
Additional results can be found in the supplementary material.

\section{Conclusion}
This paper proposed a single image defocus deblurring framework based on inverse kernels.
To effectively simulate spatially varying inverse kernels, we proposed the Kernel-Sharing Parallel Atrous Convolutions (KPAC) block.
KPAC provides an effective way to handle spatially varying defocus blur with a small number of atrous convolution layers that share the same convolutional kernel weights but with different dilation rates.
KPAC is also equipped with the per-pixel scale attention to further facilitate the handling of spatially varying blur.
Thanks to the effective and light-weight structure of KPAC, we can simply stack multiple blocks of KPAC and achieve state-of-the-art deblurring performance.
We experimentally validated the effectiveness of KPAC and showed that our method clearly outperforms previous methods with much fewer parameters.

\vspace{-12pt}
\paragraph{Limitations and future work}
\change{While our method outperforms previous state-of-the-art methods, it may still fail in challenging cases such as large-scale blur, blur with irregular shapes, and bokeh with sharp boundaries
(refer to the supplementary material for our deblurring results on images containing such cases).}
Handling these challenging cases would be an interesting future direction. 



\vspace{-13pt}
\paragraph{Acknowledgements}
This work was supported by the Ministry of Science and ICT, Korea,
through 
IITP grants
(SW Star Lab, 2015-0-00174;
Artificial Intelligence Graduate School Program (POSTECH), 2019-0-01906)
and
NRF grants (2018R1A5A1060031; 2020R1C1C1014863).

{\small
\bibliographystyle{ieee_fullname}
\bibliography{egbib}

\begin{thebibliography}{1}\itemsep=-1pt

\bibitem{Ashikaga:2014:FFT}
Hiroshi Ashikaga, Heidi Estner, Daniel Herzka, Elliot Mcveigh, and Henry
  Halperin.
\newblock Quantitative assessment of single-image super-resolution in
  myocardial scar imaging.
\newblock {\em IEEE Journal of Translational Engineering in Health and
  Medicine}, 2:1--12, 01 2014.

\bibitem{Garg2019ICCV}
Rahul Garg, Neal Wadhwa, Sameer Ansari, and Jonathan~T. Barron.
\newblock Learning single camera depth estimation using dual-pixels.
\newblock In {\em Proc. ICCV}, 2019.

\bibitem{Smith:2007:FFT}
J.~O. Smith.
\newblock {\em Mathematics of the Discrete Fourier Transform (DFT), with Audio
  Applications}.
\newblock W3K Publishing, second edition, 2007.

\end{thebibliography}


\begin{thebibliography}{10}\itemsep=-1pt

\bibitem{Abuolaim:2020:DPDNet}
A. Abuolaim and M.S. Brown.
\newblock Defocus deblurring using dual-pixel data.
\newblock In {\em Proc. ECCV}, 2020.

\bibitem{Bando2007refocus}
Y. Bando and T. Nishita.
\newblock Towards digital refocusing from a single photograph.
\newblock In {\em Proc. Pacific Graphics}, 2007.

\bibitem{Chen2018Atrous}
L. {Chen}, G. {Papandreou}, I. {Kokkinos}, K. {Murphy}, and A.~L. {Yuille}.
\newblock {DeepLab}: Semantic image segmentation with deep convolutional nets,
  atrous convolution, and fully connected crfs.
\newblock {\em IEEE Trans. Pattern Analysis and Machine Intelligence (TPAMI)},
  40(4):834--848, 2018.

\bibitem{Cho:2017:Convergence}
S. {Cho} and S. {Lee}.
\newblock Convergence analysis of {MAP} based blur kernel estimation.
\newblock In {\em Proc. ICCV}, pages 4818--4826, 2017.

\bibitem{Andres:2016:RTF}
L. D'Andrès, J. Salvador, A. Kochale, and S. Susstrunk.
\newblock Non-parametric blur map regression for depth of field extension.
\newblock {\em IEEE Trans. Image Processing (TIP)}, 25(4):1660--1673, 2016.

\bibitem{Dong2020Wiener}
J. Dong, S. Roth, and B. Schiele.
\newblock Deep wiener deconvolution: Wiener meets deep learning for image
  deblurring.
\newblock In {\em Proc. NeurIPS}, 2020.

\bibitem{Duchon1979Lanczos}
Claude~E. Duchon.
\newblock Lanczos filtering in one and two dimensions.
\newblock {\em Journal of Applied Meteorology and Climatology},
  18(8):1016–1022, 1979.

\bibitem{Fish:95:BD}
D.~A. Fish, A.~M. Brinicombe, E.~R. Pike, and J.~G. Walker.
\newblock Blind deconvolution by means of the richardson-lucy algorithm.
\newblock {\em Journal of the Optical Society of America A (JOSA A)},
  12(1):58--65, 1995.

\bibitem{Holschneider1990Atrous}
M. Holschneider, R. Kronland-Martinet, J. Morlet, and Ph. Tchamitchian.
\newblock A real-time algorithm for signal analysis with the help of the
  wavelet transform.
\newblock In {\em Proc. Wavelets}, 1990.

\bibitem{Johnson2016Perceptual}
Justin Johnson, Alexandre Alahi, and Li Fei-Fei.
\newblock Perceptual losses for real-time style transfer and super-resolution.
\newblock In {\em Proc. ECCV}, 2016.

\bibitem{Karaali:2018:DMEAdaptive}
A. Karaali and C. Jung.
\newblock Edge-based defocus blur estimation with adaptive scale selection.
\newblock {\em IEEE Trans. Image Processing (TIP)}, 27(3):1126--1137, 2018.

\bibitem{Krishnan:2008:deconvolution}
D. Krishnan and R. Fergus.
\newblock Fast image deconvolution using hyper-laplacian priors.
\newblock In {\em Proc. NIPS}, 2009.

\bibitem{Lee2019locally}
Guang-He Lee, David Alvarez-Melis, and Tommi~S. Jaakkola.
\newblock Towards robust, locally linear deep networks.
\newblock In {\em Proc. ICLR}, 2019.

\bibitem{Lee:2019:DMENet}
J. Lee, S. Lee, S. Cho, and S. Lee.
\newblock Deep defocus map estimation using domain adaptation.
\newblock In {\em Proc. CVPR}, 2019.

\bibitem{Levin:2007:Coded}
A. Levin, R. Fergus, F. Durand, and W. Freeman.
\newblock Image and depth from a conventional camera with a coded aperture.
\newblock {\em ACM Trans. Graphics (TOG)}, 26:70, 2007.

\bibitem{Maas:2013:Leaky}
Andrew~L. Maas, Awni~Y. Hannun, and Andrew~Y. Ng.
\newblock Rectifier nonlinearities improve neural network acoustic models.
\newblock In {\em Proc. ICML}, 2013.

\bibitem{Montufar2014linear}
Guido Montúfar, Razvan Pascanu, Kyunghyun Cho, and Yoshua Bengio.
\newblock On the number of linear regions of deep neural networks.
\newblock In {\em Proc. NeurIPS}, 2014.

\bibitem{Park:2017:unified}
J. Park, Y. Tai, D. Cho, and I.~S. Kweon.
\newblock A unified approach of multi-scale deep and hand-crafted features for
  defocus estimation.
\newblock In {\em Proc. CVPR}, 2017.

\bibitem{paszke2017automatic}
A. Paszke, S. Gross, S. Chintala, G. Chanan, E. Yang, Z. DeVito, Z. Lin, A.
  Desmaison, L. Antiga, and A. Lerer.
\newblock Automatic differentiation in pytorch.
\newblock In {\em Proc. NIPSW}, 2017.

\bibitem{Ren2018deconv}
Wenqi Ren, Jiawei Zhang, Lin Ma, Jinshan Pan, Xiaochun Cao, Wangmeng Zuo, Wei
  Liu, and Ming-Hsuan Yang.
\newblock Deep non-blind deconvolution via generalized low-rank approximation.
\newblock In {\em Proc. NeurIPS}, 2018.

\bibitem{Ronneberger:2015:Unet}
O. Ronneberger, P. Fischer, and T. Brox.
\newblock {U-Net}: Convolutional networks for biomedical image segmentation.
\newblock In {\em Proc. MICCAI}, 2015.

\bibitem{Shi:2014:CUHK}
J. Shi, L. Xu, and J. Jia.
\newblock Discriminative blur detection features.
\newblock In {\em Proc. CVPR}, 2014.

\bibitem{Shi:2015:JNB}
J. Shi, L. Xu, and J. Jia.
\newblock Just noticeable defocus blur detection and estimation.
\newblock In {\em Proc. CVPR}, 2015.

\bibitem{Simonyan15}
Karen Simonyan and Andrew Zisserman.
\newblock Very deep convolutional networks for large-scale image recognition.
\newblock In {\em Proc. ICLR}, 2015.

\bibitem{Son2017}
H. {Son} and S. {Lee}.
\newblock Fast non-blind deconvolution via regularized residual networks with
  long/short skip-connections.
\newblock In {\em Proc. ICCP}, 2017.

\bibitem{wang2004ssim}
Z. Wang, A.~C. Bovik, H.~R. Sheikh, and E.~P. Simoncelli.
\newblock Image quality assessment: from error visibility to structural
  similarity.
\newblock {\em IEEE Trans. Image Processing (TIP)}, 13(4):600--612, 2004.

\bibitem{Wiener}
Norbert Wiener.
\newblock {\em Extrapolation, Interpolation, and Smoothing of Stationary Time
  Series}.
\newblock The MIT Press, 1964.

\bibitem{Kelvin2015Spatial}
Kelvin Xu, Jimmy~Lei Ba, Ryan Kiros, Kyunghyun Cho, Aaron Courville, Ruslan
  Salakhutdinov, Richard~S. Zemel, and Yoshua Bengio.
\newblock {Show, Attend and Tell}: Neural image caption generation with visual
  attention.
\newblock In {\em Proc. ICML}, 2015.

\bibitem{NIPS2014inverse}
Li Xu, Jimmy~SJ Ren, Ce Liu, and Jiaya Jia.
\newblock Deep convolutional neural network for image deconvolution.
\newblock In {\em Proc. NeurIPS}, 2014.

\bibitem{Xu:2014:DECONV}
L. Xu, X. Tao, and J. Jia.
\newblock Inverse kernels for fast spatial deconvolution.
\newblock In {\em Proc. ECCV}, 2014.

\bibitem{Yuan2008progressive}
Lu Yuan, Jian Sun, Long Quan, and Heung-Yeung Shum.
\newblock Progressive inter-scale and intra-scale non-blind image
  deconvolution.
\newblock {\em ACM Transactions on Graphics}, 27(3):1–10, 2008.

\bibitem{zhang2018perceptual}
R. Zhang, P. Isola, A.~A. Efros, E. Shechtman, and O. Wang.
\newblock The unreasonable effectiveness of deep features as a perceptual
  metric.
\newblock In {\em Proc. CVPR}, 2018.

\bibitem{zhang2018rcan}
Yulun Zhang, Kunpeng Li, Kai Li, Lichen Wang, Bineng Zhong, and Yun Fu.
\newblock Image super-resolution using very deep residual channel attention
  networks.
\newblock In {\em Proc. ECCV}, 2018.

\bibitem{Zhou:2019:STFAN}
S. Zhou, J. Zhang, J. Pan, H. Xie, W. Zuo, and J. Ren.
\newblock Spatio-temporal filter adaptive network for video deblurring.
\newblock In {\em Proc. ICCV}, 2019.

\end{thebibliography}
}

\clearpage
\appendix
\begin{center}
\textbf{\LARGE Supplementary Material}
\end{center}

In this supplementary material, we first present the details of the network architecture (\Sec{architecture}) and then provide additional analyses and experiments: a proof of Eq. (4) in the main paper (\Sec{proof}), \change{more discussion on kernel upsampling (\Sec{kernel_upsampling})}, more validation examples of Eq. (5) in the main paper (\Sec{approx_ex}), more examples of inverse kernel sampling using dilated kernels (\Sec{dilated_ex}), blur detection using a scale attention map (\Sec{blur_detection}), sensitivity to noise (\Sec{noise}), \change{handling irregular blur (\Sec{irregular_blur})}, additional results (\Sec{additional}), and our model extended to use dual-pixel images (\Sec{dualpixel}).

\section{Detailed Network Architecture}
\label{sec:architecture}
Detailed architectures of overall deblurring network and KPAC block can be found in \Tbls{deblurnet} and \ref{tbl:KPACb}, respectively.

\section{Proof of $(\frac{1}{s^{2}}k_{\uparrow{s}})^{\dagger}=\frac{1}{s^{2}}({k^{\dagger}}_{\uparrow{s}})$}
\label{sec:proof}

In this section, we present a formal discussion on Eq. (4) in the main paper.
For a 2D image $l$ of size $w \times h$, the spatial upsampling can be performed by zero padding to the discrete Fourier transform of $l$~\citeSM{Smith:2007:FFT}\citeSM{Ashikaga:2014:FFT}.
Let $\uparrow\!\!{s}$ denote the upsampling operation by a scaling factor $s$,
and let $l_{\uparrow{s}}$ be the upsampled result of image $l$ by the scaling factor $s$.
Then, the discrete Fourier transform $L_{\uparrow{s}}$ of $l_{\uparrow{s}}$ is defined in the range \(-\frac{sw}{2}\leq u<\frac{sw}{2}, -\frac{sh}{2}\leq v<\frac{sh}{2}\), and can be obtained as:
\vspace{2pt}
\begin{equation}
    L(u,v)_{\uparrow{s}} = 
    \begin{cases}
    L(u,v) & -\frac{w}{2} \leq u < \frac{w}{2}, -\frac{h}{2} \leq v< \frac{h}{2}\\
    0 & \textrm{otherwise,}\\
    \end{cases}
\label{eq:fourier_upsample}
\vspace{2pt}
\end{equation}
where \(L(u,v)\) is the discrete Fourier transform of $l$,
and $u$ and $v$ are pixel indices in the frequency domain.
$L(0,0)$ corresponds to the DC component of $L$. 
This zero padding-based upsampling is mathematically equivalent to convolution with a sinc kernel~\citeSM{Ashikaga:2014:FFT}.

\begin{table}%
\centering
\begin{tabular}{cccccc}
  \hline
  \multicolumn{2}{c}{\emph{layer type}(\#)}&size & stride & out  & act.\\
  \hline
  \hline
  \multicolumn{6}{c}{\textbf{Encoder} (3-level)}\\
  \hline
  \hline
  \multicolumn{2}{c}{\emph{Conv}1\_1} & $5\times5$  & (1, 1) & 48 & \emph{lrelu} \\
  \multicolumn{2}{c}{\emph{Conv}1\_2} & $3\times3$  & (1, 1) & 48 & \emph{lrelu} \\
  \multicolumn{2}{c}{\emph{Conv}2\_1} & $3\times3$  & (2, 2) & 48 & \emph{lrelu} \\
  \multicolumn{2}{c}{\emph{Conv}2\_2} & $3\times3$ & (1, 1) & 48 & \emph{lrelu} \\
  \multicolumn{2}{c}{\emph{Conv}3\_1} & $3\times3$ & (2, 2) & 96 & \emph{lrelu} \\
  \multicolumn{2}{c}{\emph{Conv}3\_2} & $3\times3$ & (1, 1) & 96 & \emph{lrelu} \\
  \multicolumn{2}{c}{\emph{Conv}4\_1} & $3\times3$ & (2, 2) & 96 & \emph{lrelu} \\
  \multicolumn{2}{c}{\emph{Conv}4\_2} & $3\times3$ & (1, 1) & 96 & \emph{lrelu} \\

  \hline
  \hline
  \multicolumn{6}{c}{\textbf{KPAC blocks}}\\
  \hline
  \hline
    \multicolumn{2}{c}{\emph{KPAC}1} & $5\times5$ & (1, 1) & 96 & \emph{lrelu} \\
    \multicolumn{2}{c}{\emph{KPAC}2} & $5\times5$ & (1, 1) & 96 & \emph{lrelu} \\
    \multicolumn{2}{c}{\emph{concat}} & \multicolumn{4}{c}{\emph{Conv}4\_2, \emph{KPAC}1, \emph{KPAC}2} \\
  \hline
  \hline
  \multicolumn{6}{c}{\textbf{Decoder} (3-level)}\\
  \hline
  \hline
  
  \multicolumn{2}{c}{\emph{Conv}5\_1} & $3\times3$ & (1, 1) & 96 & \emph{lrelu} \\
  \multicolumn{2}{c}{\emph{Conv}5\_2} & $3\times3$ & (1, 1) & 96 & \emph{lrelu} \\
  \multicolumn{2}{c}{\emph{Deconv}1} & $4\times4$  & (2, 2) & 96 & \emph{lrelu} \\
  \multicolumn{2}{c}{\emph{cocnat}} & \multicolumn{4}{c}{\emph{Deconv}1, \emph{Conv}3\_2} \\
  \multicolumn{2}{c}{\emph{Conv}6} & $3\times3$ & (1, 1) & 96 & \emph{lrelu} \\
  \multicolumn{2}{c}{\emph{Deconv}2} & $4\times4$  & (2, 2) & 48 & \emph{lrelu} \\
  \multicolumn{2}{c}{\emph{cocnat}} & \multicolumn{4}{c}{\emph{Deconv}2, \emph{Conv}2\_2} \\
  \multicolumn{2}{c}{\emph{Conv}7} & $3\times3$ & (1, 1) & 48 & \emph{lrelu} \\
  \multicolumn{2}{c}{\emph{Deconv}3} & $4\times4$  & (2, 2) & 48 & \emph{lrelu} \\
  \multicolumn{2}{c}{\emph{cocnat}} & \multicolumn{4}{c}{\emph{Deconv}3, \emph{Conv}1\_2} \\
  \multicolumn{2}{c}{\emph{Conv}8} & $5\times5$ & (1, 1) & 3 & \emph{lrelu} \\
  \multicolumn{2}{c}{\emph{add}} & \multicolumn{3}{c}{\emph{Conv}8, \emph{Input}} & - \\
  \hline
\end{tabular}
\vspace{0.1cm}
\caption{Architecture of our deblurring network.}
\label{tbl:deblurnet}
\vspace{-7pt}
\end{table}

In the remaining of this section, for notational simplicity, we use $\uparrow\!\!{s}$ to indicate the zero-padding operation for upsampling in the frequency domain as well as the upsampling operation in the spatial domain.
We also omit the pixel coordinates $(u,v)$, e.g., representing $L(u,v)_{\uparrow{s}}$ as $L_{\uparrow{s}}$.

We use the Wiener deconvolution~\cite{Wiener} to compute the inverse kernel $k^{\dagger}$ of a blur kernel $k$, \ie,
\vspace{2pt}
\begin{equation}
    k^{\dagger} =F^{-1}\left({\frac{\overline{F(k)}}{|F(k)|^{2}+\epsilon}}\right),
\label{eq:inverse_kernel}
\vspace{2pt}
\end{equation}
where $F(k)$ is the discrete Fourier transform of $k$, and 
$\overline{F(k)}$ is the complex conjugate of $F(k)$.
The division operation is done in an element-wise manner.
$\epsilon$ is a noise parameter.
Note that when $\epsilon=0$, this inverse kernel is equivalent to the direct inverse kernel $F^{-1}(1/F(k))$.

We first prove the equality relation:
\vspace{2pt}
\begin{equation}
    (k_{\uparrow{s}})^{\dagger}={k^{\dagger}}_{\uparrow{s}}.
    \label{eq:equality}
\vspace{2pt}
\end{equation}
Regarding the left side of \Eq{equality}, the upsampling of a blur kernel $k$ can be obtained 
using the zero padding-based upsampling as:
\vspace{2pt}
\begin{equation}
\begin{split}
    k_{\uparrow{s}} = F^{-1}(F(k)_{\uparrow{s}}).
\end{split}
\label{eq:zeropad_upsample}
\vspace{2pt}
\end{equation}
Then, in the frequency domain, from \Eq{fourier_upsample}, we have
\vspace{2pt}
\begin{equation}
\begin{split}
    F(k_{\uparrow{s}}) =  F(k)_{\uparrow{s}} =
    \begin{cases}
    F(k) & -\frac{w}{2} \leq u < \frac{w}{2}, -\frac{h}{2} \leq v< \frac{h}{2}\\
    0 & \textrm{otherwise.}\\
    \end{cases}
    \label{eq:upsample_blur_kernel}
\end{split}
\vspace{2pt}
\end{equation}
From \Eq{inverse_kernel}, the inverse kernel $(k_{\uparrow{s}})^{\dagger}$ for the upsampled kernel $k_{\uparrow{s}}$ can be derived by
\vspace{2pt}
\begin{equation}
    (k_{\uparrow{s}})^{\dagger} =F^{-1}\left({\frac{\overline{F(k_{\uparrow{s}})}}{|F(k_{\uparrow{s}})|^{2}+\epsilon}}\right).
\label{eq:inverse_kernel_up1}
\vspace{2pt}
\end{equation}
In the frequency domain, from \Eq{upsample_blur_kernel}, we have
\vspace{2pt}
\begin{equation}
\begin{split}
    F((k_{\uparrow{s}})^{\dagger}) = \frac{\overline{F(k_{\uparrow{s}})}}{|F(k_{\uparrow{s}})|^{2}+\epsilon} \  \  \  \  \  \  \  \  \  \  \  \  \  \  \  \  \  \  \  \  \  \  \  \  \  \  \  \  \  \  \  \  \  \  \  \  \  \  \  \  \  
    \\ =\begin{cases}
    \frac{\overline{F(k)}}{|F(k)|^{2}+\epsilon} & -\frac{w}{2} \leq u < \frac{w}{2}, -\frac{h}{2} \leq v< \frac{h}{2}\\
    0 & \textrm{otherwise.}\\
    \end{cases}
\end{split}
    \label{eq:inverse_upsample_blur_kernel}
\vspace{2pt}
\end{equation}
Note that the zero padded area in $F(k_{\uparrow{s}})$ remains zero in $F((k_{\uparrow{s}})^{\dagger})$.

Regarding the right side of \Eq{equality}, from \Eqs{zeropad_upsample} and (\ref{eq:inverse_kernel}), the upsampled inverse kernel ${k^{\dagger}}_{\uparrow{s}}$ for kernel $k$ can be derived as
\vspace{2pt}
\begin{equation}
\begin{split}
    {k^{\dagger}}_{\uparrow{s}} = F^{-1}(F(k^{\dagger})_{\uparrow{s}})  \  \  \  \  \  \  \  \  \  \  \  \  \  \  \  \  \  \  \  \  \  \  \  \  \  \  \  \  \  \  \  \  \  \  \  \  \  \  \  \  \  \  \  \  \ \\
    = F^{-1}\left(\left(F\left(F^{-1}\left(\frac{\overline{F(k)}}{|F(k)|^{2}+\epsilon}\right)\right)\right)_{\uparrow{s}}\right)\\
    = F^{-1}\left(\left(\frac{\overline{F(k)}}{|F(k)|^{2}+\epsilon}\right)_{\uparrow{s}}\right).  \  \  \  \  \  \  \  \  \  \  \  \  \  \  \  \  \  \  \  \  \  \  \
\end{split}
\label{eq:inverse_kernel_up2}
\vspace{2pt}
\end{equation}
Then, in the frequency domain, from \Eq{fourier_upsample}, we have
\vspace{2pt}
\begin{flalign}
\begin{split}
    F({k^{\dagger}}_{\uparrow{s}}) = \left(\frac{\overline{F(k)}}{|F(k)|^{2}+\epsilon}\right)_{\uparrow{s}} \  \  \  \  \  \  \  \  \  \  \  \  \  \  \  \  \  \  \  \  \  \  \  \  \  \  \  \  \  \  \  \  \  \  \
    \\ = \begin{cases}
    \frac{\overline{F(k)}}{|F(k)|^{2}+\epsilon} & -\frac{w}{2} \leq u < \frac{w}{2}, -\frac{h}{2} \leq v< \frac{h}{2}\\
    0 & \textrm{otherwise.}\\
    \end{cases}
\end{split}
    \label{eq:upsample_inverse_blur_kernel}
\vspace{2pt}
\end{flalign}
As \Eqs{inverse_upsample_blur_kernel} and (\ref{eq:upsample_inverse_blur_kernel}) are equivalent to each other,
their spatial domain counterparts $(k_{\uparrow{s}})^{\dagger}$ and ${k^{\dagger}}_{\uparrow{s}}$ are equivalent too.
This proves \Eq{equality}.

Eq.~(4) in the main paper has a scaling factor in both left and right sides.
The upsampling operation in the left side of \Eq{equality} scales up the total intensity of kernel $k$ by $s^2$ times.
Similarly, the upsampling operation in the right side of \Eq{equality} also scales up the total intensity of inverse kernel $k^{\dagger}$ by $s^2$ times.
Thus, to obtain a properly normalized inverse kernel in the left and right sides, we apply a scaling factor $\frac{1}{s^2}$ to $k_{\uparrow{s}}$ in the left side, and to ${k^{\dagger}}_{\uparrow{s}}$ in the right side.
Then, we obtain Eq.~(4) in the main paper.

\begin{table}%
\centering
\begin{tabular}{cccccc}
  \hline
  \multicolumn{2}{c}{\emph{layer type}(\#)}&size & dilation & out  & act.\\
  \hline
  \hline
  \multicolumn{6}{c}{\textbf{Scale attention module}}\\
  \hline
  \hline
  \multicolumn{2}{c}{\emph{AC}} & $5\times5$  & (2, 2) & 32 & \emph{lrelu} \\
  \multicolumn{2}{c}{\emph{AC}} & $5\times5$  & (2, 2) & 32 & \emph{lrelu}
  \\
  \multicolumn{2}{c}{\emph{AC}} & $5\times5$  & (2, 2) & 16 & \emph{lrelu}
  \\
  \multicolumn{2}{c}{\emph{AC}} & $5\times5$  & (2, 2) & 16 & \emph{lrelu}
  \\
  \multicolumn{2}{c}{\emph{Conv} \(\{\alpha_{1}, ..., \alpha_{5}\}\)} & $5\times5$  & (1, 1) & 5 & \emph{sigmoid}
  \\

  \hline
  \hline
  \multicolumn{6}{c}{\textbf{Shape attention module}}\\
  \hline
  \hline
  
  \multicolumn{4}{c}{global average pooling} & 96 & - \\ 
  \multicolumn{4}{c}{fully connected layer} & 16 & \emph{lrelu} \\
  \multicolumn{4}{c}{fully connected layer~(\(\beta\))} & 48 & \emph{sigmoid}
  \\

  \hline
  \hline
  \multicolumn{6}{c}{\textbf{Multiple atrous convolutions}}\\
  \hline
  \hline
    \multicolumn{2}{c}{\emph{AC}1} & $5\times5$ & (1, 1) & 48 & \emph{lrelu} \\
    \multicolumn{2}{c}{\emph{AC}2} & $5\times5$ & (2, 2) & 48 & \emph{lrelu} \\
    \multicolumn{2}{c}{\emph{AC}3} & $5\times5$ & (3, 3) & 48 & \emph{lrelu} \\
    \multicolumn{2}{c}{\emph{AC}4} & $5\times5$ & (4, 4) & 48 & \emph{lrelu} \\
    \multicolumn{2}{c}{\emph{AC}5} & $5\times5$ & (5, 5) & 48 & \emph{lrelu} \\
  \hline
  \hline
  \multicolumn{6}{c}{\textbf{Fusion}}\\
  \hline
  \hline
    \multicolumn{2}{c}{\emph{cocnat}} & \multicolumn{4}{c}{\(\alpha_{1}\times\beta\times AC1, ..., \alpha_{5}\times\beta\times AC5\)} \\
    \multicolumn{2}{c}{\emph{Conv}} & $3\times3$  & (1, 1) & 96 & \emph{lrelu}\\

  \hline
\end{tabular}
\vspace{0.01cm}
\caption{Architecture of our KPAC block. AC denotes an atrous convolution layer.}
\label{tbl:KPACb}
\vspace{-0pt}
\end{table}

\change{
\section{Discussion on Kernel Upsampling}
\label{sec:kernel_upsampling}
We consider general upsampling operation in Eq.\ 4 in the main paper for the commutative property between upsampling and inversion of a kernel.
For validating the property, in \Sec{proof} of this supplementary material, we used a specific upsampling method using the sinc filter to change the spatial scale of a kernel.
Then, there could be a concern whether upsampling of a blur kernel can model actual scale changes of the blur kernel.
In our observation, for Gaussian blur kernels with different standard deviations, which is an often-used assumption in existing defocus deblurring approaches, upsampling a kernel is the same as changing the standard deviation of the kernel (Fig. \ref{fig:gaussian}).

\begin{figure}[t]
\begin{center}
\includegraphics [width=1.01\linewidth] {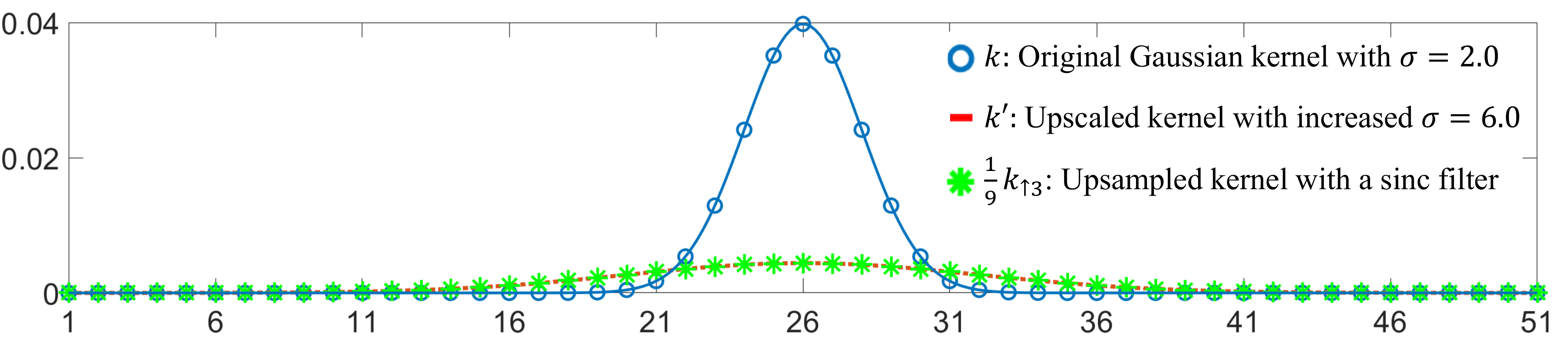}
\end{center}
\vspace{-12pt}
  \caption{1D plot of $3\times$ upscaling of a 2D Gaussian kernel. 
  Upsampling ($\frac{1}{9}k_{\uparrow{3}}$) the original Gaussian kernel ($k$) is the same as upscaling ($k'$) the kernel by increasing the standard deviation ($\sigma$).
}
\label{fig:gaussian}
\vspace{8pt}
\end{figure}

Nonetheless, the upsampling method may cause a gap in accurately modeling the scale changes of real-world blur when the blur kernel is arbitrary, other than Gaussian.
This potential modeling gap in upsampling would be handled by shape and scale attentions in our KPAC blocks together with other convolution layers in our network.
}

\setcounter{footnote}{-1}
\begin{figure*}[tp]
\centering
\small
\setlength\tabcolsep{1 pt}
  \begin{tabular}{cccccccccccc}

    \multicolumn{2}{c}{\includegraphics[width=0.160\linewidth]{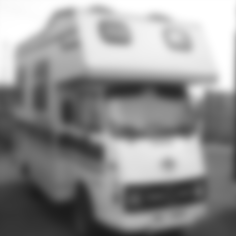}} &
    \multicolumn{2}{c}{\includegraphics[width=0.160\linewidth]{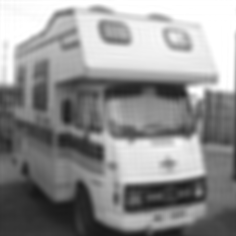}} &
    \multicolumn{2}{c}{\includegraphics[width=0.160\linewidth]{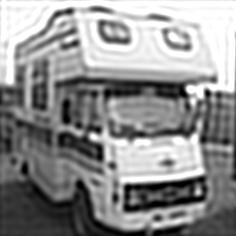}} &
    \multicolumn{2}{c}{\includegraphics[width=0.160\linewidth]{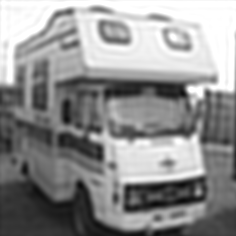}} &
    \multicolumn{2}{c}{\includegraphics[width=0.160\linewidth]{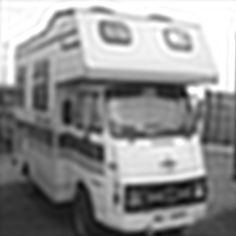}} &
    \multicolumn{2}{c}{\includegraphics[width=0.160\linewidth]{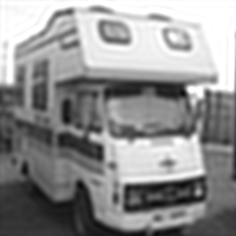}} \\

    \multicolumn{2}{c}{\includegraphics[width=0.160\linewidth]{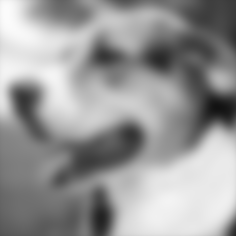}} &
    \multicolumn{2}{c}{\includegraphics[width=0.160\linewidth]{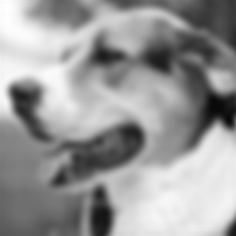}} &
    \multicolumn{2}{c}{\includegraphics[width=0.160\linewidth]{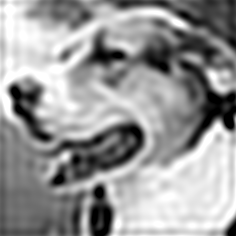}} &
    \multicolumn{2}{c}{\includegraphics[width=0.160\linewidth]{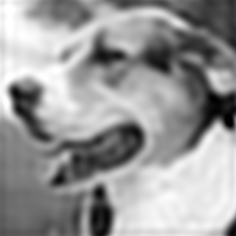}} &
    \multicolumn{2}{c}{\includegraphics[width=0.160\linewidth]{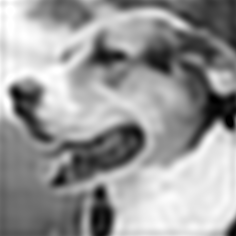}} &
    \multicolumn{2}{c}{\includegraphics[width=0.160\linewidth]{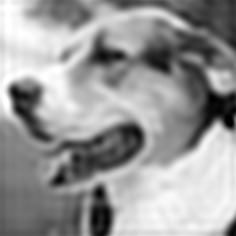}} \\

    \multicolumn{2}{c}{\includegraphics[width=0.160\linewidth]{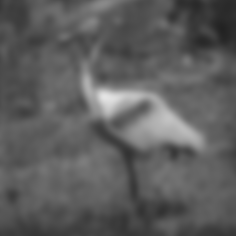}} &
    \multicolumn{2}{c}{\includegraphics[width=0.160\linewidth]{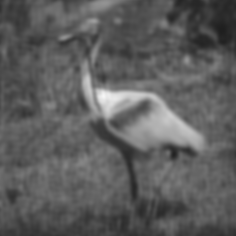}} &
    \multicolumn{2}{c}{\includegraphics[width=0.160\linewidth]{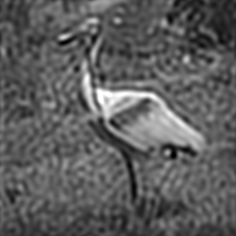}} &
    \multicolumn{2}{c}{\includegraphics[width=0.160\linewidth]{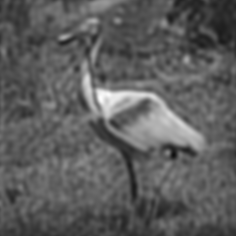}} &
    \multicolumn{2}{c}{\includegraphics[width=0.160\linewidth]{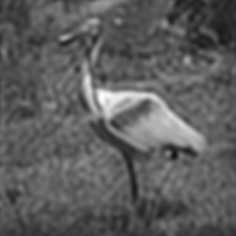}} &
    \multicolumn{2}{c}{\includegraphics[width=0.160\linewidth]{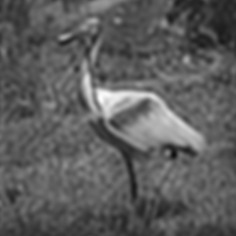}} \\

    \multicolumn{2}{c}{\includegraphics[width=0.160\linewidth]{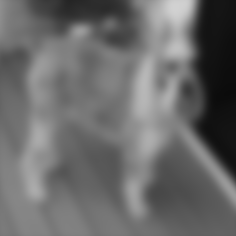}} &
    \multicolumn{2}{c}{\includegraphics[width=0.160\linewidth]{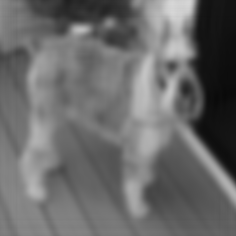}} &
    \multicolumn{2}{c}{\includegraphics[width=0.160\linewidth]{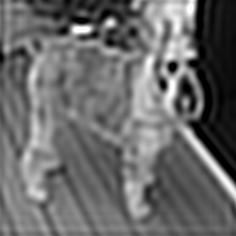}} &
    \multicolumn{2}{c}{\includegraphics[width=0.160\linewidth]{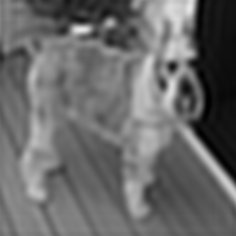}} &
    \multicolumn{2}{c}{\includegraphics[width=0.160\linewidth]{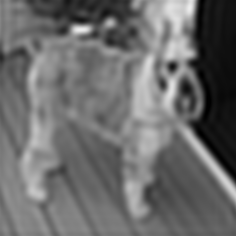}} &
    \multicolumn{2}{c}{\includegraphics[width=0.160\linewidth]{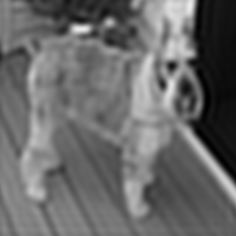}} \\

    \multicolumn{2}{c}{\includegraphics[width=0.160\linewidth]{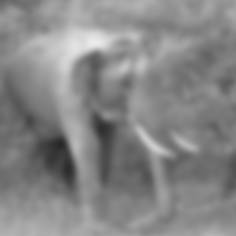}} &
    \multicolumn{2}{c}{\includegraphics[width=0.160\linewidth]{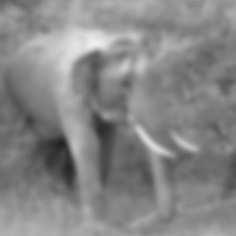}} &
    \multicolumn{2}{c}{\includegraphics[width=0.160\linewidth]{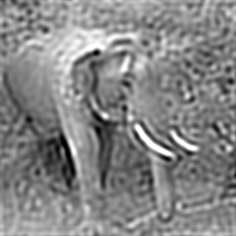}} &
    \multicolumn{2}{c}{\includegraphics[width=0.160\linewidth]{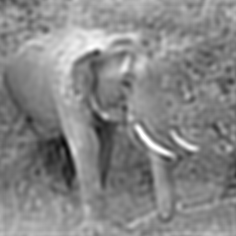}} &
    \multicolumn{2}{c}{\includegraphics[width=0.160\linewidth]{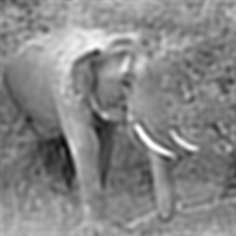}} &
    \multicolumn{2}{c}{\includegraphics[width=0.160\linewidth]{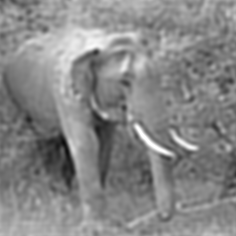}} \\

    \multicolumn{2}{c}{(a) blurred by }  & \multicolumn{2}{c}{(b) deblurred by}
     & \multicolumn{2}{c}{(c) deblurred by} & \multicolumn{2}{c}{(d) blended using} & \multicolumn{2}{c}{(e) blended using} & \multicolumn{2}{c}{(f) deblurred by} \\
    \multicolumn{2}{c}{\tiny{\(\frac{1}{{s_{t}}^2}k_{\uparrow{s_{t}}}\)}} & \multicolumn{2}{c}{\tiny{\(\frac{1}{{s_{1}}^{2}}({k^{\dagger}}_{\uparrow{s_{1}}})\)}} & \multicolumn{2}{c}{\tiny{\(\frac{1}{{s_{2}}^{2}}({k^{\dagger}}_{\uparrow{s_{2}}})\)}} & \multicolumn{2}{c}{Eq. (5)} & \multicolumn{2}{c}{modified Eq. (5)} & \multicolumn{2}{c}{\tiny{\(\frac{1}{{s_{t}}^{2}}({k^{\dagger}}_{\uparrow{s_{t}}})\)}} \\
  \end{tabular}
  \vspace{4pt}
  \caption{Deconvolution results using Eq. (5) in the main paper and its modified version using dilated inverse kernels.
  In the examples, $s_{1}$ and $s_{2}$ are the sampled scale factors and $s_{t}$ is the target scale factor, where $s_{1}<s_{t}<s_{2}$. 
  From top to bottom, we used (3.0, 3.4, 4.0), (3.0, 3.5, 4.0), (3.0, 3.6, 4.0), (2.0, 2.6, 3.0), and (1.0, 1.7, 2.0) for the scale factors ($s_{1}$, $s_{t}$, $s_{2}$).
  We found the blending weights for producing (d) and (e) using non-negative least squares to fit the result of the target scale factor $s_{t}$ in (f) using those of the sampled scale factors $s_{1}$ and $s_{2}$ in (b) and (c).
  Approximation accuracies\protect\footnotemark
of (d) from top to bottom are 98.68\%, 97.84\%, 98.96\%, 95.37\%, and 97.03\%, respectively. Approximation accuracies of (e) from top to bottom are 98.55\%, 97.87\%, 99.10\%, 96.12\%, and 97.64\%, respectively.}
  \vspace{-15pt}
\label{fig:approx}
\end{figure*}

\begin{figure*}[tp]
\centering
\small
\setlength\tabcolsep{1 pt}
  \begin{tabular}{cccccccccccc}

    \multicolumn{2}{c}{\includegraphics[width=0.160\linewidth]{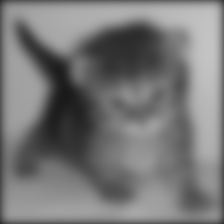}} &
    \multicolumn{2}{c}{\includegraphics[width=0.160\linewidth]{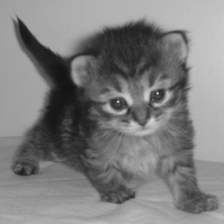}} &
    \multicolumn{2}{c}{\includegraphics[width=0.160\linewidth]{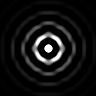}} &
    \multicolumn{2}{c}{\includegraphics[width=0.160\linewidth]{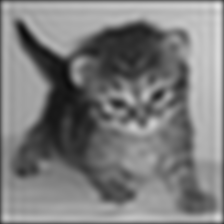}} &
    \multicolumn{2}{c}{\includegraphics[width=0.160\linewidth]{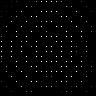}} &
    \multicolumn{2}{c}{\includegraphics[width=0.160\linewidth]{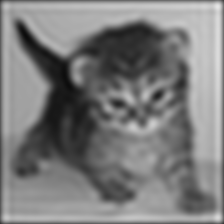}} \\

    \multicolumn{2}{c}{\includegraphics[width=0.160\linewidth]{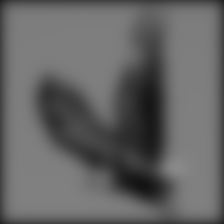}} &
    \multicolumn{2}{c}{\includegraphics[width=0.160\linewidth]{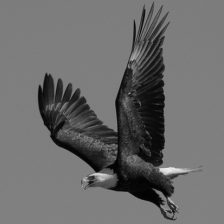}} &
    \multicolumn{2}{c}{\includegraphics[width=0.160\linewidth]{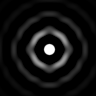}} &
    \multicolumn{2}{c}{\includegraphics[width=0.160\linewidth]{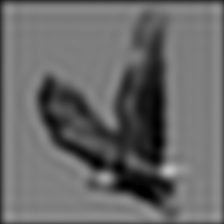}} &
    \multicolumn{2}{c}{\includegraphics[width=0.160\linewidth]{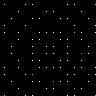}} &
    \multicolumn{2}{c}{\includegraphics[width=0.160\linewidth]{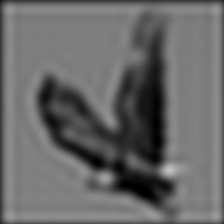}} \\
    
    \multicolumn{2}{c}{\includegraphics[width=0.160\linewidth]{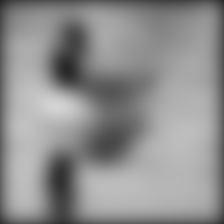}} &
    \multicolumn{2}{c}{\includegraphics[width=0.160\linewidth]{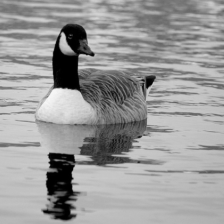}} &
    \multicolumn{2}{c}{\includegraphics[width=0.160\linewidth]{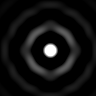}} &
    \multicolumn{2}{c}{\includegraphics[width=0.160\linewidth]{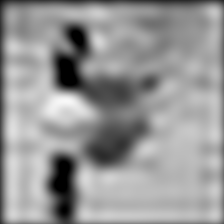}} &
    \multicolumn{2}{c}{\includegraphics[width=0.160\linewidth]{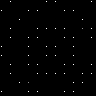}} &
    \multicolumn{2}{c}{\includegraphics[width=0.160\linewidth]{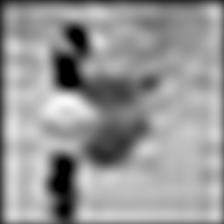}} \\

    \multicolumn{2}{c}{\includegraphics[width=0.160\linewidth]{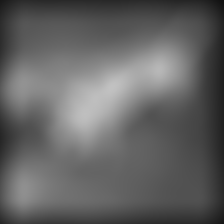}} &
    \multicolumn{2}{c}{\includegraphics[width=0.160\linewidth]{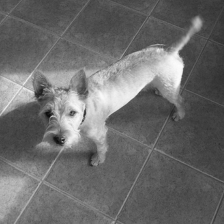}} &
    \multicolumn{2}{c}{\includegraphics[width=0.160\linewidth]{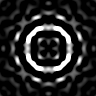}} &
    \multicolumn{2}{c}{\includegraphics[width=0.160\linewidth]{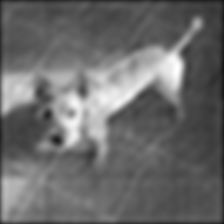}} &
    \multicolumn{2}{c}{\includegraphics[width=0.160\linewidth]{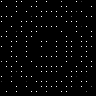}} &
    \multicolumn{2}{c}{\includegraphics[width=0.160\linewidth]{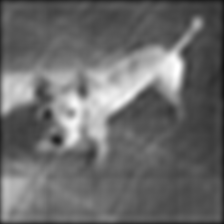}} \\
    
    \multicolumn{2}{c}{\includegraphics[width=0.160\linewidth]{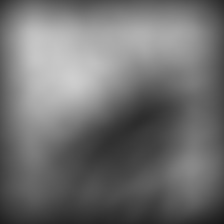}} &
    \multicolumn{2}{c}{\includegraphics[width=0.160\linewidth]{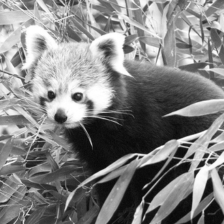}} &
    \multicolumn{2}{c}{\includegraphics[width=0.160\linewidth]{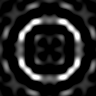}} &
    \multicolumn{2}{c}{\includegraphics[width=0.160\linewidth]{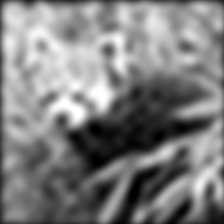}} &
    \multicolumn{2}{c}{\includegraphics[width=0.160\linewidth]{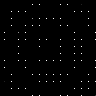}} &
    \multicolumn{2}{c}{\includegraphics[width=0.160\linewidth]{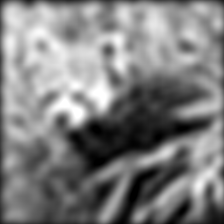}} \\
    
    \multicolumn{2}{c}{\includegraphics[width=0.160\linewidth]{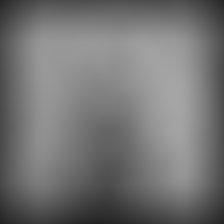}} &
    \multicolumn{2}{c}{\includegraphics[width=0.160\linewidth]{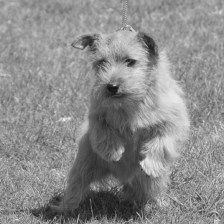}} &
    \multicolumn{2}{c}{\includegraphics[width=0.160\linewidth]{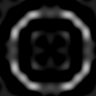}} &
    \multicolumn{2}{c}{\includegraphics[width=0.160\linewidth]{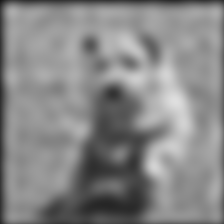}} &
    \multicolumn{2}{c}{\includegraphics[width=0.160\linewidth]{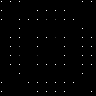}} &
    \multicolumn{2}{c}{\includegraphics[width=0.160\linewidth]{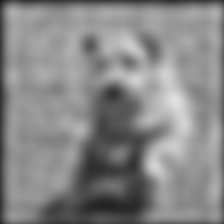}} \\
    
    \multicolumn{2}{c}{(a) blurred by \(\frac{1}{s^2}k_{\uparrow{s}}\)}  & \multicolumn{2}{c}{(b) ground-truth}  & \multicolumn{2}{c}{(c) \(\frac{1}{s^{2}}({k^{\dagger}}_{\uparrow{s}})\)}
     & \multicolumn{2}{c}{(d) deblurred by (c)} & \multicolumn{2}{c}{(e) \({k^{\dagger}}_{\uparrow{/s}}\)} & \multicolumn{2}{c}{(f) deblurred by (e)} \\
  \end{tabular}
  \caption{Visual examples showing that \(\frac{1}{s^{2}}({k^{\dagger}}_{\uparrow{s}})\) and \({k^{\dagger}}_{\uparrow{/s}}\) produce similar results. From top to bottom, we used 3, 4, 5, 3, 4, and 5 for the scale factor $s$. All PSNR values between (d) and (f) are over 40dB.}
  \vspace{-15pt}
\label{fig:dilated_inverse_kernel}
\end{figure*}

\section{Inverse Kernel-based Deconvolution for\\Spatially Varying Defocus Blur}
\label{sec:approx_ex}
In Eq. (5) in the main paper, we approximate spatially varying image deconvolution by combining the results obtained from inverse kernels with a discrete set of sizes.
In this section, we present additional visual examples to show the validity of our approximated deconvolution.
\Figs{approx}b and \ref{fig:approx}c are the deblurred results obtained by convolving inverse kernels of different sizes. While neither kernel fits the actual blur size, 
a linear combination of the deconvolution results still produces a visually pleasing result (\Fig{approx}d), almost equivalent to the deconvolution result (\Fig{approx}f) using the inverse kernel of the target scale.

\footnotetext{The accuracy is computed by 
$1\!-\!\mathit{MAE}(1, \hat{x}/{x_s})$,
where $\mathit{MAE}$ is the mean absolute error, $/$ is pixel-wise division, $x_s$ is the deconvolution result using an inverse kernel of a target scale (\textit{e.g.}, \Fig{approx}f), and $\hat{x}$ is the approximated deconvolution result computed using Eq. (5) (\textit{e.g.}, \Fig{approx}d).}
\begin{figure*}[tp]
\centering
\setlength\tabcolsep{0.5 pt}
  \begin{tabular}{cccccccc}
    \multicolumn{2}{c}{\includegraphics[width=0.245\linewidth]{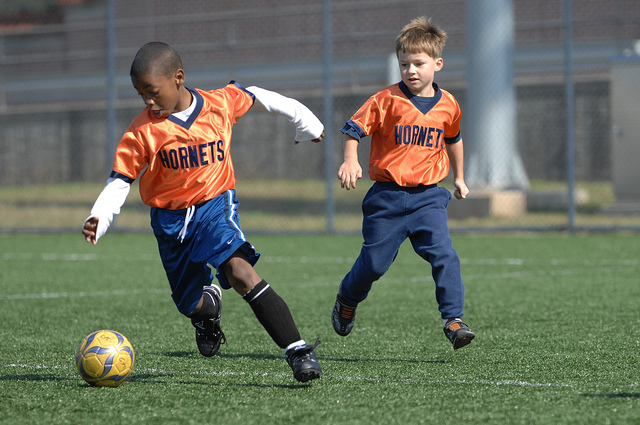}} & 
    \multicolumn{2}{c}{\includegraphics[trim=0 0 0 13px, clip, width=0.245\linewidth]{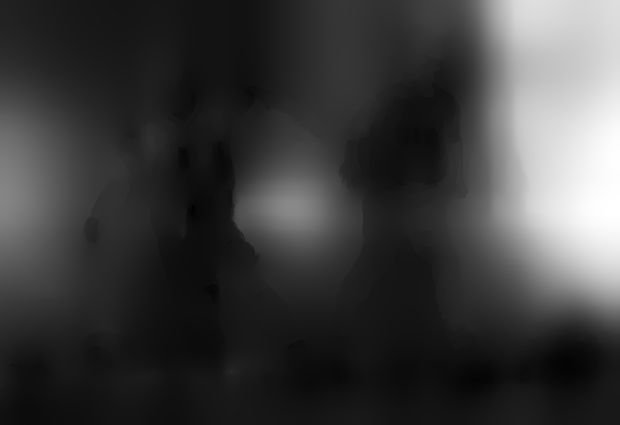}} & 
    \multicolumn{2}{c}{\includegraphics[width=0.245\linewidth]{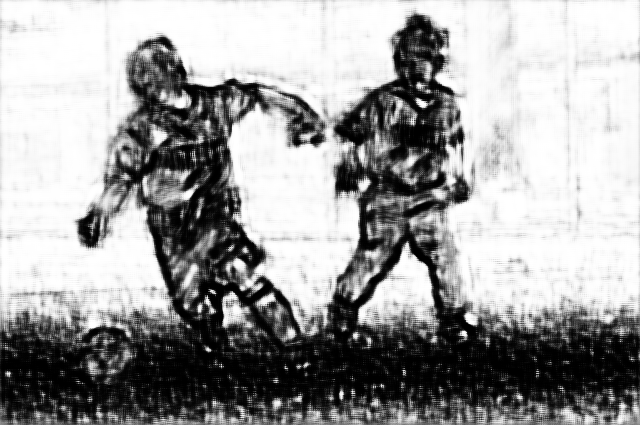}} &
    \multicolumn{2}{c}{\includegraphics[width=0.245\linewidth]{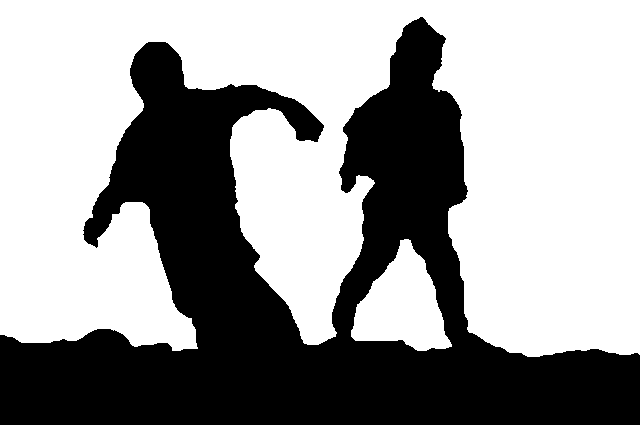}} \\[-0.02in]
    
    \multicolumn{2}{c}{\includegraphics[width=0.245\linewidth]{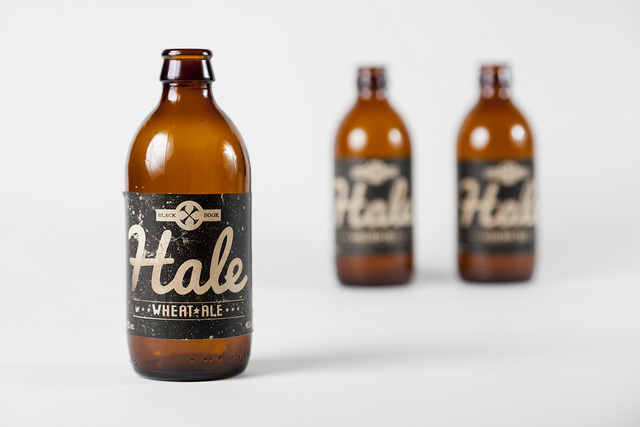}} & 
    \multicolumn{2}{c}{\includegraphics[trim=0 0 0 13px, clip,width=0.245\linewidth]{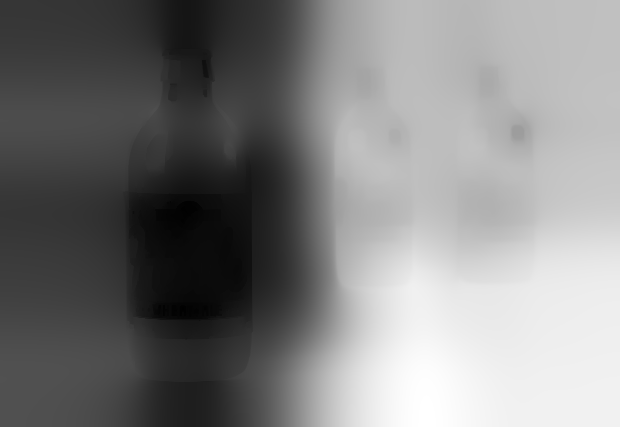}} & 
    \multicolumn{2}{c}{\includegraphics[width=0.245\linewidth]{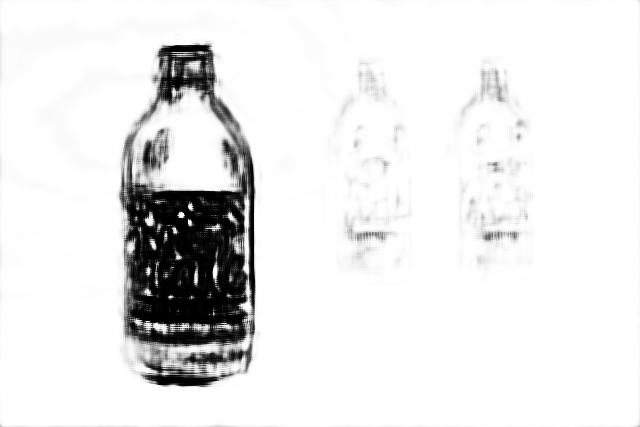}} &
    \multicolumn{2}{c}{\includegraphics[width=0.245\linewidth]{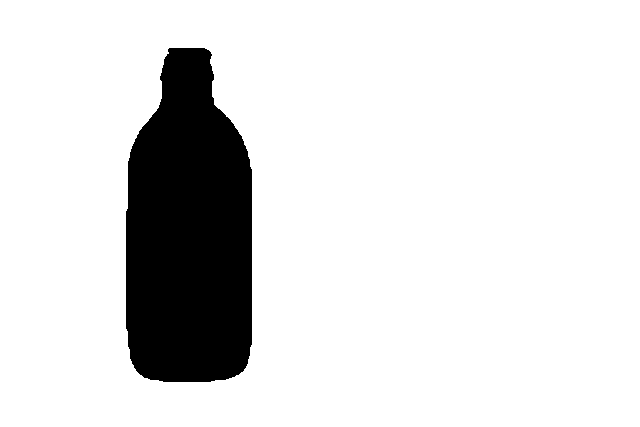}} \\[-0.02in]

    \multicolumn{2}{c}{(a) input}  & \multicolumn{2}{c}{(b) EBDB~\cite{Karaali:2018:DMEAdaptive} }
     & \multicolumn{2}{c}{(c) our attention map} & \multicolumn{2}{c}{(d) ground-truth }\\

  \end{tabular}
  \vspace{-0.05cm}
  \caption{Blur detection using our scale attention map. The attention map used for blur detection is for the atrous convolution layer with a dilation rate 1 in the first KPAC block of our model.}
\label{fig:blur_detection}
\vspace{-10pt}
\end{figure*}

\section{Inverse Kernel Sampling for Atrous Convolution}
\label{sec:dilated_ex}


In Sec. 4.1 of the main paper, we claimed that for a blurred region, which is spatially smooth, filtering operations using sparsely sampled pixels (with a dilated kernel) and using densely sampled pixels (with a rescaled kernel) produce similar results.
\Fig{dilated_inverse_kernel} presents additional visual examples to show that a dense sampling of an inverse kernel can be replaced by its sparse sampling in terms of deblurring performance.

Moreover, we also claimed that a modified version of Eq. (5) in the main paper using the dilated inverse kernels produces results almost equivalent to those from the original Eq. (5).
In \Fig{approx},
we present additional examples to show the validity of the claim,
where the modified and original versions of Eq. (5) produce almost same results in \Figs{approx}e and \ref{fig:approx}d, respectively.

\section{Blur Detection Using a Scale Attention Map}
\label{sec:blur_detection}
In Sec. 5.1 of the main paper, we showed that the scale attention map for the atrous convolution layer with a dilation rate of 1 in the first KPAC block of our deblurring network captures blur of almost any size.
In this section, we evaluate the blur detection performance of the scale attention map using the CUHK dataset~\cite{Shi:2014:CUHK}.
We use 200 test images in the CUHK dataset and measure F-measure and accuracy.
Since our attention is computed in the low resolution feature space, small blurs that can be removed in the encoder network would be ignored.
Therefore, we upsample input images four times for obtaining attention maps in this test.
The result shows that our attention map (F-measure: 0.832 and accuracy: 78.4\%) can detect blur comparably to the recent defocus map estimation method~\cite{Karaali:2018:DMEAdaptive} (F-measure: 0.839 and accuracy: 76.5\%).
\Fig{blur_detection} shows qualitative examples.

\begin{table}[t]
\centering
\begin{tabular}{ c c c c }
\toprule
&\multicolumn{3}{c}{Noise level ($\sigma$) of testset}\\
  & 1\% & 3\% & 5\%  \\
 \cmidrule(r){1-1} \cmidrule(l){2-4}
 w/o noise augmentation & 25.13 & 24.76 & 24.23 \\
 w/ noise augmentation & 25.06 & 25.01  & 24.85 \\
\bottomrule

\end{tabular}
\vspace{0.1cm}
\caption{Performance (PSNR) under different noise conditions.}
\label{tbl:noise}
\vspace{-5pt}
\end{table}

\section{Sensitivity to Noise}
\label{sec:noise}
We investigate the sensitivity of our model to different noise levels, as the shape of a desirable inverse kernel can be affected by noise.
\Tbl{noise} quantitatively shows the effect of the noise level on our model.
For the experiment, we prepare two models.
One model is trained with the original dataset of \cite{Abuolaim:2020:DPDNet} (top row of the table).
The other model is trained with defocused images augmented with additive Gaussian noise controlled by $\sigma$ within a range $[0\%, 3\%]$ (bottom row of the table).
Compared to the model trained without the noise augmentation, the model trained with the noise augmentation is more robust to noise, and shows more consistent PSNRs around 25 dB.

\begin{figure}[t]
\begin{center}
\includegraphics [width=1.0\linewidth] {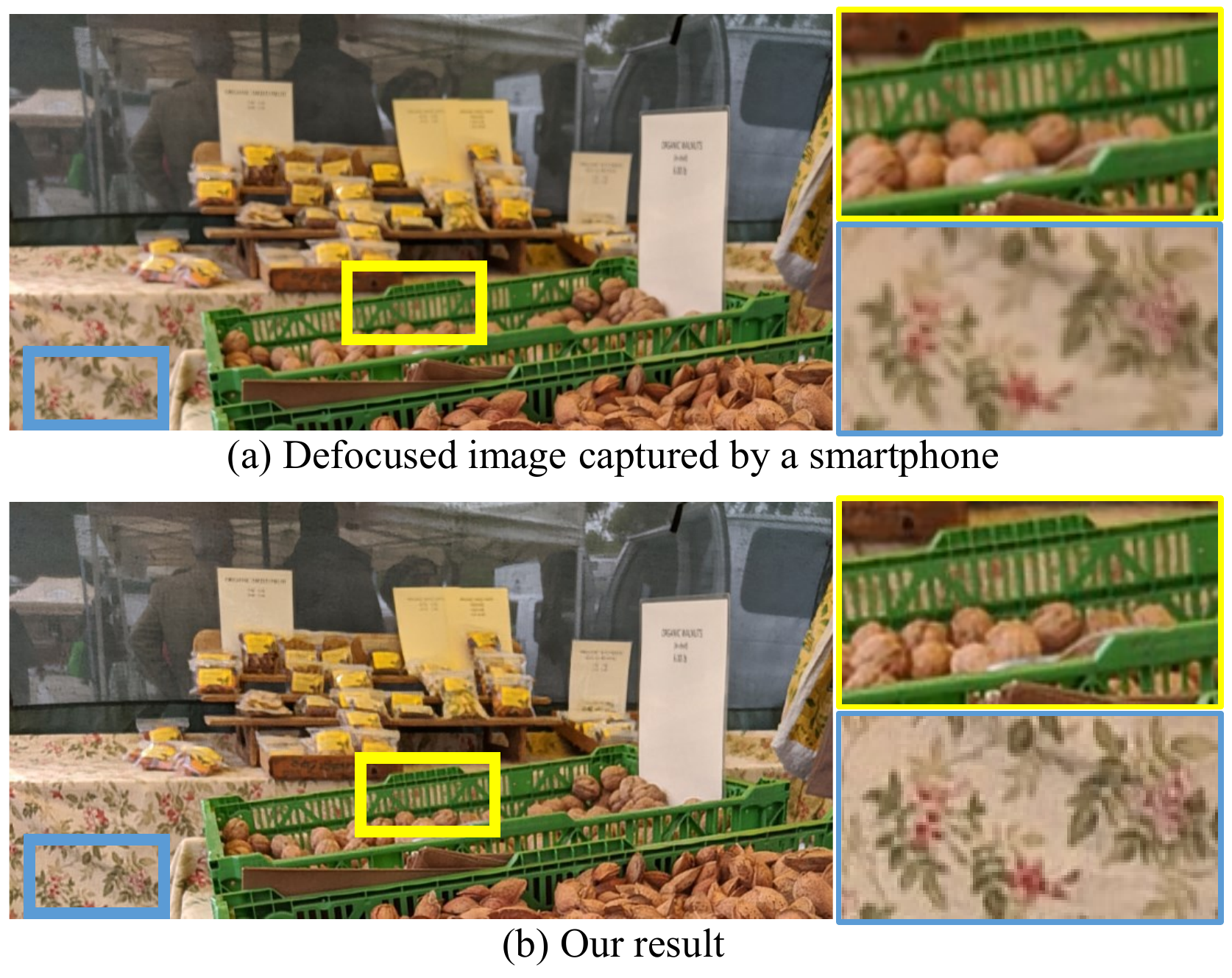}
\end{center}
\vspace{-13pt}
\caption{Deblurring on the smartphone dataset~\protect\citeSM{Garg2019ICCV}. Our network properly removes the blur both in center (yellow box) and peripheral (blue box) regions of an image.
}
\label{fig:smartphone}
\vspace{-0pt}
\end{figure}

\begin{figure}[t]
\begin{center}
\includegraphics [width=1.0\linewidth] {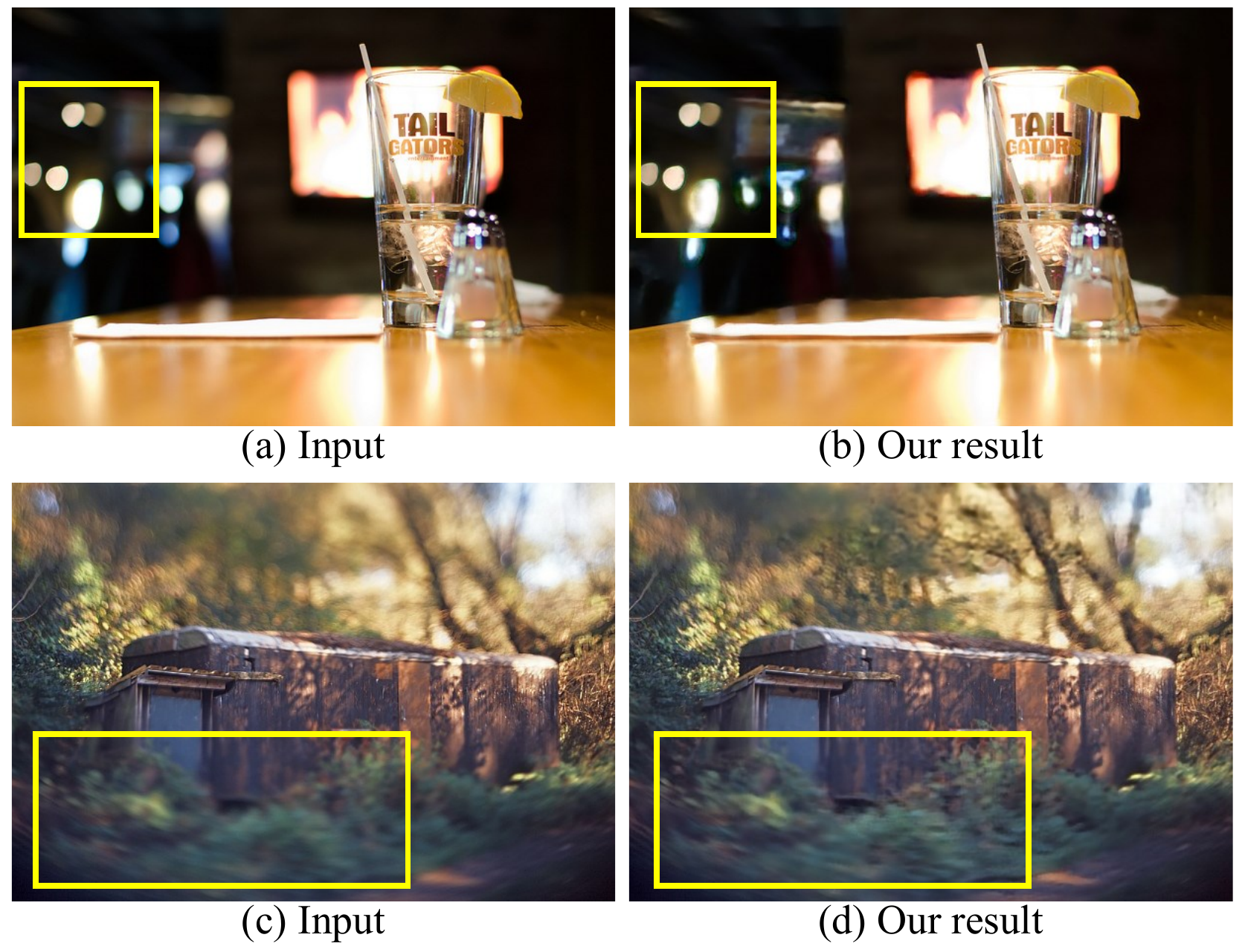}
\end{center}
\vspace{-13pt}
  \caption{Failure cases on the CUHK dataset \cite{Shi:2014:CUHK}. Our network fails for irregular blurs such as bokeh with sharp boundaries (a) and swirly bokeh (c), not contained in the training dataset \cite{Abuolaim:2020:DPDNet}.
}
\label{fig:failure}
\vspace{-8pt}
\end{figure}

\change{
\section{Handling Irregular Blur}
\label{sec:irregular_blur}
Due to the network design based on kernel weight sharing, 
our method would be more effective for the case where the majority of blur variation happens in the size.
In practice, blur shape can spatially vary as well, e.g., due to lens distortion in a smartphone camera. 
Still, we observed that our network moderately works on defocused images captured by smartphones in an unseen dataset \citeSM{Garg2019ICCV}, which usually contain small-sized blurs (Fig. \ref{fig:smartphone}).
However, as we discussed as limitations in Sec. 6 of the main paper, our network may not properly handle blur with severely irregular shapes or strong highlights (Fig. \ref{fig:failure}), which are rarely included in the training set \cite{Abuolaim:2020:DPDNet}.
}

\section{Additional Results}
\label{sec:additional}
\begin{figure*}[tp]
\centering
\setlength\tabcolsep{1 pt}
  \begin{tabular}{cccccccccccc}
    \multicolumn{2}{c}{\includegraphics[width=0.196\linewidth]{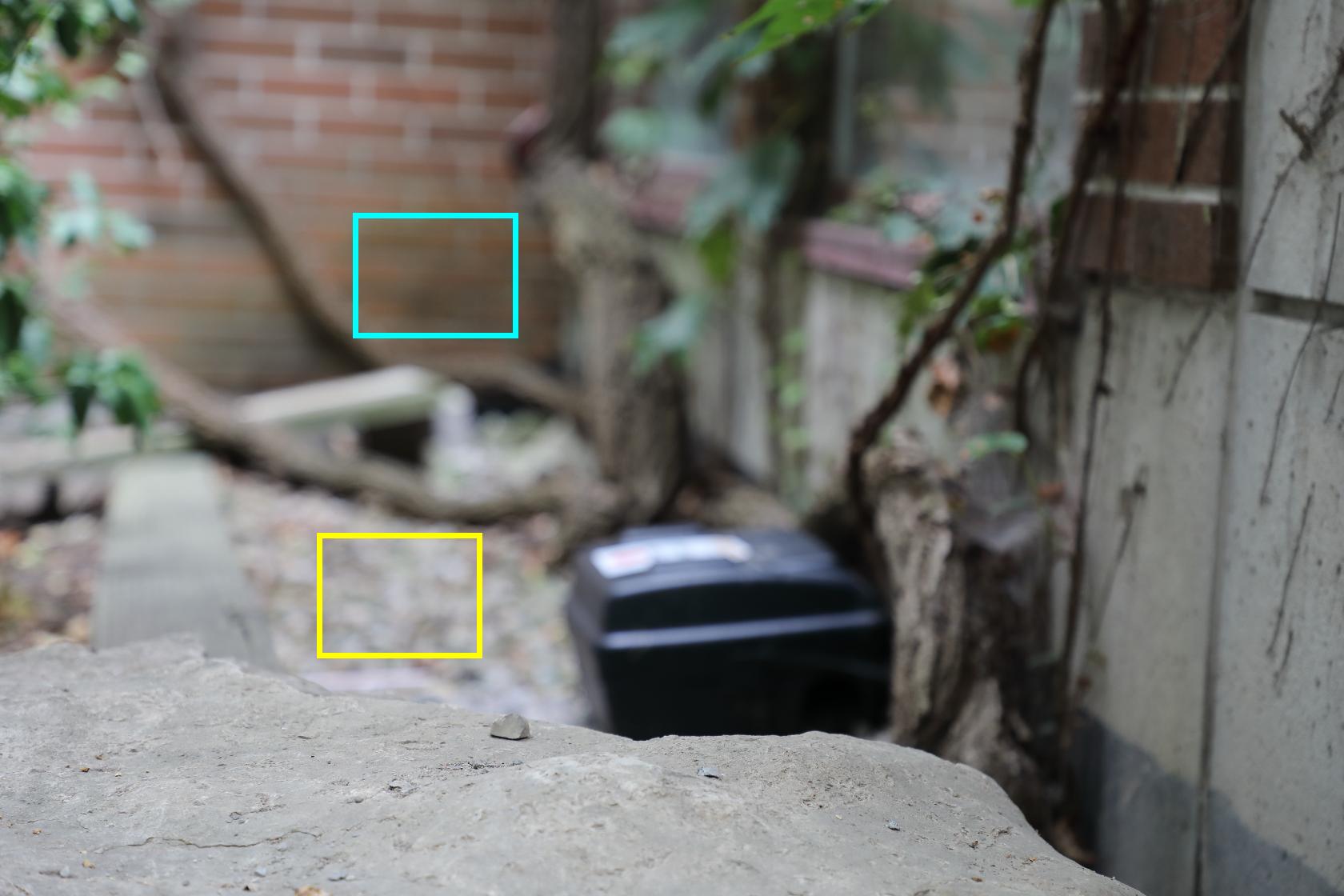}} &
    \multicolumn{2}{c}{\includegraphics[width=0.196\linewidth]{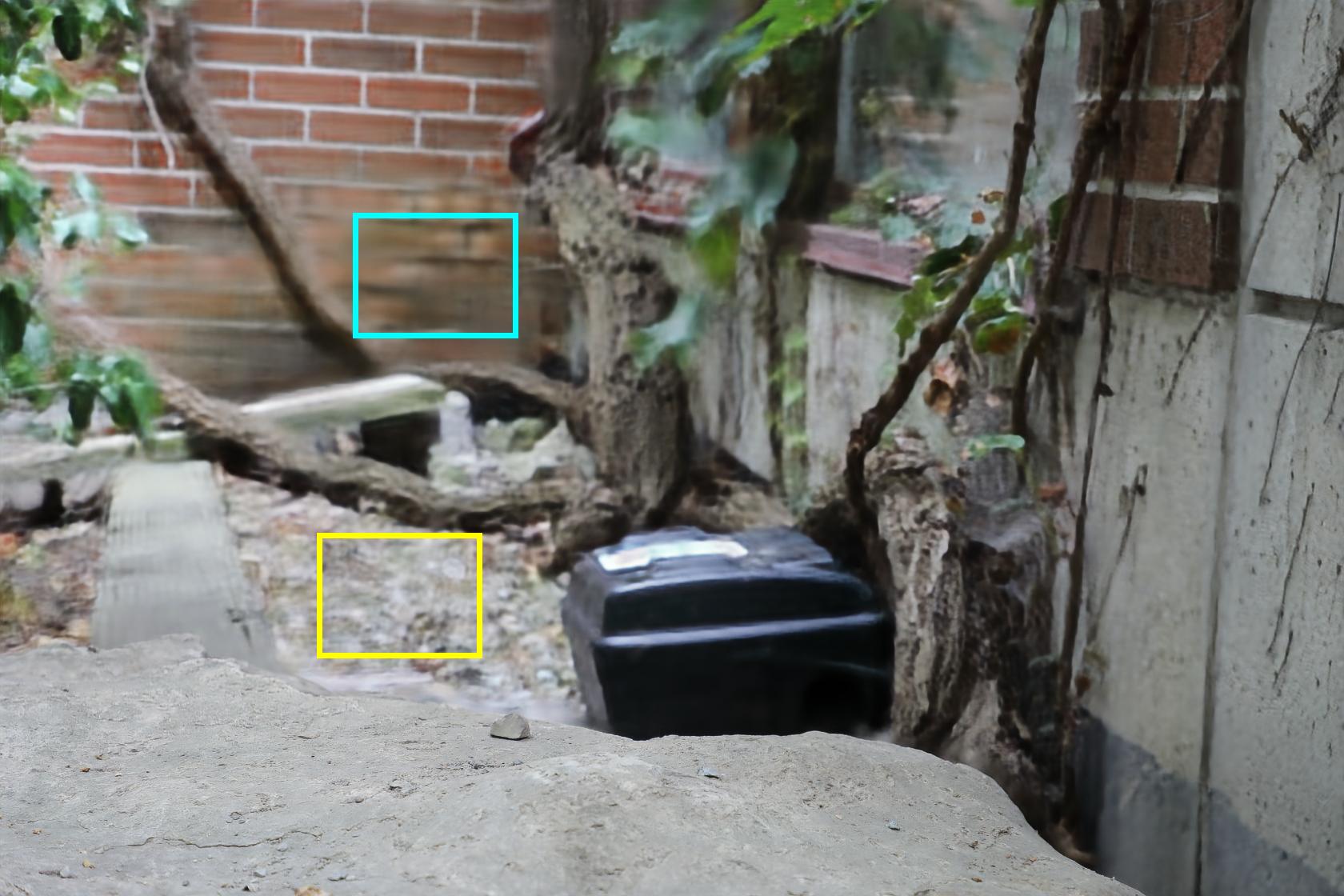}} &
    \multicolumn{2}{c}{\includegraphics[width=0.196\linewidth]{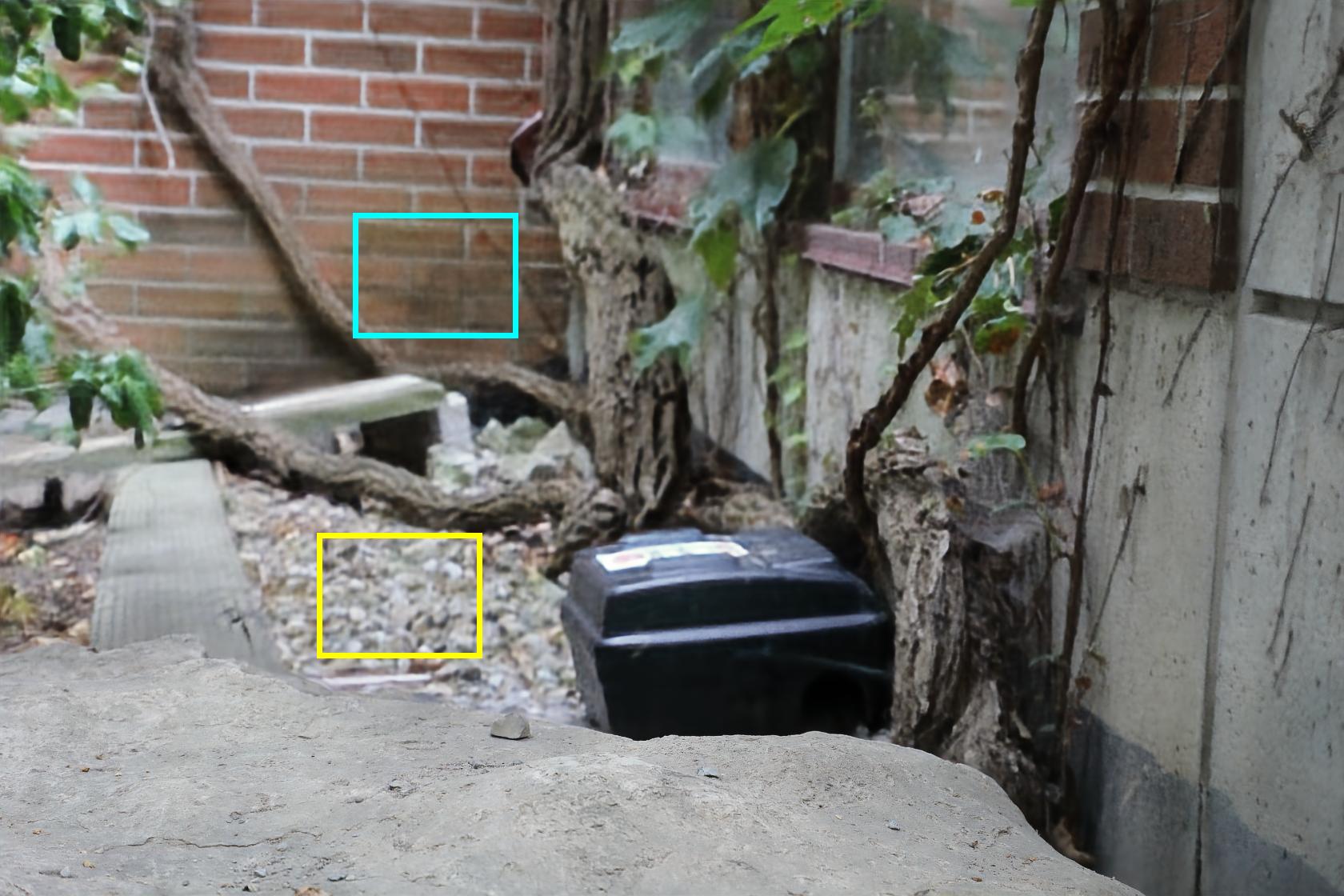}} &
    \multicolumn{2}{c}{\includegraphics[width=0.196\linewidth]{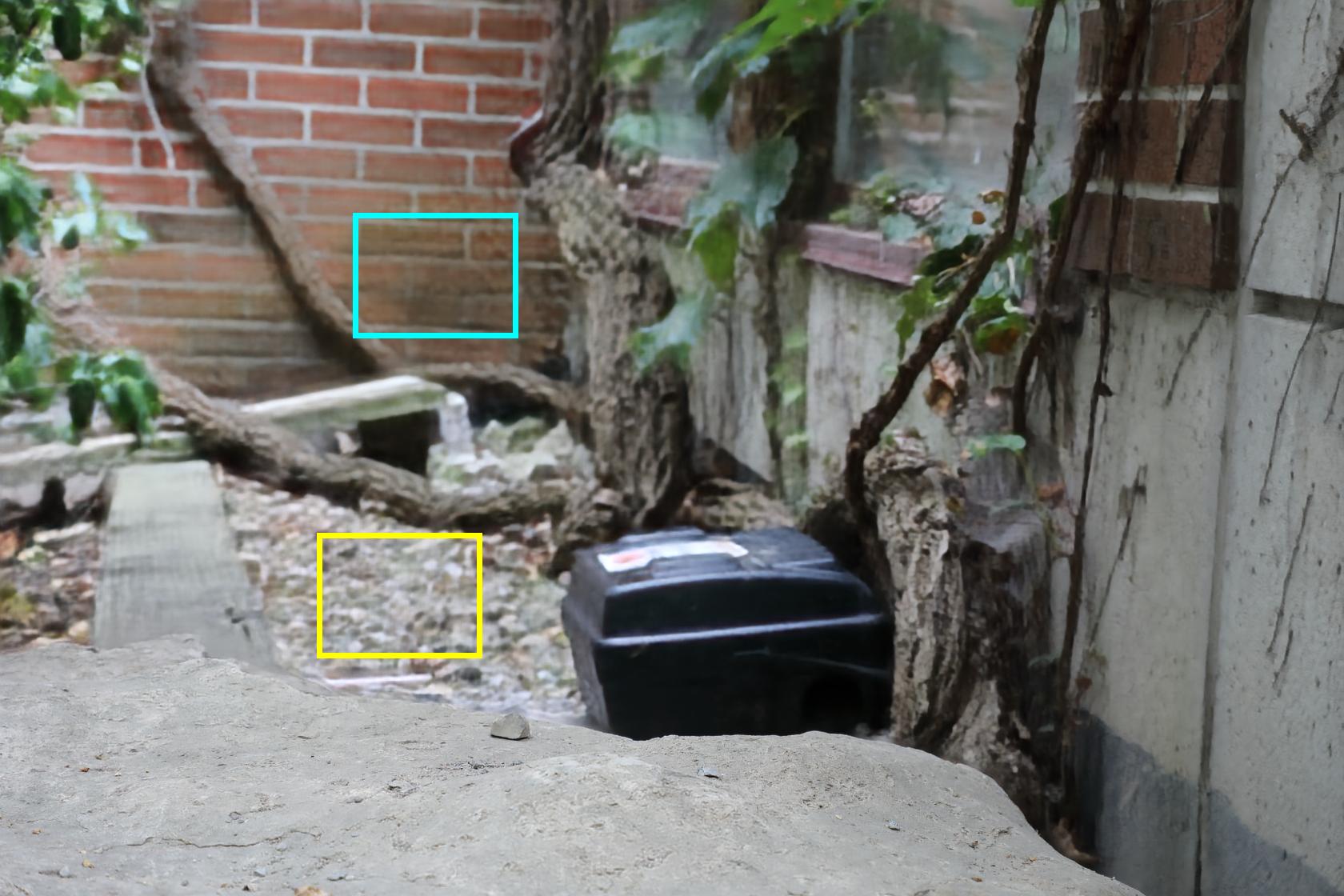}} &
    \multicolumn{2}{c}{\includegraphics[width=0.196\linewidth]{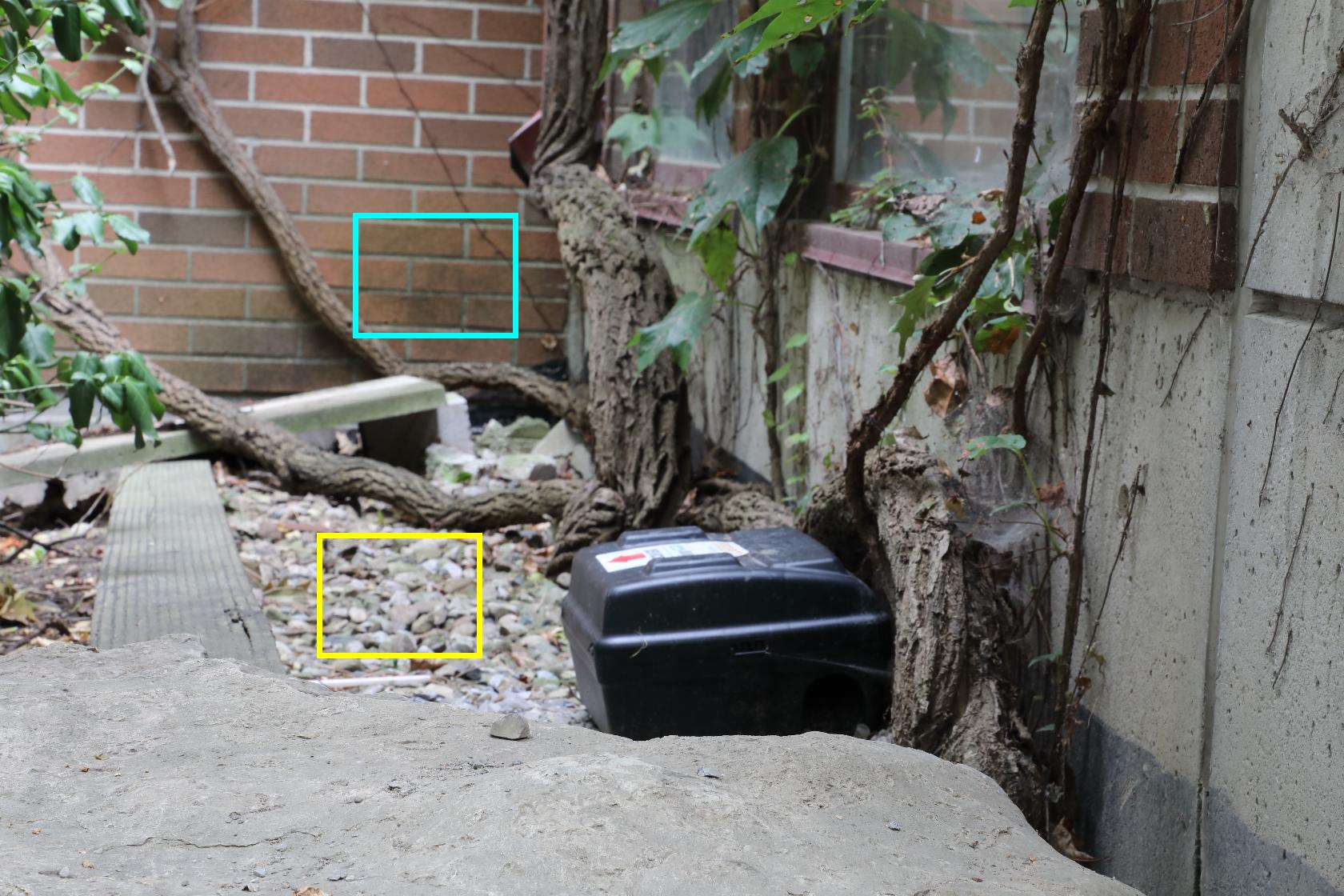}} \\ 
    \multicolumn{1}{c}{\includegraphics[width=0.096\linewidth]{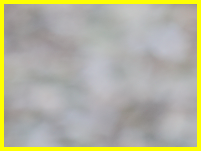}} &
    \multicolumn{1}{c}{\includegraphics[width=0.096\linewidth]{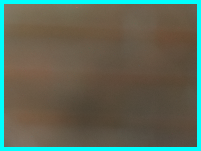}} &
    \multicolumn{1}{c}{\includegraphics[width=0.096\linewidth]{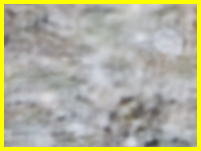}} &
    \multicolumn{1}{c}{\includegraphics[width=0.096\linewidth]{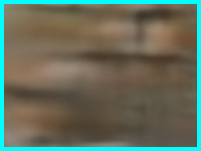}} &
    \multicolumn{1}{c}{\includegraphics[width=0.096\linewidth]{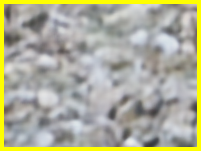}} &
    \multicolumn{1}{c}{\includegraphics[width=0.096\linewidth]{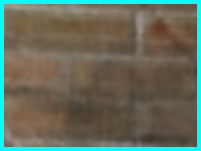}} &
    \multicolumn{1}{c}{\includegraphics[width=0.096\linewidth]{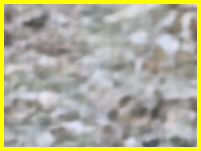}} &
    \multicolumn{1}{c}{\includegraphics[width=0.096\linewidth]{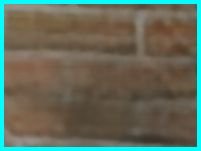}} &
    \multicolumn{1}{c}{\includegraphics[width=0.096\linewidth]{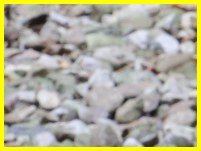}} &
    \multicolumn{1}{c}{\includegraphics[width=0.096\linewidth]{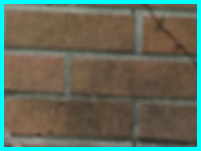}} \\

    \multicolumn{2}{c}{\includegraphics[width=0.196\linewidth]{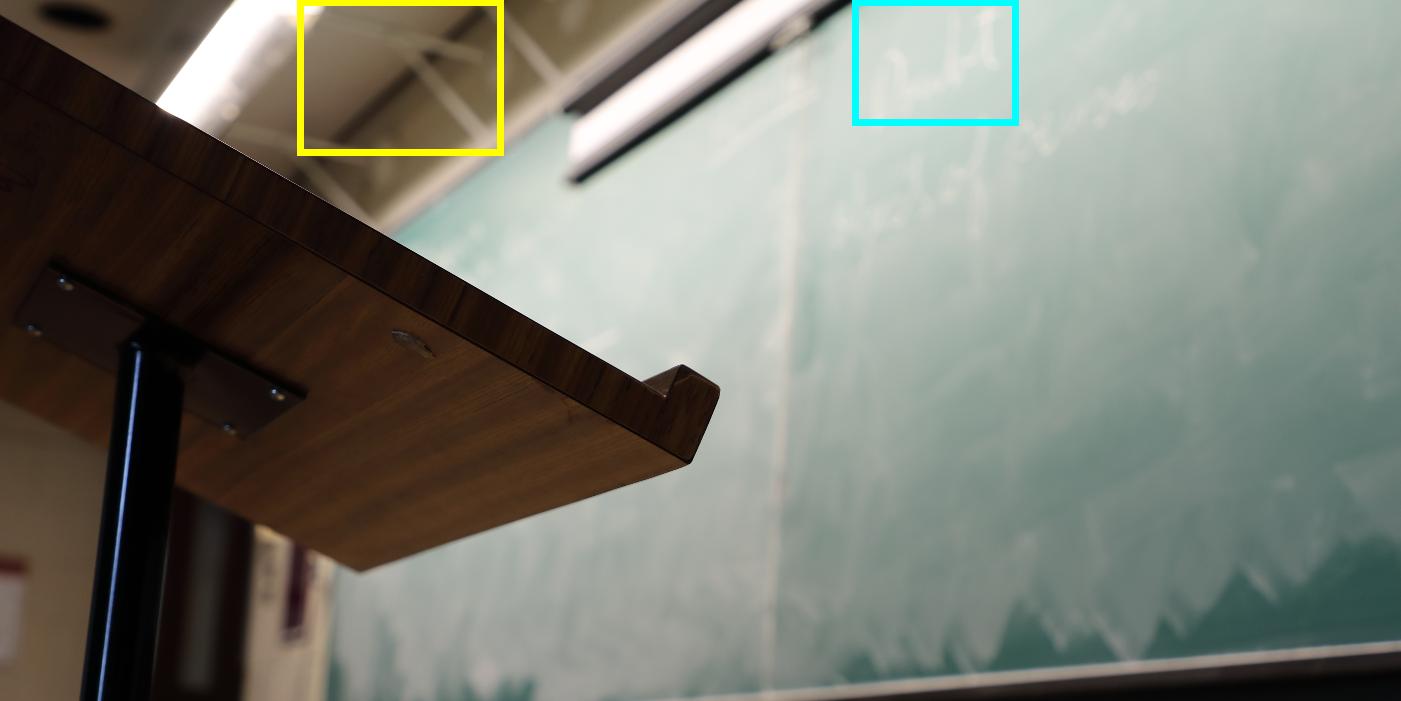}} &
    \multicolumn{2}{c}{\includegraphics[width=0.196\linewidth]{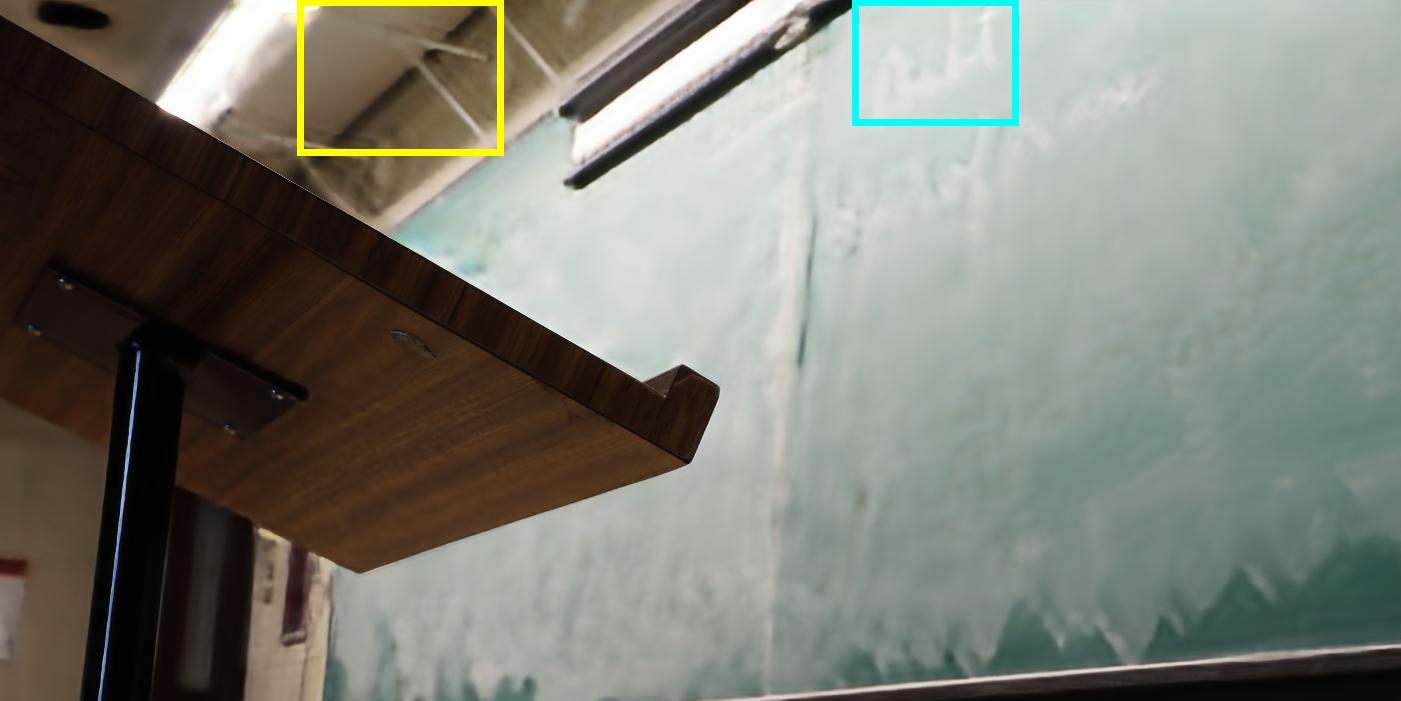}} &
    \multicolumn{2}{c}{\includegraphics[width=0.196\linewidth]{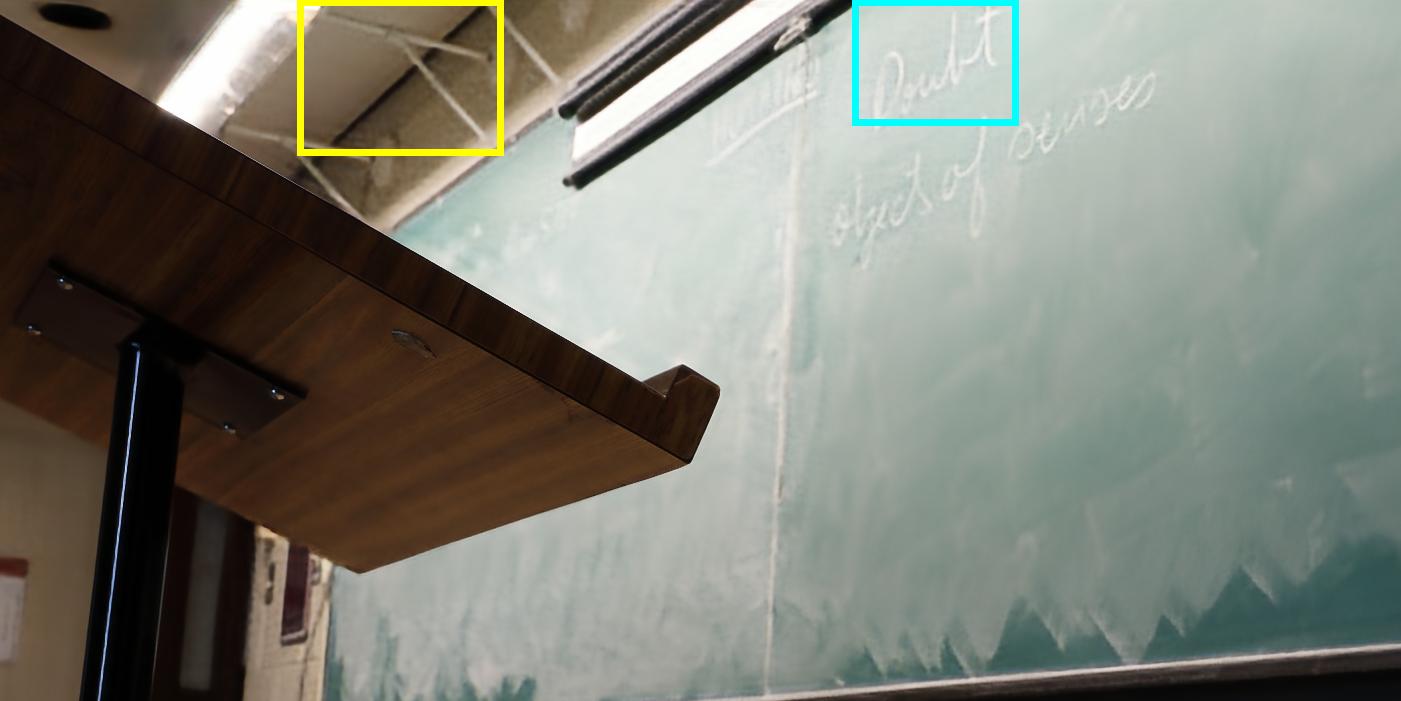}} &
    \multicolumn{2}{c}{\includegraphics[width=0.196\linewidth]{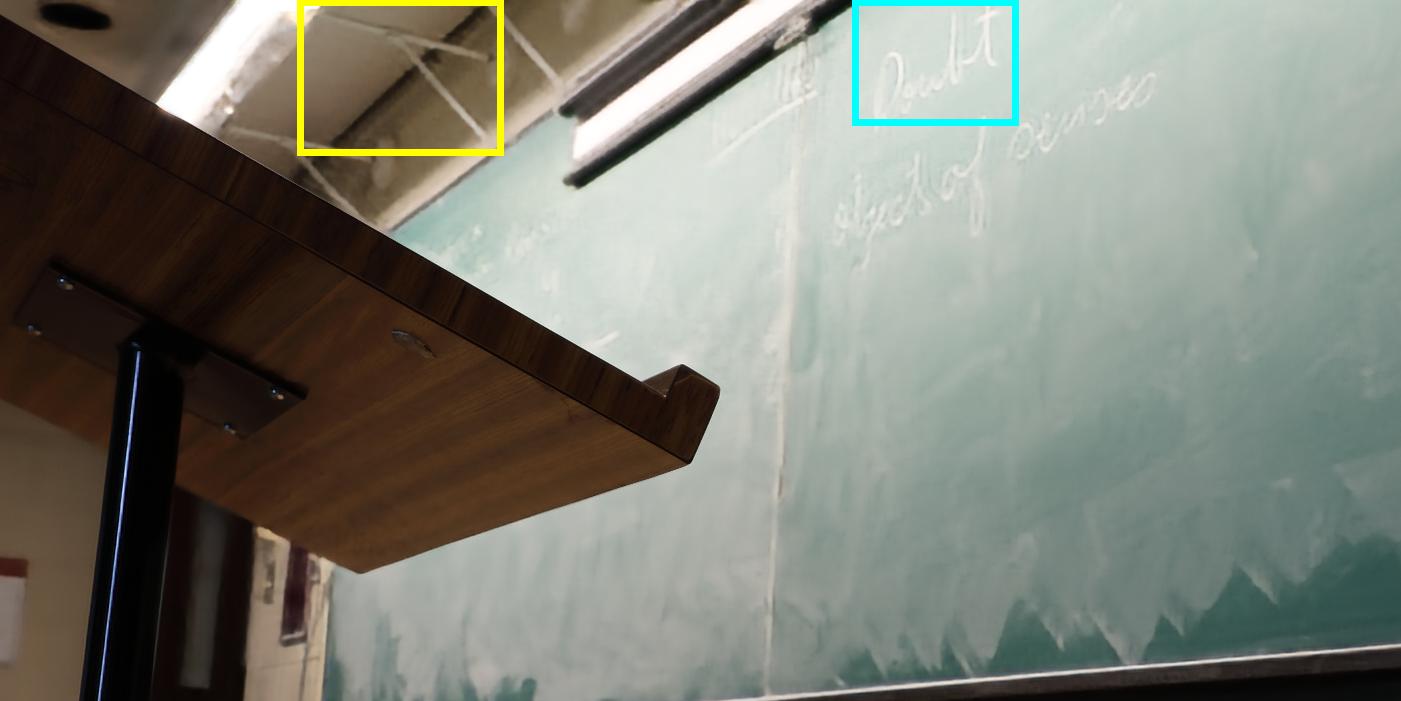}} &
    \multicolumn{2}{c}{\includegraphics[width=0.196\linewidth]{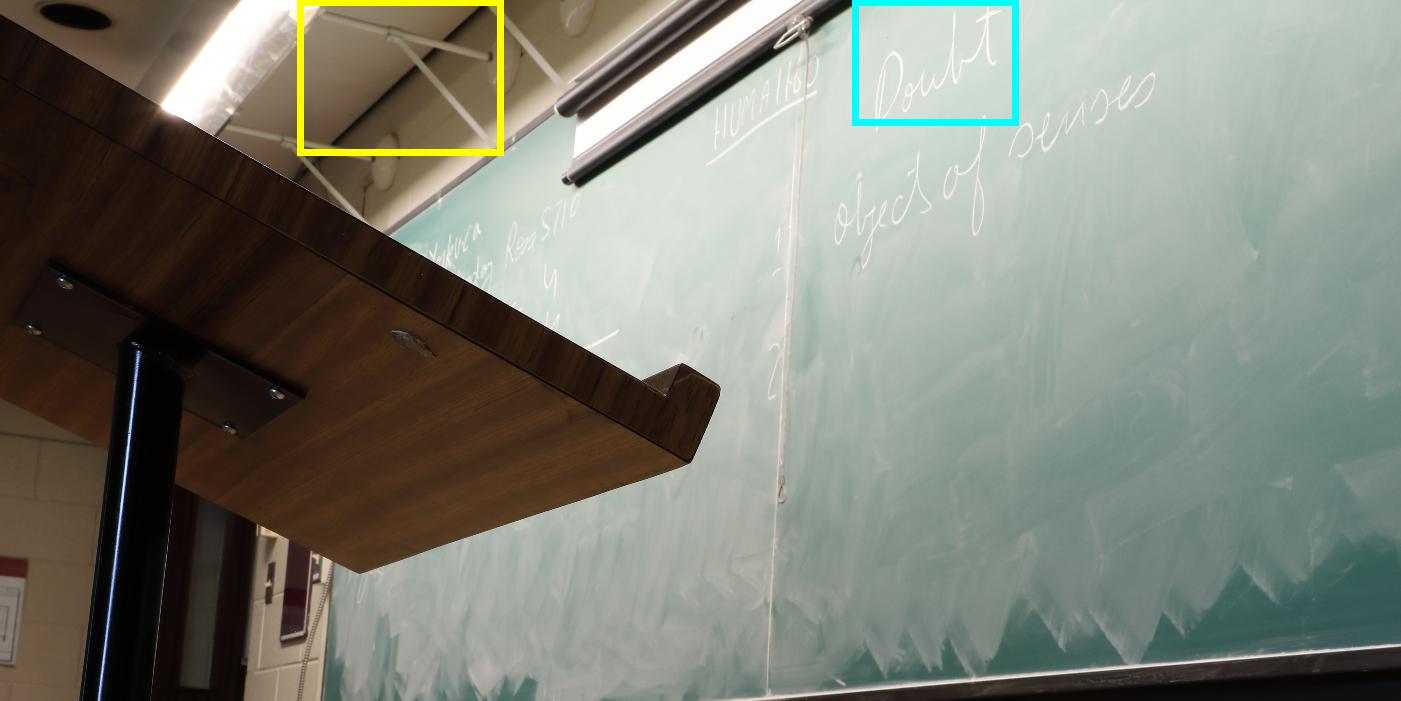}} \\[-0.02in]
    \multicolumn{1}{c}{\includegraphics[width=0.096\linewidth]{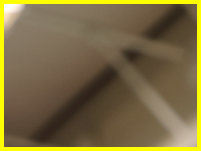}} &
    \multicolumn{1}{c}{\includegraphics[width=0.096\linewidth]{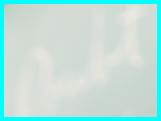}} &
    \multicolumn{1}{c}{\includegraphics[width=0.096\linewidth]{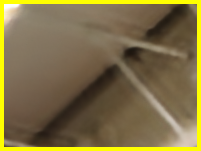}} &
    \multicolumn{1}{c}{\includegraphics[width=0.096\linewidth]{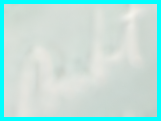}} &
    \multicolumn{1}{c}{\includegraphics[width=0.096\linewidth]{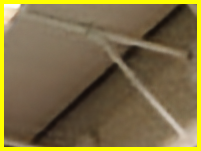}} &
    \multicolumn{1}{c}{\includegraphics[width=0.096\linewidth]{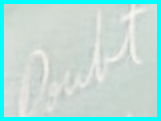}} &
    \multicolumn{1}{c}{\includegraphics[width=0.096\linewidth]{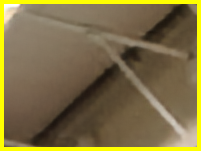}} &
    \multicolumn{1}{c}{\includegraphics[width=0.096\linewidth]{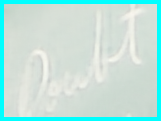}} &
    \multicolumn{1}{c}{\includegraphics[width=0.096\linewidth]{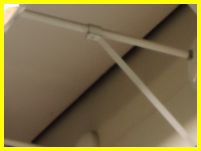}} &
    \multicolumn{1}{c}{\includegraphics[width=0.096\linewidth]{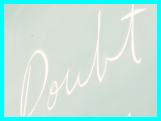}} \\
  
    \multicolumn{2}{c}{\includegraphics[width=0.196\linewidth]{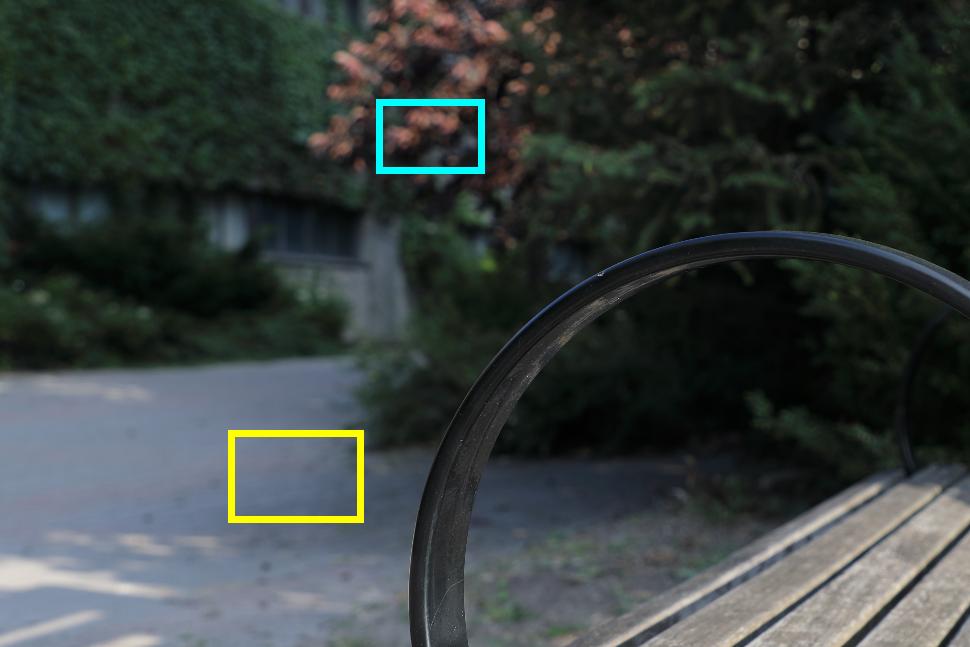}} &
    \multicolumn{2}{c}{\includegraphics[width=0.196\linewidth]{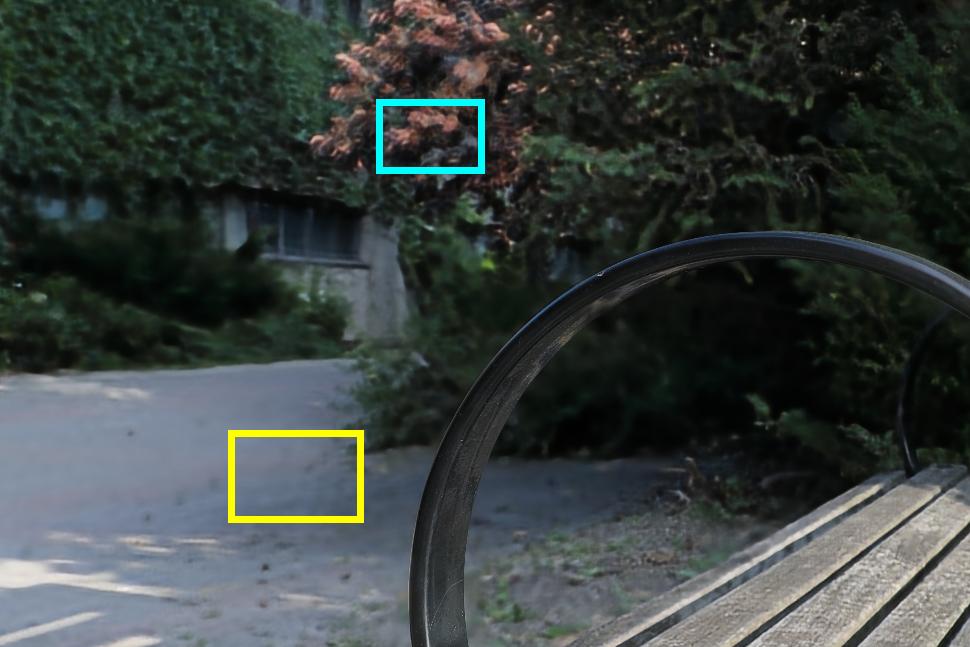}} &
    \multicolumn{2}{c}{\includegraphics[width=0.196\linewidth]{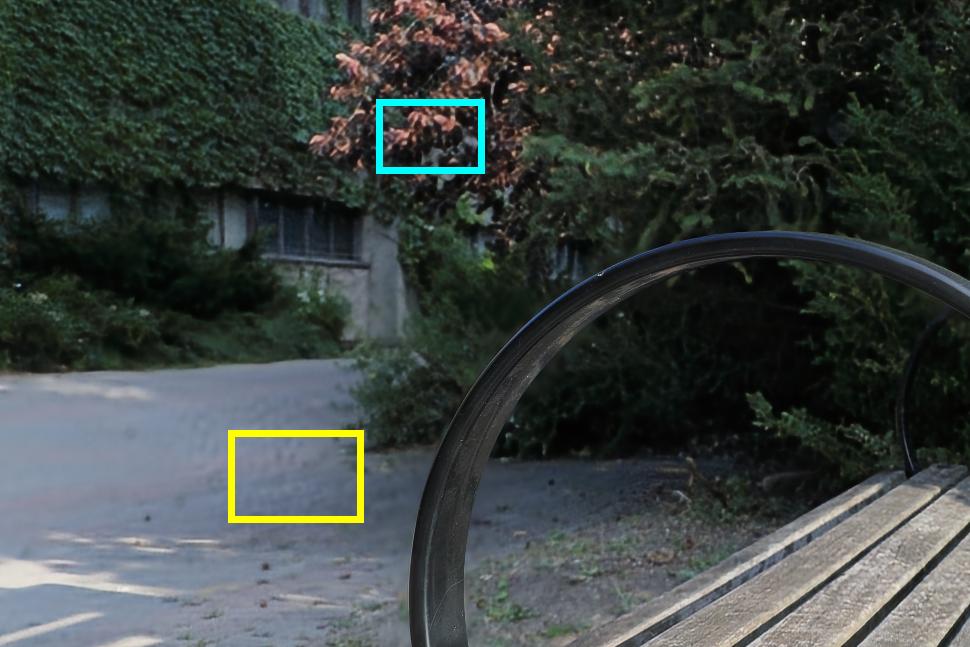}} &
    \multicolumn{2}{c}{\includegraphics[width=0.196\linewidth]{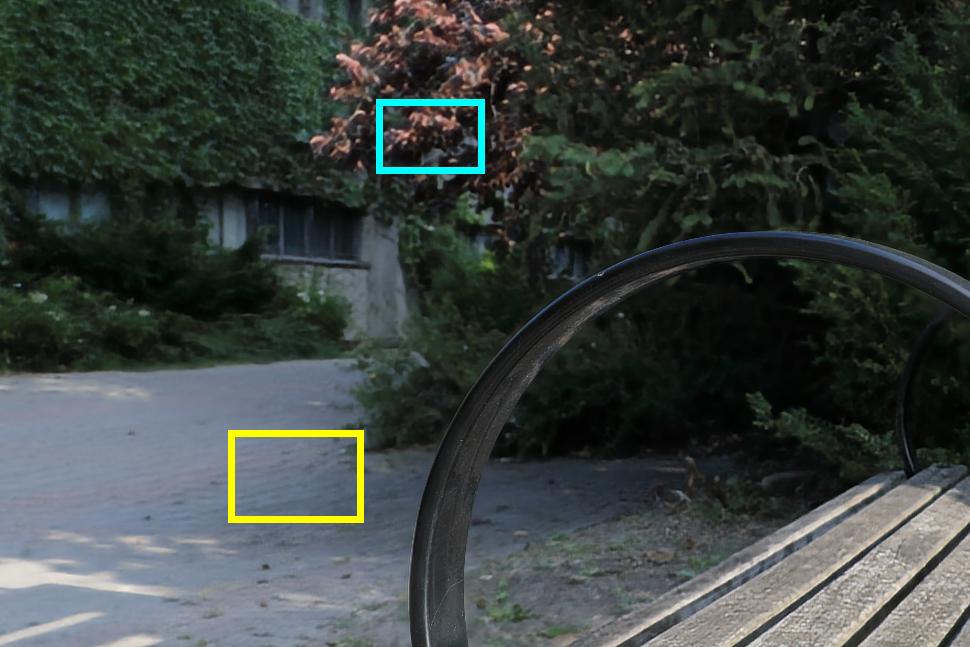}} &
    \multicolumn{2}{c}{\includegraphics[width=0.196\linewidth]{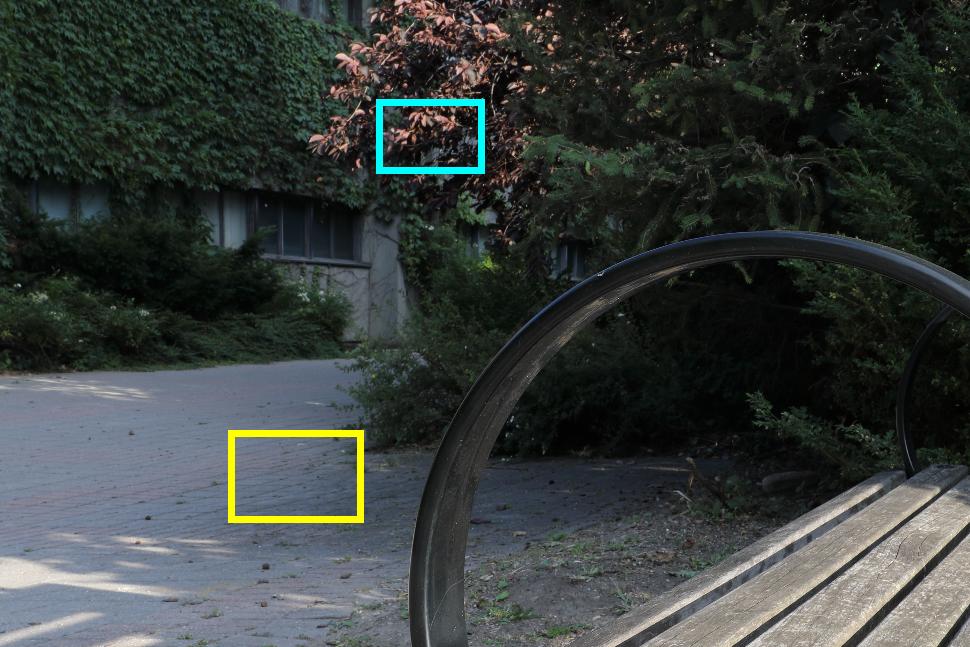}} \\
    \multicolumn{1}{c}{\includegraphics[width=0.096\linewidth]{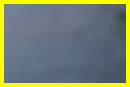}} &
    \multicolumn{1}{c}{\includegraphics[width=0.096\linewidth]{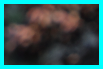}} &
    \multicolumn{1}{c}{\includegraphics[width=0.096\linewidth]{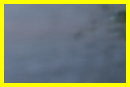}} &
    \multicolumn{1}{c}{\includegraphics[width=0.096\linewidth]{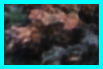}} &
    \multicolumn{1}{c}{\includegraphics[width=0.096\linewidth]{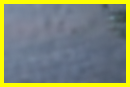}} &
    \multicolumn{1}{c}{\includegraphics[width=0.096\linewidth]{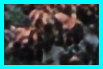}} &
    \multicolumn{1}{c}{\includegraphics[width=0.096\linewidth]{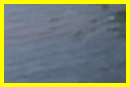}} &
    \multicolumn{1}{c}{\includegraphics[width=0.096\linewidth]{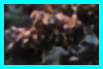}} &
    \multicolumn{1}{c}{\includegraphics[width=0.096\linewidth]{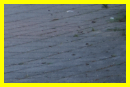}} &
    \multicolumn{1}{c}{\includegraphics[width=0.096\linewidth]{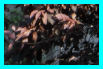}} \\

    \multicolumn{2}{c}{\includegraphics[width=0.196\linewidth]{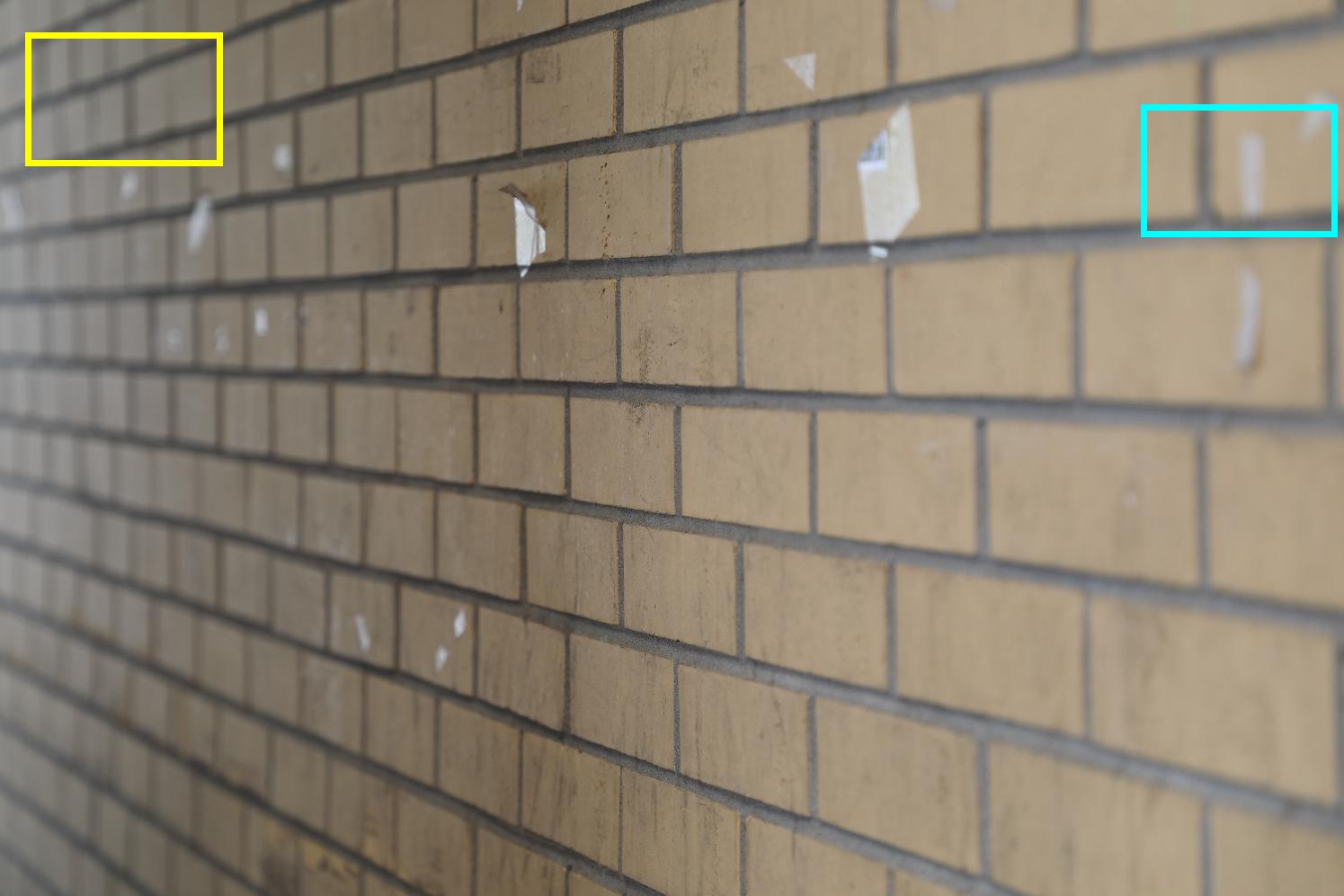}} &
    \multicolumn{2}{c}{\includegraphics[width=0.196\linewidth]{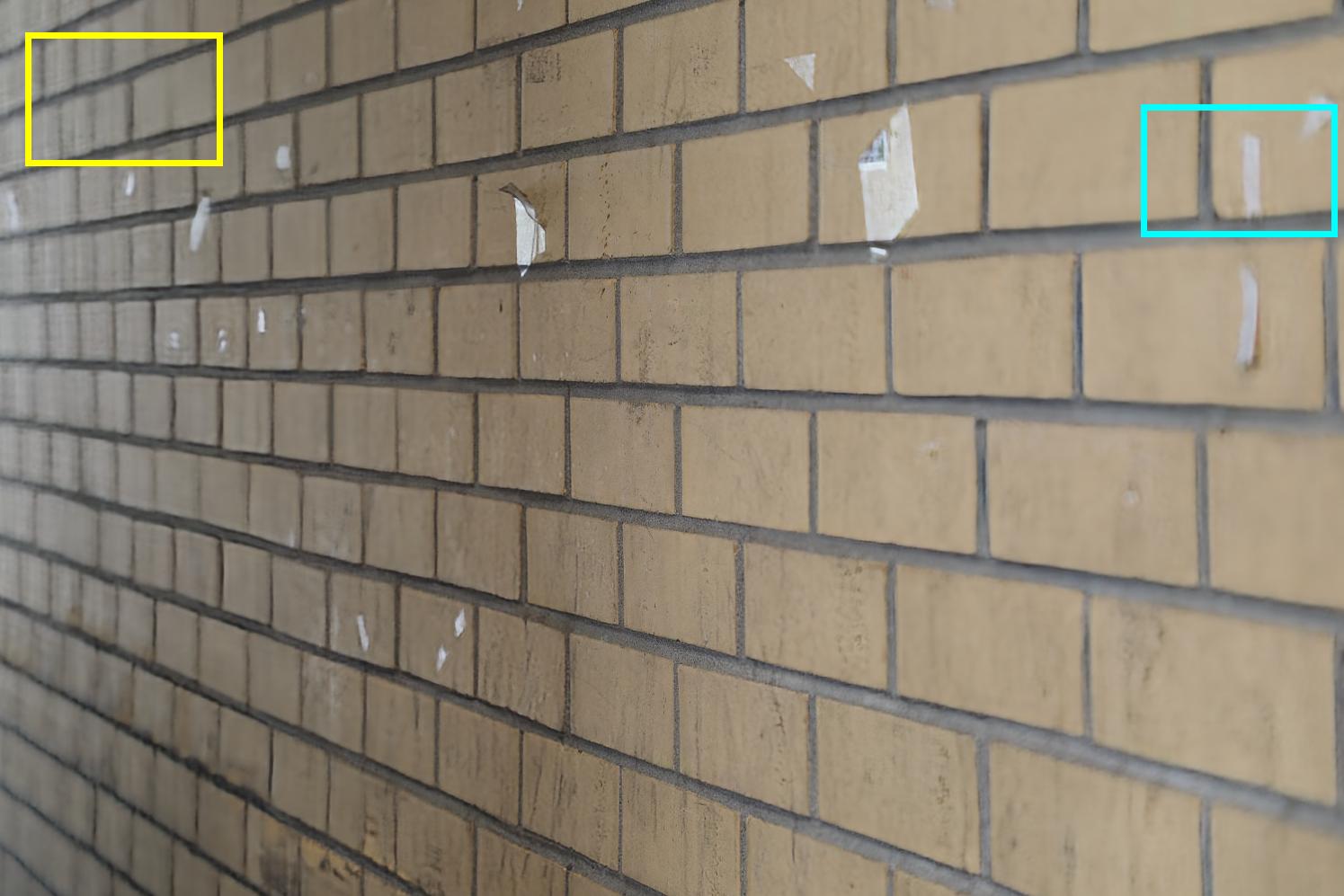}} &
    \multicolumn{2}{c}{\includegraphics[width=0.196\linewidth]{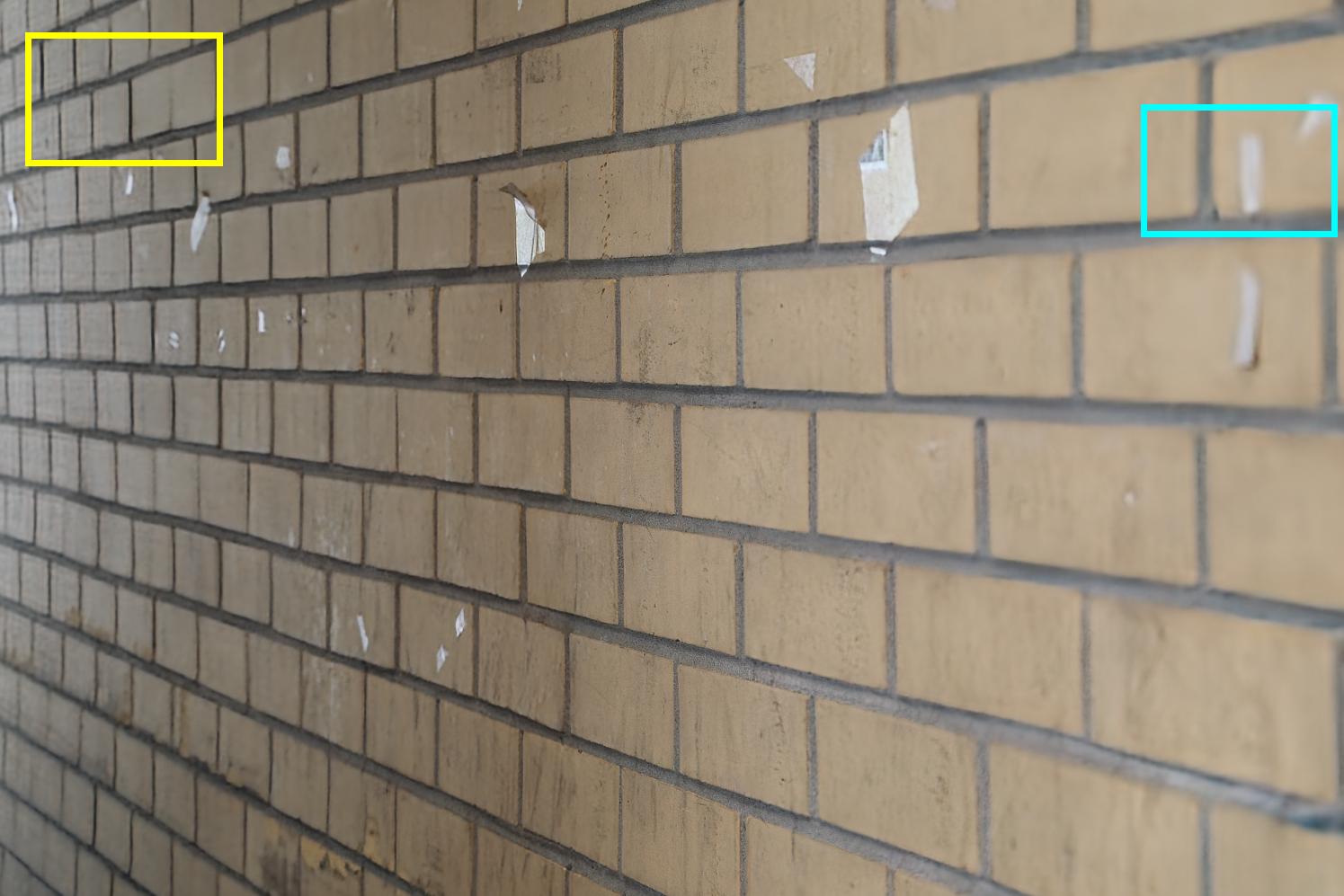}} &
    \multicolumn{2}{c}{\includegraphics[width=0.196\linewidth]{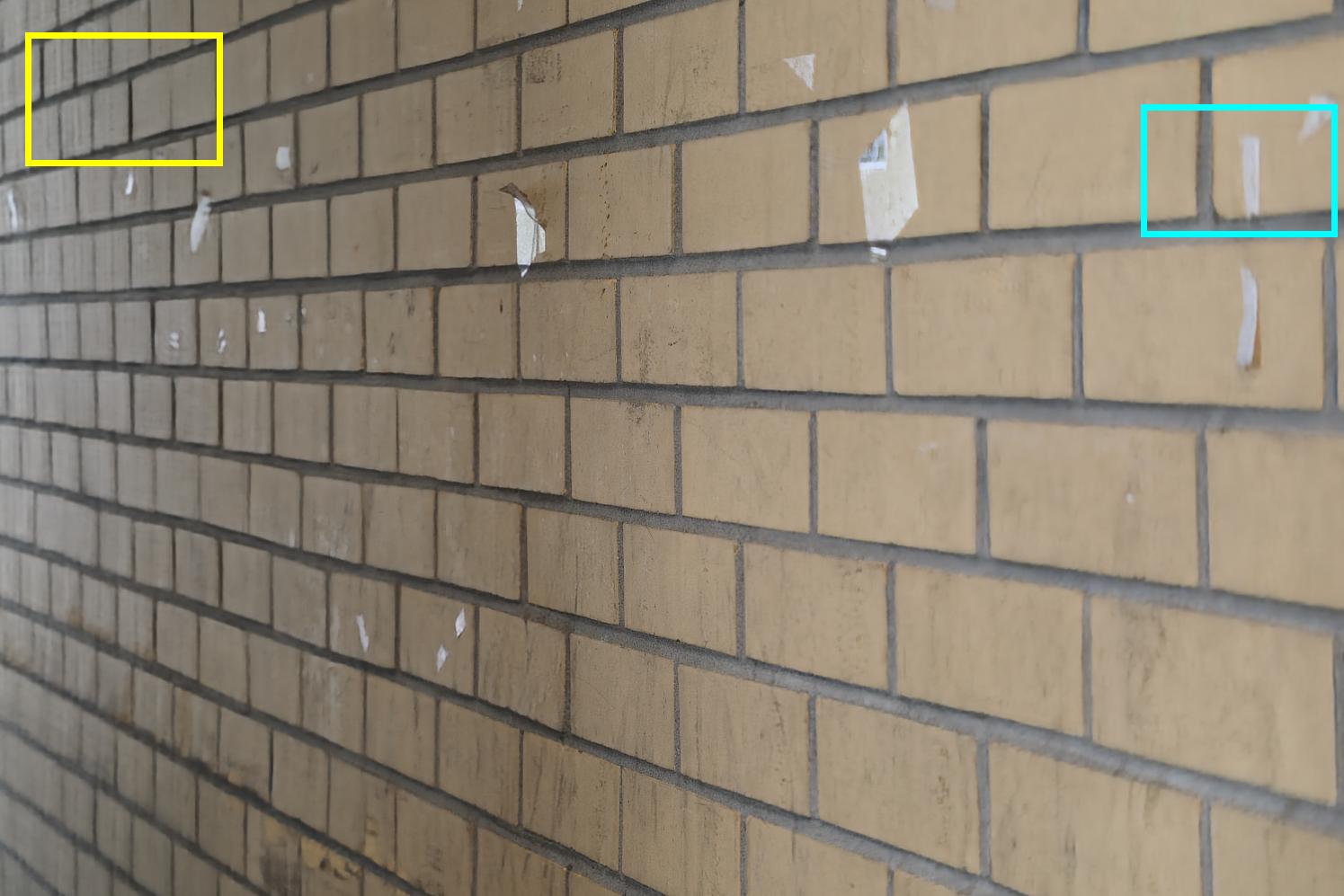}} &
    \multicolumn{2}{c}{\includegraphics[width=0.196\linewidth]{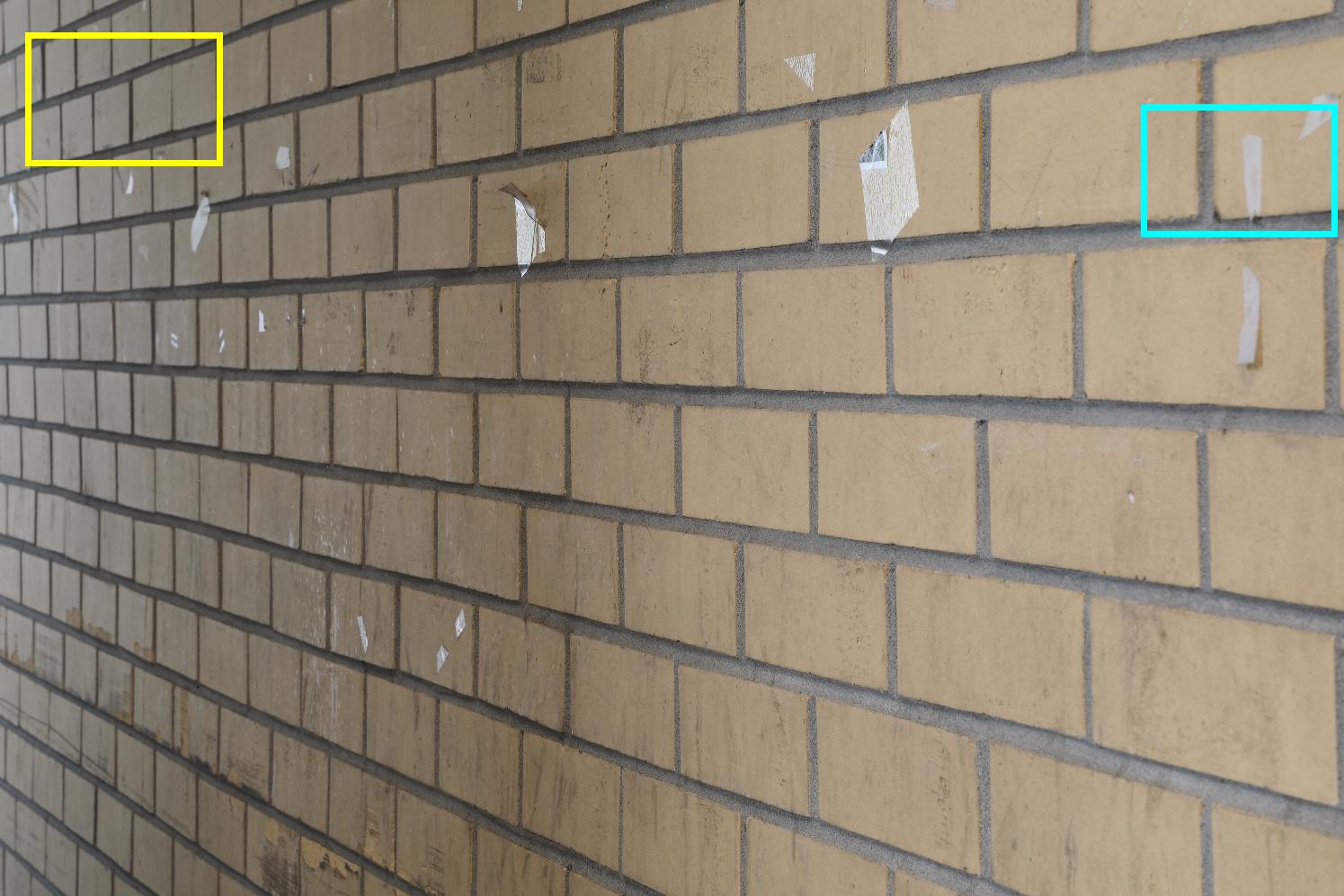}} \\
    \multicolumn{1}{c}{\includegraphics[width=0.096\linewidth]{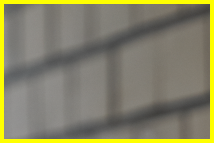}} &
    \multicolumn{1}{c}{\includegraphics[width=0.096\linewidth]{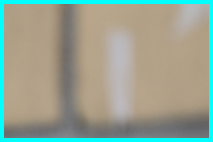}} &
    \multicolumn{1}{c}{\includegraphics[width=0.096\linewidth]{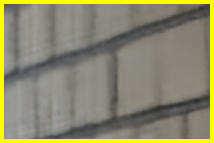}} &
    \multicolumn{1}{c}{\includegraphics[width=0.096\linewidth]{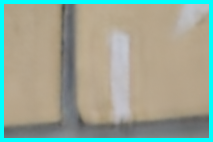}} &
    \multicolumn{1}{c}{\includegraphics[width=0.096\linewidth]{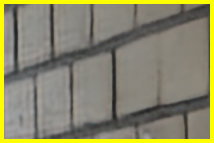}} &
    \multicolumn{1}{c}{\includegraphics[width=0.096\linewidth]{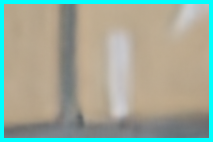}} &
    \multicolumn{1}{c}{\includegraphics[width=0.096\linewidth]{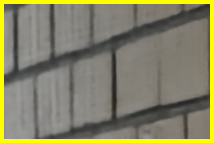}} &
    \multicolumn{1}{c}{\includegraphics[width=0.096\linewidth]{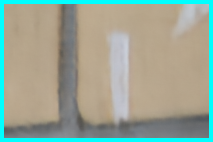}} &
    \multicolumn{1}{c}{\includegraphics[width=0.096\linewidth]{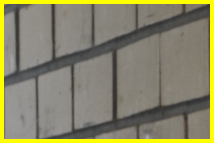}} &
    \multicolumn{1}{c}{\includegraphics[width=0.096\linewidth]{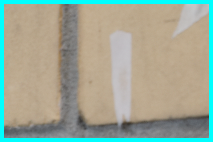}} \\

    \multicolumn{2}{c}{\includegraphics[width=0.196\linewidth]{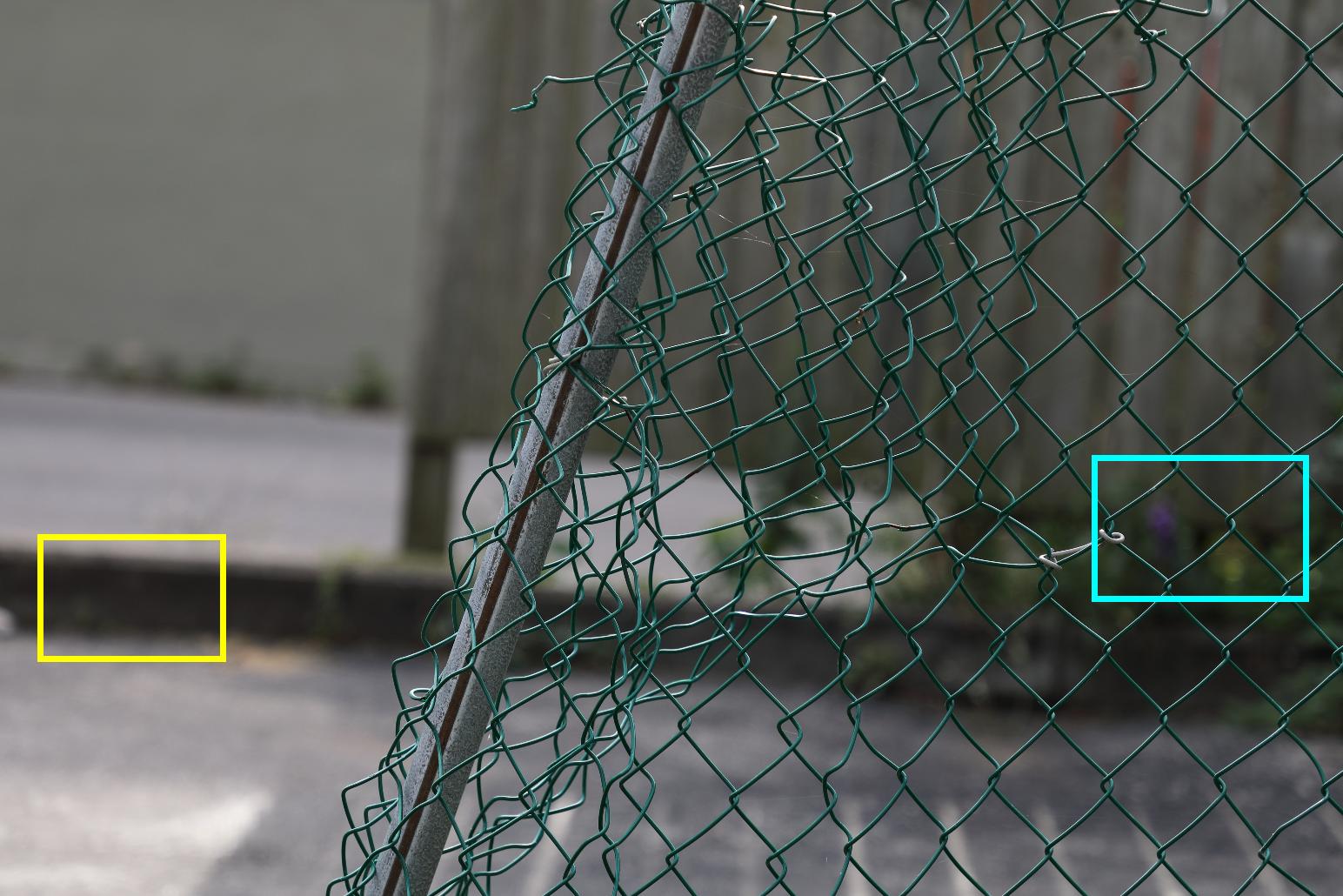}} &
    \multicolumn{2}{c}{\includegraphics[width=0.196\linewidth]{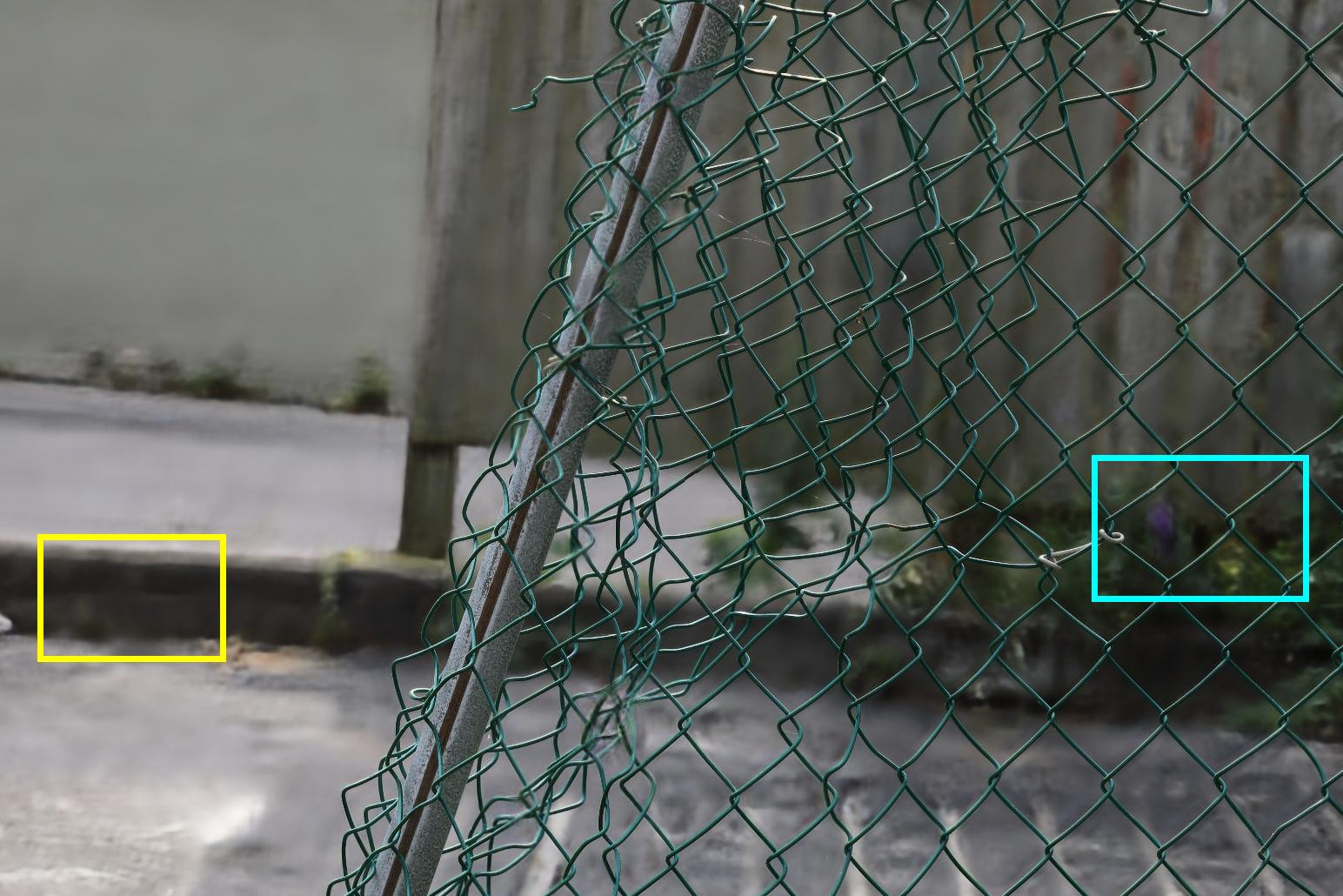}} &
    \multicolumn{2}{c}{\includegraphics[width=0.196\linewidth]{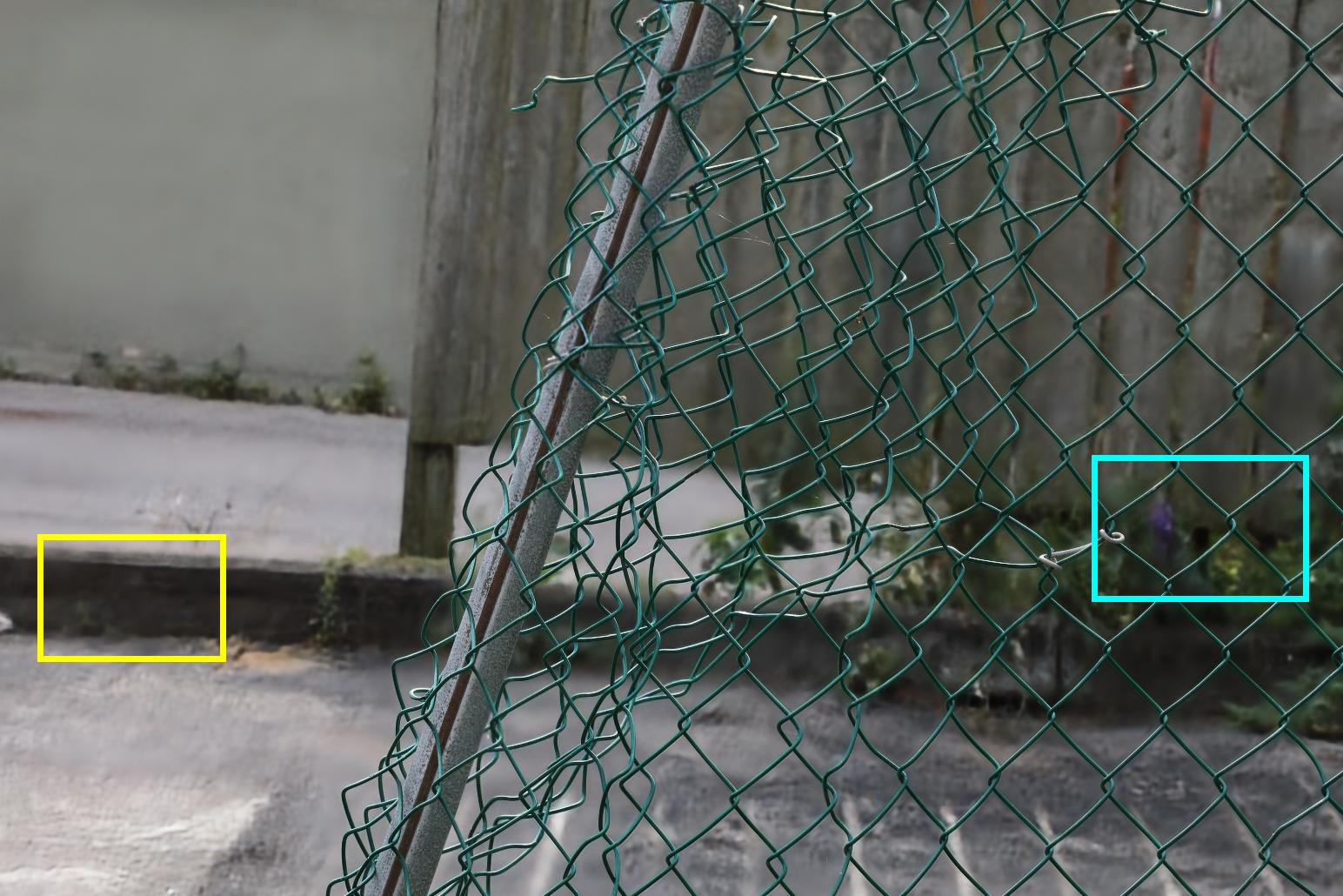}} &
    \multicolumn{2}{c}{\includegraphics[width=0.196\linewidth]{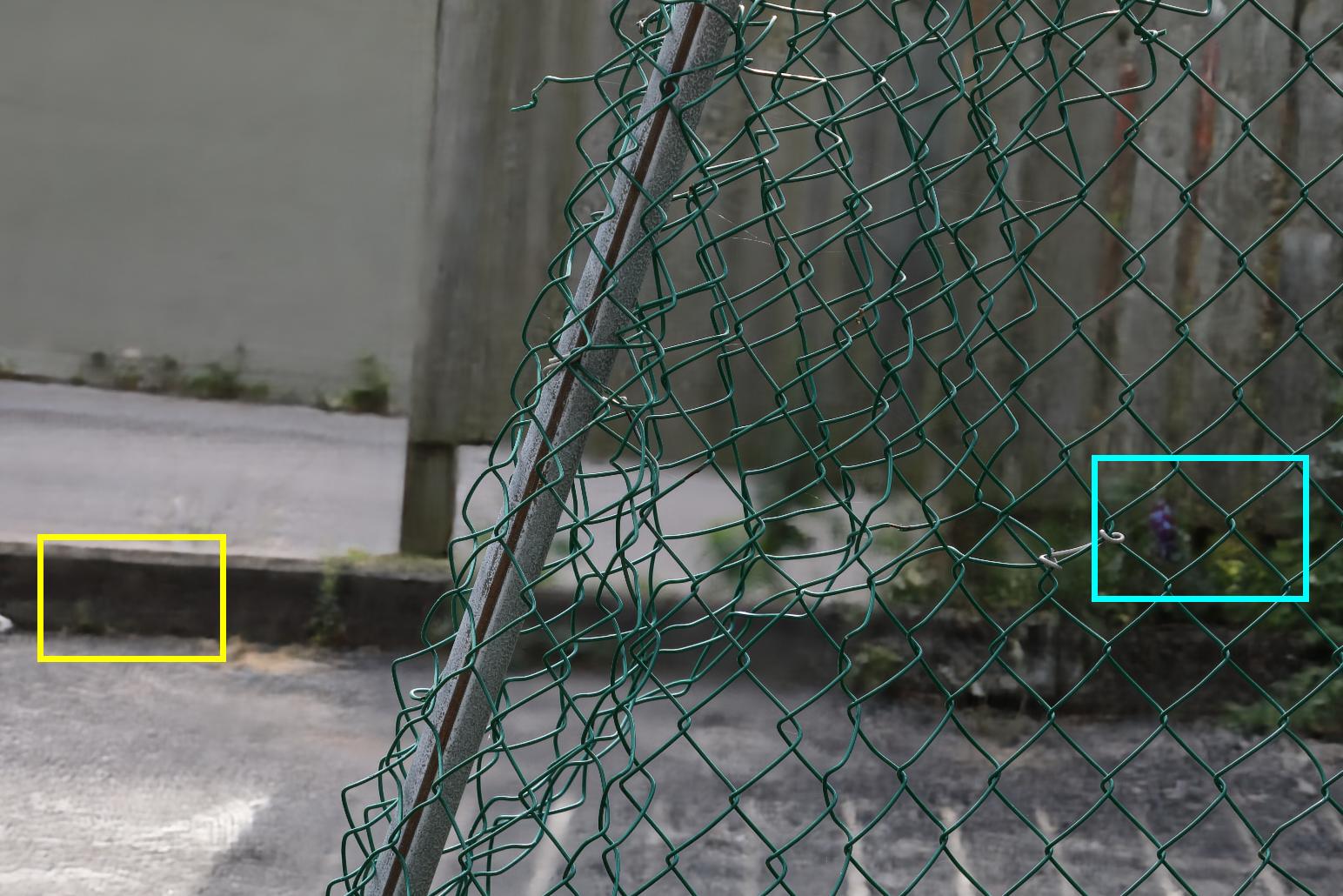}} &
    \multicolumn{2}{c}{\includegraphics[width=0.196\linewidth]{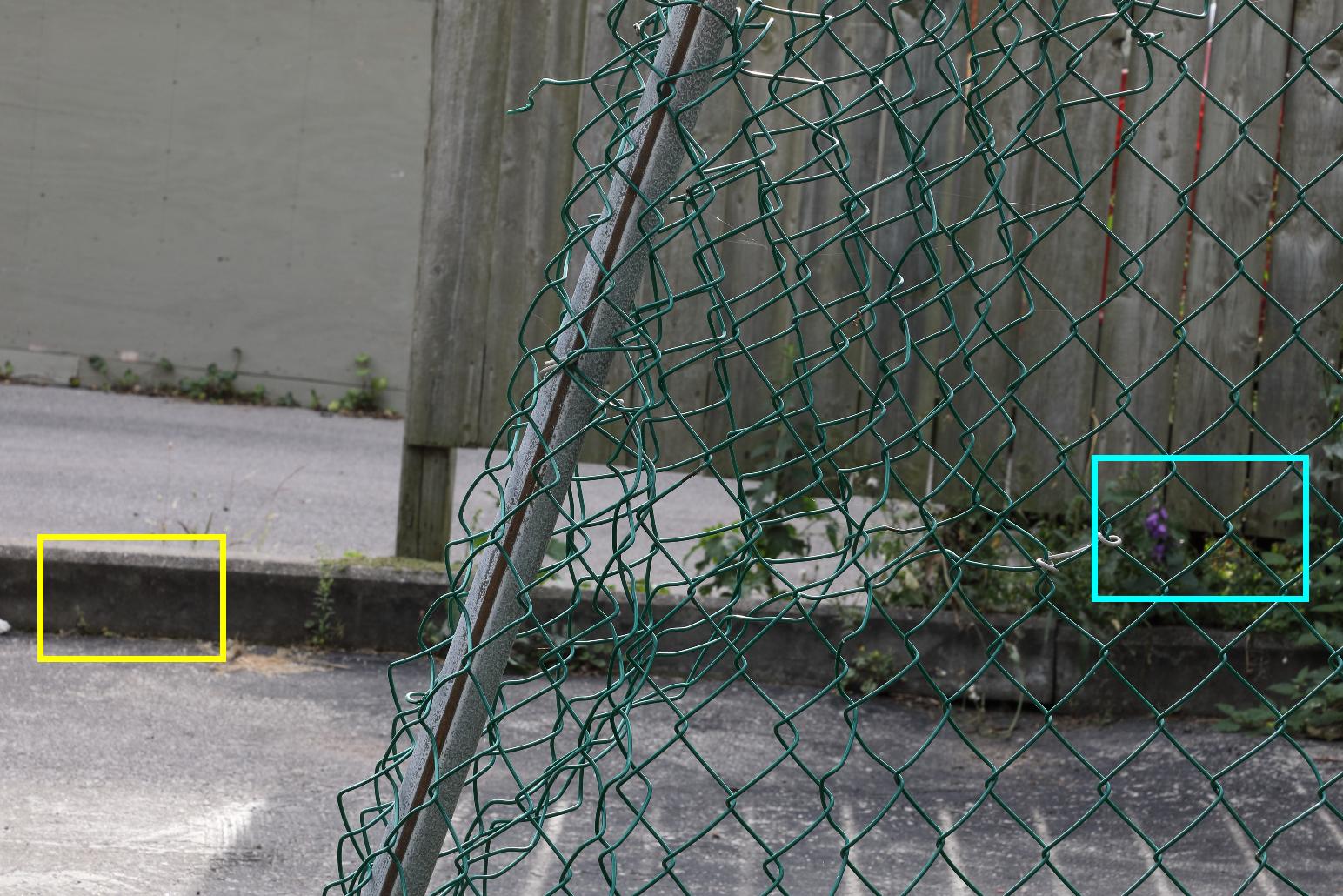}} \\
    \multicolumn{1}{c}{\includegraphics[width=0.096\linewidth]{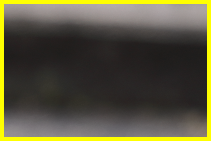}} &
    \multicolumn{1}{c}{\includegraphics[width=0.096\linewidth]{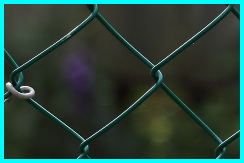}} &
    \multicolumn{1}{c}{\includegraphics[width=0.096\linewidth]{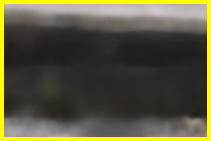}} &
    \multicolumn{1}{c}{\includegraphics[width=0.096\linewidth]{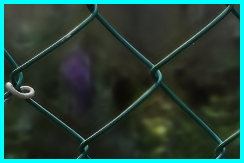}} &
    \multicolumn{1}{c}{\includegraphics[width=0.096\linewidth]{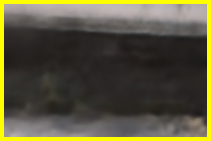}} &
    \multicolumn{1}{c}{\includegraphics[width=0.096\linewidth]{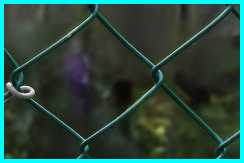}} &
    \multicolumn{1}{c}{\includegraphics[width=0.096\linewidth]{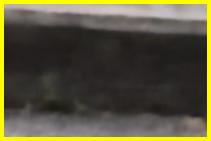}} &
    \multicolumn{1}{c}{\includegraphics[width=0.096\linewidth]{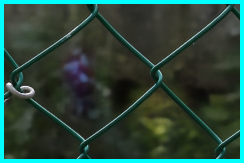}} &
    \multicolumn{1}{c}{\includegraphics[width=0.096\linewidth]{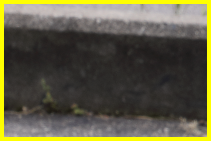}} &
    \multicolumn{1}{c}{\includegraphics[width=0.096\linewidth]{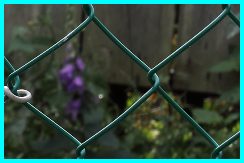}} \\

    \multicolumn{2}{c}{\includegraphics[width=0.196\linewidth]{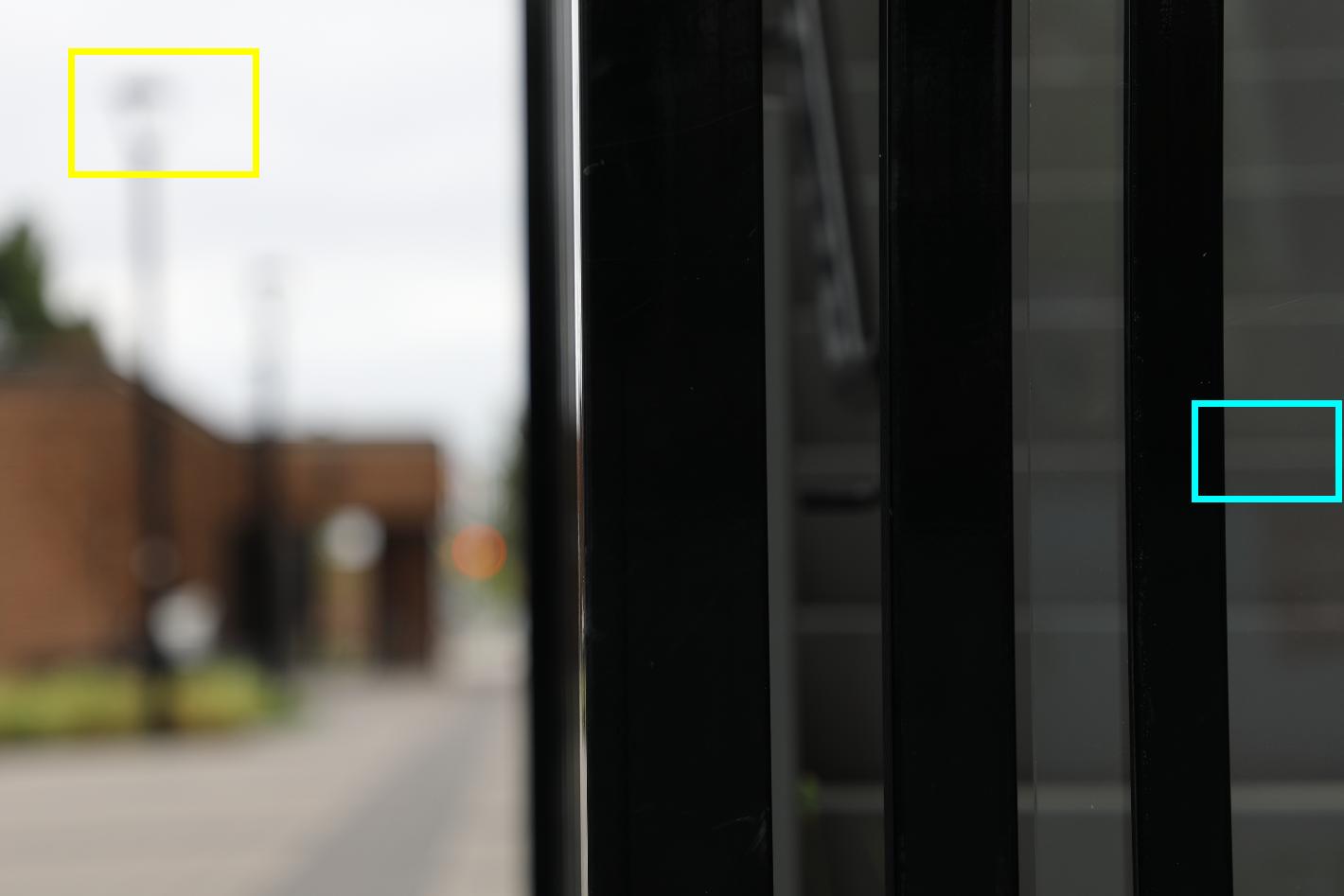}} &
    \multicolumn{2}{c}{\includegraphics[width=0.196\linewidth]{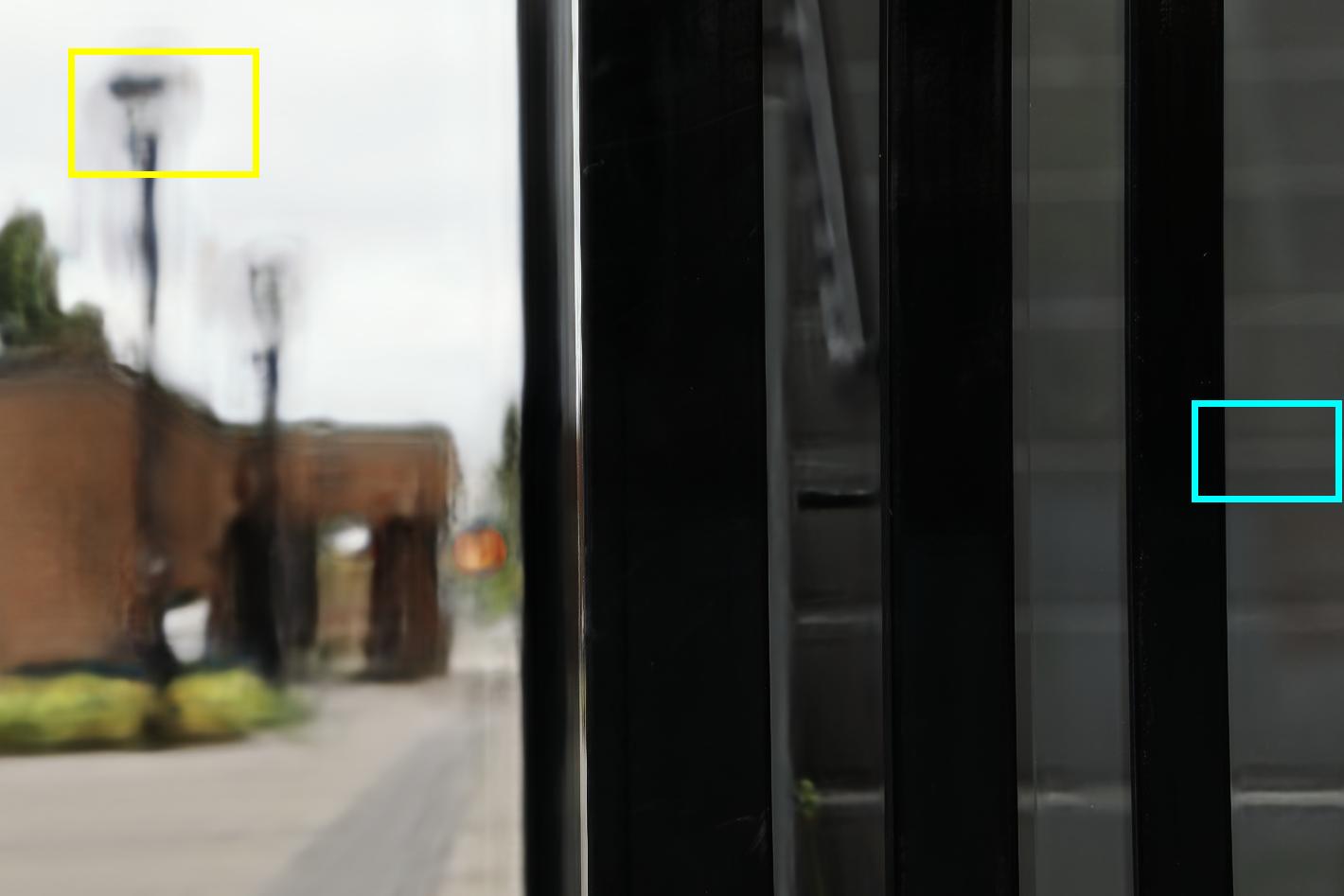}} &
    \multicolumn{2}{c}{\includegraphics[width=0.196\linewidth]{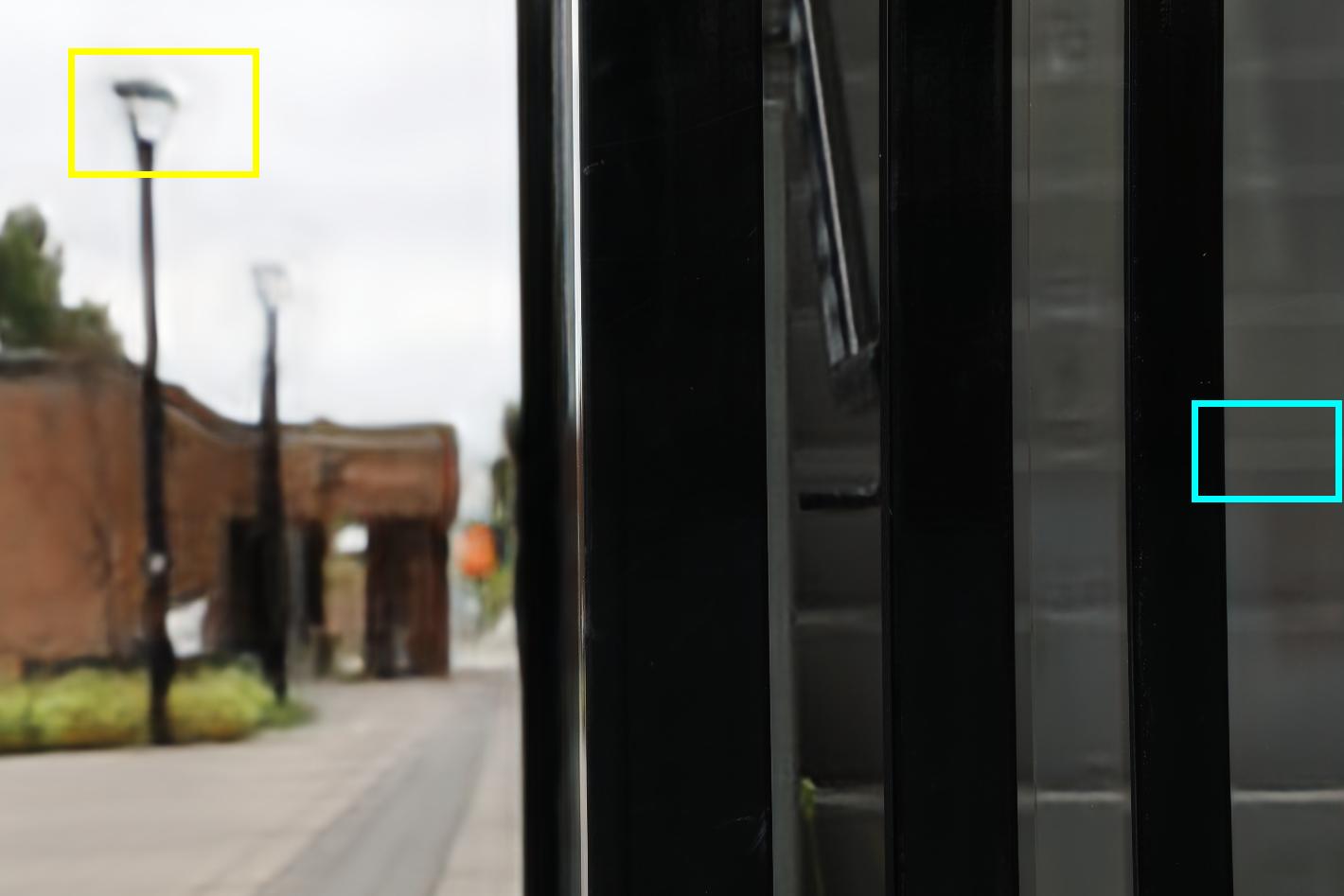}} &
    \multicolumn{2}{c}{\includegraphics[width=0.196\linewidth]{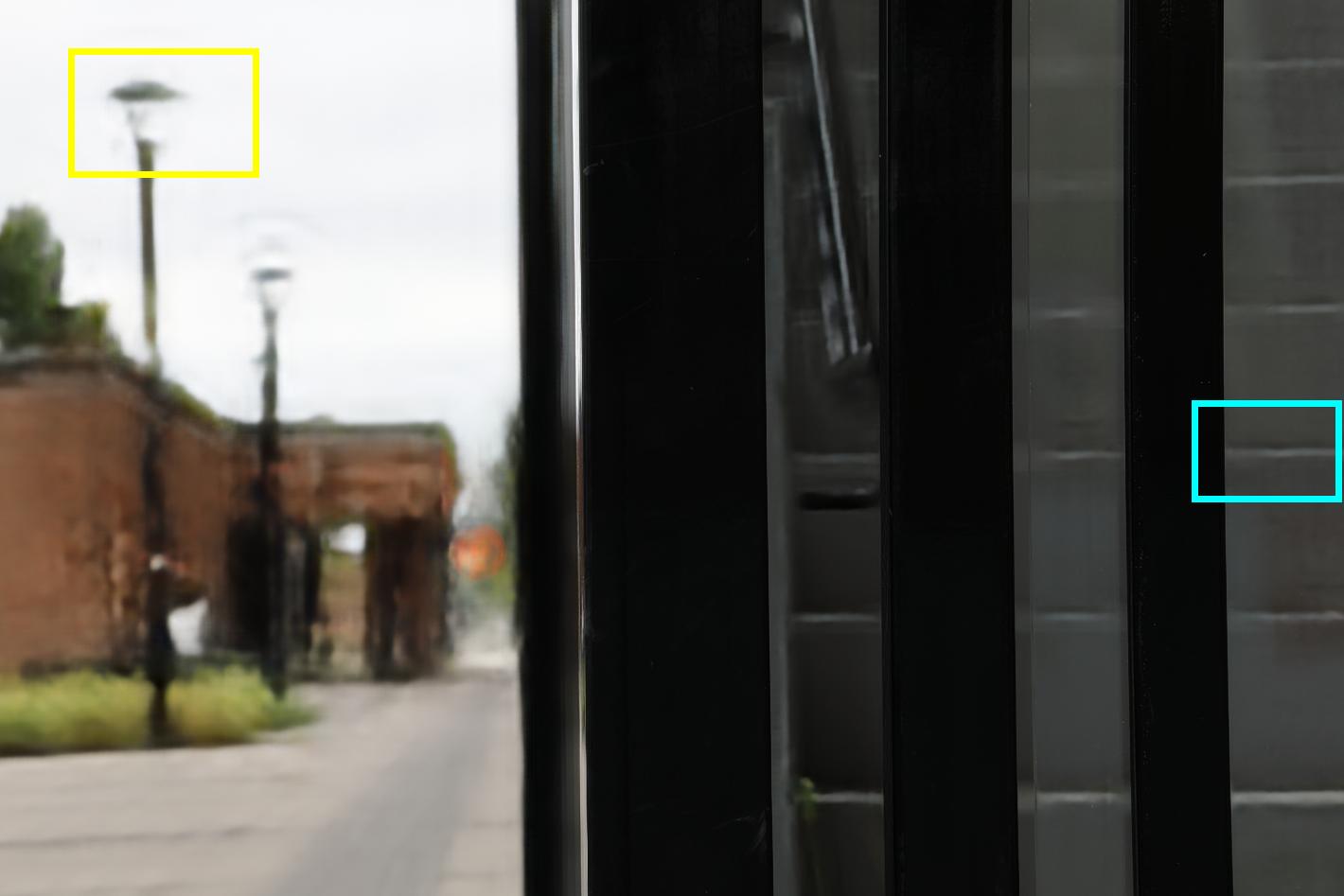}} &
    \multicolumn{2}{c}{\includegraphics[width=0.196\linewidth]{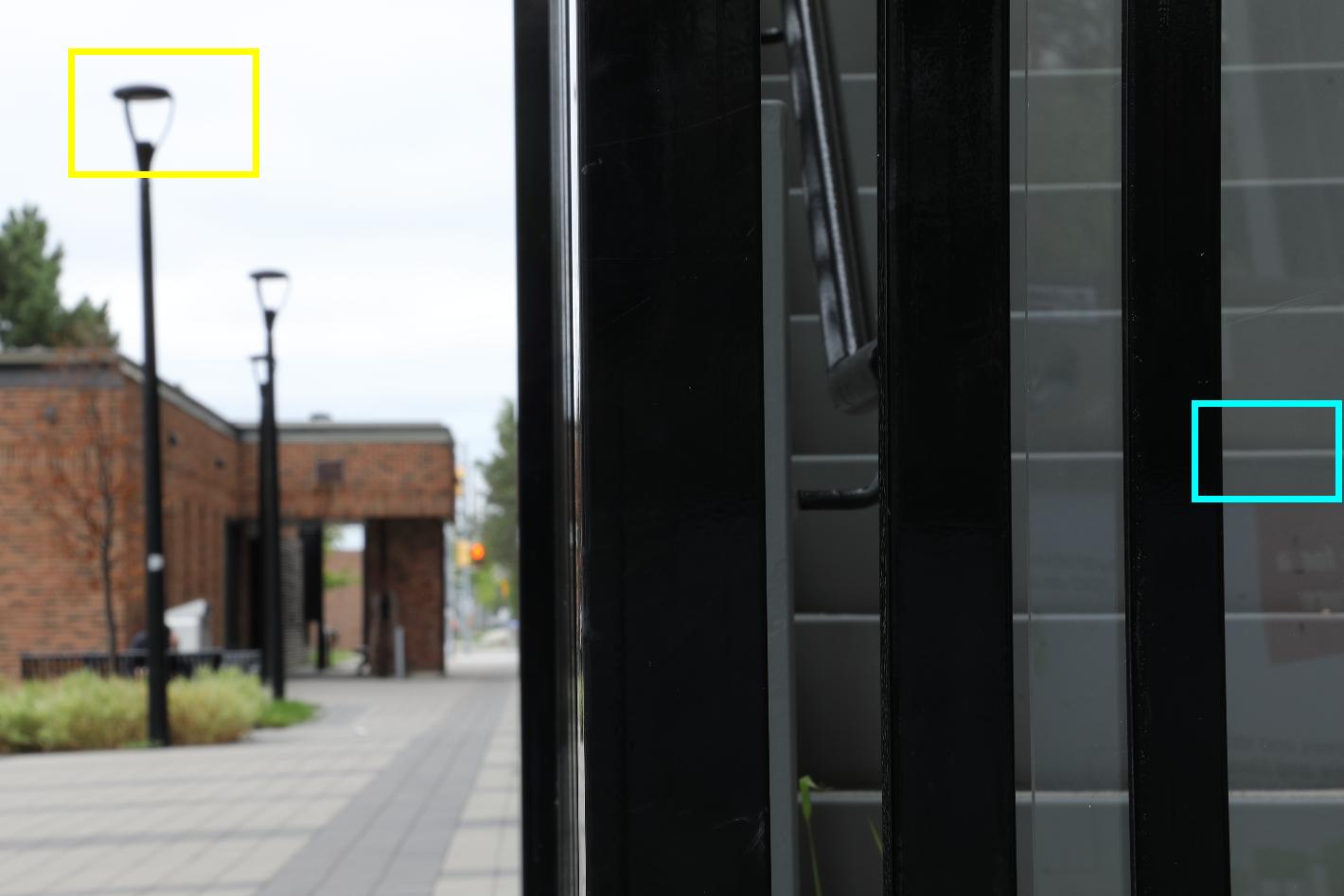}} \\
    \multicolumn{1}{c}{\includegraphics[width=0.096\linewidth]{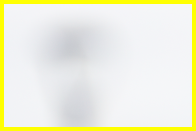}} &
    \multicolumn{1}{c}{\includegraphics[width=0.096\linewidth]{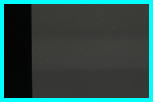}} &
    \multicolumn{1}{c}{\includegraphics[width=0.096\linewidth]{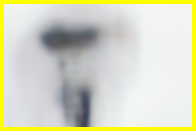}} &
    \multicolumn{1}{c}{\includegraphics[width=0.096\linewidth]{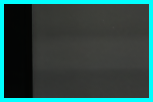}} &
    \multicolumn{1}{c}{\includegraphics[width=0.096\linewidth]{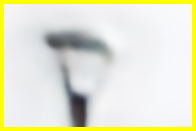}} &
    \multicolumn{1}{c}{\includegraphics[width=0.096\linewidth]{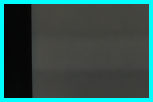}} &
    \multicolumn{1}{c}{\includegraphics[width=0.096\linewidth]{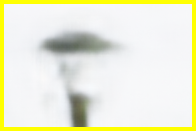}} &
    \multicolumn{1}{c}{\includegraphics[width=0.096\linewidth]{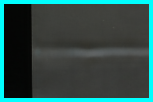}} &
    \multicolumn{1}{c}{\includegraphics[width=0.096\linewidth]{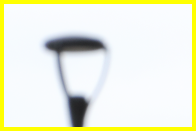}} &
    \multicolumn{1}{c}{\includegraphics[width=0.096\linewidth]{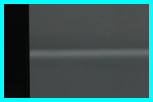}} \\

    \multicolumn{2}{c}{(a) input}  & \multicolumn{2}{c}{(b) DPDNet (single)~\cite{Abuolaim:2020:DPDNet}}
     & \multicolumn{2}{c}{(c) DPDNet (dual)~\cite{Abuolaim:2020:DPDNet} } & \multicolumn{2}{c}{(d) ours } & \multicolumn{2}{c}{(e) GT} \\

  \end{tabular}
  \vspace{-0.05cm}
  \caption{Additional qualitative comparisons with DPDNet~\cite{Abuolaim:2020:DPDNet} on the test set of the DPDD dataset \cite{Abuolaim:2020:DPDNet}.}
\label{fig:dpdd1}
\vspace{-10pt}
\end{figure*}

\begin{figure*}[tp]
\centering
\setlength\tabcolsep{1 pt}
  \begin{tabular}{cccccccccccc}

    \multicolumn{2}{c}{\includegraphics[width=0.196\linewidth]{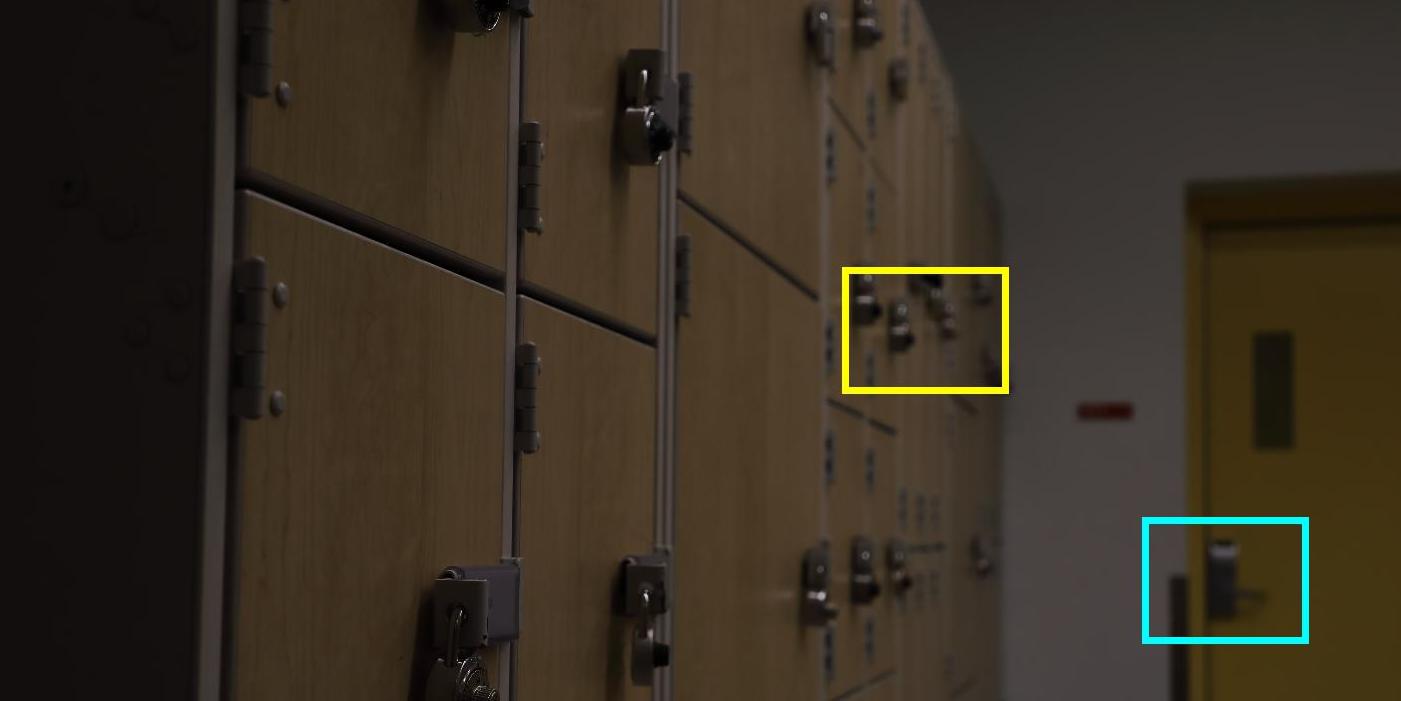}} &
    \multicolumn{2}{c}{\includegraphics[width=0.196\linewidth]{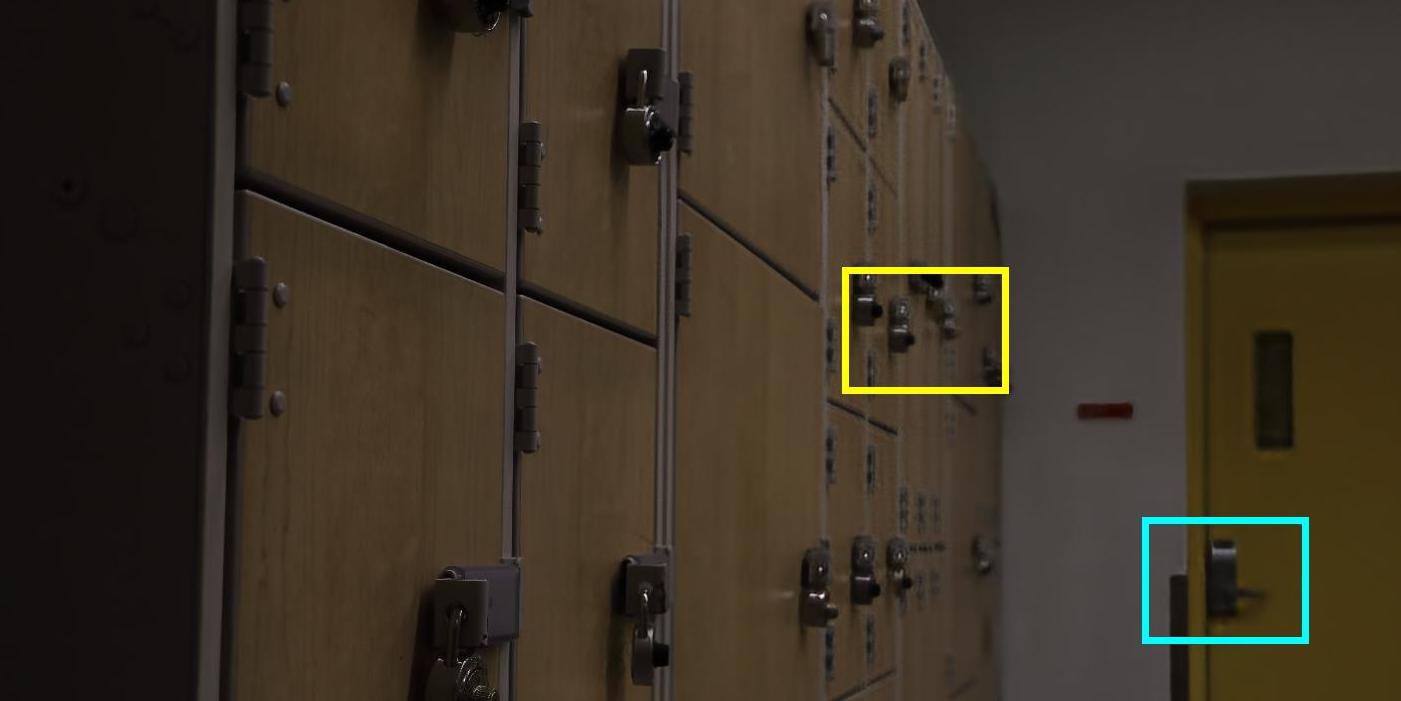}} &
    \multicolumn{2}{c}{\includegraphics[width=0.196\linewidth]{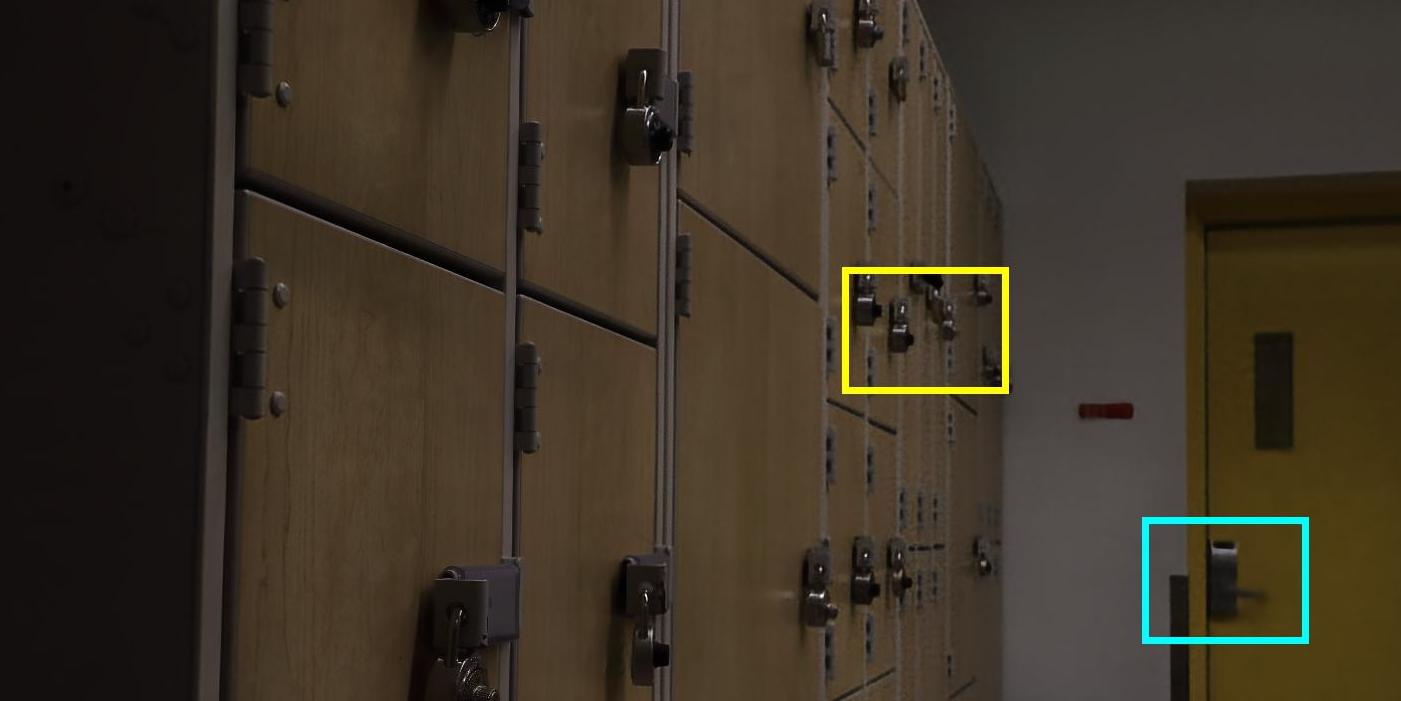}} &
    \multicolumn{2}{c}{\includegraphics[width=0.196\linewidth]{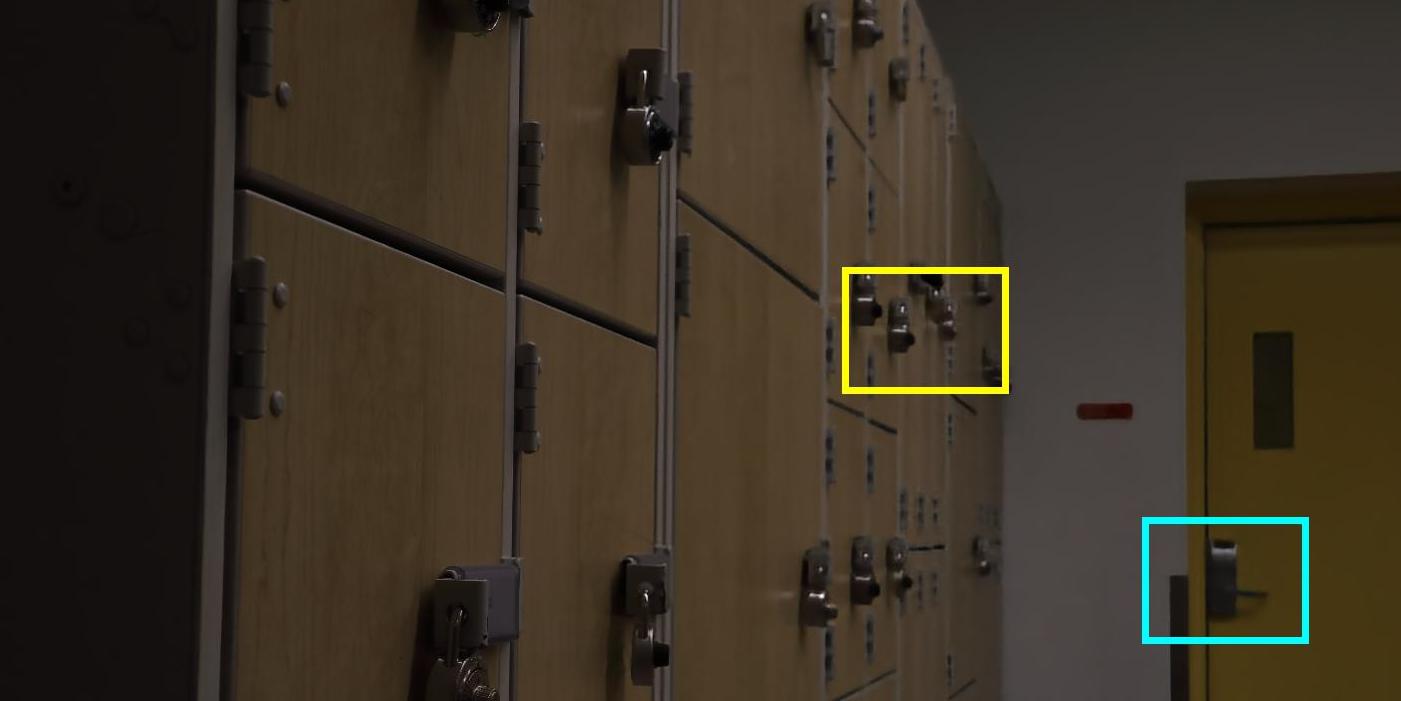}} &
    \multicolumn{2}{c}{\includegraphics[width=0.196\linewidth]{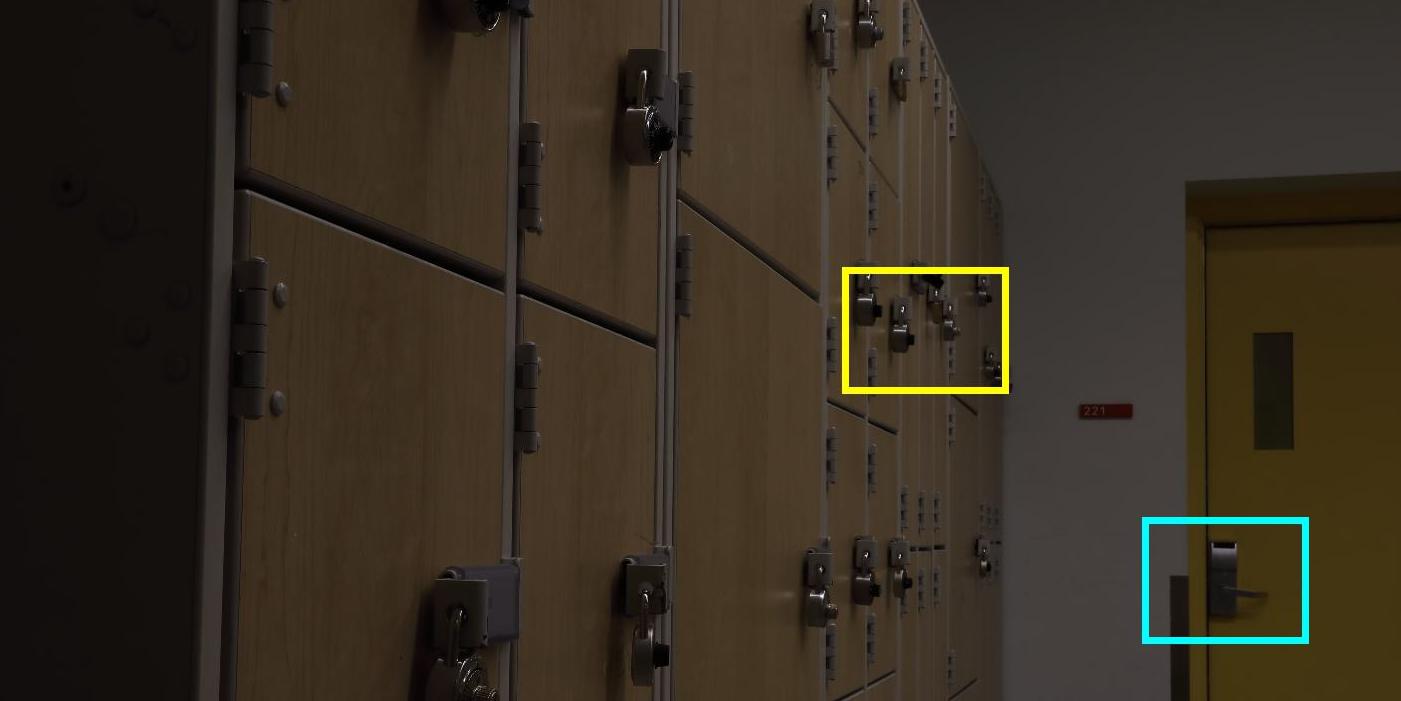}} \\
    \multicolumn{1}{c}{\includegraphics[width=0.096\linewidth]{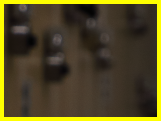}} &
    \multicolumn{1}{c}{\includegraphics[width=0.096\linewidth]{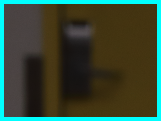}} &
    \multicolumn{1}{c}{\includegraphics[width=0.096\linewidth]{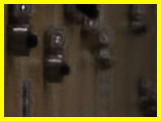}} &
    \multicolumn{1}{c}{\includegraphics[width=0.096\linewidth]{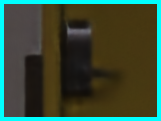}} &
    \multicolumn{1}{c}{\includegraphics[width=0.096\linewidth]{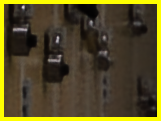}} &
    \multicolumn{1}{c}{\includegraphics[width=0.096\linewidth]{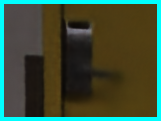}} &
    \multicolumn{1}{c}{\includegraphics[width=0.096\linewidth]{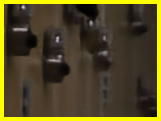}} &
    \multicolumn{1}{c}{\includegraphics[width=0.096\linewidth]{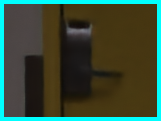}} &
    \multicolumn{1}{c}{\includegraphics[width=0.096\linewidth]{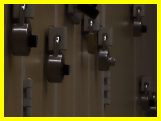}} &
    \multicolumn{1}{c}{\includegraphics[width=0.096\linewidth]{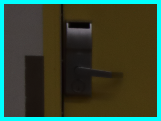}} \\

    \multicolumn{2}{c}{\includegraphics[width=0.196\linewidth]{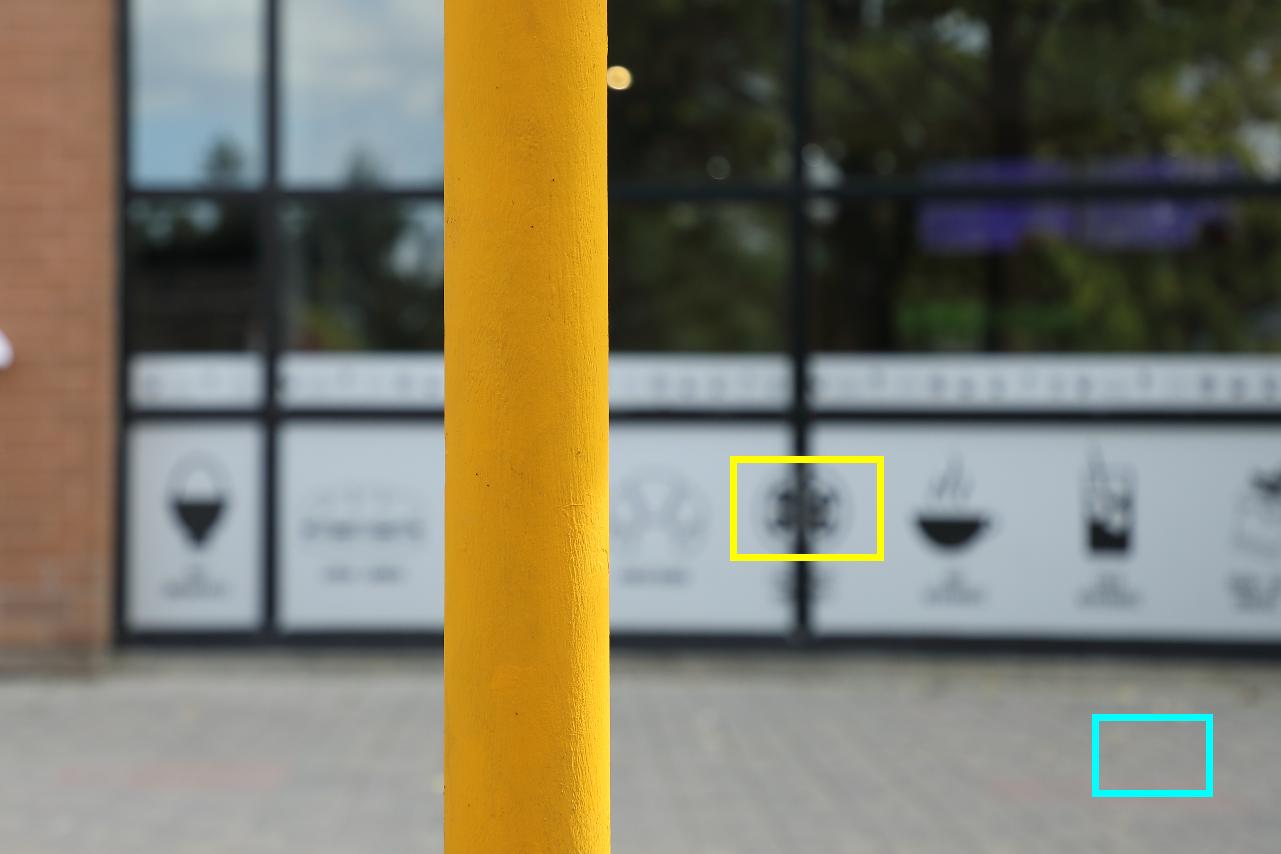}} &
    \multicolumn{2}{c}{\includegraphics[width=0.196\linewidth]{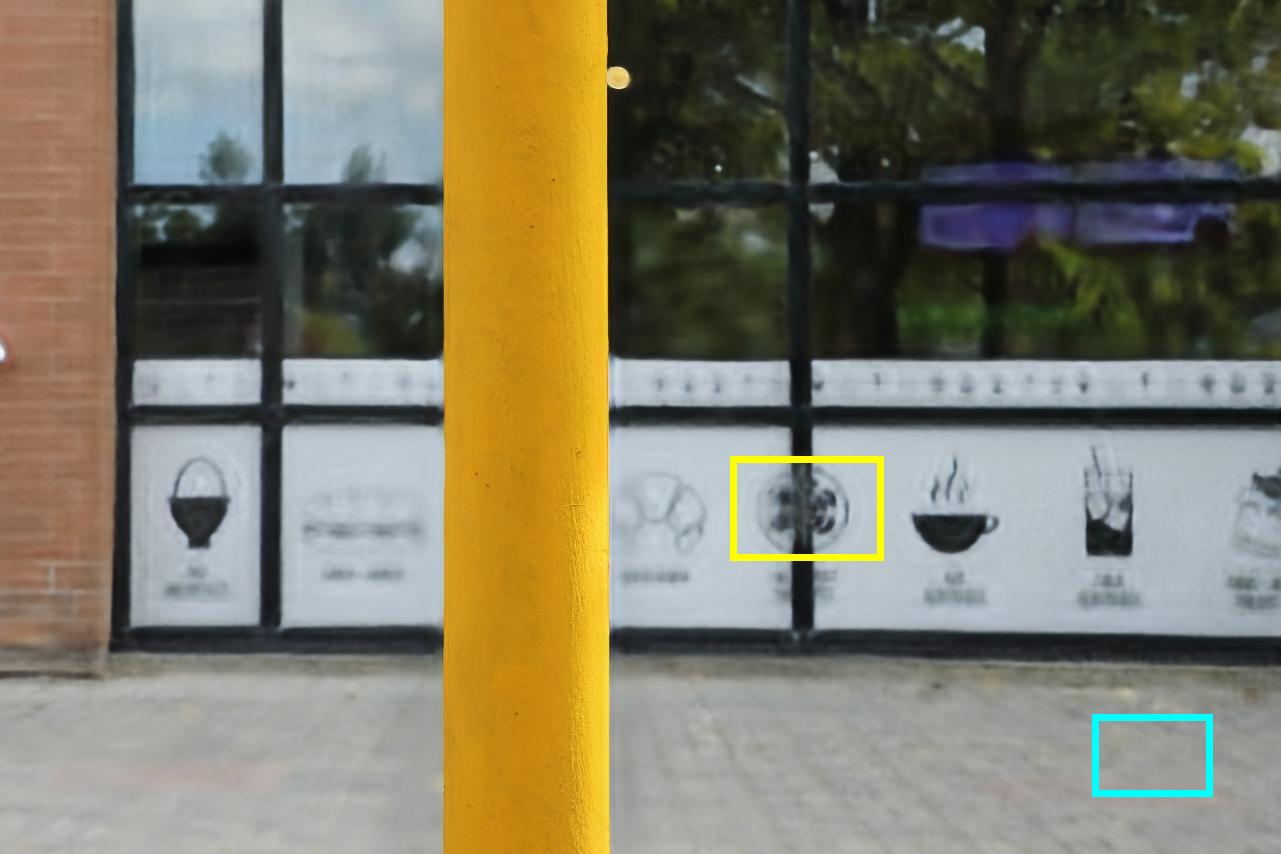}} &
    \multicolumn{2}{c}{\includegraphics[width=0.196\linewidth]{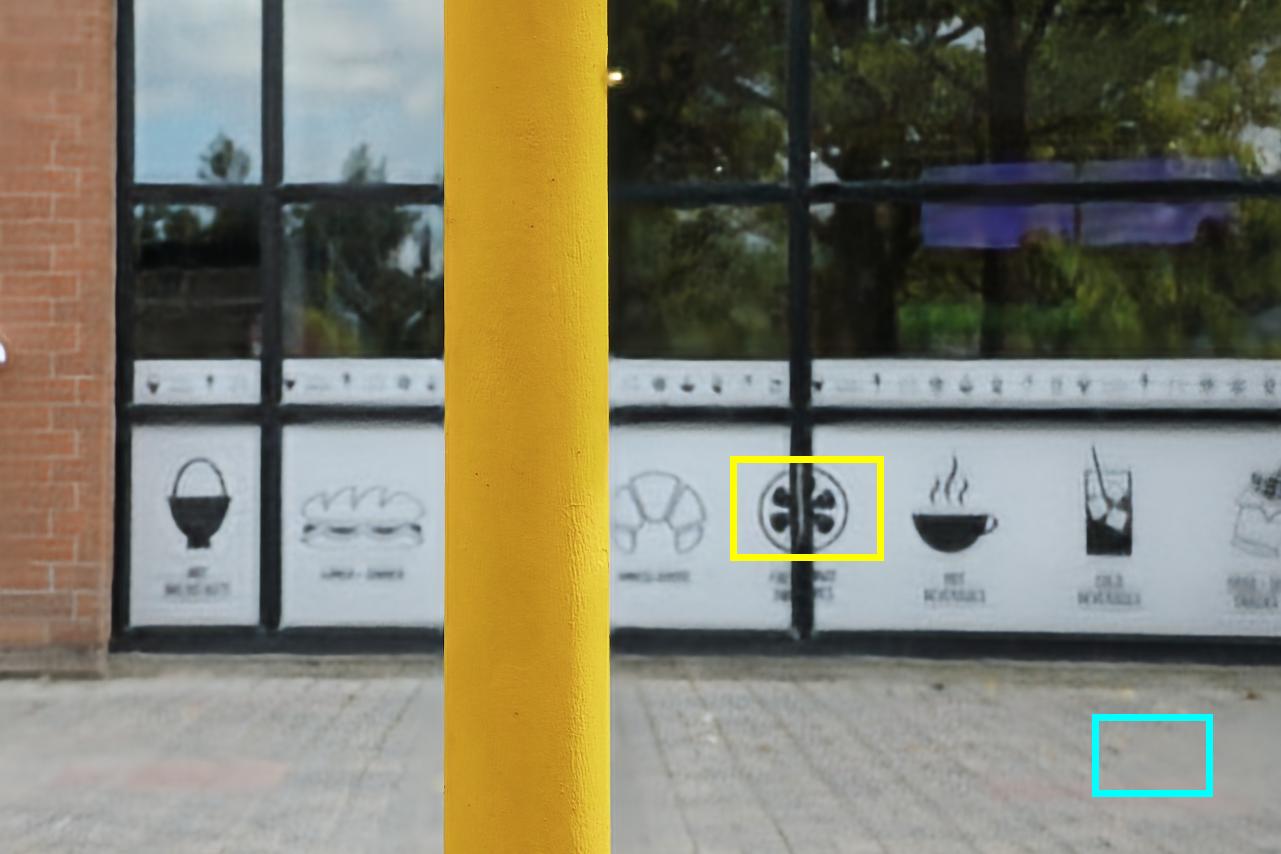}} &
    \multicolumn{2}{c}{\includegraphics[width=0.196\linewidth]{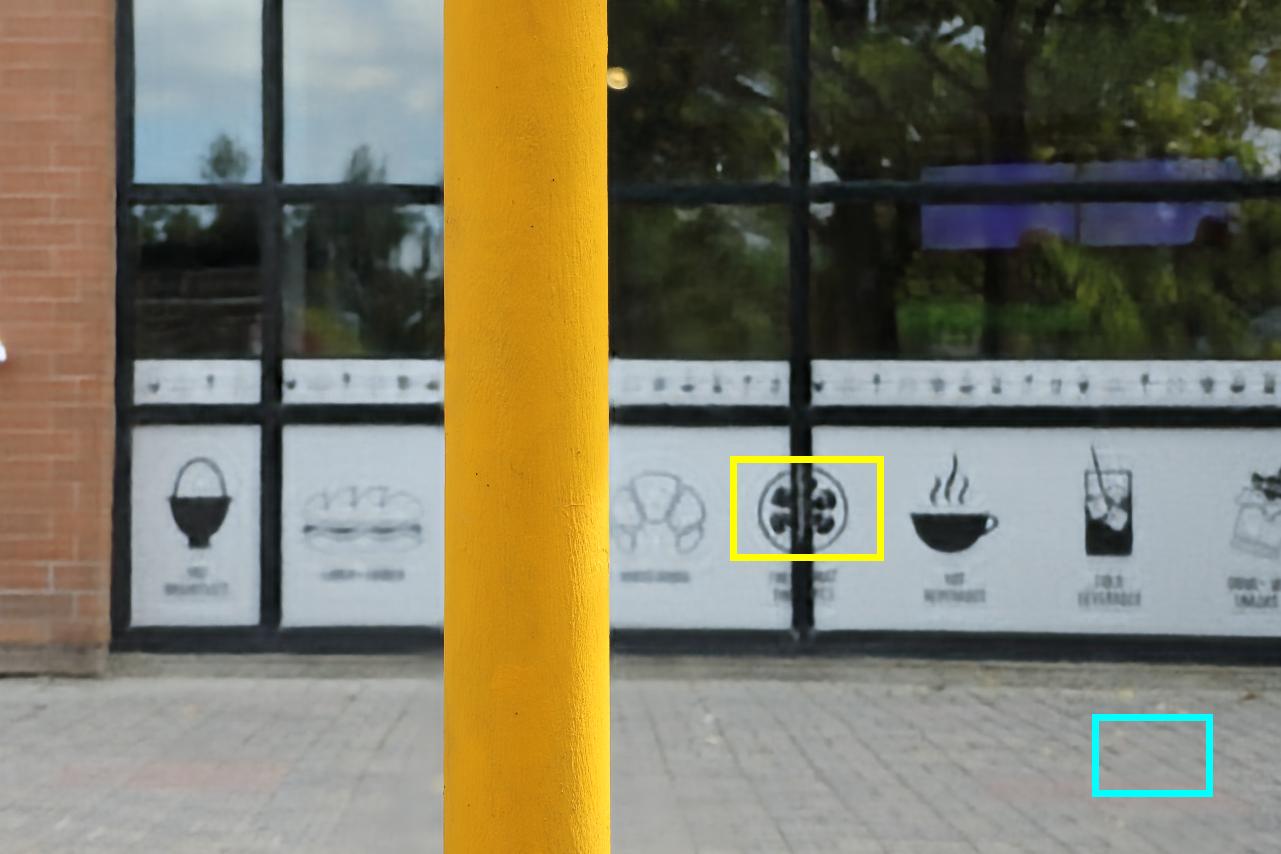}} &
    \multicolumn{2}{c}{\includegraphics[width=0.196\linewidth]{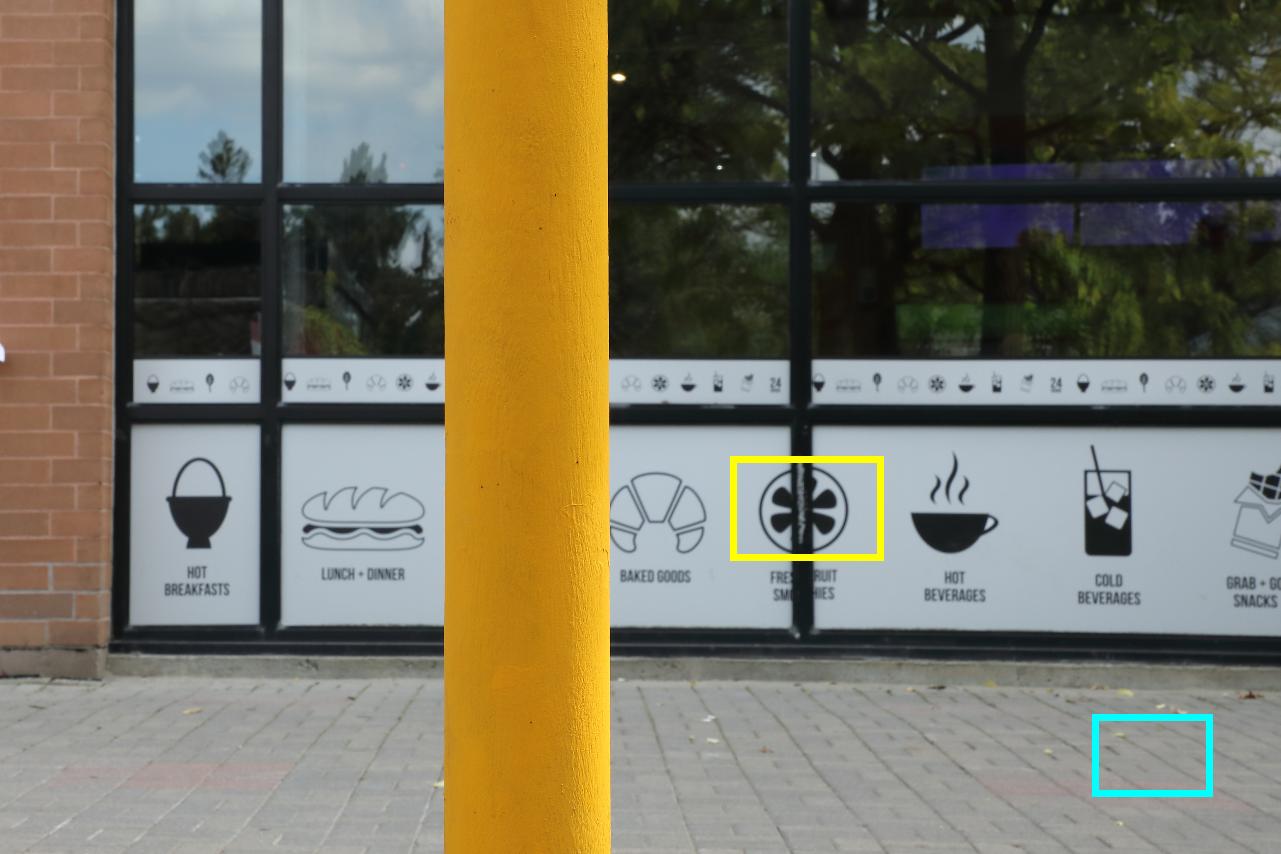}} \\
    \multicolumn{1}{c}{\includegraphics[width=0.096\linewidth]{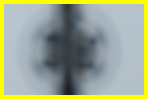}} &
    \multicolumn{1}{c}{\includegraphics[width=0.096\linewidth]{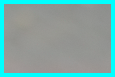}} &
    \multicolumn{1}{c}{\includegraphics[width=0.096\linewidth]{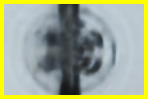}} &
    \multicolumn{1}{c}{\includegraphics[width=0.096\linewidth]{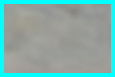}} &
    \multicolumn{1}{c}{\includegraphics[width=0.096\linewidth]{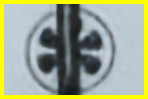}} &
    \multicolumn{1}{c}{\includegraphics[width=0.096\linewidth]{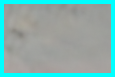}} &
    \multicolumn{1}{c}{\includegraphics[width=0.096\linewidth]{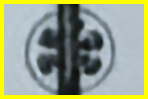}} &
    \multicolumn{1}{c}{\includegraphics[width=0.096\linewidth]{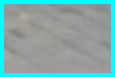}} &
    \multicolumn{1}{c}{\includegraphics[width=0.096\linewidth]{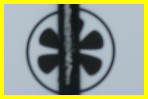}} &
    \multicolumn{1}{c}{\includegraphics[width=0.096\linewidth]{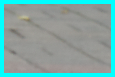}} \\

    \multicolumn{2}{c}{\includegraphics[width=0.196\linewidth]{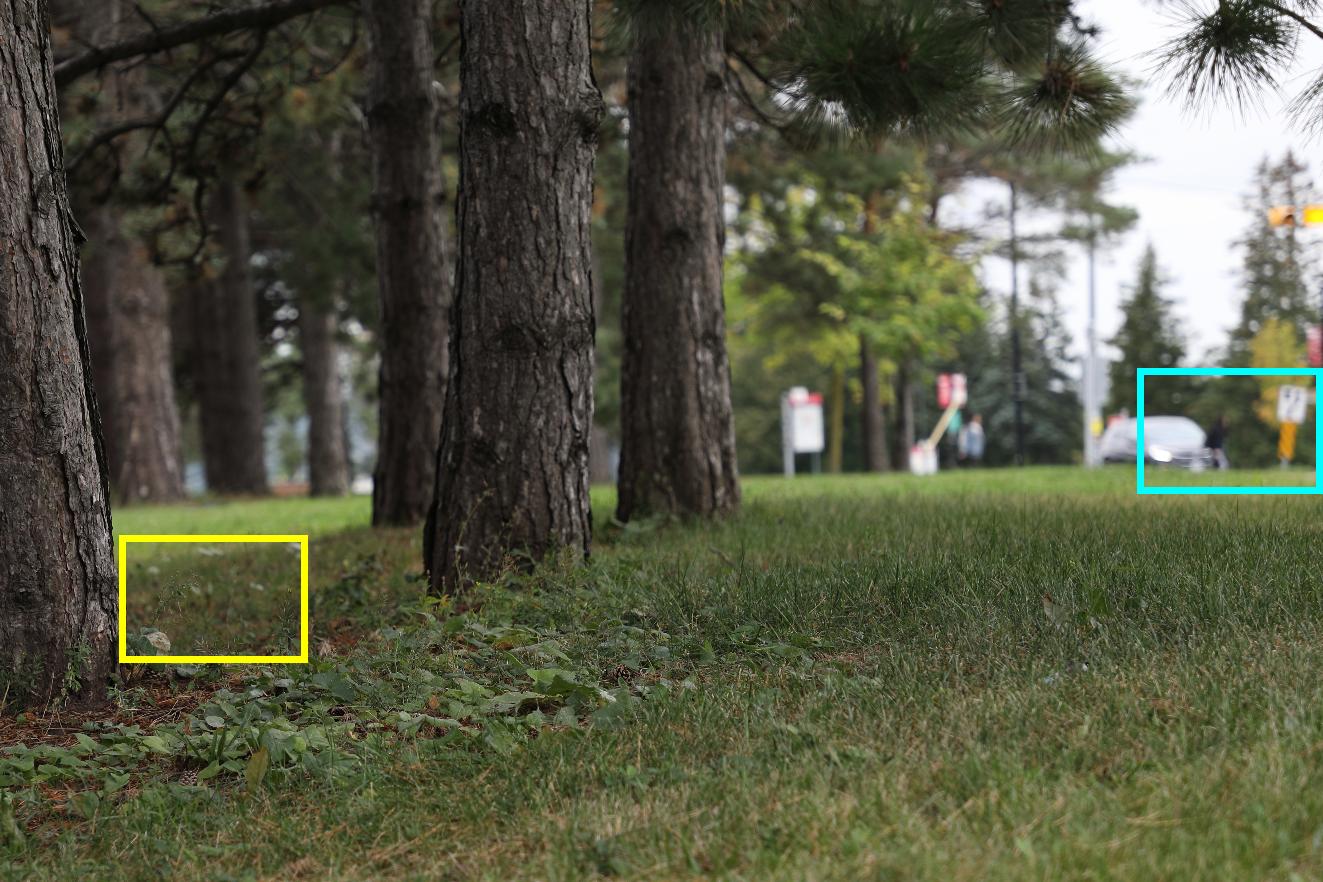}} &
    \multicolumn{2}{c}{\includegraphics[width=0.196\linewidth]{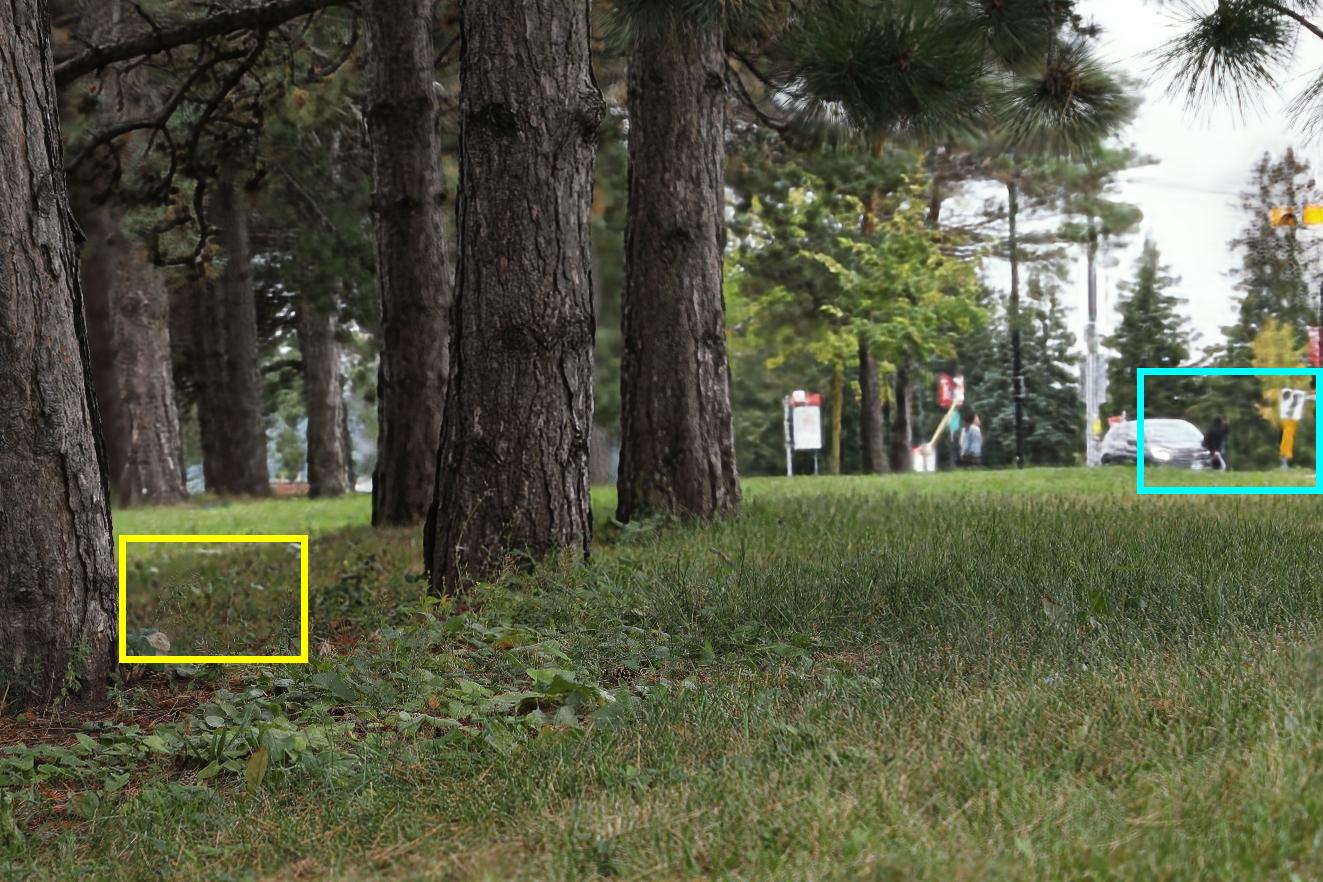}} &
    \multicolumn{2}{c}{\includegraphics[width=0.196\linewidth]{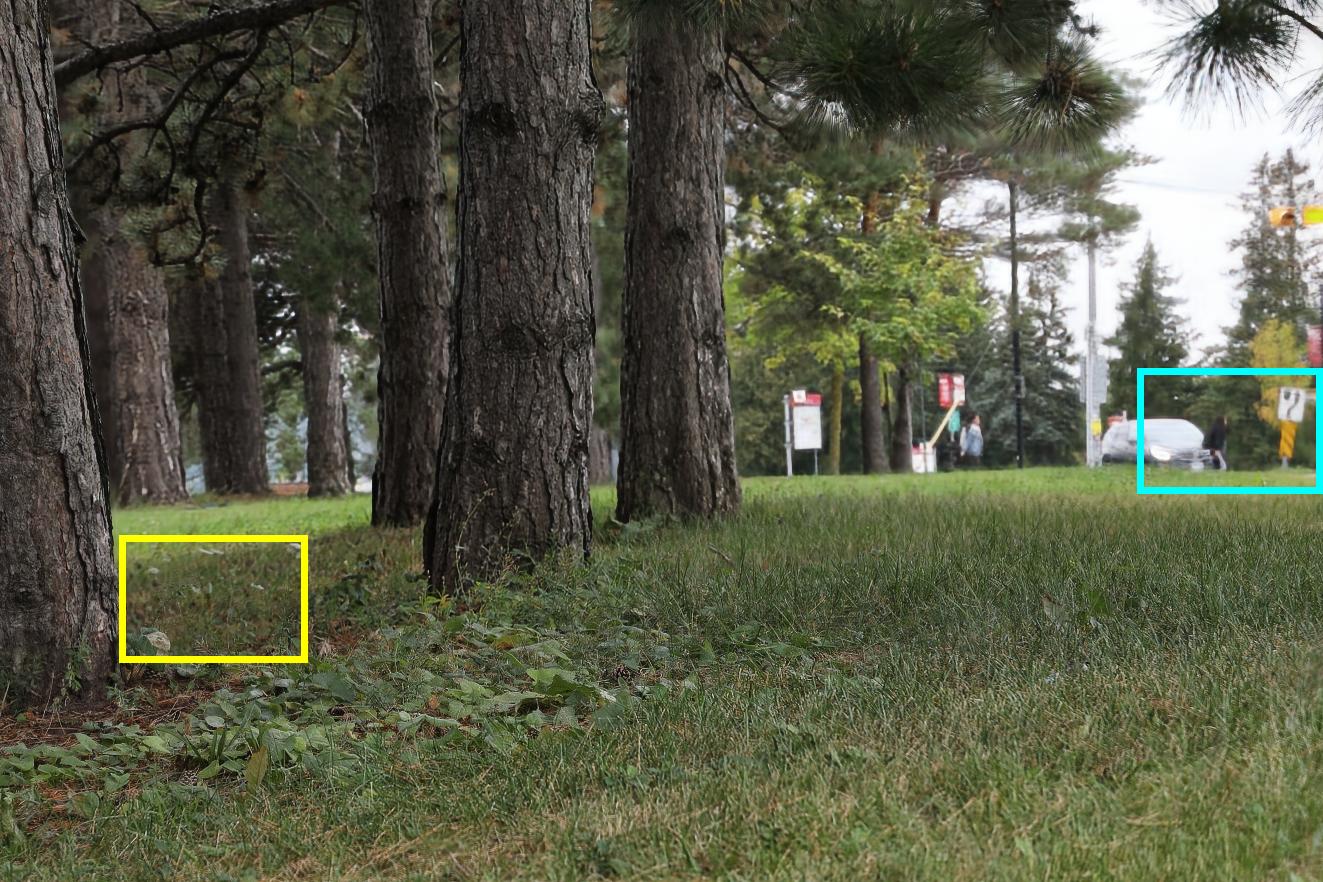}} &
    \multicolumn{2}{c}{\includegraphics[width=0.196\linewidth]{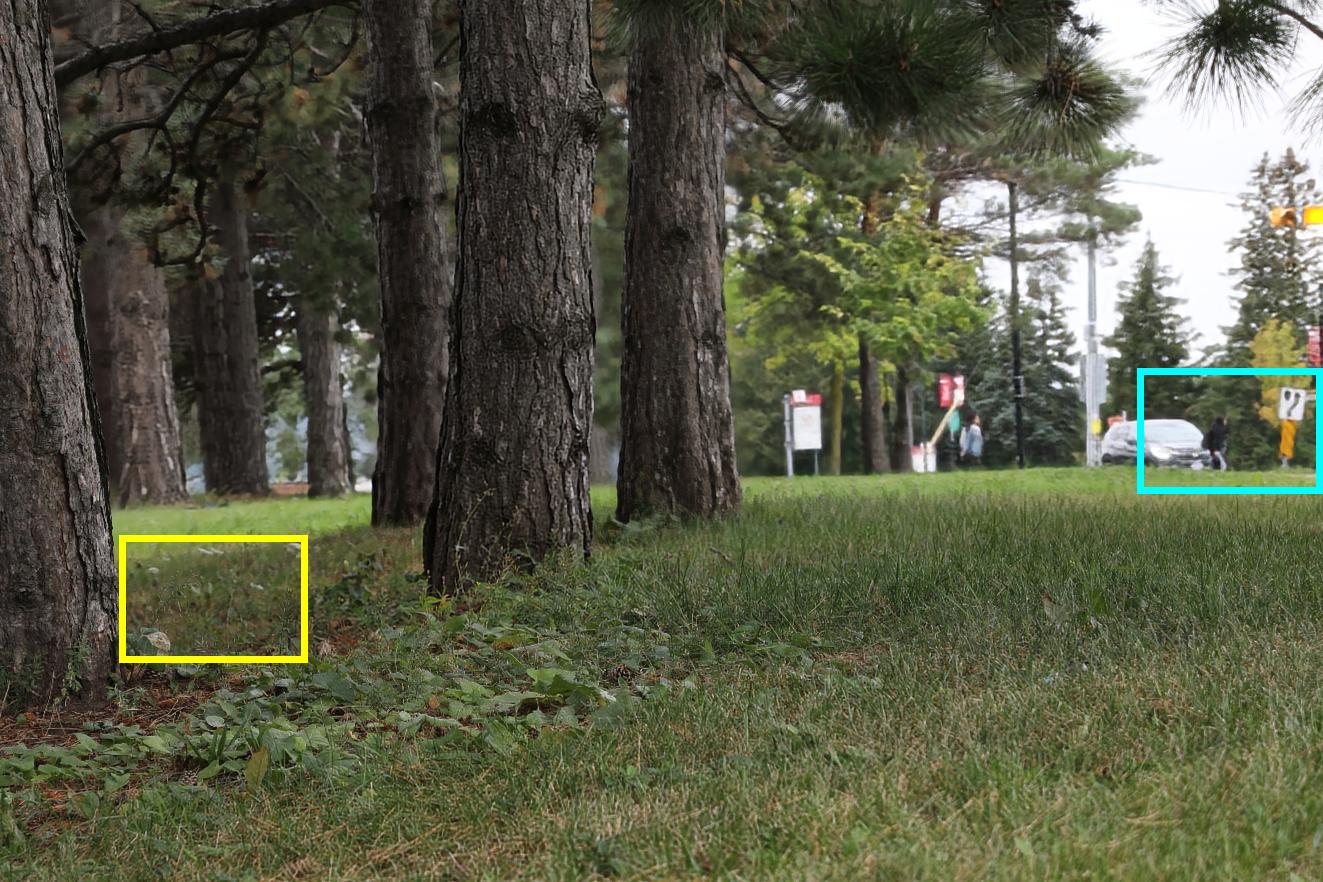}} &
    \multicolumn{2}{c}{\includegraphics[width=0.196\linewidth]{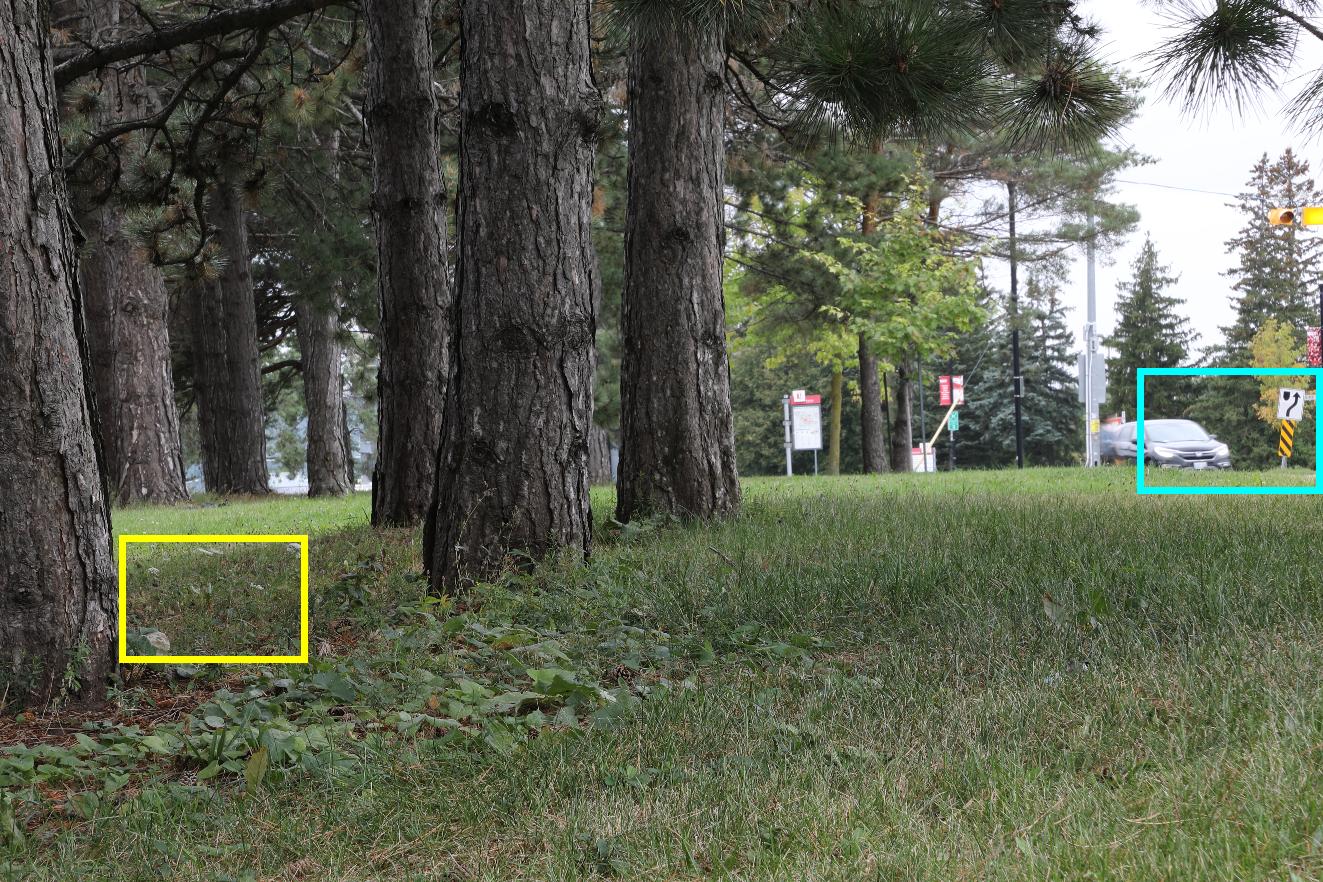}} \\
    \multicolumn{1}{c}{\includegraphics[width=0.096\linewidth]{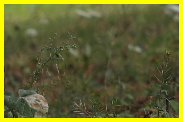}} &
    \multicolumn{1}{c}{\includegraphics[width=0.096\linewidth]{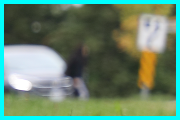}} &
    \multicolumn{1}{c}{\includegraphics[width=0.096\linewidth]{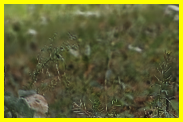}} &
    \multicolumn{1}{c}{\includegraphics[width=0.096\linewidth]{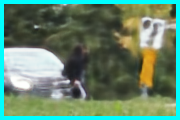}} &
    \multicolumn{1}{c}{\includegraphics[width=0.096\linewidth]{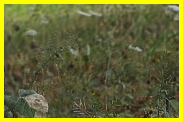}} &
    \multicolumn{1}{c}{\includegraphics[width=0.096\linewidth]{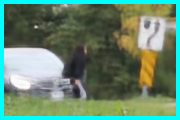}} &
    \multicolumn{1}{c}{\includegraphics[width=0.096\linewidth]{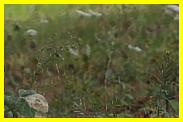}} &
    \multicolumn{1}{c}{\includegraphics[width=0.096\linewidth]{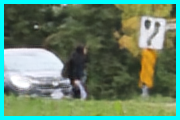}} &
    \multicolumn{1}{c}{\includegraphics[width=0.096\linewidth]{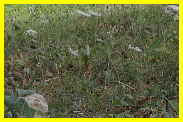}} &
    \multicolumn{1}{c}{\includegraphics[width=0.096\linewidth]{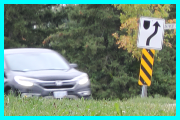}} \\

    \multicolumn{2}{c}{\includegraphics[width=0.196\linewidth]{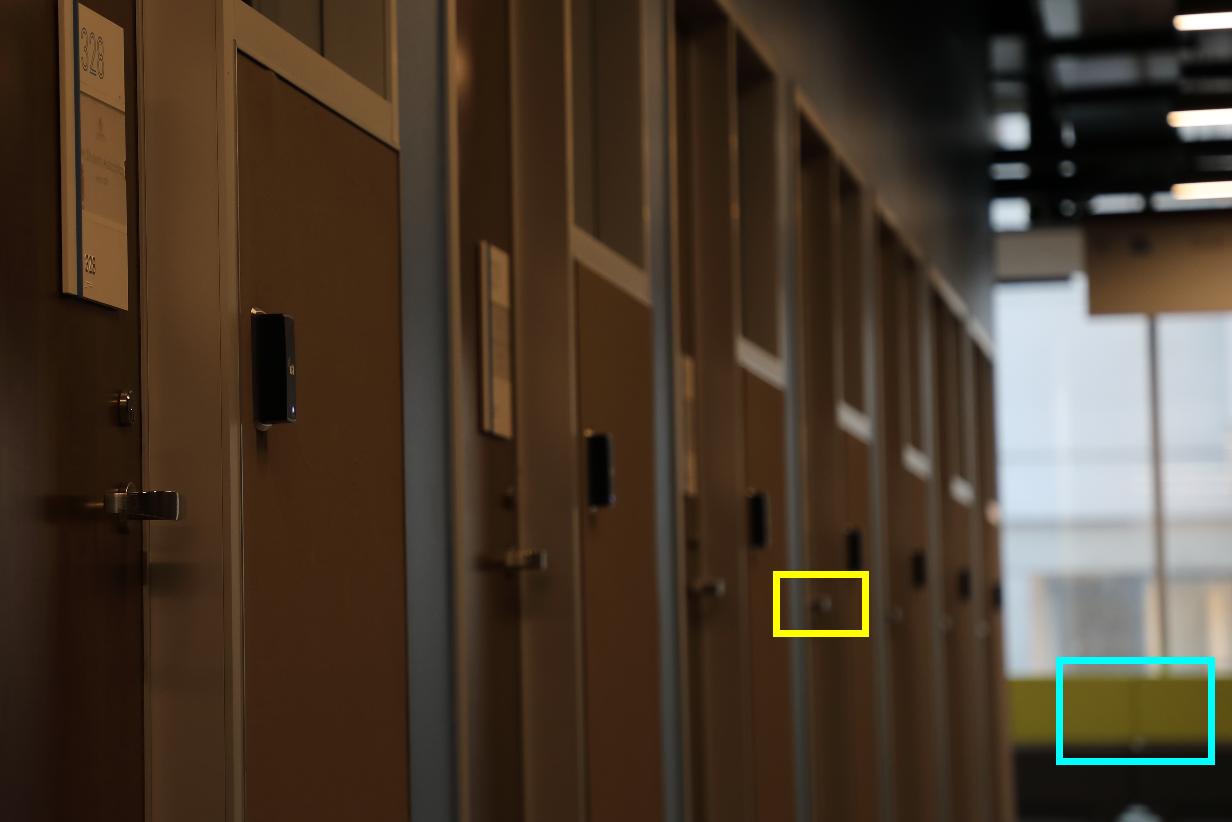}} &
    \multicolumn{2}{c}{\includegraphics[width=0.196\linewidth]{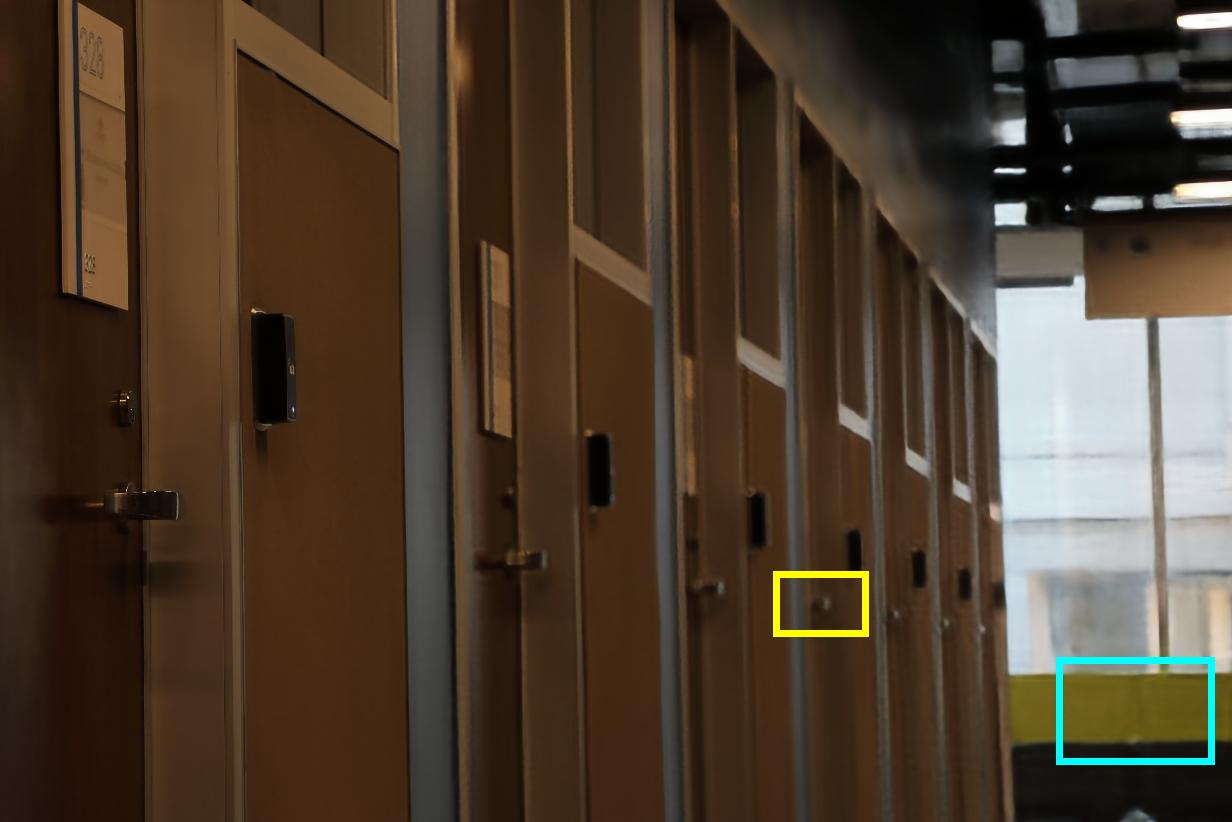}} &
    \multicolumn{2}{c}{\includegraphics[width=0.196\linewidth]{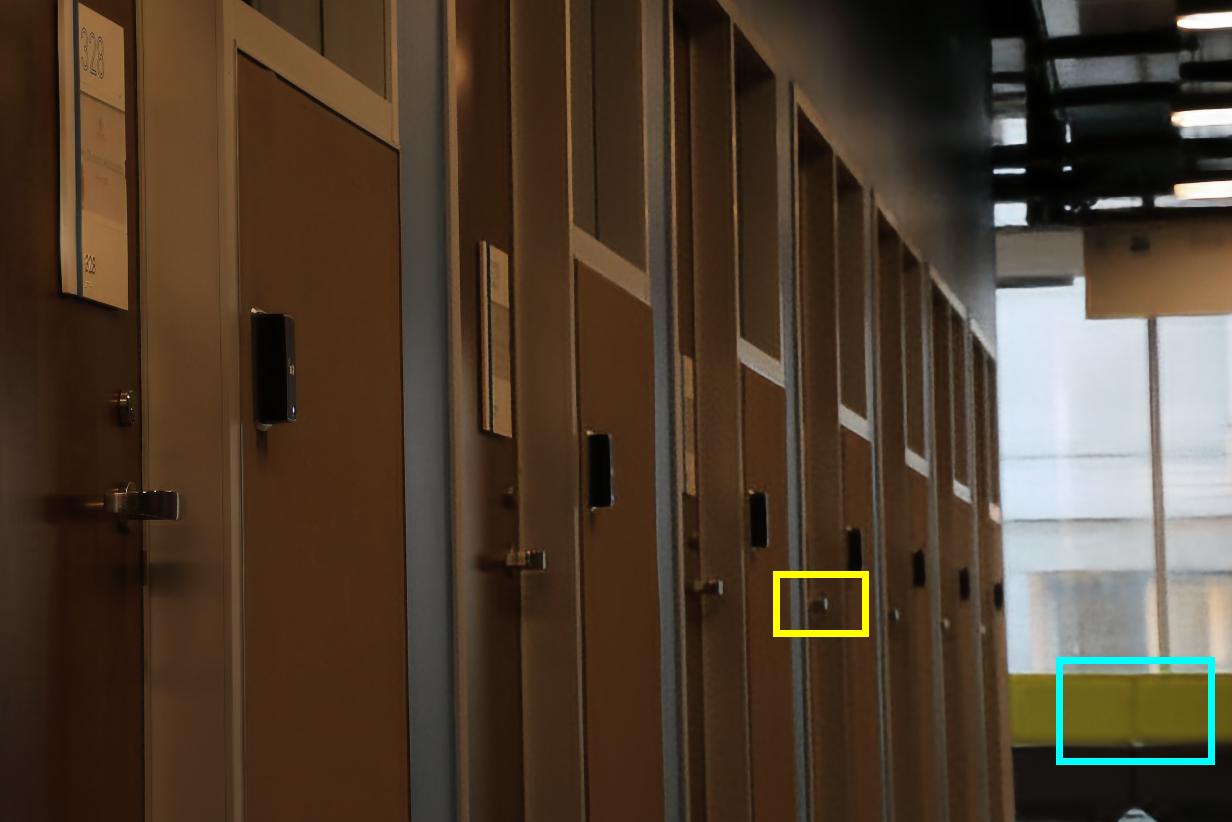}} &
    \multicolumn{2}{c}{\includegraphics[width=0.196\linewidth]{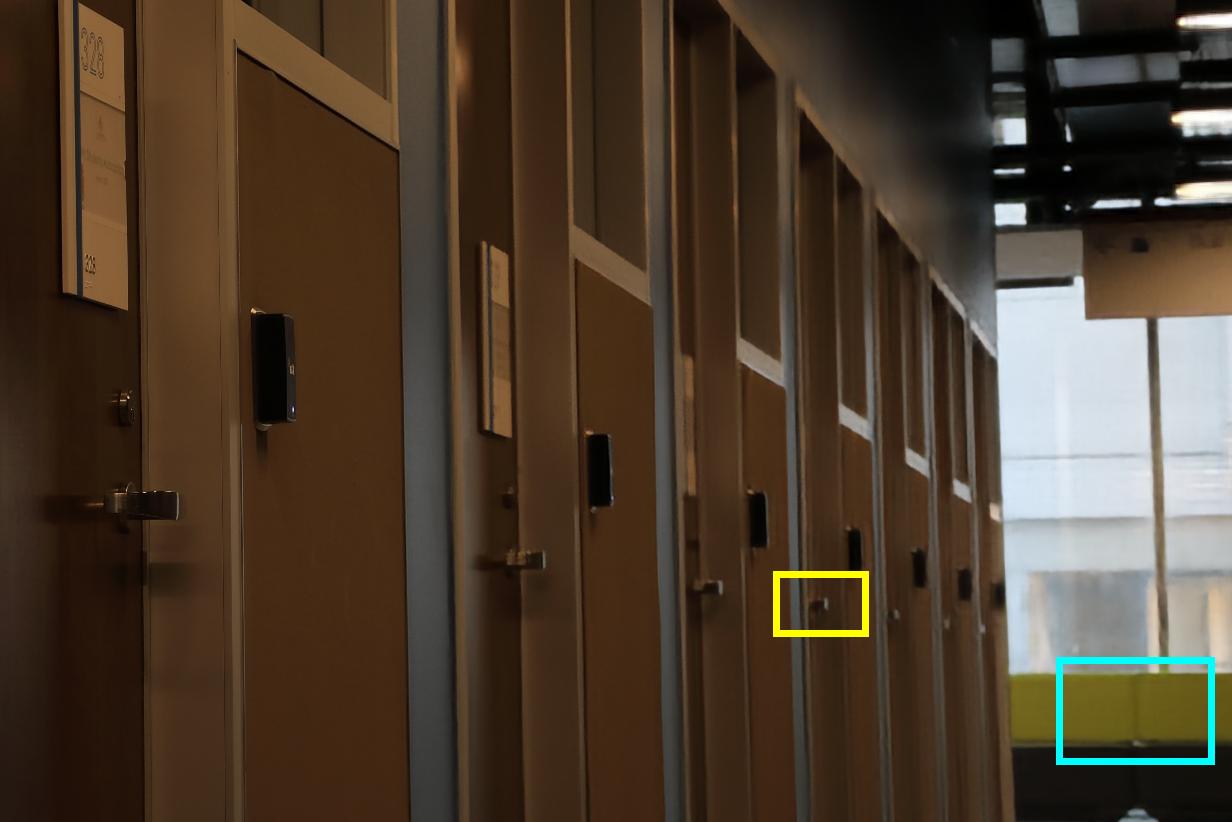}} &
    \multicolumn{2}{c}{\includegraphics[width=0.196\linewidth]{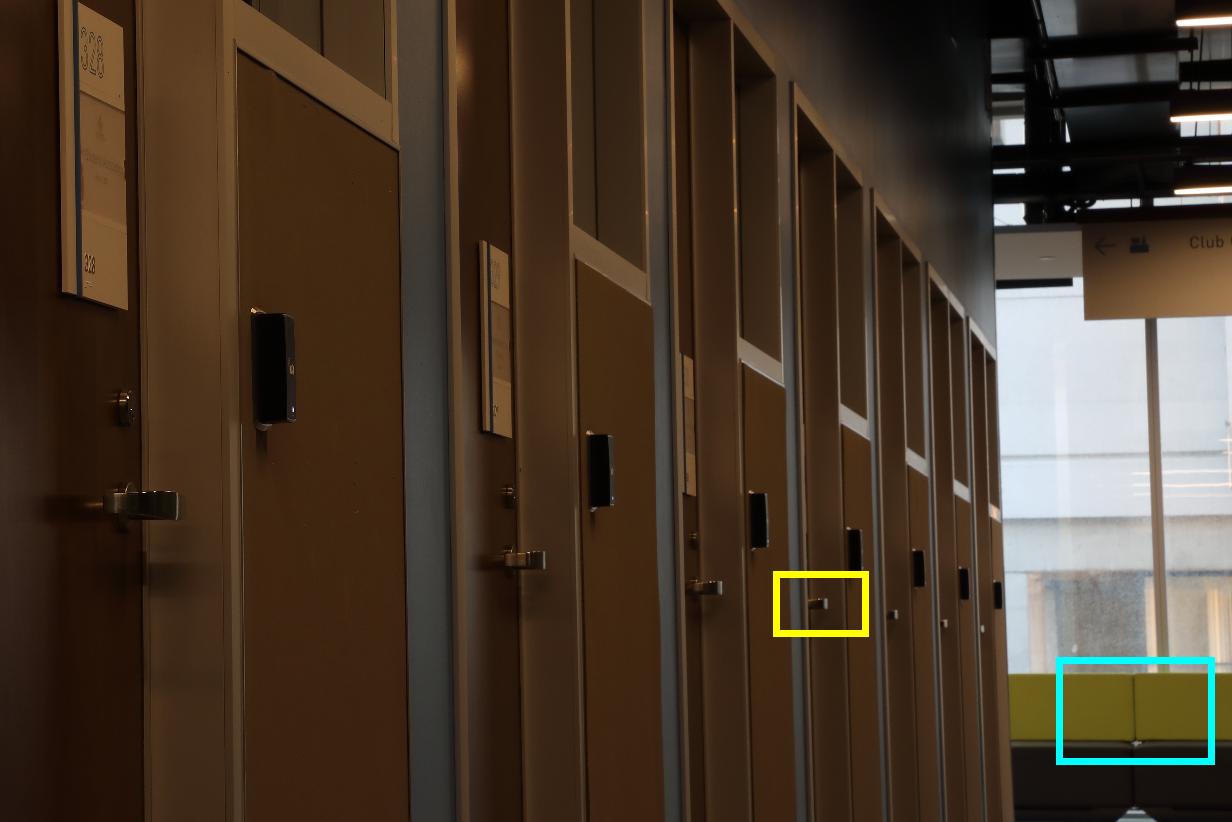}} \\
    \multicolumn{1}{c}{\includegraphics[width=0.096\linewidth]{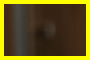}} &
    \multicolumn{1}{c}{\includegraphics[width=0.096\linewidth]{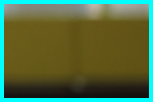}} &
    \multicolumn{1}{c}{\includegraphics[width=0.096\linewidth]{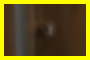}} &
    \multicolumn{1}{c}{\includegraphics[width=0.096\linewidth]{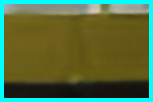}} &
    \multicolumn{1}{c}{\includegraphics[width=0.096\linewidth]{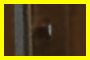}} &
    \multicolumn{1}{c}{\includegraphics[width=0.096\linewidth]{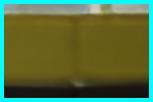}} &
    \multicolumn{1}{c}{\includegraphics[width=0.096\linewidth]{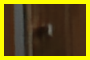}} &
    \multicolumn{1}{c}{\includegraphics[width=0.096\linewidth]{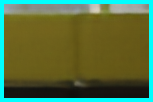}} &
    \multicolumn{1}{c}{\includegraphics[width=0.096\linewidth]{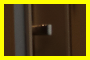}} &
    \multicolumn{1}{c}{\includegraphics[width=0.096\linewidth]{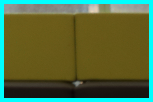}} \\

    \multicolumn{2}{c}{\includegraphics[width=0.196\linewidth]{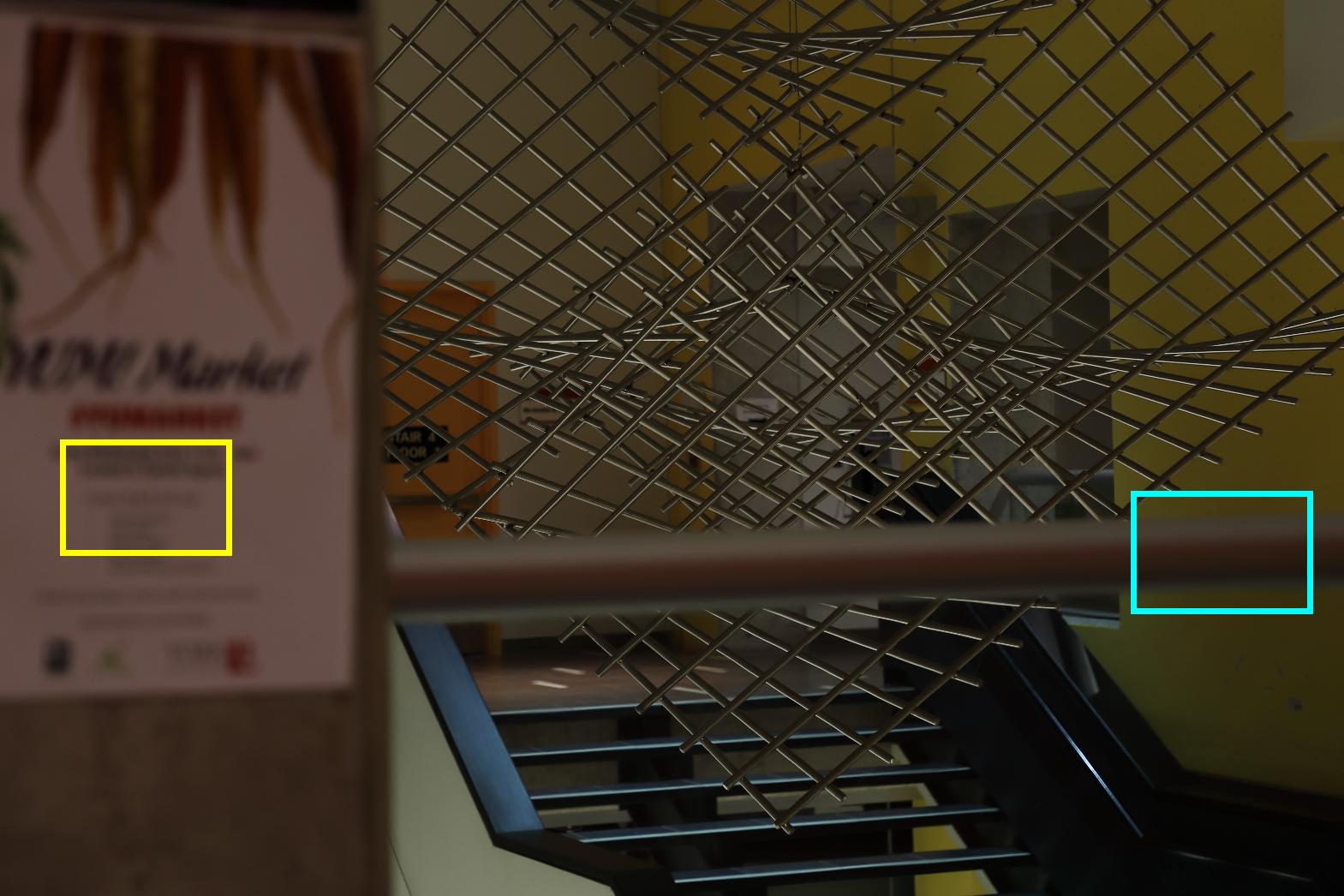}} &
    \multicolumn{2}{c}{\includegraphics[width=0.196\linewidth]{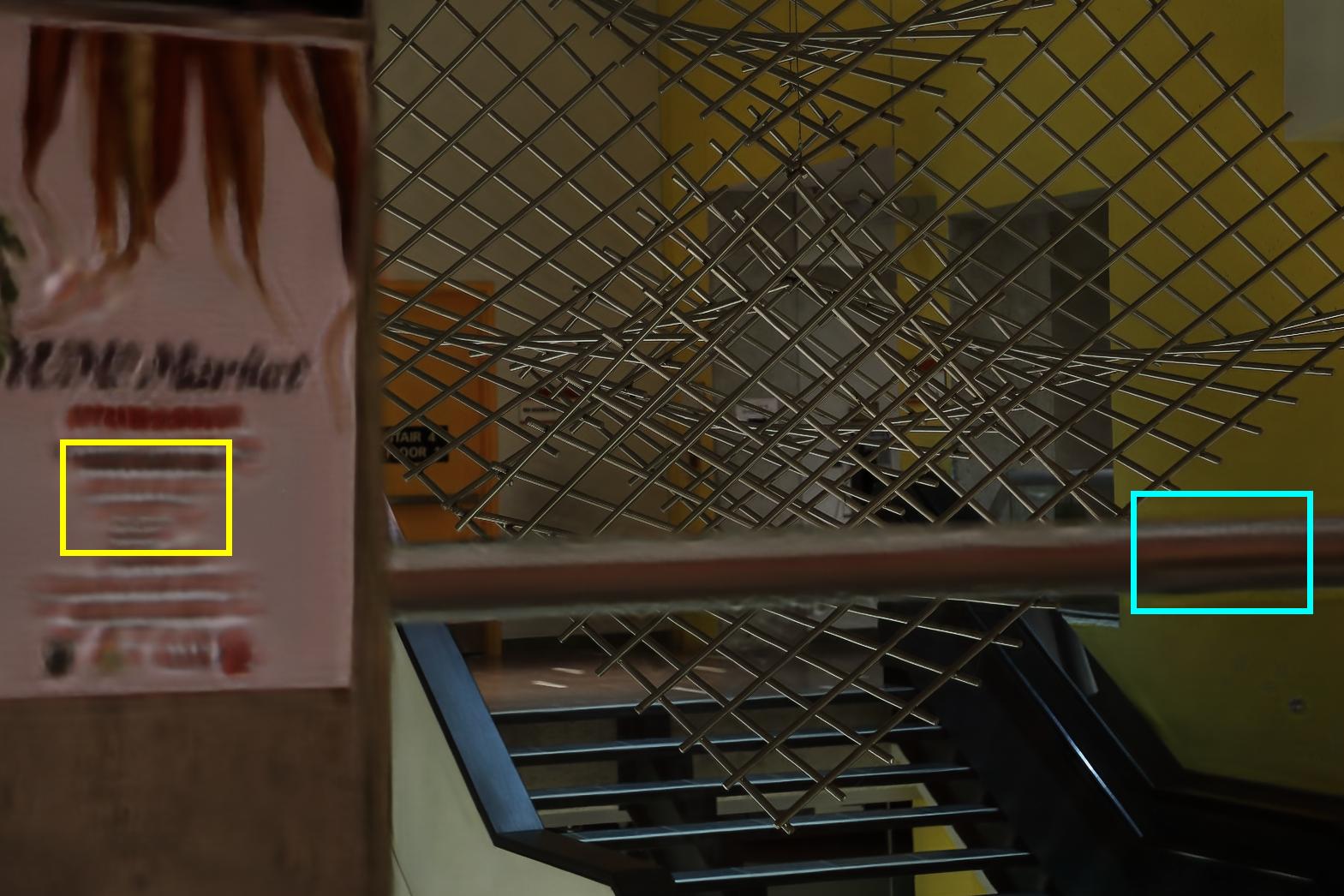}} &
    \multicolumn{2}{c}{\includegraphics[width=0.196\linewidth]{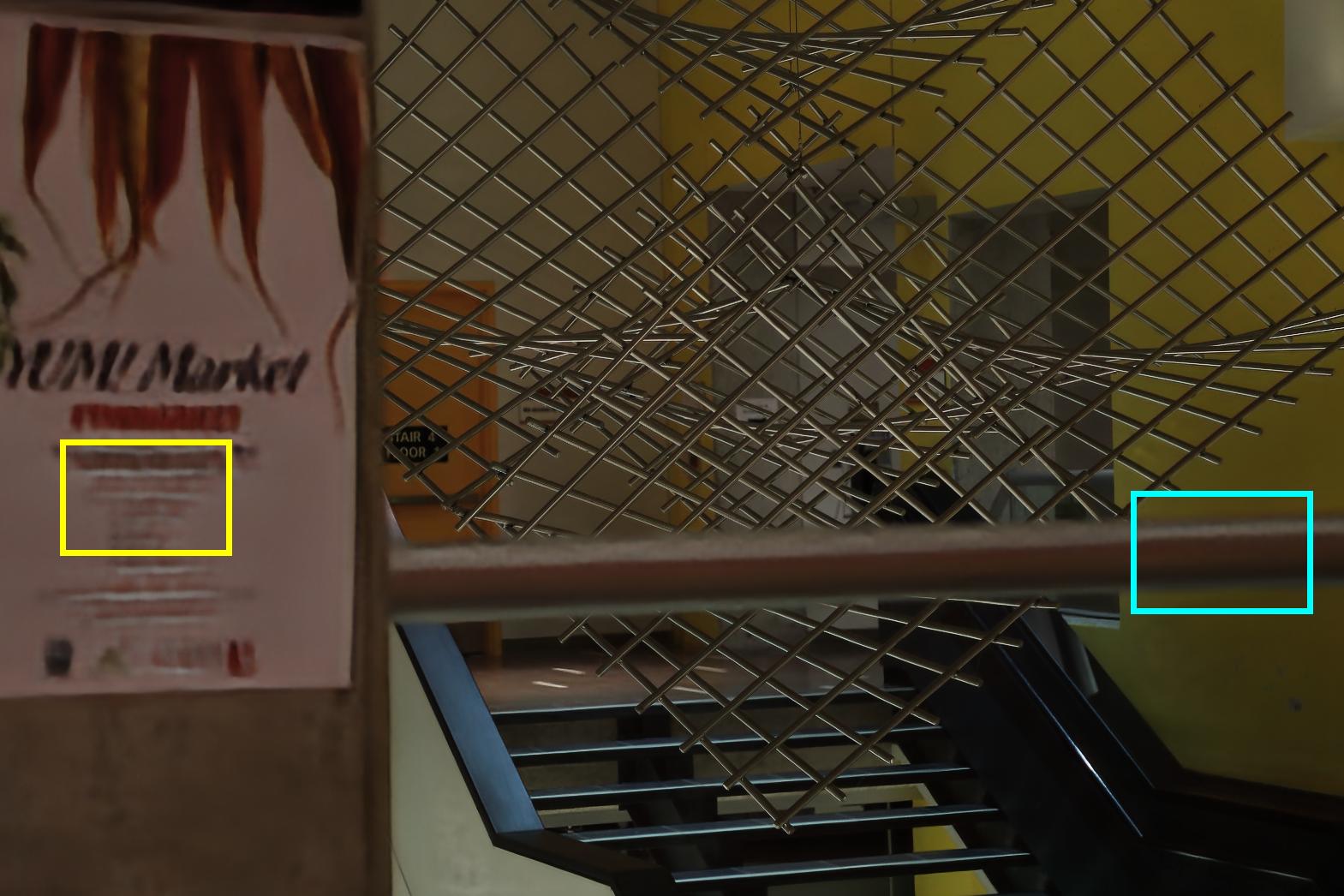}} &
    \multicolumn{2}{c}{\includegraphics[width=0.196\linewidth]{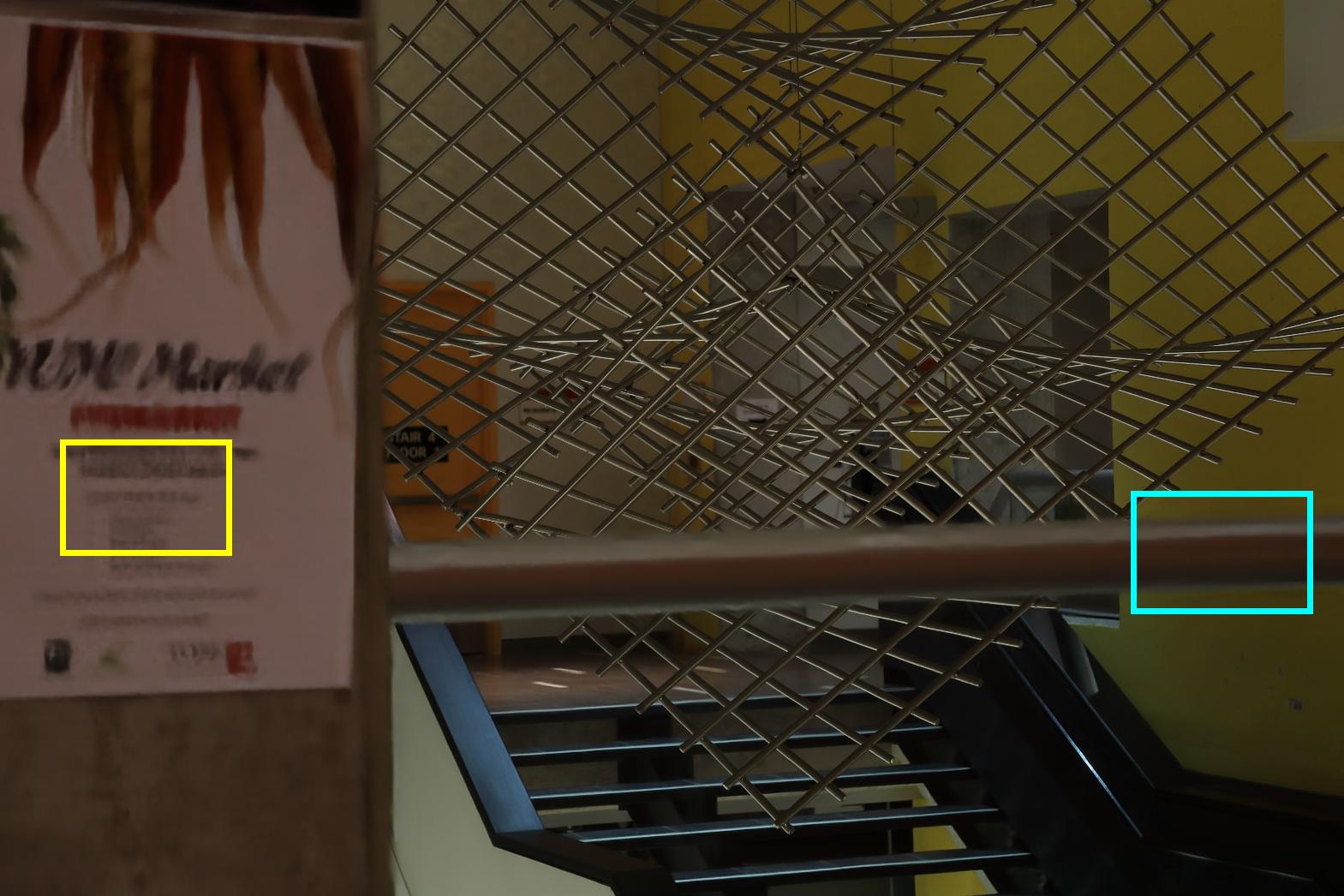}} &
    \multicolumn{2}{c}{\includegraphics[width=0.196\linewidth]{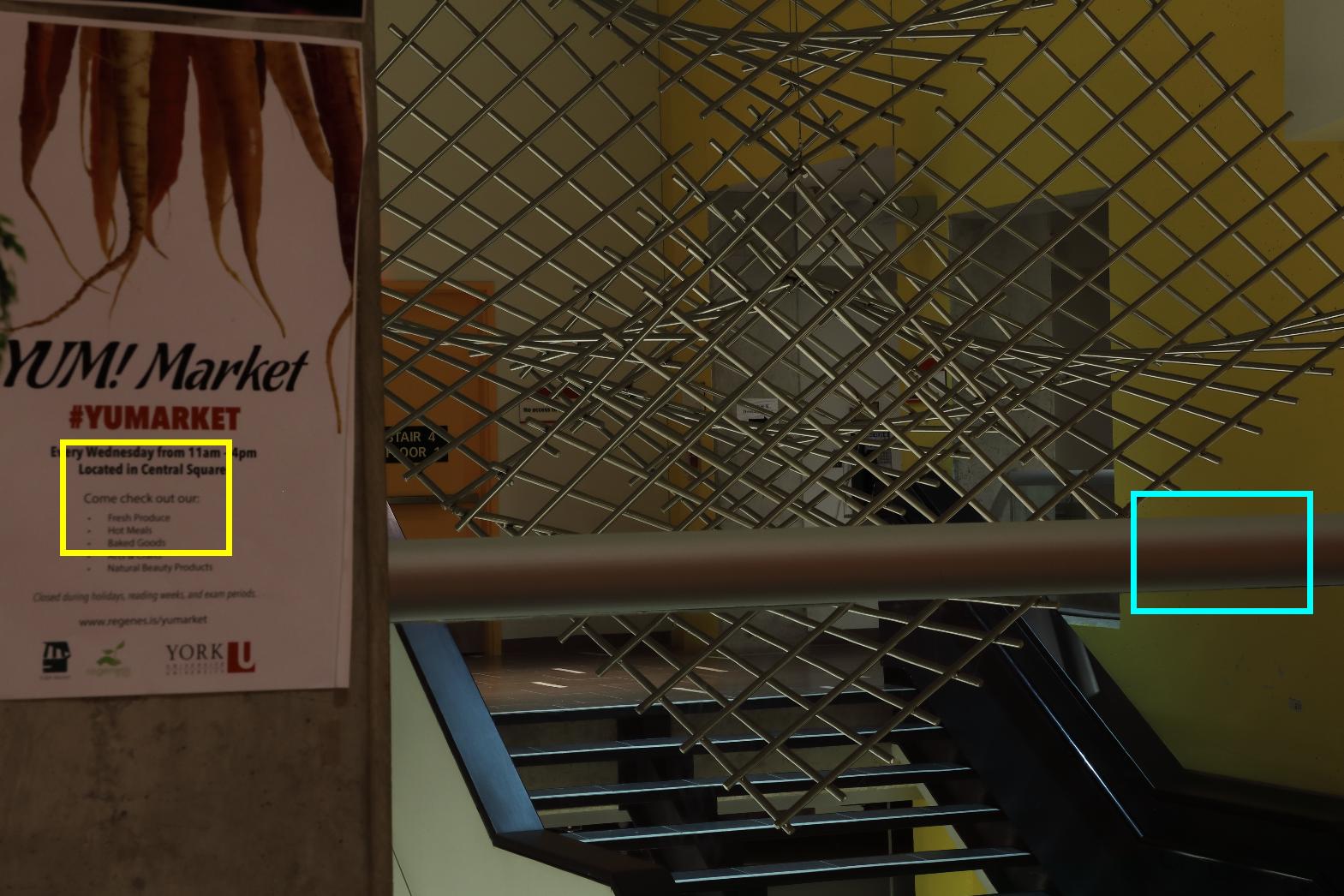}} \\
    \multicolumn{1}{c}{\includegraphics[width=0.096\linewidth]{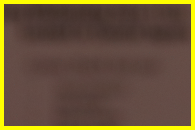}} &
    \multicolumn{1}{c}{\includegraphics[width=0.096\linewidth]{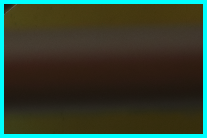}} &
    \multicolumn{1}{c}{\includegraphics[width=0.096\linewidth]{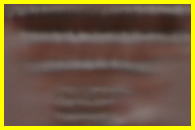}} &
    \multicolumn{1}{c}{\includegraphics[width=0.096\linewidth]{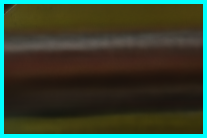}} &
    \multicolumn{1}{c}{\includegraphics[width=0.096\linewidth]{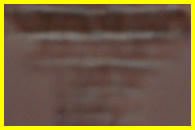}} &
    \multicolumn{1}{c}{\includegraphics[width=0.096\linewidth]{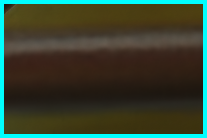}} &
    \multicolumn{1}{c}{\includegraphics[width=0.096\linewidth]{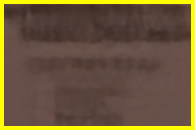}} &
    \multicolumn{1}{c}{\includegraphics[width=0.096\linewidth]{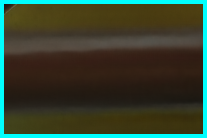}} &
    \multicolumn{1}{c}{\includegraphics[width=0.096\linewidth]{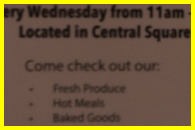}} &
    \multicolumn{1}{c}{\includegraphics[width=0.096\linewidth]{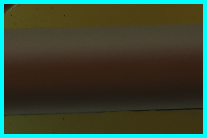}} \\

    \multicolumn{2}{c}{\includegraphics[width=0.196\linewidth]{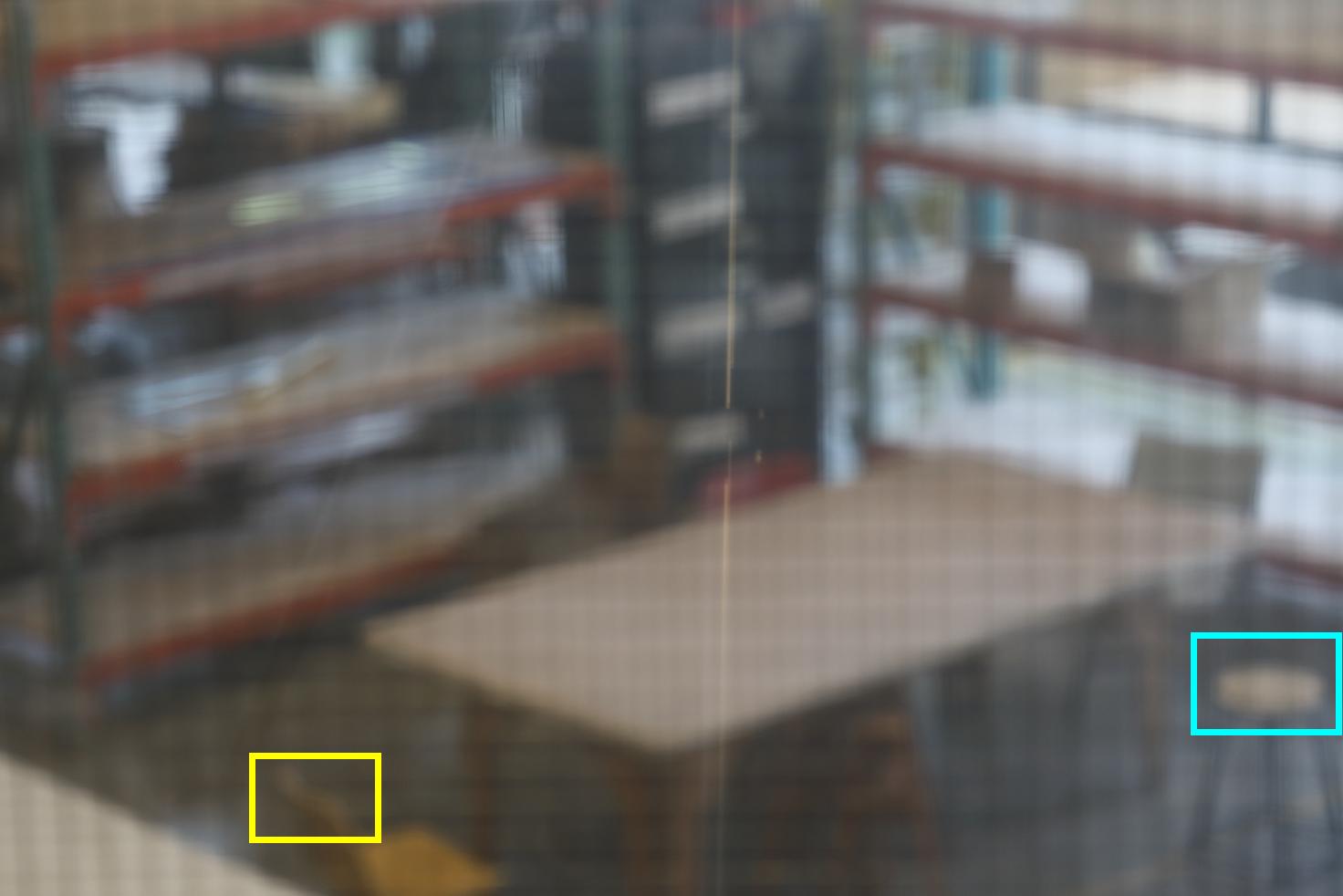}} &
    \multicolumn{2}{c}{\includegraphics[width=0.196\linewidth]{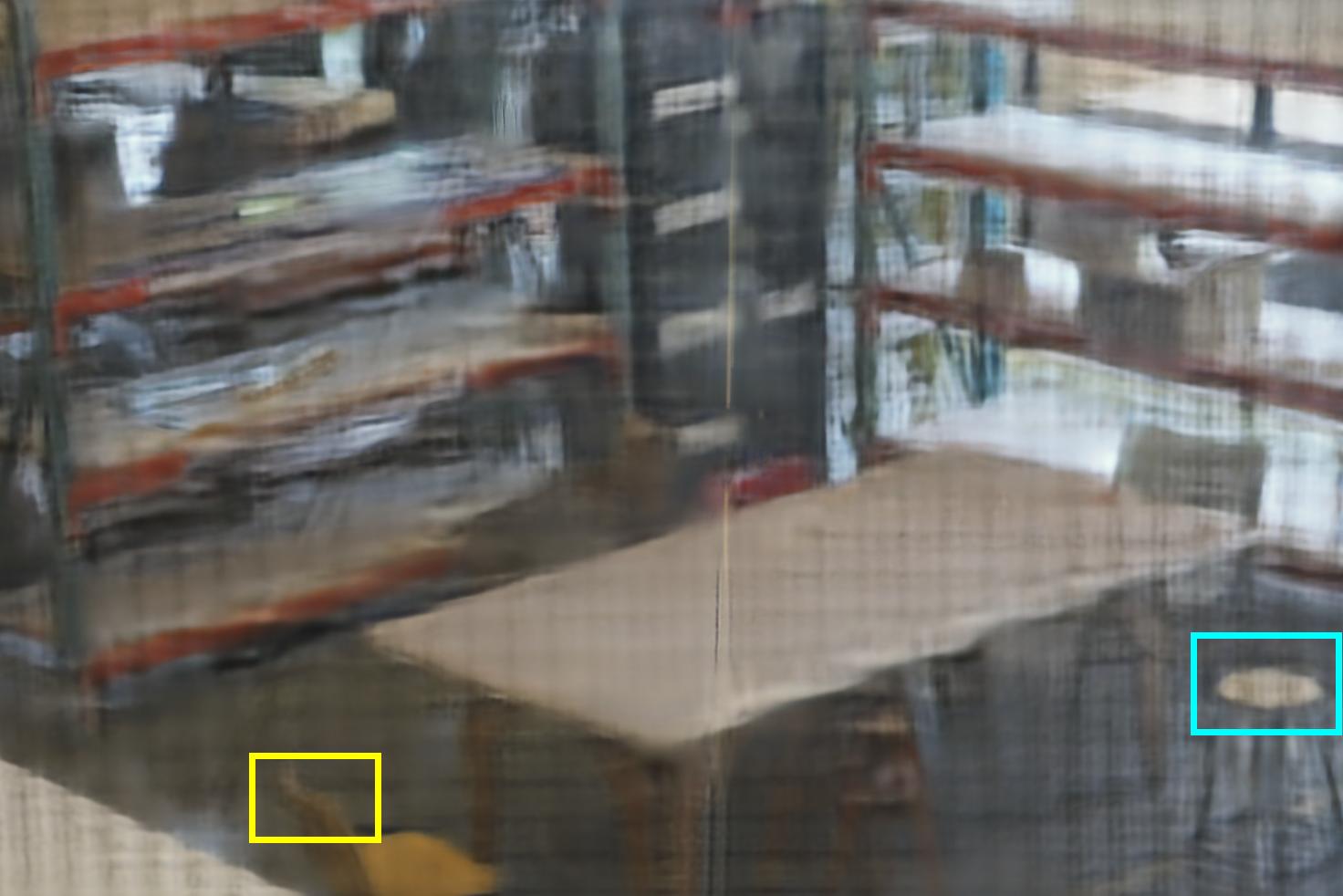}} &
    \multicolumn{2}{c}{\includegraphics[width=0.196\linewidth]{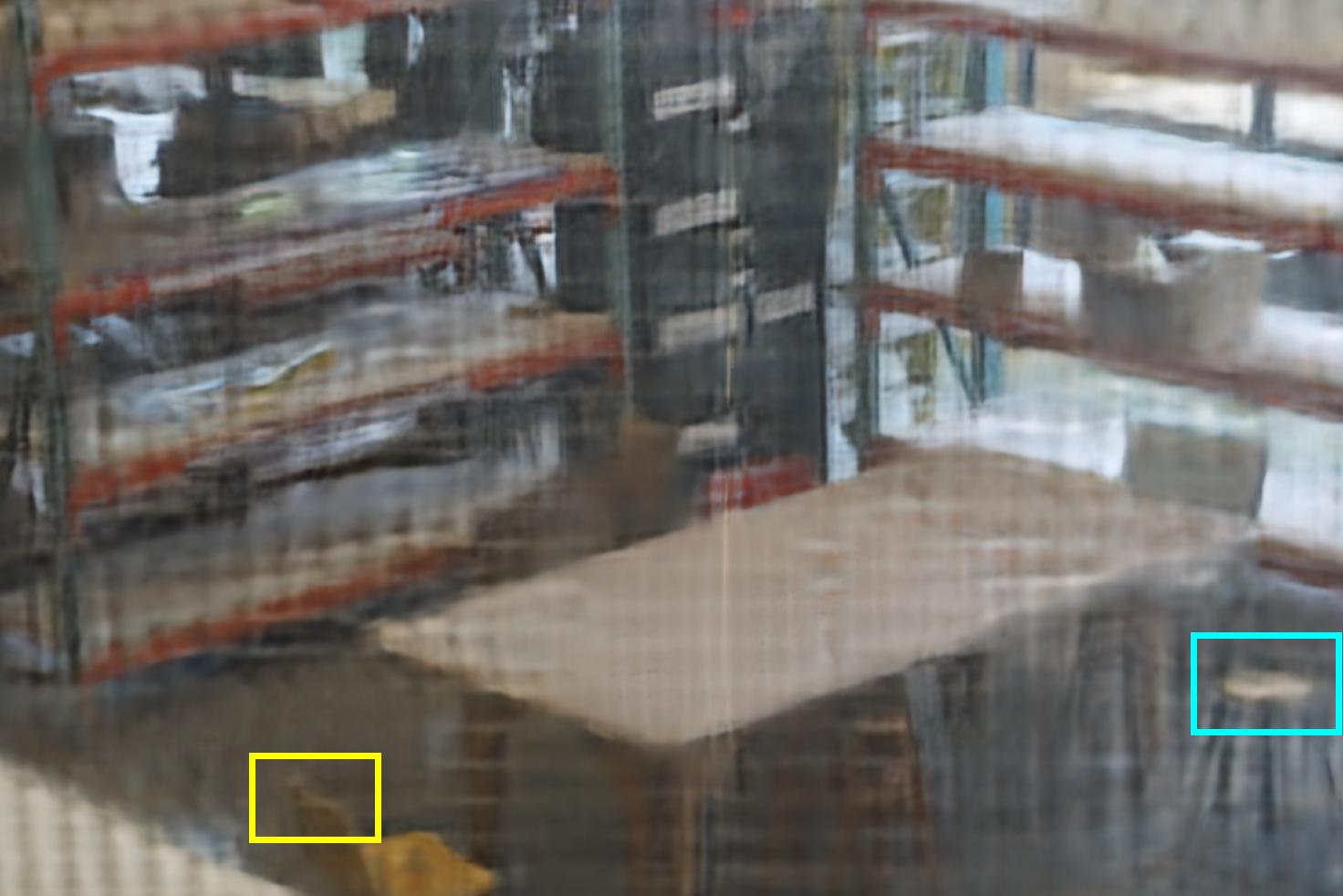}} &
    \multicolumn{2}{c}{\includegraphics[width=0.196\linewidth]{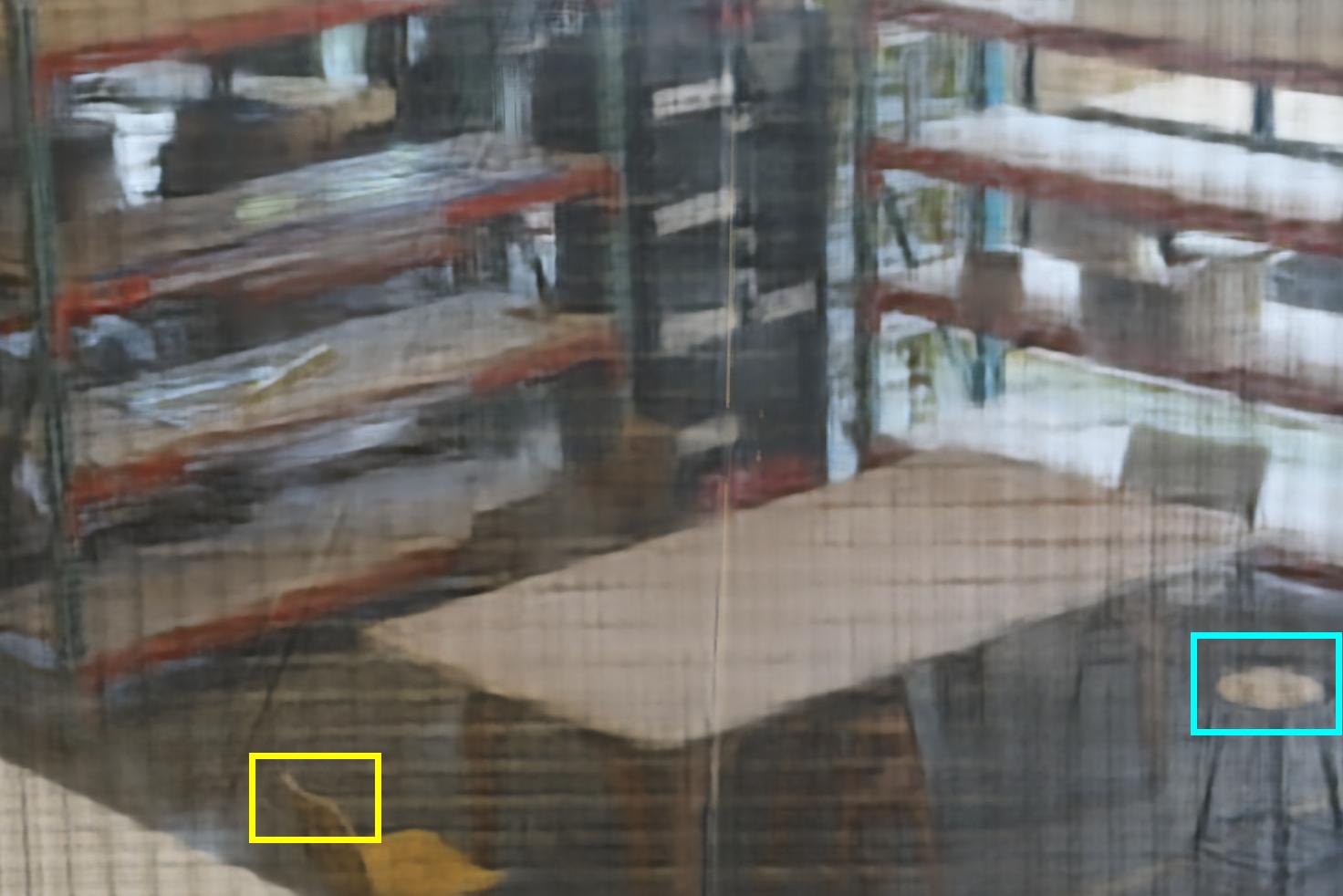}} &
    \multicolumn{2}{c}{\includegraphics[width=0.196\linewidth]{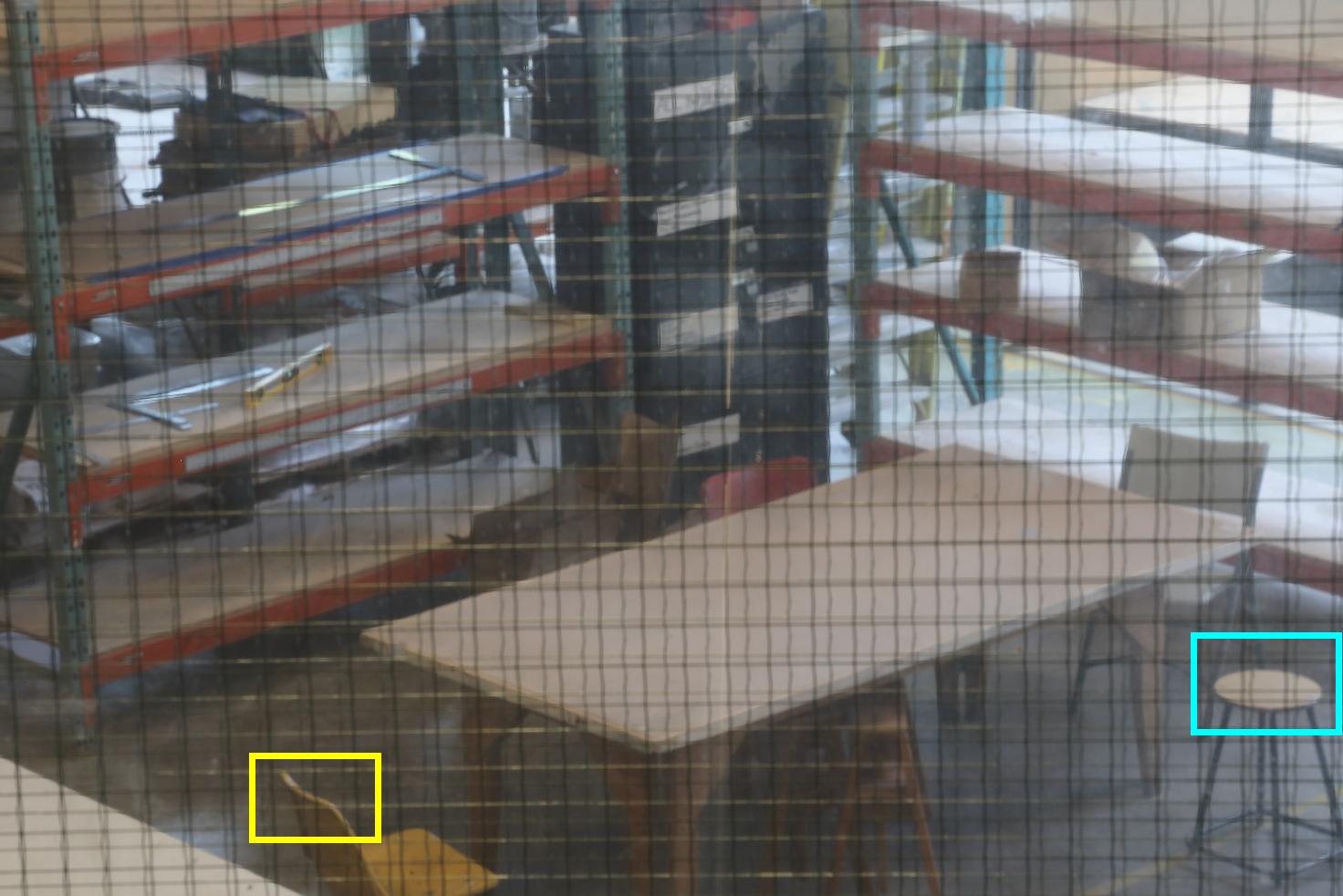}} \\
    \multicolumn{1}{c}{\includegraphics[width=0.096\linewidth]{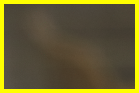}} &
    \multicolumn{1}{c}{\includegraphics[width=0.096\linewidth]{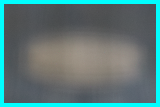}} &
    \multicolumn{1}{c}{\includegraphics[width=0.096\linewidth]{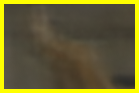}} &
    \multicolumn{1}{c}{\includegraphics[width=0.096\linewidth]{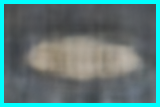}} &
    \multicolumn{1}{c}{\includegraphics[width=0.096\linewidth]{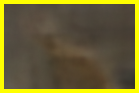}} &
    \multicolumn{1}{c}{\includegraphics[width=0.096\linewidth]{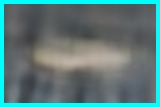}} &
    \multicolumn{1}{c}{\includegraphics[width=0.096\linewidth]{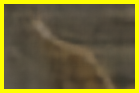}} &
    \multicolumn{1}{c}{\includegraphics[width=0.096\linewidth]{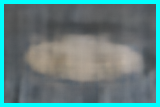}} &
    \multicolumn{1}{c}{\includegraphics[width=0.096\linewidth]{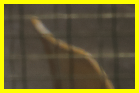}} &
    \multicolumn{1}{c}{\includegraphics[width=0.096\linewidth]{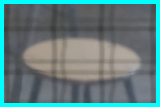}} \\


    \multicolumn{2}{c}{(a) input}  & \multicolumn{2}{c}{(b) DPDNet (single)~\cite{Abuolaim:2020:DPDNet}}
     & \multicolumn{2}{c}{(c) DPDNet (dual)~\cite{Abuolaim:2020:DPDNet} } & \multicolumn{2}{c}{(d) ours } & \multicolumn{2}{c}{(e) GT} \\

  \end{tabular}
  \vspace{-0.05cm}
  \caption{Additional qualitative comparisons with DPDNet~\cite{Abuolaim:2020:DPDNet} on the test set of the DPDD dataset \cite{Abuolaim:2020:DPDNet}.}
\label{fig:dpdd2}
\vspace{-10pt}
\end{figure*}

\begin{figure*}[tp]
\centering
\setlength\tabcolsep{1 pt}
  \begin{tabular}{cccccc}
    \multicolumn{2}{c}{\includegraphics[width=0.331\linewidth]{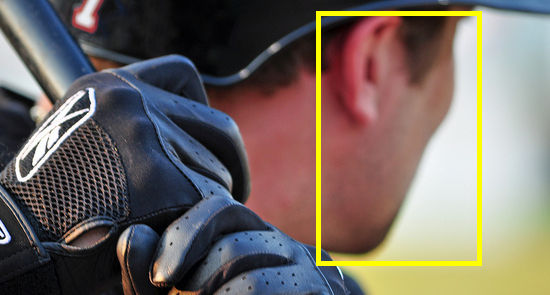}} & 
    \multicolumn{2}{c}{\includegraphics[width=0.331\linewidth]{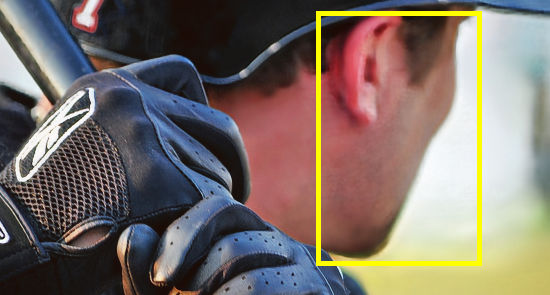}} & 
    \multicolumn{2}{c}{\includegraphics[width=0.331\linewidth]{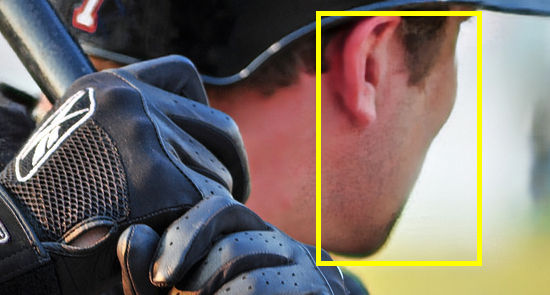}} \\

    \multicolumn{2}{c}{\includegraphics[width=0.331\linewidth]{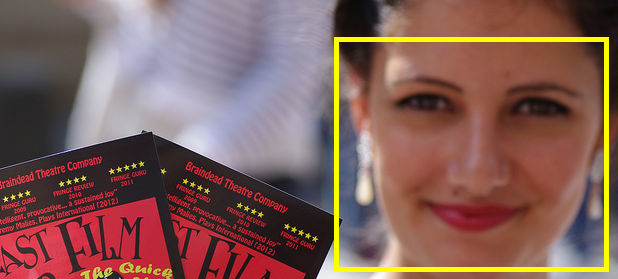}} & 
    \multicolumn{2}{c}{\includegraphics[width=0.331\linewidth]{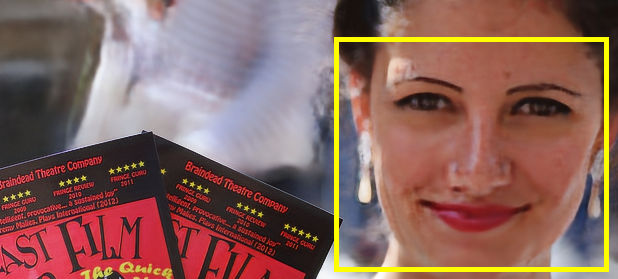}} & 
    \multicolumn{2}{c}{\includegraphics[width=0.331\linewidth]{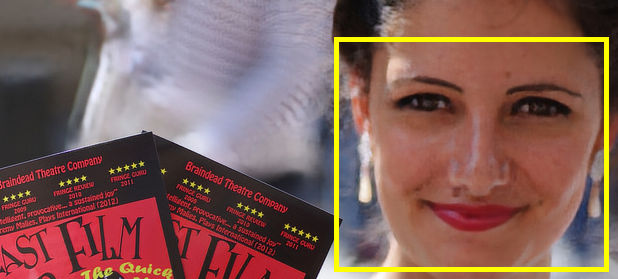}} \\

    \multicolumn{2}{c}{\includegraphics[width=0.331\linewidth]{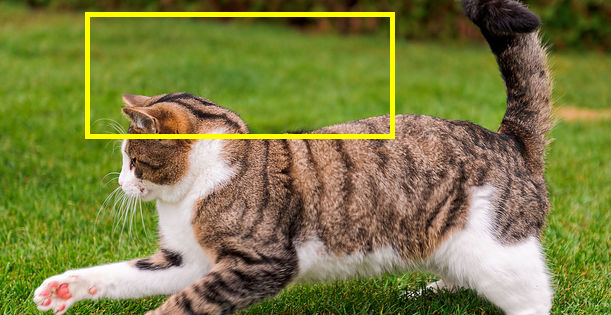}} & 
    \multicolumn{2}{c}{\includegraphics[width=0.331\linewidth]{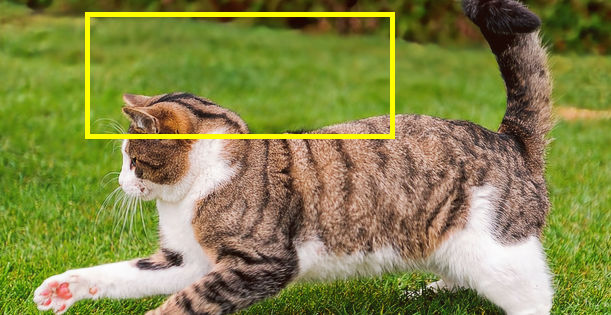}} & 
    \multicolumn{2}{c}{\includegraphics[width=0.331\linewidth]{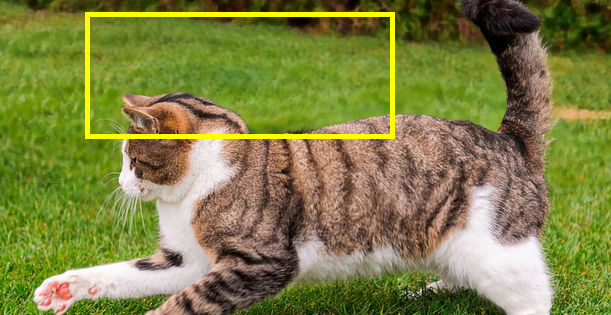}} \\

    \multicolumn{2}{c}{\includegraphics[width=0.331\linewidth]{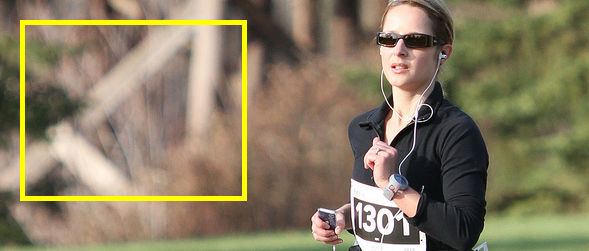}} & 
    \multicolumn{2}{c}{\includegraphics[width=0.331\linewidth]{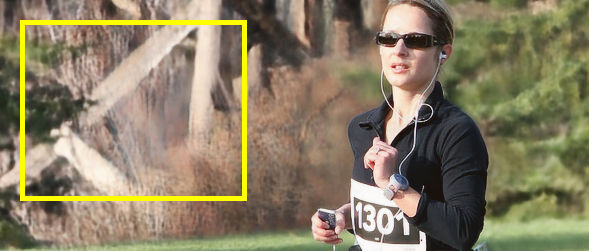}} & 
    \multicolumn{2}{c}{\includegraphics[width=0.331\linewidth]{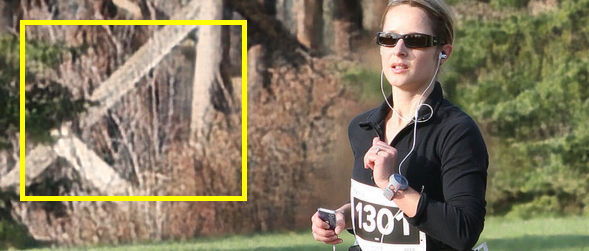}} \\

    \multicolumn{2}{c}{\includegraphics[width=0.331\linewidth]{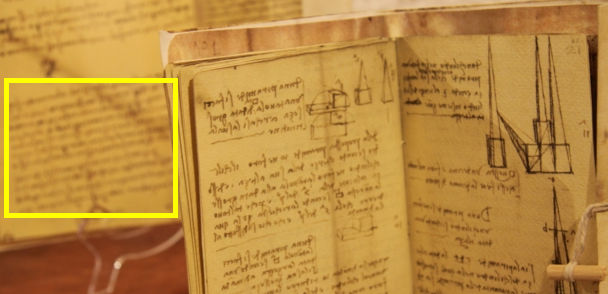}} & 
    \multicolumn{2}{c}{\includegraphics[width=0.331\linewidth]{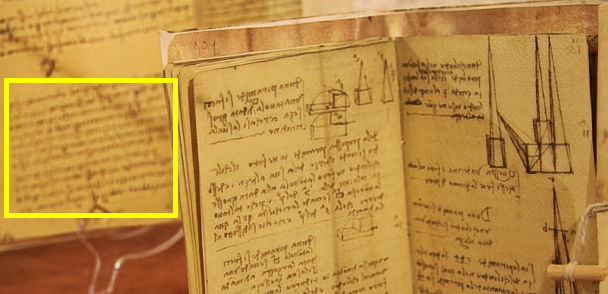}} & 
    \multicolumn{2}{c}{\includegraphics[width=0.331\linewidth]{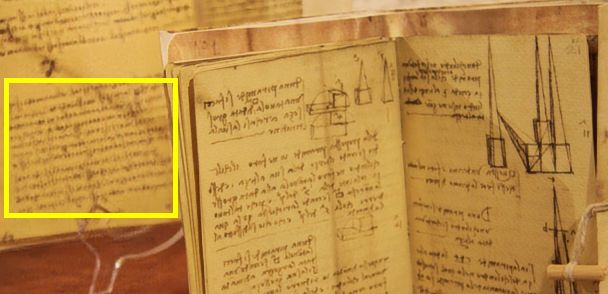}} \\
    
    \multicolumn{2}{c}{\includegraphics[width=0.331\linewidth]{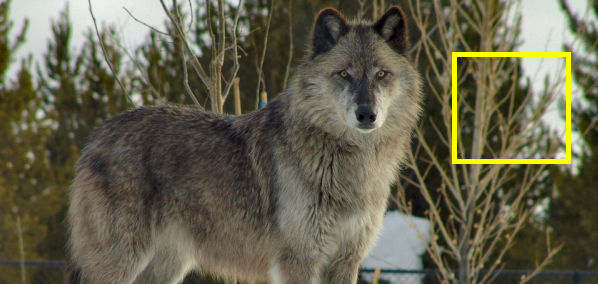}} & 
    \multicolumn{2}{c}{\includegraphics[width=0.331\linewidth]{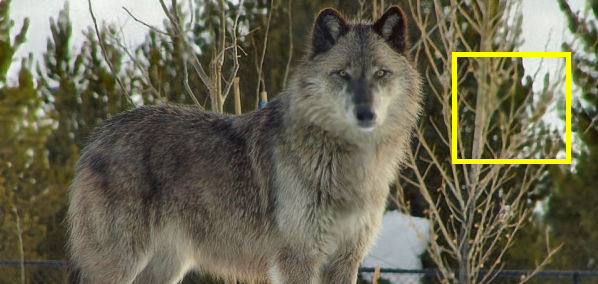}} & 
    \multicolumn{2}{c}{\includegraphics[width=0.331\linewidth]{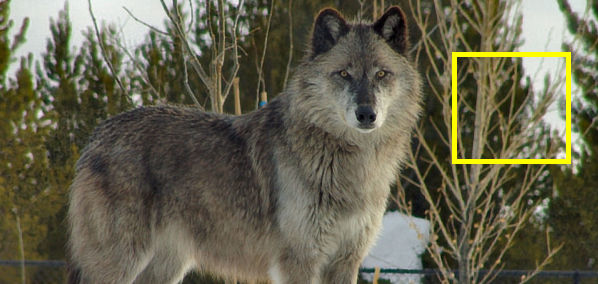}} \\

    \multicolumn{2}{c}{\includegraphics[width=0.331\linewidth]{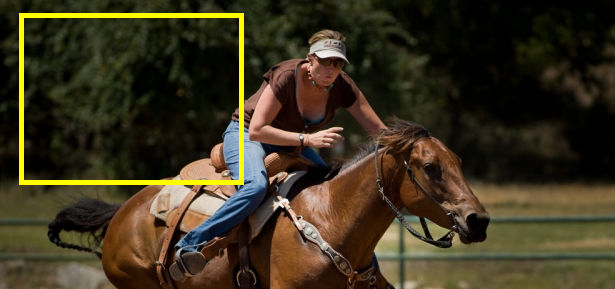}} & 
    \multicolumn{2}{c}{\includegraphics[width=0.331\linewidth]{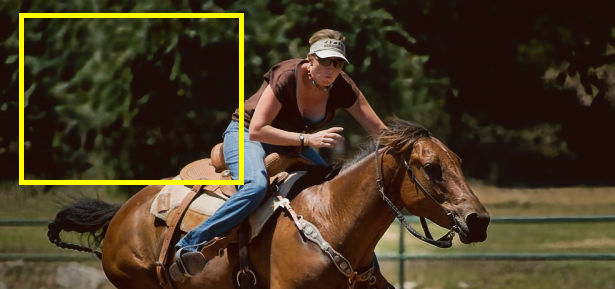}} & 
    \multicolumn{2}{c}{\includegraphics[width=0.331\linewidth]{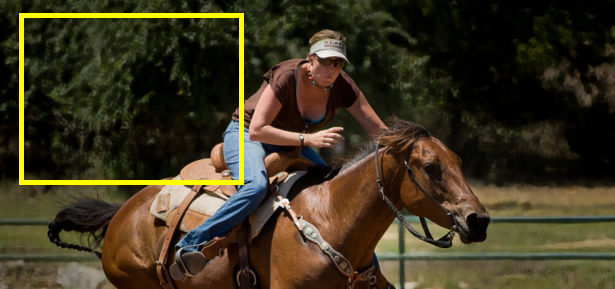}} \\
    
    \multicolumn{2}{c}{(a) input}  & \multicolumn{2}{c}{(b) DPDNet~\cite{Abuolaim:2020:DPDNet}  }
     & \multicolumn{2}{c}{(c) ours } \\
  \end{tabular}
  \vspace{-0.05cm}
  \caption{Additional qualitative comparisons with DPDNet~\cite{Abuolaim:2020:DPDNet} on the defocused images in the CUHK blur detection dataset~\cite{Shi:2014:CUHK}.}
\label{fig:CUHK1}
\end{figure*}

\begin{figure*}[tp]
\centering
\setlength\tabcolsep{1 pt}
  \begin{tabular}{cccccc}

    \multicolumn{2}{c}{\includegraphics[width=0.331\linewidth]{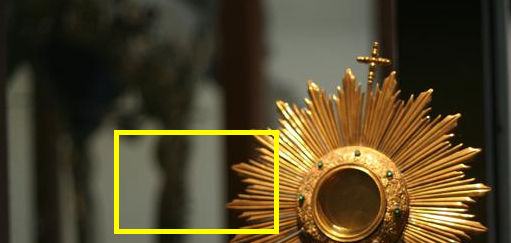}} & 
    \multicolumn{2}{c}{\includegraphics[width=0.331\linewidth]{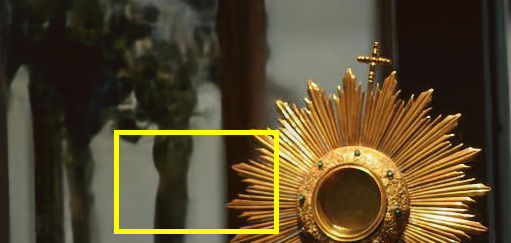}} & 
    \multicolumn{2}{c}{\includegraphics[width=0.331\linewidth]{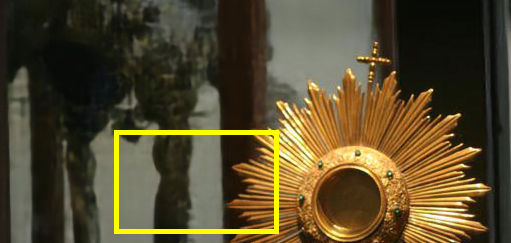}} \\

    \multicolumn{2}{c}{\includegraphics[width=0.331\linewidth]{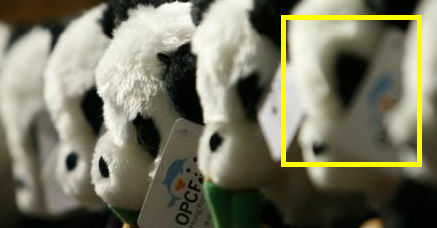}} & 
    \multicolumn{2}{c}{\includegraphics[width=0.331\linewidth]{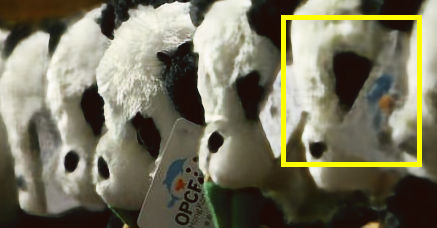}} & 
    \multicolumn{2}{c}{\includegraphics[width=0.331\linewidth]{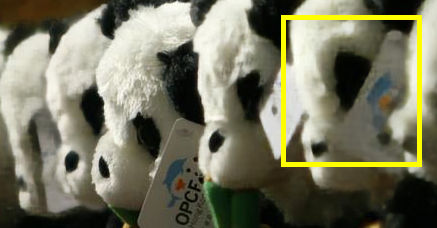}} \\

    \multicolumn{2}{c}{\includegraphics[width=0.331\linewidth]{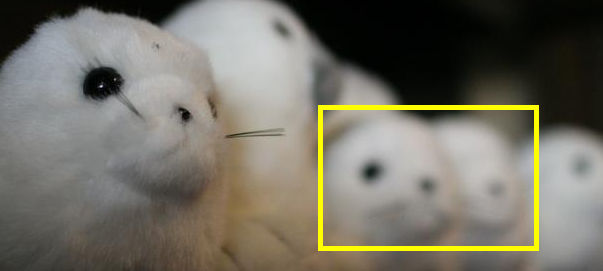}} & 
    \multicolumn{2}{c}{\includegraphics[width=0.331\linewidth]{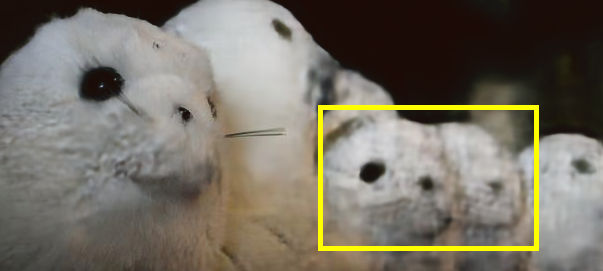}} & 
    \multicolumn{2}{c}{\includegraphics[width=0.331\linewidth]{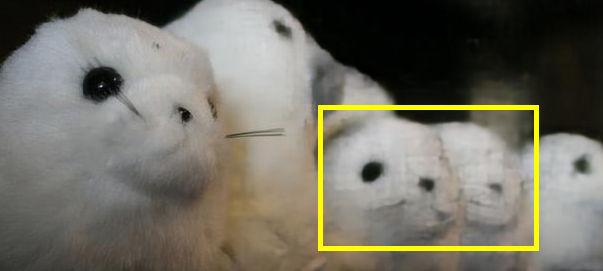}} \\
    
    \multicolumn{2}{c}{\includegraphics[width=0.331\linewidth]{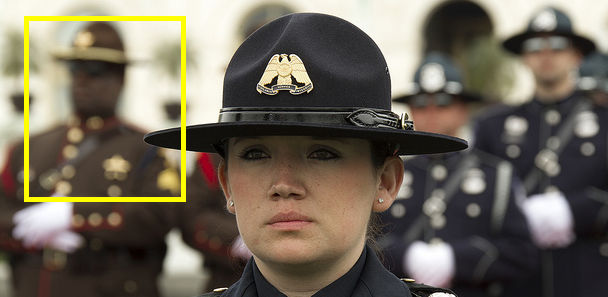}} & 
    \multicolumn{2}{c}{\includegraphics[width=0.331\linewidth]{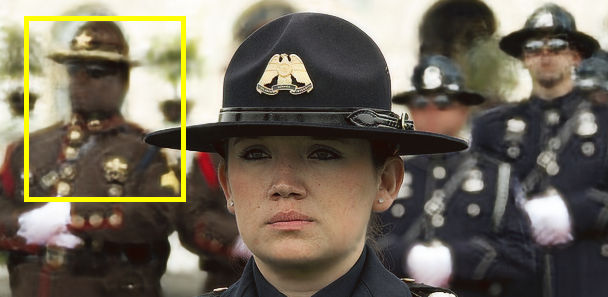}} & 
    \multicolumn{2}{c}{\includegraphics[width=0.331\linewidth]{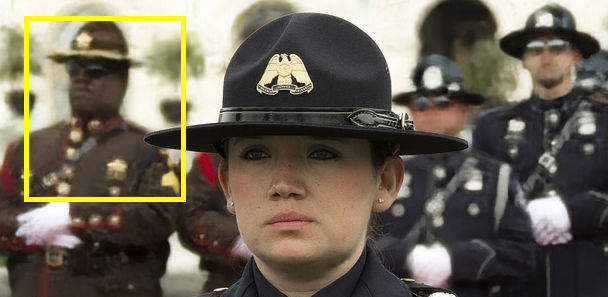}} \\

    
    \multicolumn{2}{c}{\includegraphics[width=0.331\linewidth]{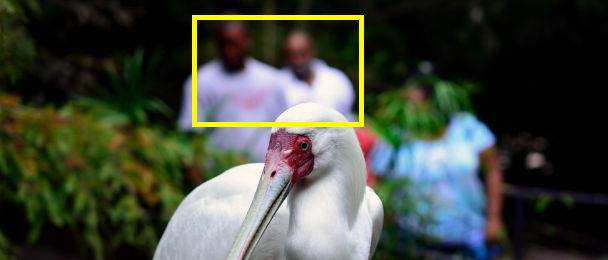}} & 
    \multicolumn{2}{c}{\includegraphics[width=0.331\linewidth]{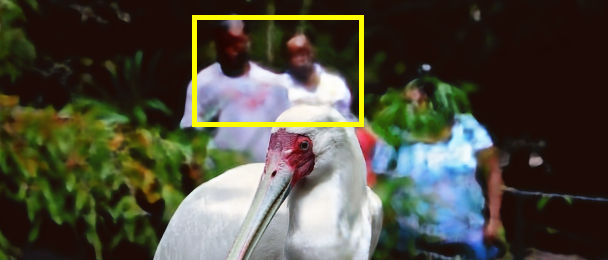}} & 
    \multicolumn{2}{c}{\includegraphics[width=0.331\linewidth]{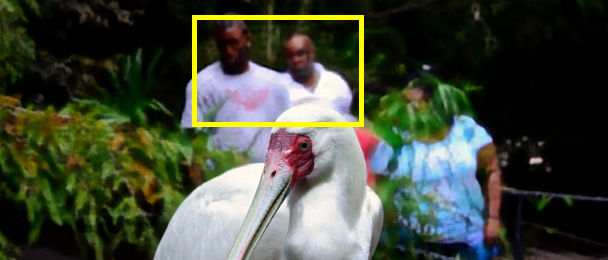}} \\
    
    \multicolumn{2}{c}{\includegraphics[width=0.331\linewidth]{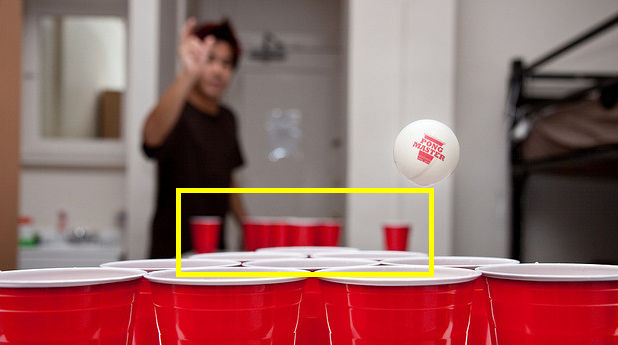}} & 
    \multicolumn{2}{c}{\includegraphics[width=0.331\linewidth]{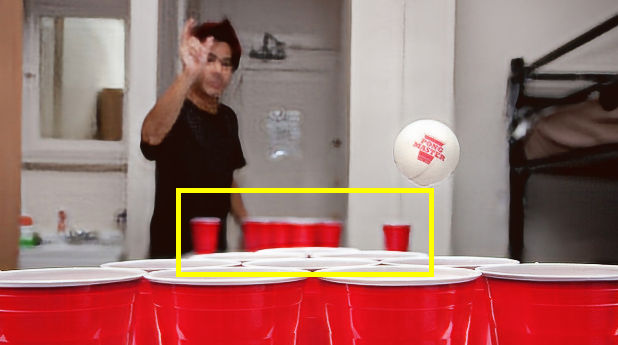}} & 
    \multicolumn{2}{c}{\includegraphics[width=0.331\linewidth]{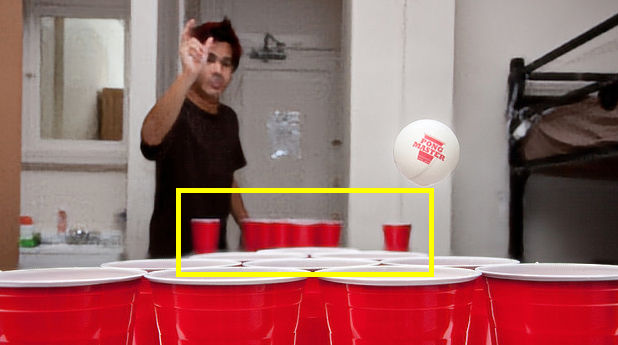}} \\

    \multicolumn{2}{c}{\includegraphics[width=0.331\linewidth]{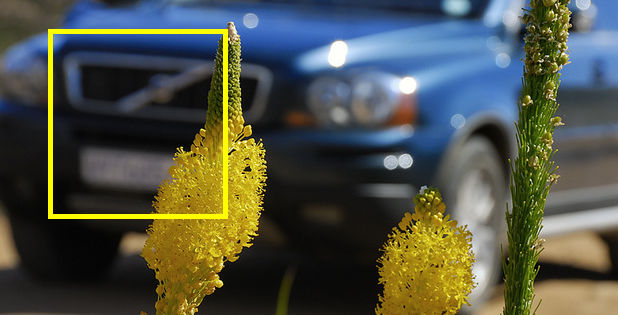}} & 
    \multicolumn{2}{c}{\includegraphics[width=0.331\linewidth]{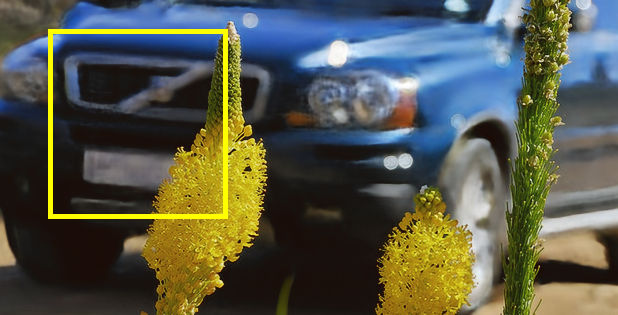}} & 
    \multicolumn{2}{c}{\includegraphics[width=0.331\linewidth]{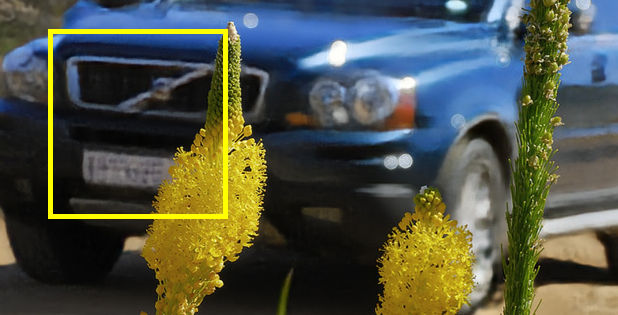}} \\

    \multicolumn{2}{c}{(a) input}  & \multicolumn{2}{c}{(b) DPDNet~\cite{Abuolaim:2020:DPDNet}  }
     & \multicolumn{2}{c}{(c) ours } \\

  \end{tabular}
  \vspace{-0.05cm}
  \caption{Additional qualitative comparisons with DPDNet~\cite{Abuolaim:2020:DPDNet} on the defocused images in the CUHK blur detection dataset~\cite{Shi:2014:CUHK}.}
\label{fig:CUHK2}
\end{figure*}

We present additional qualitative results on the DPDD dataset~\cite{Abuolaim:2020:DPDNet} (\Figs{dpdd1} and \ref{fig:dpdd2}) and the CUHK blur detection dataset~\cite{Shi:2014:CUHK} (\Figs{CUHK1} and \ref{fig:CUHK2}).

\section{Our Model Using Dual-Pixel Images}
\label{sec:dualpixel}
While our model with a single image input shows state-of-the-art performance, 
the deblurring performance can further boosted by using dual-pixel images as the input.
For the experiment, we retrained our model by replacing a single image input as dual-pixel image input with the same training strategy in Sec. 5 of the main paper.
Specifically, we concatenate the two images of a dual-pixel image in the channel dimension and use it as the input of the network.

\begin{figure*}[tp]
\centering
\setlength\tabcolsep{1 pt}
  \begin{tabular}{cccccccccc}

    \multicolumn{2}{c}{\includegraphics[width=0.246\linewidth]{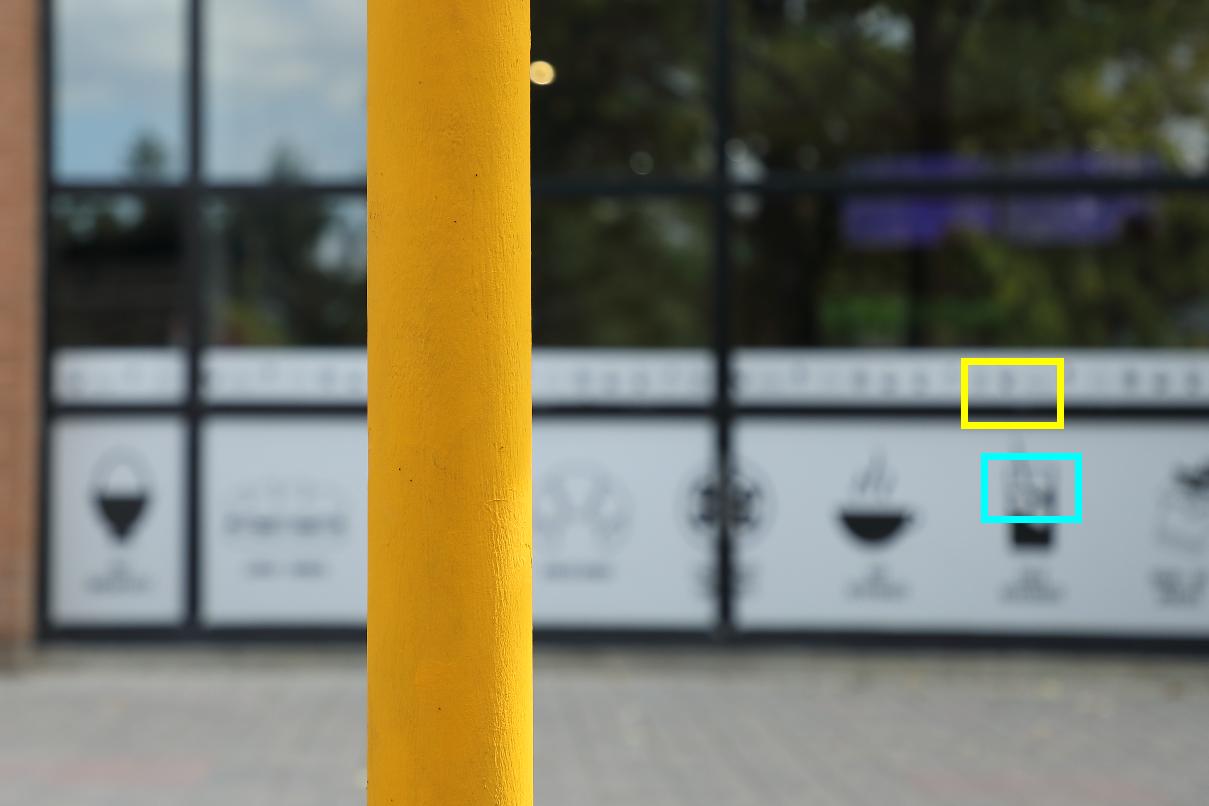}} &
    \multicolumn{2}{c}{\includegraphics[width=0.246\linewidth]{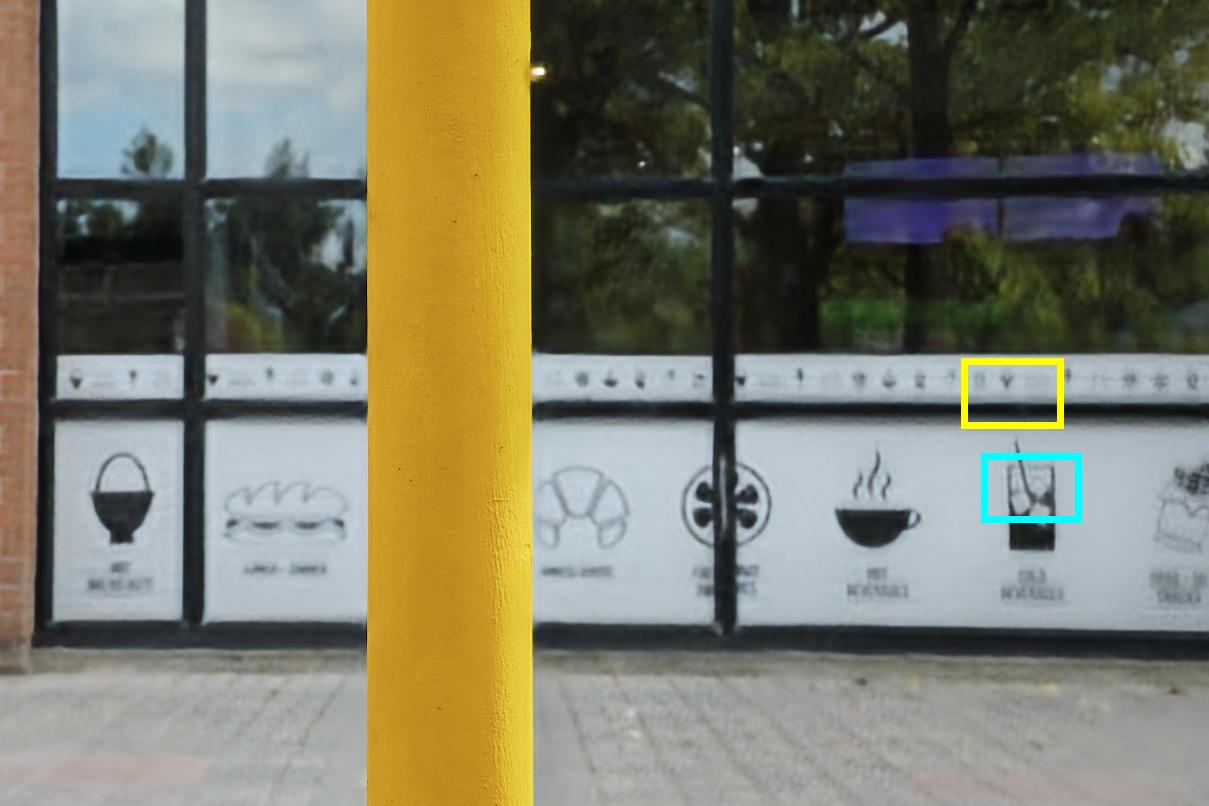}} &
    \multicolumn{2}{c}{\includegraphics[width=0.246\linewidth]{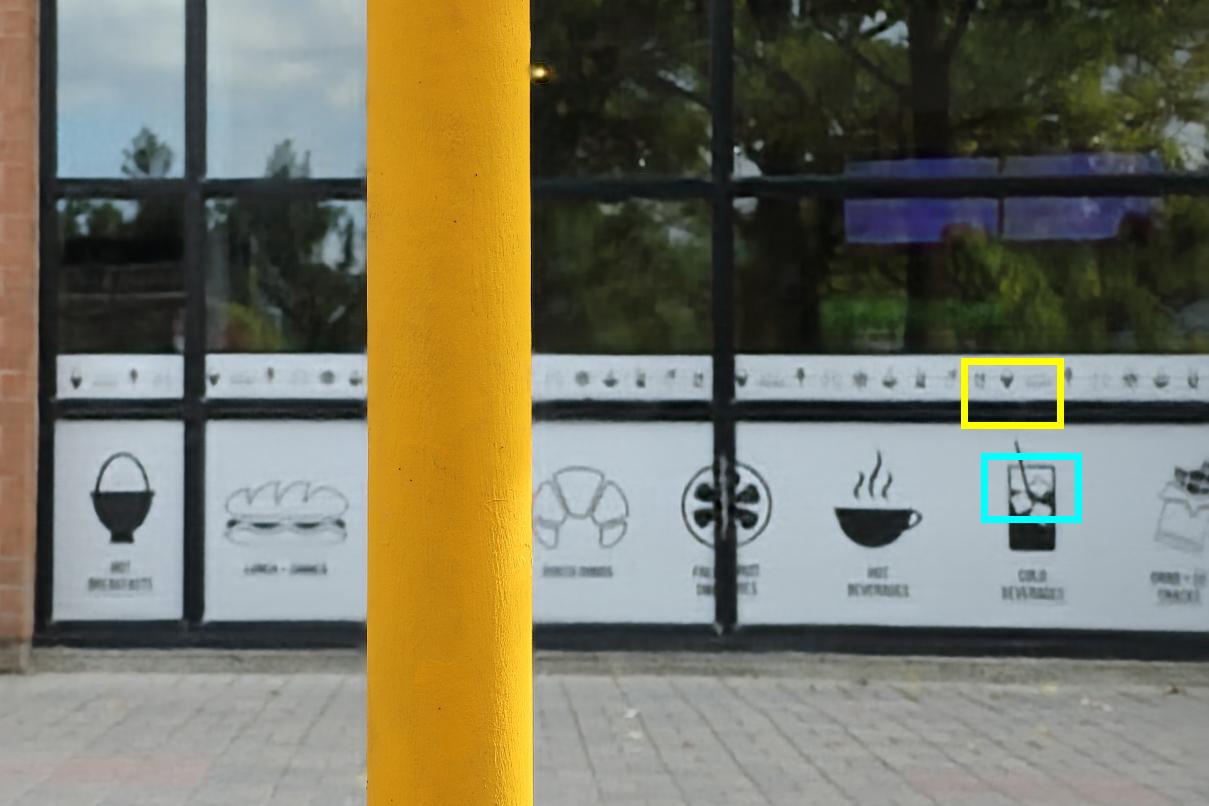}} &
    \multicolumn{2}{c}{\includegraphics[width=0.246\linewidth]{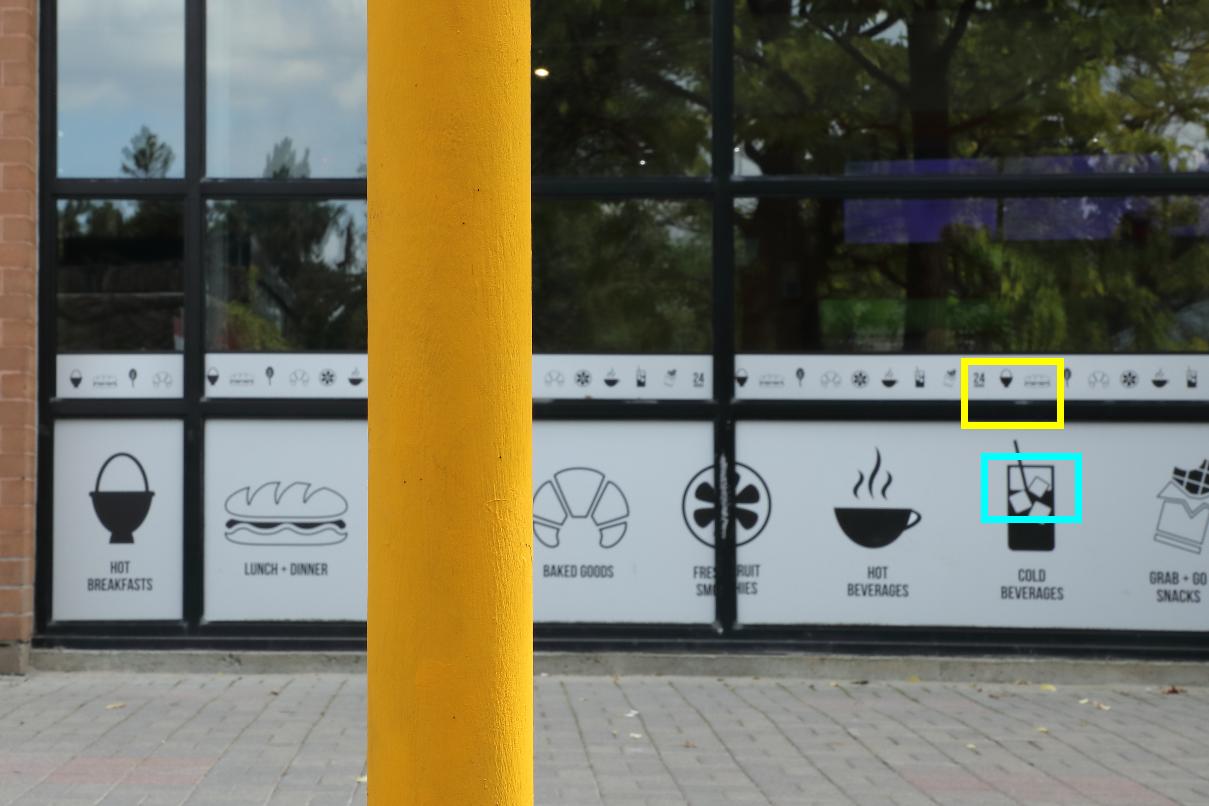}} \\
    \multicolumn{1}{c}{\includegraphics[width=0.122\linewidth]{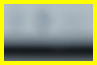}} &
    \multicolumn{1}{c}{\includegraphics[width=0.122\linewidth]{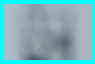}} &
    \multicolumn{1}{c}{\includegraphics[width=0.122\linewidth]{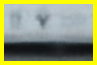}} &
    \multicolumn{1}{c}{\includegraphics[width=0.122\linewidth]{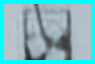}} &
    \multicolumn{1}{c}{\includegraphics[width=0.122\linewidth]{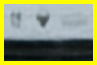}} &
    \multicolumn{1}{c}{\includegraphics[width=0.122\linewidth]{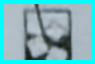}} &
    \multicolumn{1}{c}{\includegraphics[width=0.122\linewidth]{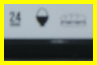}} &
    \multicolumn{1}{c}{\includegraphics[width=0.122\linewidth]{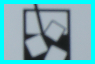}} \\

    \multicolumn{2}{c}{\includegraphics[width=0.246\linewidth]{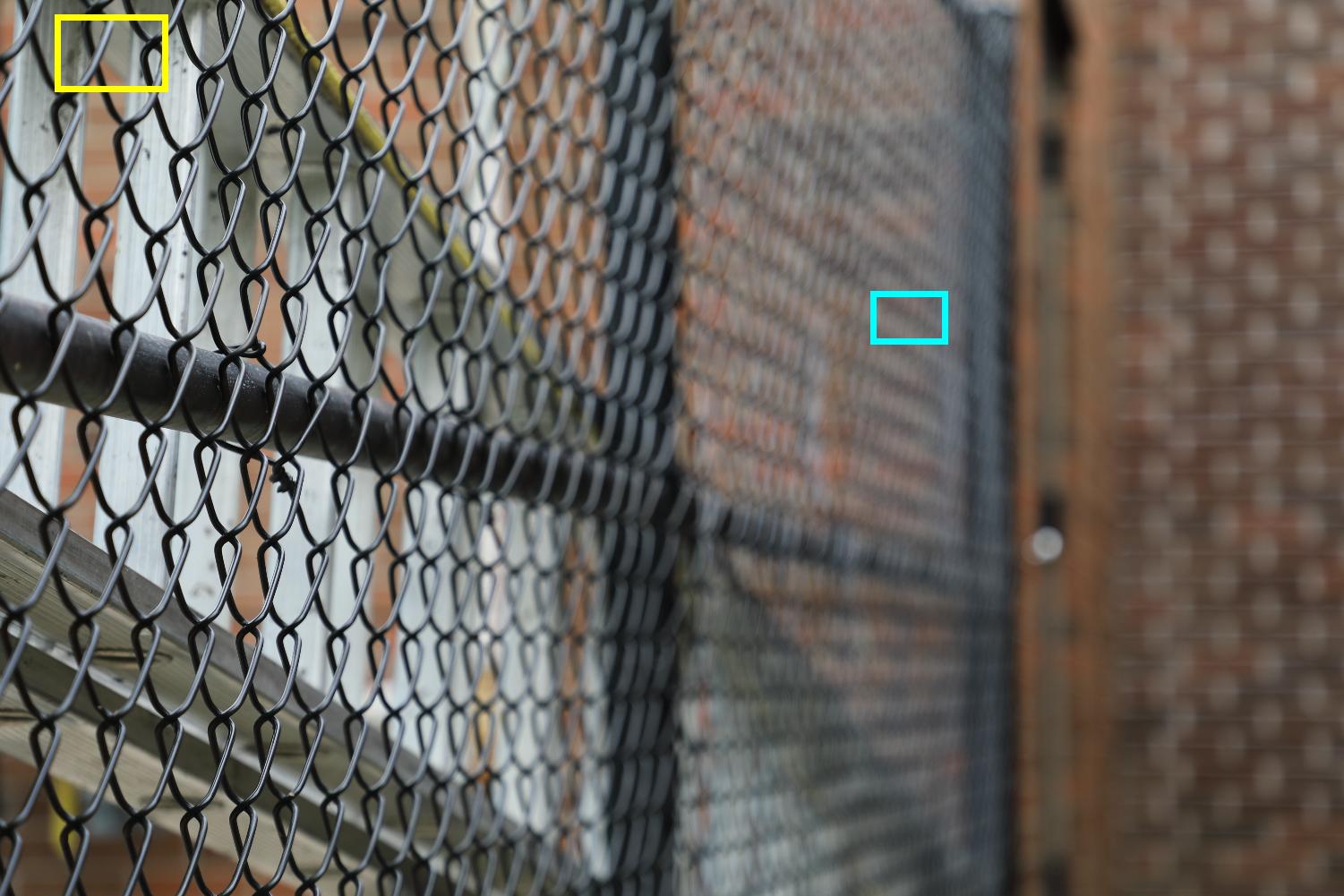}} &
    \multicolumn{2}{c}{\includegraphics[width=0.246\linewidth]{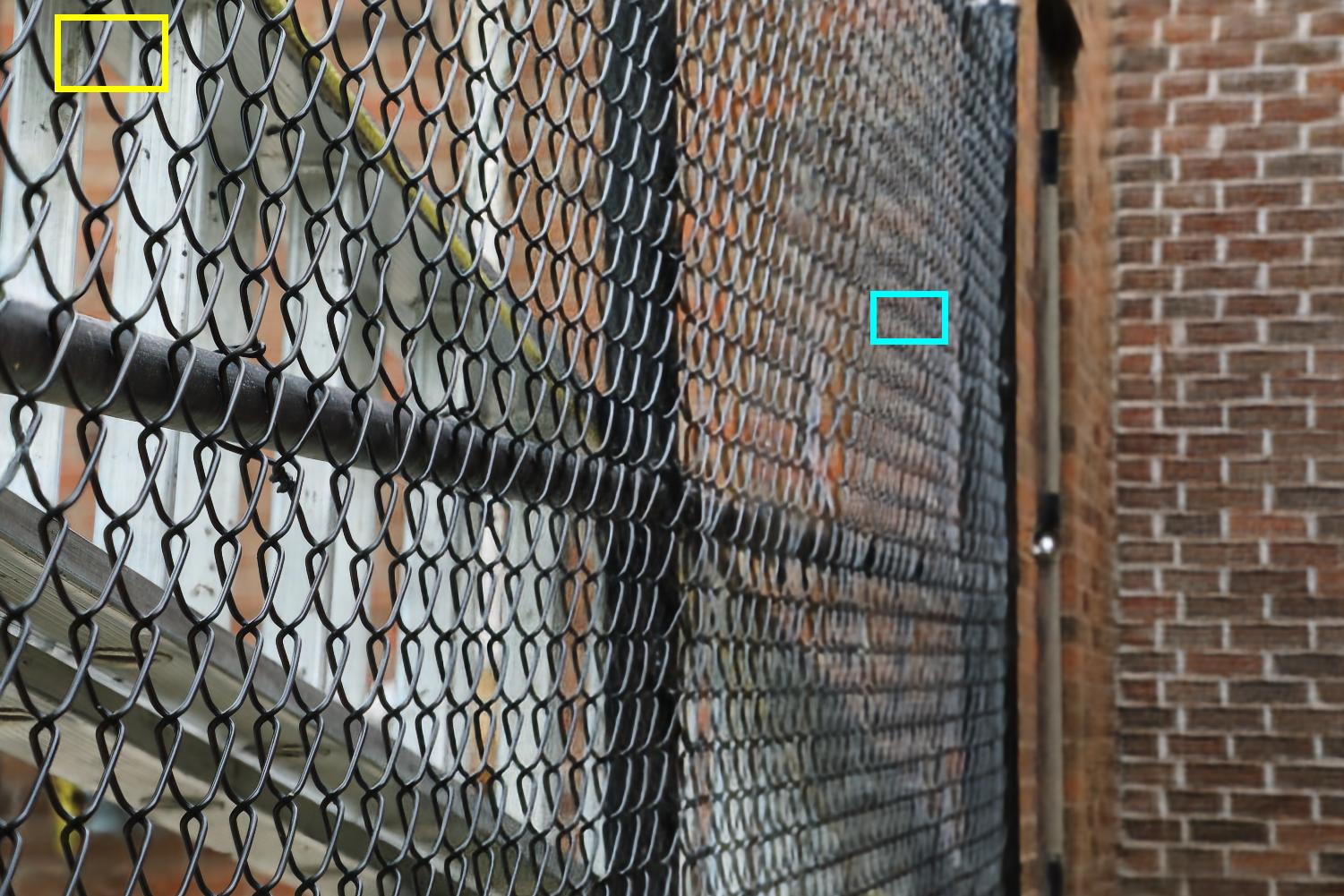}} &
    \multicolumn{2}{c}{\includegraphics[width=0.246\linewidth]{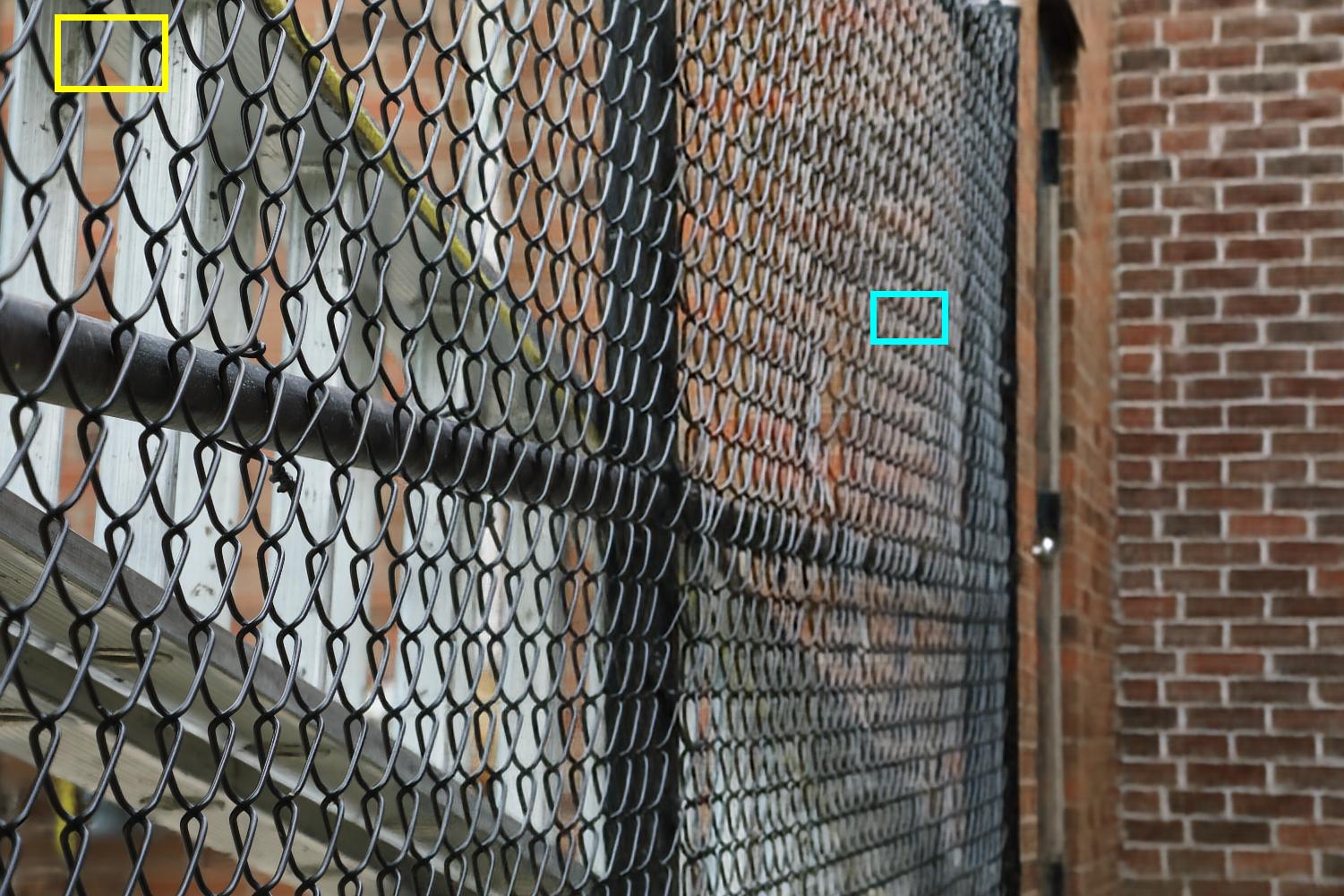}} &
    \multicolumn{2}{c}{\includegraphics[width=0.246\linewidth]{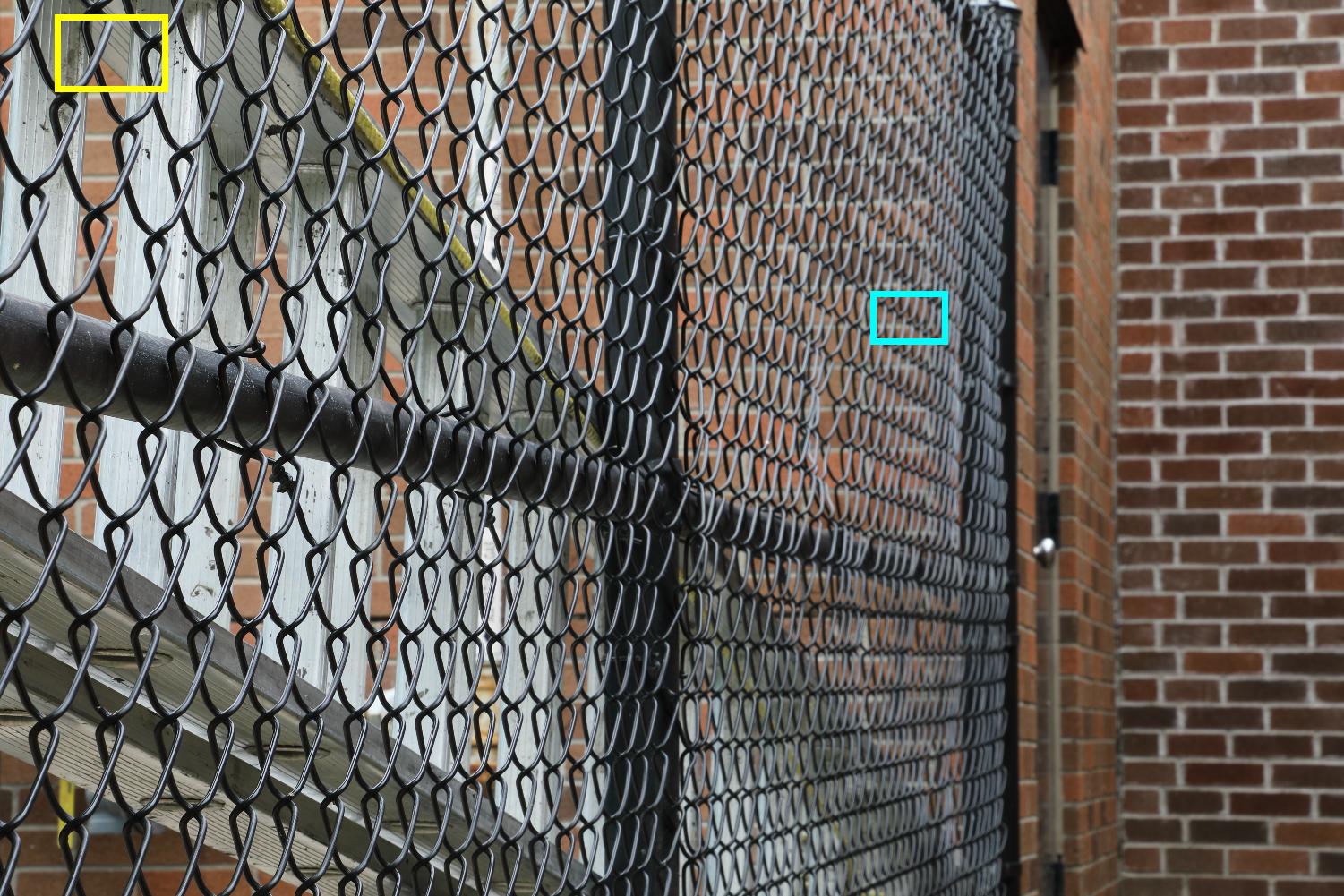}} \\
    \multicolumn{1}{c}{\includegraphics[width=0.122\linewidth]{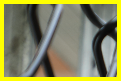}} &
    \multicolumn{1}{c}{\includegraphics[width=0.122\linewidth]{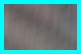}} &
    \multicolumn{1}{c}{\includegraphics[width=0.122\linewidth]{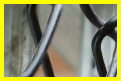}} &
    \multicolumn{1}{c}{\includegraphics[width=0.122\linewidth]{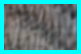}} &
    \multicolumn{1}{c}{\includegraphics[width=0.122\linewidth]{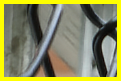}} &
    \multicolumn{1}{c}{\includegraphics[width=0.122\linewidth]{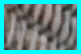}} &
    \multicolumn{1}{c}{\includegraphics[width=0.122\linewidth]{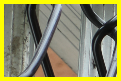}} &
    \multicolumn{1}{c}{\includegraphics[width=0.122\linewidth]{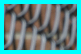}} \\

    \multicolumn{2}{c}{\includegraphics[width=0.246\linewidth]{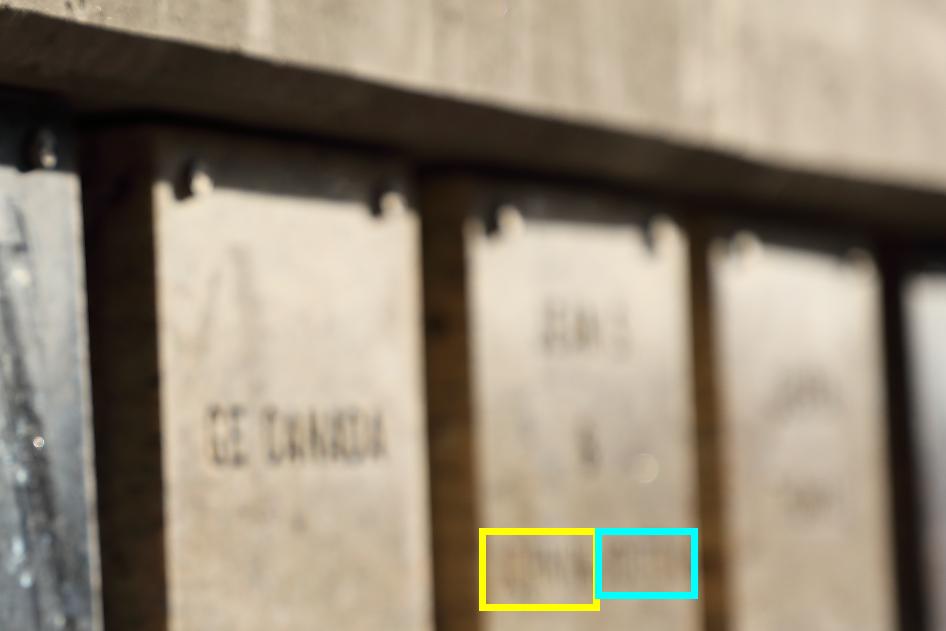}} &
    \multicolumn{2}{c}{\includegraphics[width=0.246\linewidth]{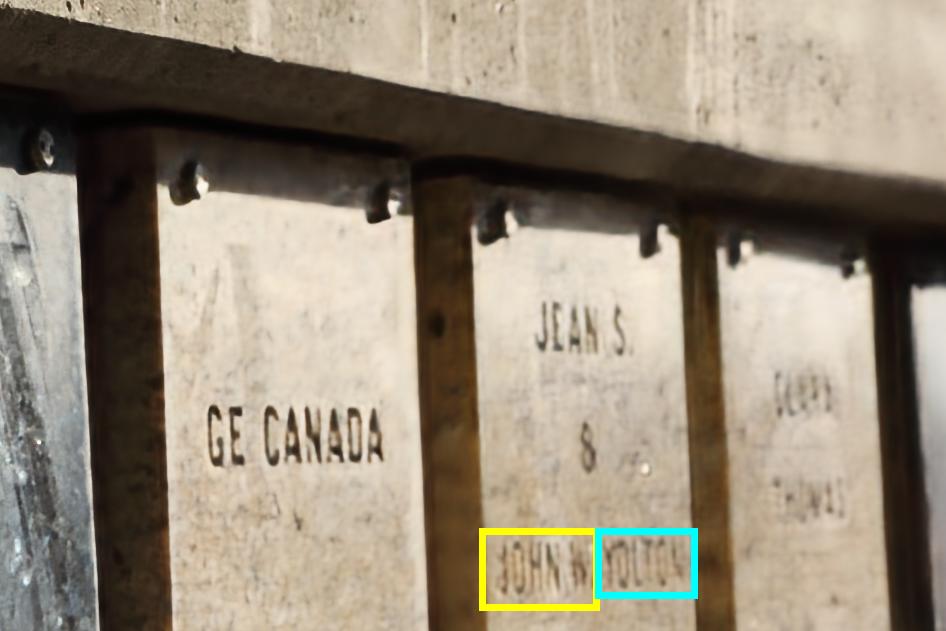}} &
    \multicolumn{2}{c}{\includegraphics[width=0.246\linewidth]{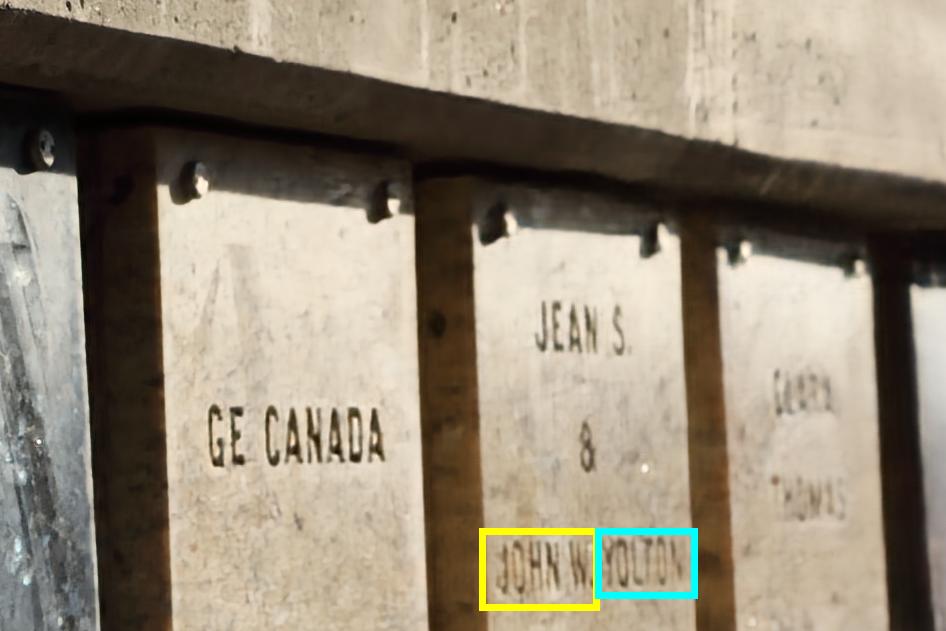}} &
    \multicolumn{2}{c}{\includegraphics[width=0.246\linewidth]{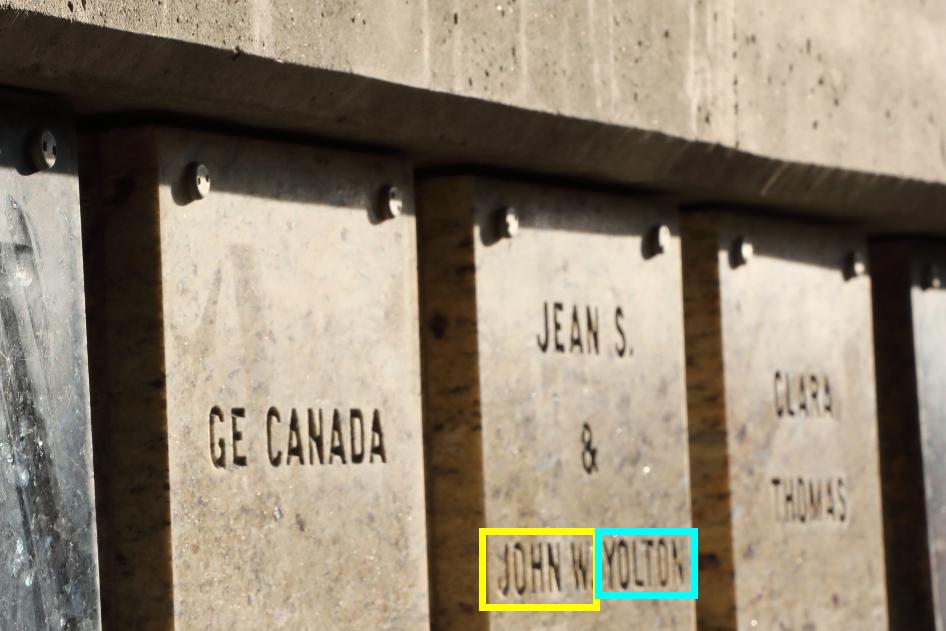}} \\
    \multicolumn{1}{c}{\includegraphics[width=0.122\linewidth]{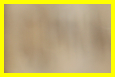}} &
    \multicolumn{1}{c}{\includegraphics[width=0.122\linewidth]{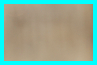}} &
    \multicolumn{1}{c}{\includegraphics[width=0.122\linewidth]{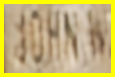}} &
    \multicolumn{1}{c}{\includegraphics[width=0.122\linewidth]{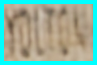}} &
    \multicolumn{1}{c}{\includegraphics[width=0.122\linewidth]{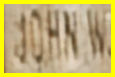}} &
    \multicolumn{1}{c}{\includegraphics[width=0.122\linewidth]{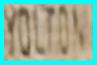}} &
    \multicolumn{1}{c}{\includegraphics[width=0.122\linewidth]{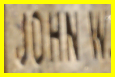}} &
    \multicolumn{1}{c}{\includegraphics[width=0.122\linewidth]{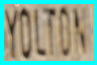}} \\
    
    \multicolumn{2}{c}{\includegraphics[width=0.246\linewidth]{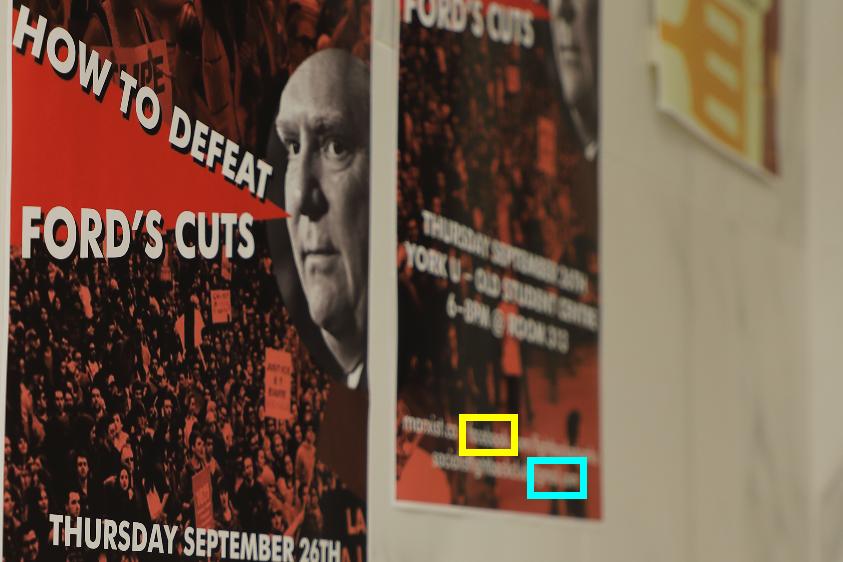}} &
    \multicolumn{2}{c}{\includegraphics[width=0.246\linewidth]{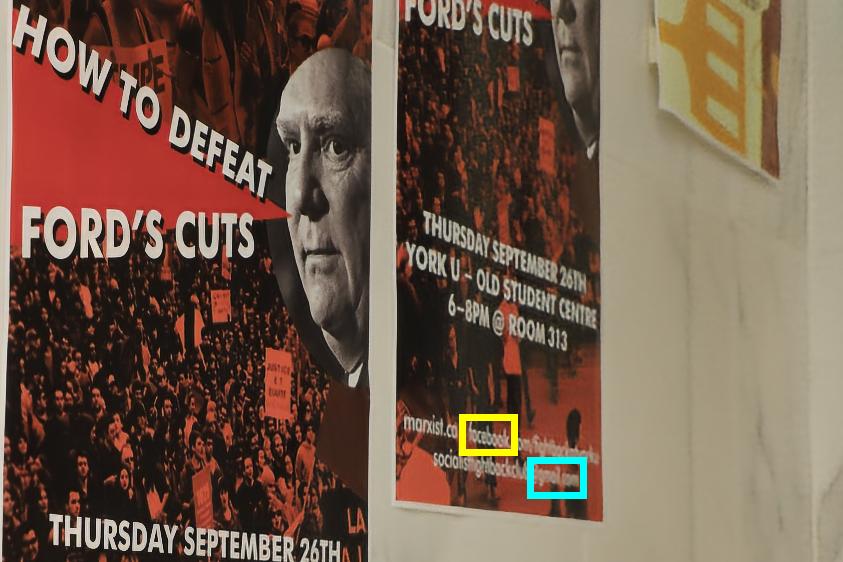}} &
    \multicolumn{2}{c}{\includegraphics[width=0.246\linewidth]{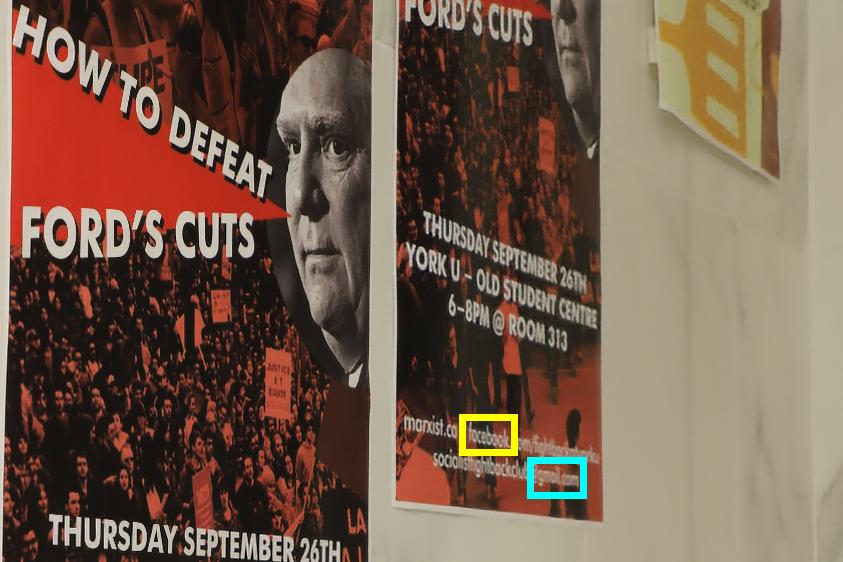}} &
    \multicolumn{2}{c}{\includegraphics[width=0.246\linewidth]{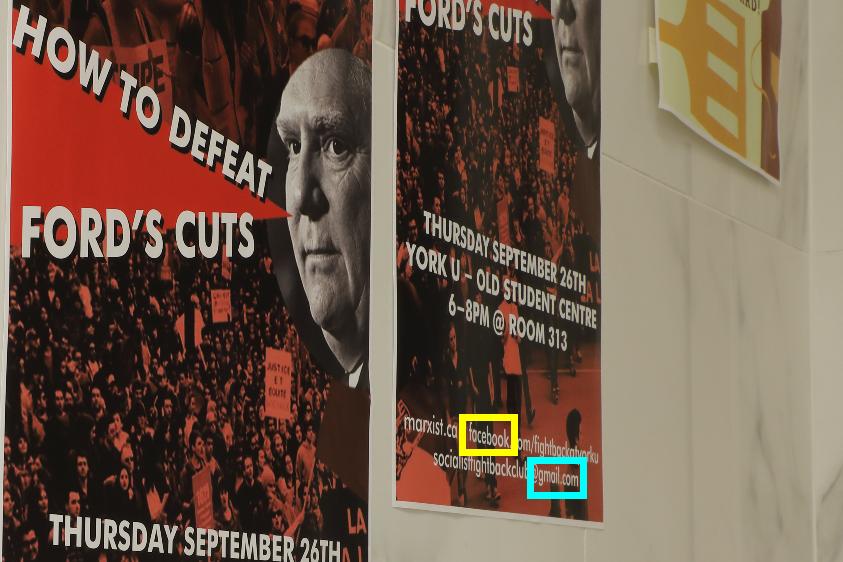}} \\
    \multicolumn{1}{c}{\includegraphics[width=0.122\linewidth]{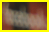}} &
    \multicolumn{1}{c}{\includegraphics[width=0.122\linewidth]{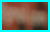}} &
    \multicolumn{1}{c}{\includegraphics[width=0.122\linewidth]{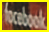}} &
    \multicolumn{1}{c}{\includegraphics[width=0.122\linewidth]{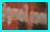}} &
    \multicolumn{1}{c}{\includegraphics[width=0.122\linewidth]{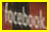}} &
    \multicolumn{1}{c}{\includegraphics[width=0.122\linewidth]{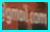}} &
    \multicolumn{1}{c}{\includegraphics[width=0.122\linewidth]{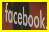}} &
    \multicolumn{1}{c}{\includegraphics[width=0.122\linewidth]{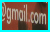}} \\

    \multicolumn{2}{c}{(a) input}  & \multicolumn{2}{c}{(b) DPDNet (dual)~\cite{Abuolaim:2020:DPDNet}}
     & \multicolumn{2}{c}{(c) ours (dual) } & \multicolumn{2}{c}{(d) GT} \\

  \end{tabular}
  \vspace{-0.05cm}
  \caption{Qualitative comparisons of defocus deblurring models using dual-pixel input on the test set of the DPDD dataset \cite{Abuolaim:2020:DPDNet}.}
\label{fig:dpdd_dual}
\vspace{-10pt}
\end{figure*}

\Tbl{dual_pixel} shows that dual-pixel input further improves the performance of our model, and our model with dual-pixel input outperforms DPDNet~\cite{Abuolaim:2020:DPDNet} with dual-pixel input by a large margin.
\Fig{dpdd_dual} shows that dual-pixel input enables our model to handle fine details better.

\begin{table}[t]
\centering
\scalebox{0.78}{
\begin{tabular}{ c  c c c c }
\toprule
Model & PSNR\(\uparrow\) & SSIM\(\uparrow\) & LPIPS\(\downarrow\) & Parameters (M)\\
\cmidrule(r){1-1} \cmidrule(l){2-5}
DPDNet (single)~\cite{Abuolaim:2020:DPDNet} & 24.42 & 0.827 & 0.277 & 32.25\\
DPDNet (dual)~\cite{Abuolaim:2020:DPDNet}& 25.12 & 0.850 & 0.223 & 32.25\\
Ours (single) & 25.24 & 0.842 & 0.225 & 2.06 \\
Ours (dual) & \textbf{25.86} & \textbf{0.859} & \textbf{0.185}  & 2.06 \\
\bottomrule
\end{tabular}
}
\vspace{0.05cm}
\caption{Quantitative performance of our dual-pixel-based model. }
\label{tbl:dual_pixel}
\vspace{-12pt}
\end{table}

{\small
\bibliographystyleSM{ieee_fullname}
\bibliographySM{egbib}
}

\end{document}